\documentclass[10pt,twocolumn,letterpaper]{article}

\usepackage{cvpr}
\usepackage{times}
\usepackage{epsfig}
\usepackage{graphicx}
\usepackage{amsmath}
\usepackage{amssymb}
\usepackage{multirow}
\usepackage{algorithm}
\usepackage{algorithmic}
\usepackage{booktabs}

\usepackage[pagebackref=true,breaklinks=true,letterpaper=true,colorlinks,citecolor=blue,linkcolor=blue,bookmarks=false]{hyperref}

\usepackage[breaklinks=true,bookmarks=false]{hyperref}

\cvprfinalcopy 

\def\cvprPaperID{****} 

\ifcvprfinal\fi
\begin{document}

\title{Gated Fusion Network for Single Image Dehazing}

\author{Wenqi Ren$^{1}$\thanks{Part of this work was done while Wenqi Ren was with Tencent AI Lab as a Visiting Scholar.}, Lin Ma$^{2}$, Jiawei Zhang$^{3}$, Jinshan Pan$^{4}$, Xiaochun Cao$^{1,5}$\thanks{Corresponding author.}, Wei Liu$^{2}$, and Ming-Hsuan Yang$^{6}$\\
	$^{1}$State Key Laboratory of Information Security (SKLOIS), IIE, CAS\\
	$^{2}$Tencent AI Lab ~~~
	$^{3}$City University of Hong Kong ~~~
	$^{4}$Nanjing University of Science and Technology\\
	$^{5}$School of Cyber Security, University of Chinese Academy of Sciences\\
	$^{6}$Electrical Engineering and Computer Science, University of California, Merced\\
	{\small\url{https://sites.google.com/site/renwenqi888/research/dehazing/gfn}}
}

\maketitle
\thispagestyle{empty}

\begin{abstract}
	In this paper, we propose an efficient algorithm to directly restore a clear image from a hazy input.
	The proposed algorithm hinges on an end-to-end trainable neural network that consists of an encoder and a decoder. The encoder is exploited to capture the context of the derived input images,
	while the decoder is employed to estimate the contribution of each input to the final dehazed result using the learned representations attributed to the encoder.
	The constructed network adopts a novel fusion-based strategy which derives three inputs from an original hazy image by applying White Balance (WB), Contrast Enhancing (CE), and Gamma Correction (GC).
	We compute pixel-wise confidence maps based on the appearance differences between these different inputs to blend the information of the derived inputs and preserve the regions with pleasant visibility.
	The final dehazed image is yielded by gating the important features of the derived inputs.
	To train the network, we introduce a multi-scale approach such that the halo artifacts can be avoided.
	Extensive experimental results on both synthetic and real-world images demonstrate that the proposed algorithm performs favorably against the state-of-the-art algorithms.
\end{abstract}

\vspace{-0.2cm}
\section{Introduction}
\vspace{-0.1cm}
The single image dehazing problem~\cite{he2011singlecvpr,zhang2015retina} aims to estimate the unknown clean image given a hazy or foggy image. This is a classical
image processing problem, which has received active research efforts in the vision communities since various high-level scene understanding tasks \cite{liu2018cross,sakaridis2017semantic,song_wacv14_decolor,yuan2017temporal} require the image dehazing to recover the clear scene. 
Early approaches focus on developing hand-crafted features based on the statistics of clear images, such as dark channel prior~\cite{he2011singlecvpr} and local max contrast~\cite{bao_icpr12_filter,tan2008visibility}.
To avoid hand-crafted priors, recent work~\cite{cai2016dehazenet,li2017aod,ren2016single,He_dehaze_2018} automatically learns haze relevant features by convolutional neural networks (CNNs).
In the dehazing literature, the hazing process is usually modeled as,
\begin{equation}
\mathbf{I}(x)=\mathbf{J}(x)t(x)+\mathbf{A}\big(1-t(x)\big),
\label{eq-math_model}
\end{equation}
where $\mathbf{I}(x)$ and $\mathbf{J}(x)$ are the observed hazy image and the haze-free scene radiance, $\mathbf{A}$ is the global atmospheric
light, and $t(x)$ is the scene transmission describing the portion of light
that is not scattered and reaches the camera sensors.
In practice, transmission and atmospheric light are unknown.
Thus, most dehazing methods try to estimate the transmission $t(x)$
and the atmospheric light $\mathbf{A}$, given a hazy image.

\begin{figure}[t]\footnotesize
	\begin{center}
		\begin{tabular}{@{}cccc@{}}
			\includegraphics[width = 0.11\textwidth]{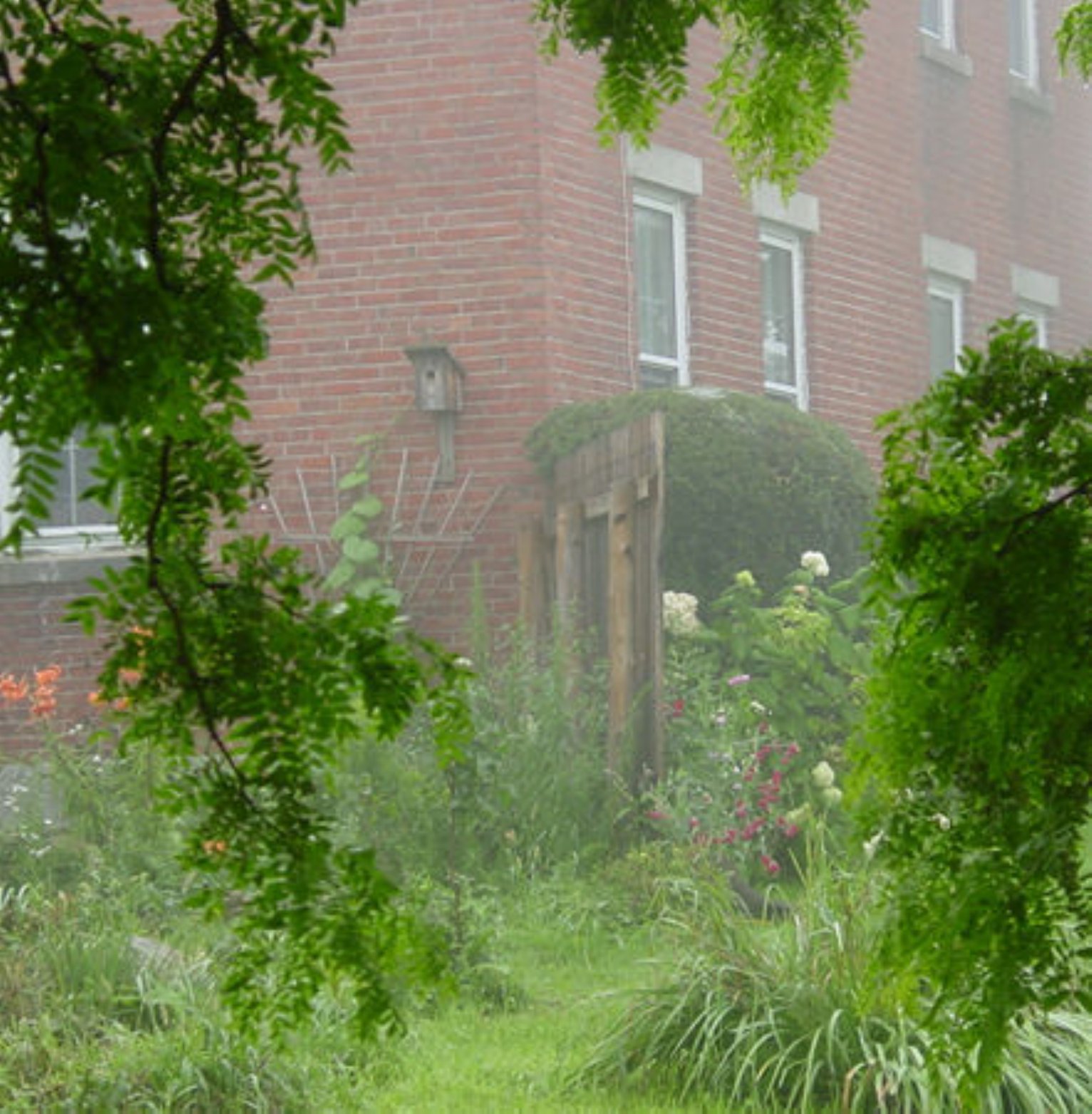} & \hspace{-0.4cm}
			\includegraphics[width = 0.11\textwidth]{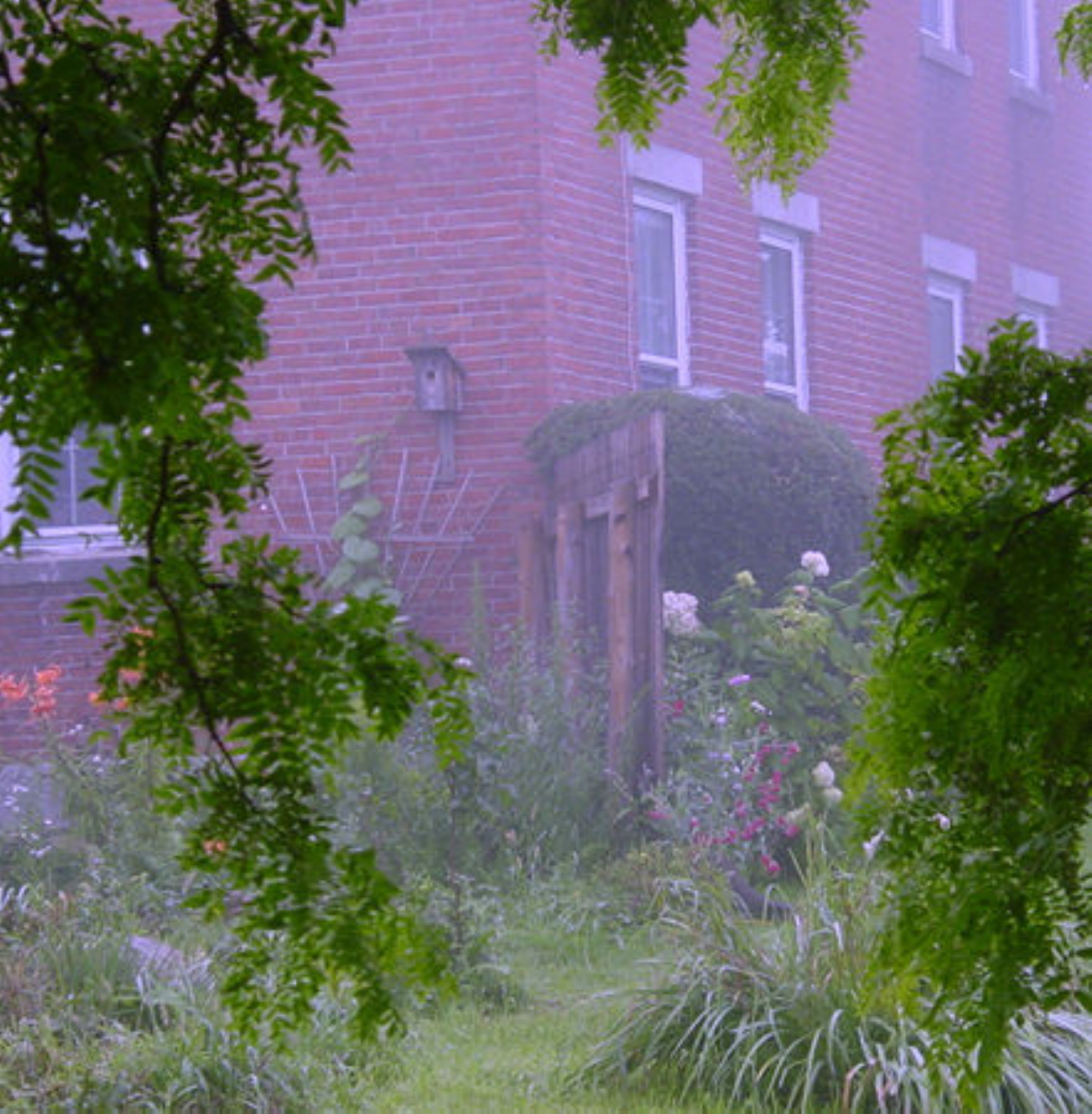} & \hspace{-0.4cm}
			\includegraphics[width = 0.11\textwidth]{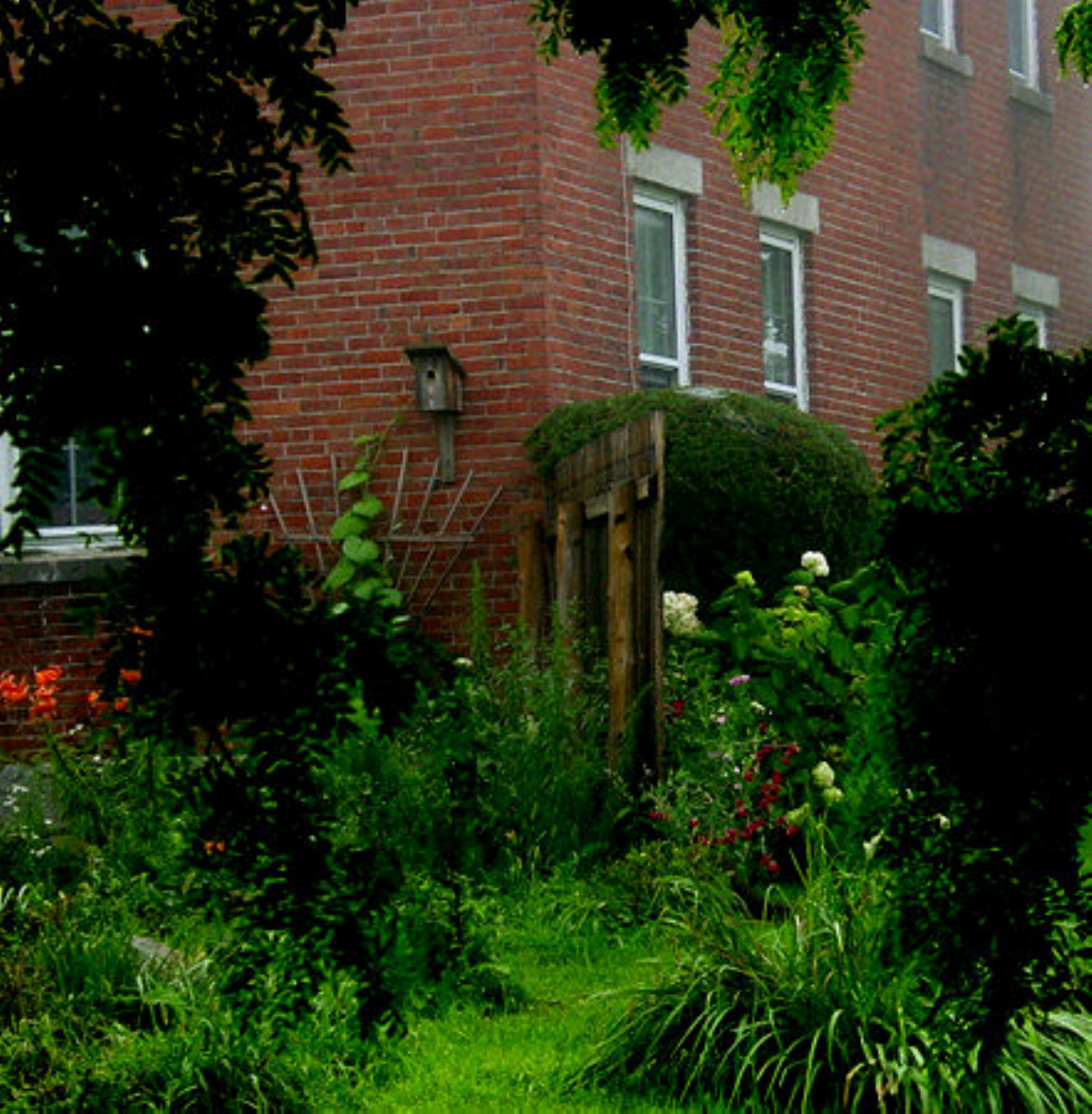} & \hspace{-0.4cm}
			\includegraphics[width = 0.11\textwidth]{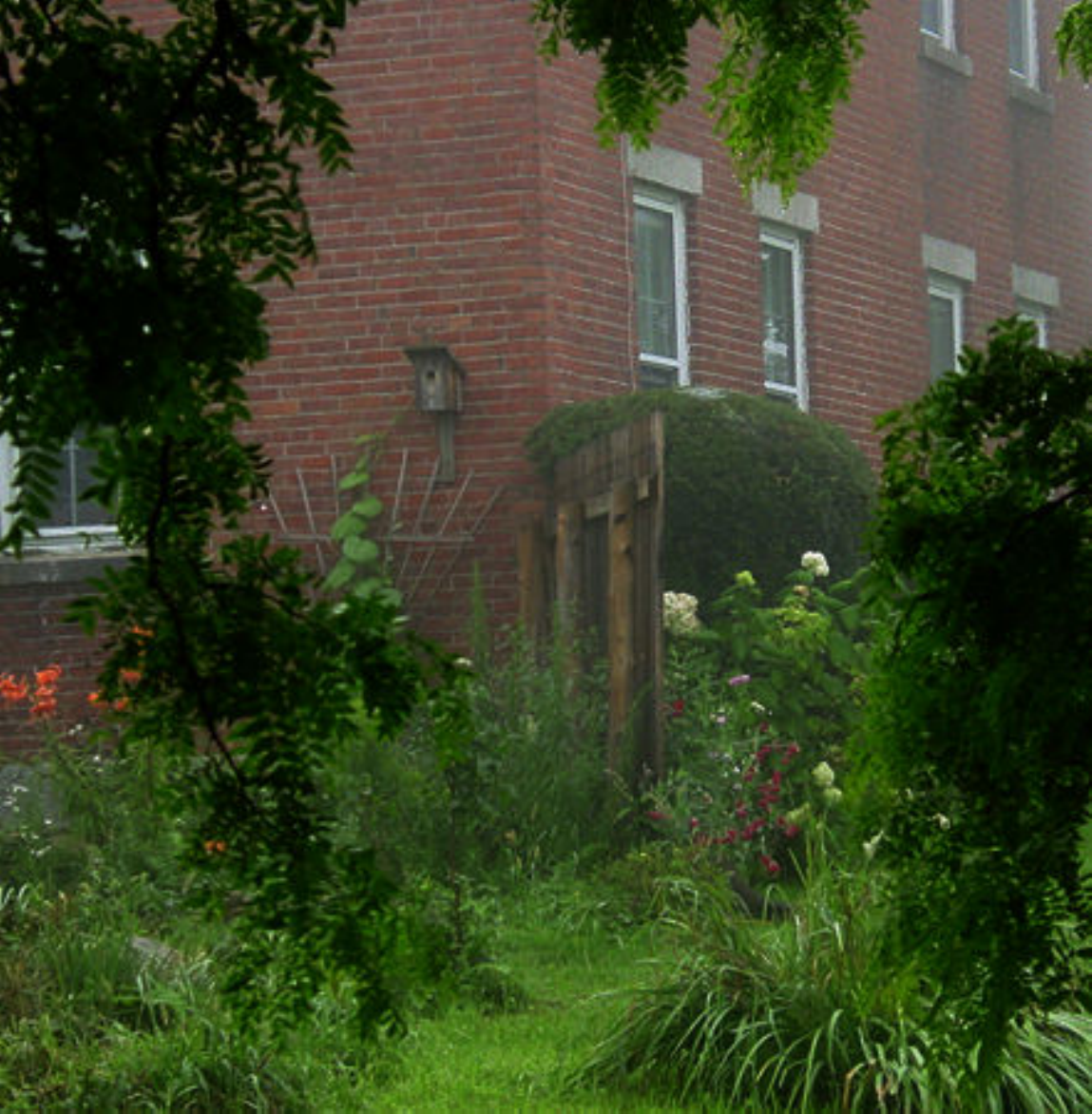} \\
			(a) Hazy input & \hspace{-0.4cm}
			(b) WB of (a) & \hspace{-0.4cm}
			(c) CE of (a) & \hspace{-0.4cm}
			(d) GC of (a) \\
			\includegraphics[width = 0.11\textwidth]{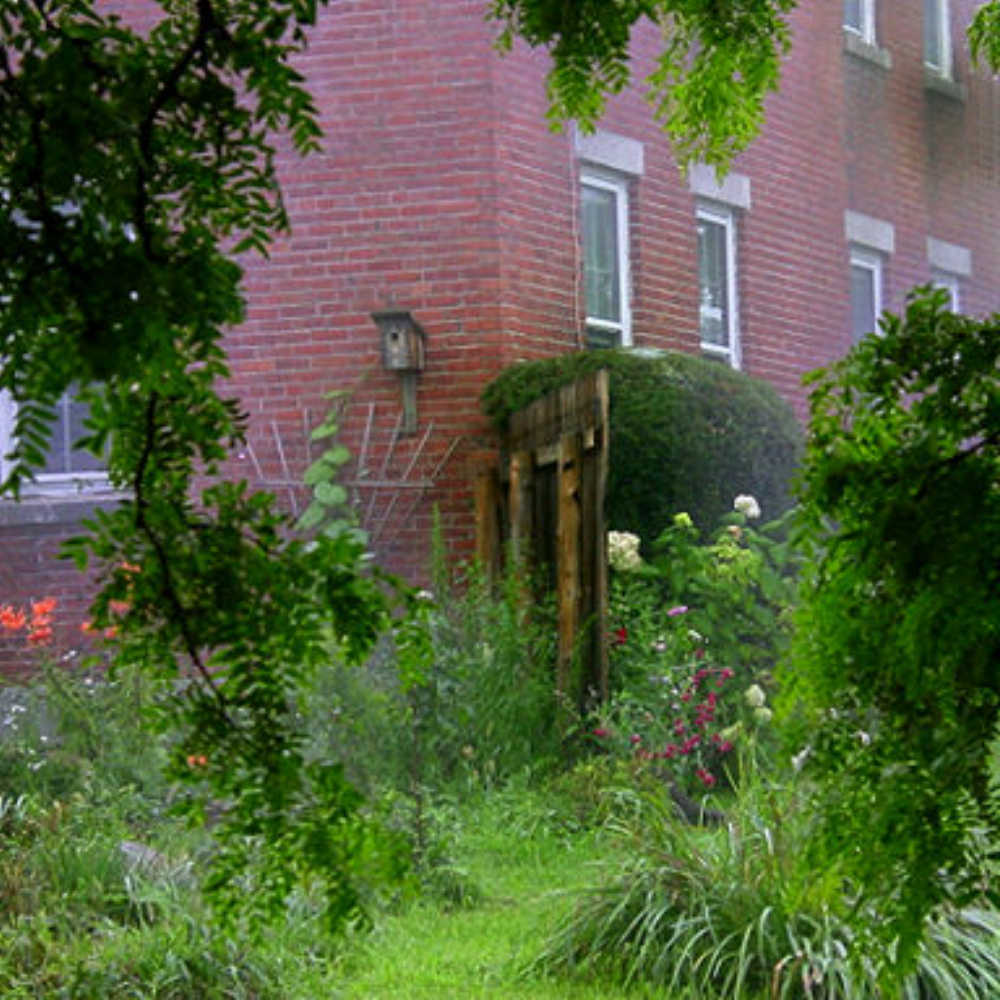} & \hspace{-0.4cm}
			\includegraphics[width = 0.11\textwidth]{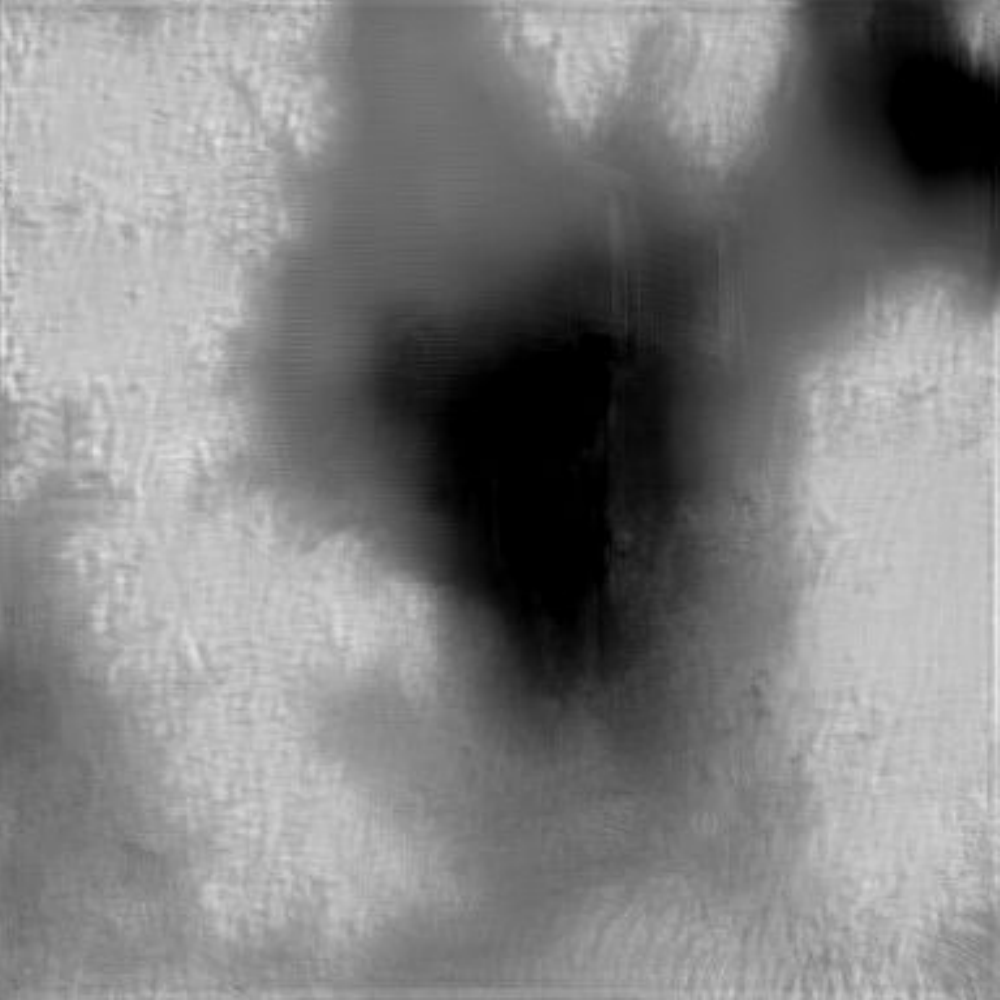} & \hspace{-0.4cm}
			\includegraphics[width = 0.11\textwidth]{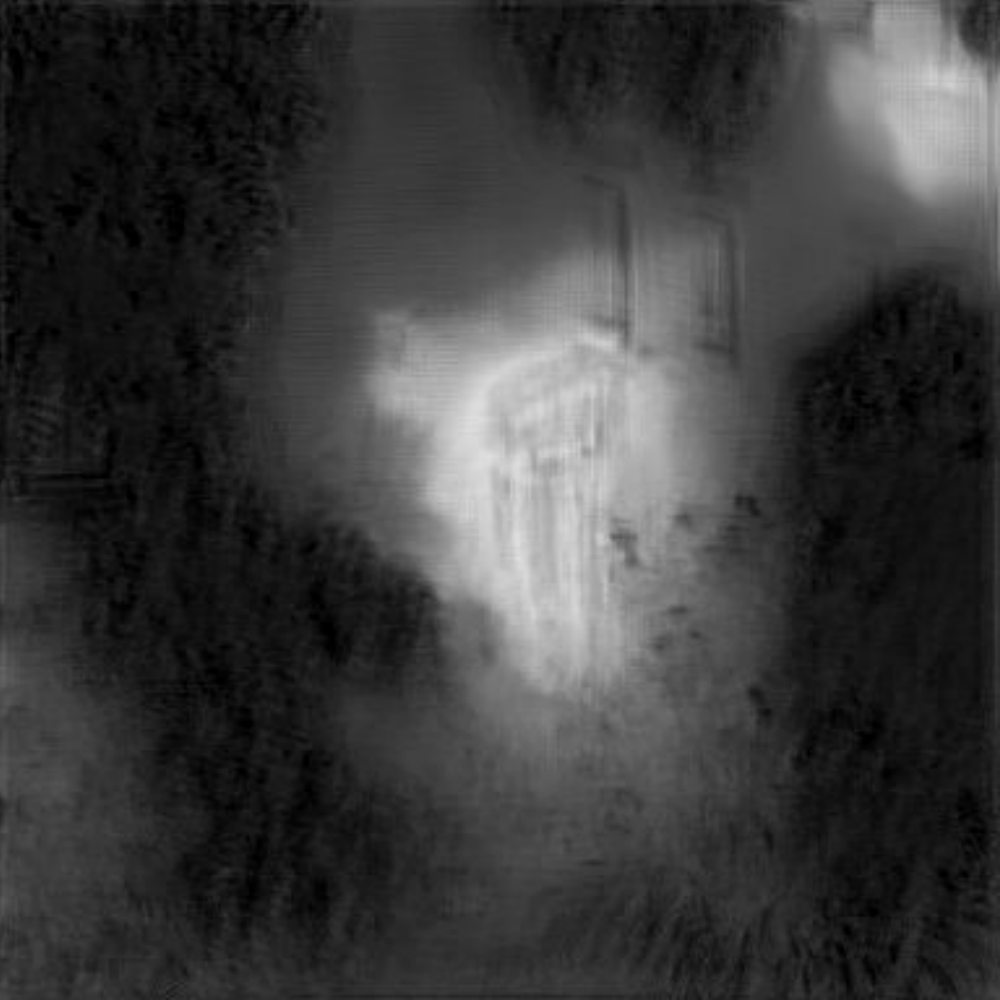} & \hspace{-0.4cm}
			\includegraphics[width = 0.11\textwidth]{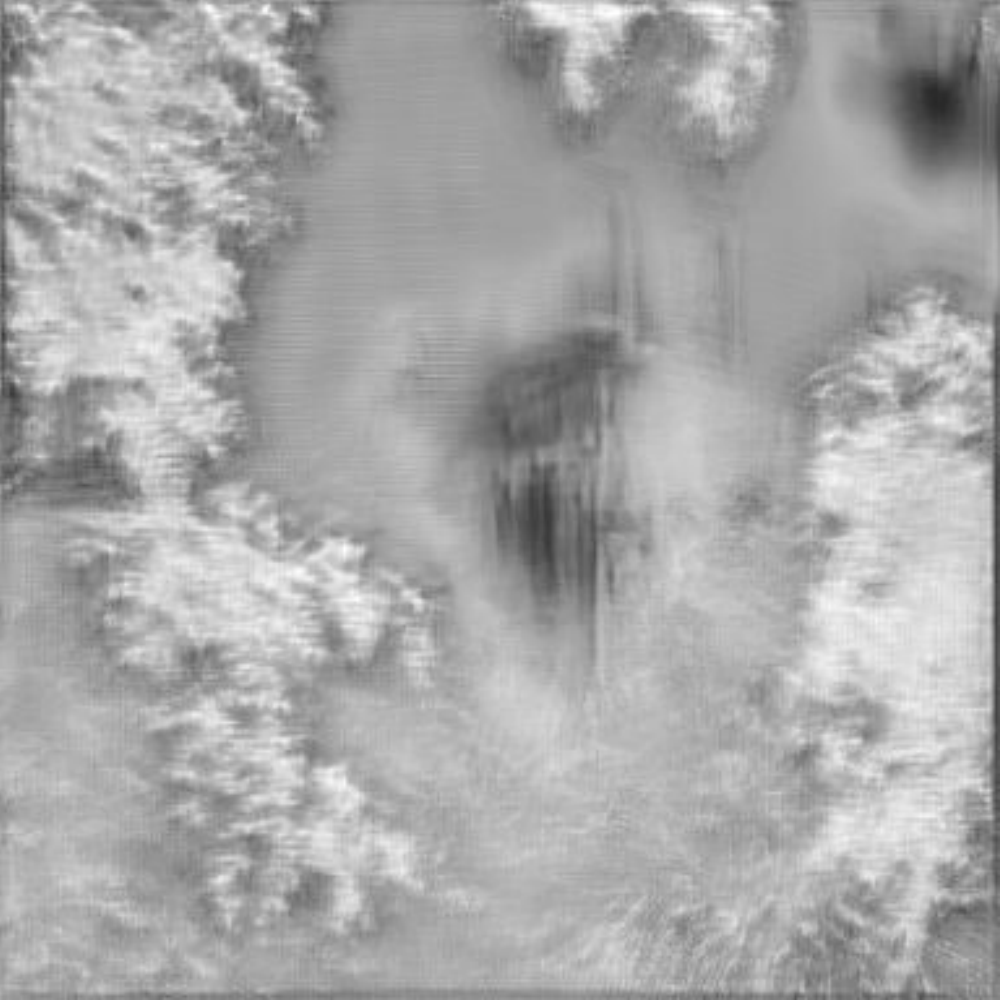} \\
			(e) Our result & \hspace{-0.4cm}
			(f) Weight of (b) & \hspace{-0.4cm}
			(g) Weight of (c) & \hspace{-0.4cm}
			(h) Weight of (d)
		\end{tabular}
	\end{center}
	\vspace{-0.2cm}
	\caption{Image dehazing result. We exploit a gated fusion network for single image deblurring. (a) Hazy input. (b)-(d) are the derived inputs. (f)-(h) are learned confidence maps for (b), (c) and (d), respectively. (e) Our result.
	}
	\vspace{-0.5cm}
	\label{fig-comp}
\end{figure}

Estimating transmission from a hazy image is a severely ill-posed problem. Some approaches try to use visual cues to capture deterministic and statistical properties of hazy images~\cite{berman2016non,fattal2008single,hautiere2007towards,schechner2001instant}.
However, these transmission approximations are inaccurate, especially in the cases of the scenes where the colors of objects are inherently similar to those of
atmospheric lights. Note that such an erroneous transmission estimation directly affects the quality of the recovered image, resulting in undesired haze artifacts.
%
%
Instead of using hand-crafted visual cues, CNN-based methods~\cite{cai2016dehazenet,ren2016single} are proposed to estimate the transmissions. 
However, these methods still follow the conventional dehazing methods in estimating atmospheric lights to recover clear images. 
%
%
Thus, if the transmissions are not estimated well, they will interfere the following atmospheric light estimation and thereby lead to low-quality results.
%
%

To address the above issues, we propose a novel end-to-end trainable neural network that does not explicitly estimate the transmission and atmospheric light. Thus, the artifacts arising from transmission estimation errors can be avoided in the final restored results.
The proposed neural network is built on a fusion strategy which aims to seamlessly blend several input images by preserving only the specific features of the composite output image.

There are two major factors in hazy images that need to be dealt with. The first one is the color cast introduced by the atmospheric light. The second one is the lack of visibility due to attenuation.
Therefore, \textit{we tackle these two problems by deriving three inputs from the original image with the aim of recovering the visibility of the scene in at least one of them.}
The first input ensures a natural rendition (Figure \ref{fig-comp}(b)) of the output by eliminating chromatic casts caused by the atmospheric light.
The second contrast enhanced input yields a better global visibility, but mainly in the thick hazy regions (\eg, the rear wall in Figure \ref{fig-comp}(c)). However, the contrast enhanced images are too dark in the light hazy regions. Hence, to recover the light hazy regions, we find that the gamma corrected images restore information of the light hazy regions well (\eg, the front lawn in Figure~\ref{fig-comp}(d)).
Consequently, the three derived inputs are gated by three confidence maps (Figure~\ref{fig-comp}(f)-(g)), which aim to preserve the regions with good visibility.
%
%

%
%
%

The contributions of this work are three-fold.
First, we propose a deep end-to-end trainable neutral network that restores clear images without assuming any restrictions on scene transmission and atmospheric light.
Second, we demonstrate the utility and effectiveness of a gated fusion network for single image dehazing by leveraging the derived inputs from an original hazy image.
Finally, we train the proposed model with a multi-scale approach to eliminate the halo artifacts that hurt image dehazing.
We show that the proposed dehazing model performs favorably against the state-of-the-arts.
%

\vspace{-0.2cm}
\section{Related Work}
\vspace{-0.1cm}
There mainly exist three kinds of methods for image dehazing: multi-image based methods, hand-crafted priors based methods,
and data-driven methods.

\vspace{-3mm}
{\flushleft \textbf{Multi-image aggregation.}}
Early methods often require
multiple images to deal with the dehazing problem~\cite{narasimhan2003contrast,kopf2008deep,treibitz2009polarization}.
Kopf~\etal~\cite{kopf2008deep} used an approximated 3D model of the scene for dehazing.
Different polarized filters were used in~\cite{treibitz2009polarization} to capture multiple images of the same scene, and then degrees of polarization were used for haze removal.
Narasimhan and Nayar~\cite{narasimhan2003contrast}
also used the differences between multiple images for estimating the haze properties.

All these methods make
the same assumption of using multiple images in the same
scene. However, there only exists one image
for a specific scene in most cases.

\vspace{-2mm}
{\flushleft \textbf{Hand-crafted priors based methods.}}
Different image priors have been explored for single image dehazing in previous methods \cite{li2015nighttime}.
Tan~\etal~\cite{tan2008visibility} enhanced the visibility of hazy images by maximizing the contrast. The dehazed results
of this method often present color distortions since this
method is not physically valid.
%
%
He~\etal~\cite{he2011singlecvpr} presented a dark channel prior (DCP) for outdoor images, which asserts that the local minimum of the dark channel of a haze-free image is close to zero.
The DCP has been shown effective for
image dehazing, and a number of methods improve \cite{he2011singlecvpr} in terms of efficiency~\cite{tarel2009fast} or quality~\cite{nishino2012bayesian}.
Fattal~\cite{fattal2014dehazing} discovered that pixels of image patches typically exhibit a one-dimensional distribution, and used it to recover the scene transmission. However, this approach cannot guarantee a correct classification of patches.
Recently, Berman~\etal~\cite{berman2016non} observed that colors of a haze-free image can be well approximated by a few hundred distinct colors, and then proposed a dehazing algorithm based on this prior.

Another line of research tries to make use of a fusion principle to restore hazy images in \cite{ancuti2013single,choi2015referenceless}. However, these methods need complex blending based on luminance, chromatic and saliency maps.
In contrast, we introduce a gated fusion based single image dehazing technique that blends only the derived three input images.

All of the above approaches strongly rely on the accuracy of the assumed image priors, so may perform poorly when the
assumed priors are insufficient to describe
real-world images. As a result, these
approaches tend to introduce undesirable artifacts such
as color distortions.

\vspace{-2mm}
{\flushleft \textbf{Data-driven methods.}}
Tang~\etal~\cite{tang2014investigating} combined four types of haze-relevant features with Random Forest
to estimate the transmission.
Zhu~\etal~\cite{zhu2015fast} created a linear model for
modeling the scene depth of the hazy image under a color attenuation prior, and learned the parameters of the model in a supervised manner.
However, these methods are still developed based on hand-crafted features.
\begin{figure*}[t]\footnotesize
	\begin{center}
		\includegraphics[width = .8\textwidth]{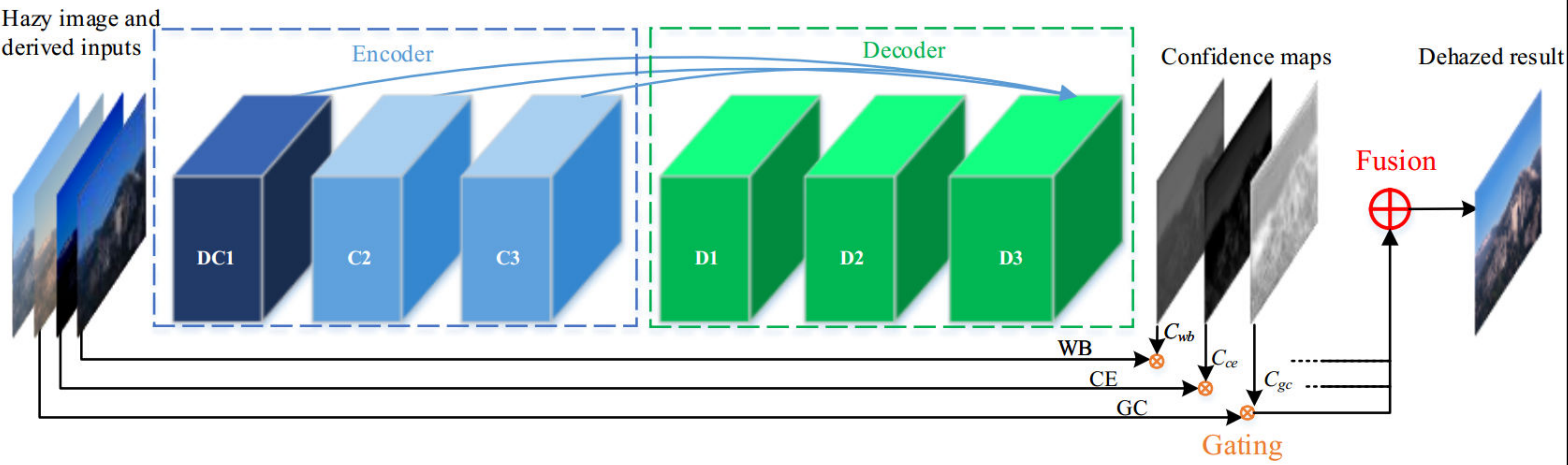}
	\end{center}
	\vspace{-0.3cm}
	\caption{The coarsest level network of GFN.
		The network contains layers of symmetric
		encoder and decoder.
		To retrieve more contextual information, we use Dilation Convolution (DC) to enlarge the receptive field in the convolutional layers in the encoder block.
		Skip shortcuts are connected from the convolutional feature maps to the deconvolutional feature maps.
		Three enhanced versions are derived from the input hazy image. Then, these three
		inputs are weighted by the three confidence maps learned by our network, respectively.
	}
	\vspace{-0.3cm}
	\label{fig-net}
\end{figure*}

Recently,
%
CNNs have also been used for image recovering problems~\cite{cai2016dehazenet,li2017aod,xu2017learning,He_derain_2018,zhang2017image,zhang2016learning}.
Cai~\etal~\cite{cai2016dehazenet} proposed a DehazeNet and a BReLU layer to estimate the transmissions from hazy inputs.
In~\cite{ren2016single}, a coarse-scale network was first used to learn the mapping between hazy inputs and their transmissions, and then a fine-scale network was exploited to refine the transmission.
One problem of these CNNs based methods~\cite{cai2016dehazenet,ren2017deep,ren2016single} is that all these methods require an accurate transmission and atmospheric light estimation step for restoring the clear image.
Although the recent AOD-Net~\cite{li2017aod} bypasses the estimation step, this method still needs to compute a newly introduced variable $\mathbf{K}(x)$ which integrates both transmission $t(x)$ and atmospheric light $\mathbf{A}$. Therefore, AOD-Net still falls into a physical model in \eqref{eq-math_model}.

Different from these CNNs based approaches,
our proposed network is built on the principle of image fusion,
and is learned to produce the sharp image
directly without estimating transmission and atmospheric light.
The main idea of image fusion is to combine several images
into a single one, retaining only the most significant features.
This idea has been successfully used in a number of applications such as image editing~\cite{perez2003poisson} and video super-resolution~\cite{liu2017robust}.
%

%

\vspace{-1mm}
\section{Gated Fusion Network}
This section presents the details of our gated fusion network
that employs an original hazy image and three derived images as inputs.
We refer to this network as \textit{Gated Fusion Network}, or \textit{GFN}, as shown in Figure~\ref{fig-net}.
The central idea is to learn the \textit{confidence maps} to
combine several input images into a
single one by keeping only the most significant features
of them.
Obviously, the choice of inputs and weights is
application-dependent. By learning the confidence map
for each input, we demonstrate that our fusion based method is
able to dehaze images effectively.

\subsection{Derived Inputs}
\label{sec-inputs}
We derive several inputs based on the following observations.
The first one is that the colors in hazy images often change due to the influence of the atmospheric light.
The second one is the lack of visibility in distant regions
due to scattering and attenuation phenomena.
Based on these observations, we generate three inputs that recover the
color and visibility of the entire image from the original hazy image.
We first estimate the White Balanced (WB) image $\mathbf{I}_{wb}$ of the hazy input $\mathbf{I}$
to recover the latent color of the scene.
Then we extract visible information including the
Contrast Enhanced (CE) image $\mathbf{I}_{ce}$ and the Gamma Corrected (GC) image $\mathbf{I}_{gc}$ to
yield a better global visibility.

\begin{figure}[t]\footnotesize
	\begin{center}
		\begin{tabular}{@{}cccc@{}}
			\includegraphics[width = 0.11\textwidth]{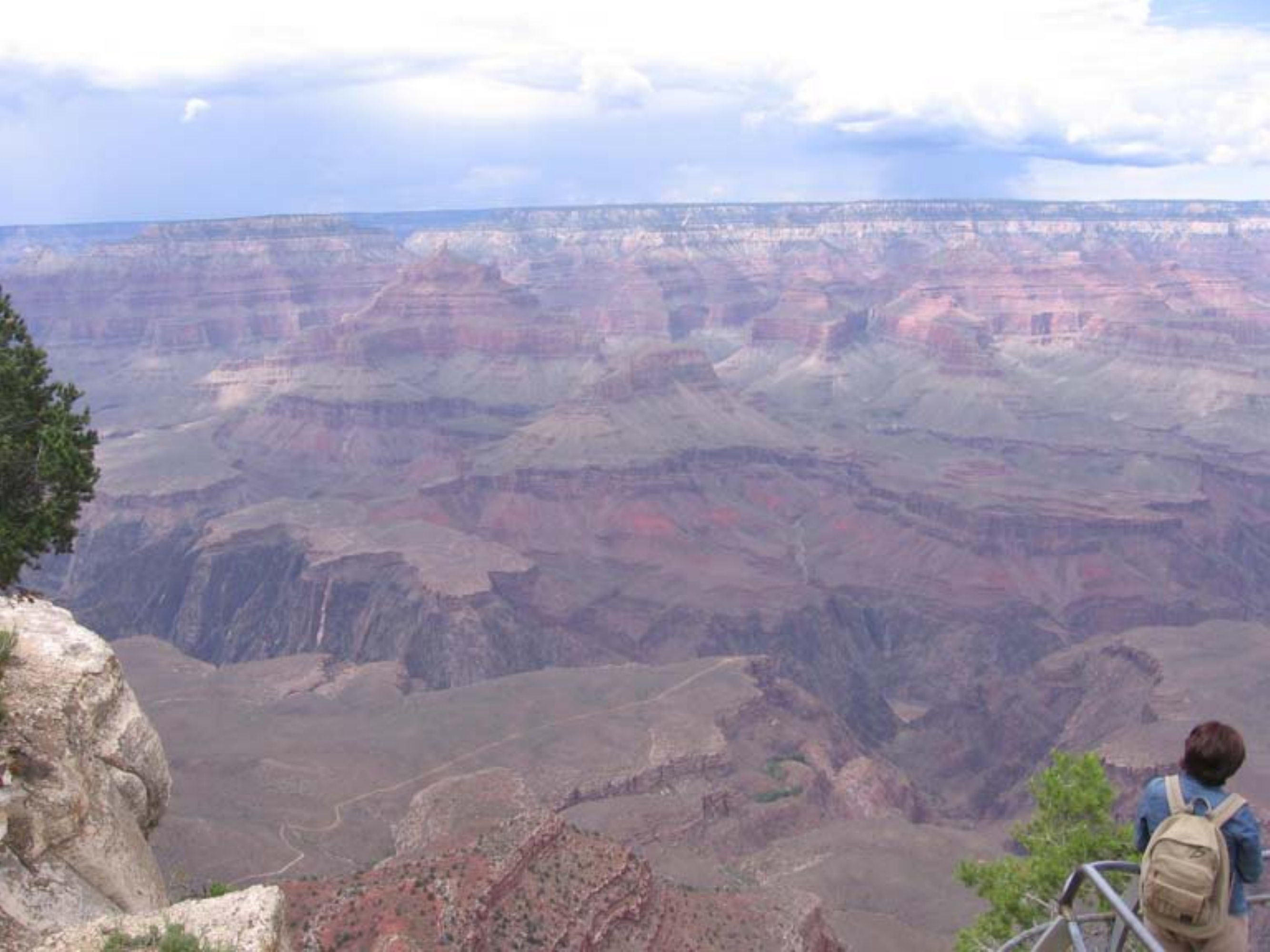} & \hspace{-0.4cm}
			\includegraphics[width = 0.11\textwidth]{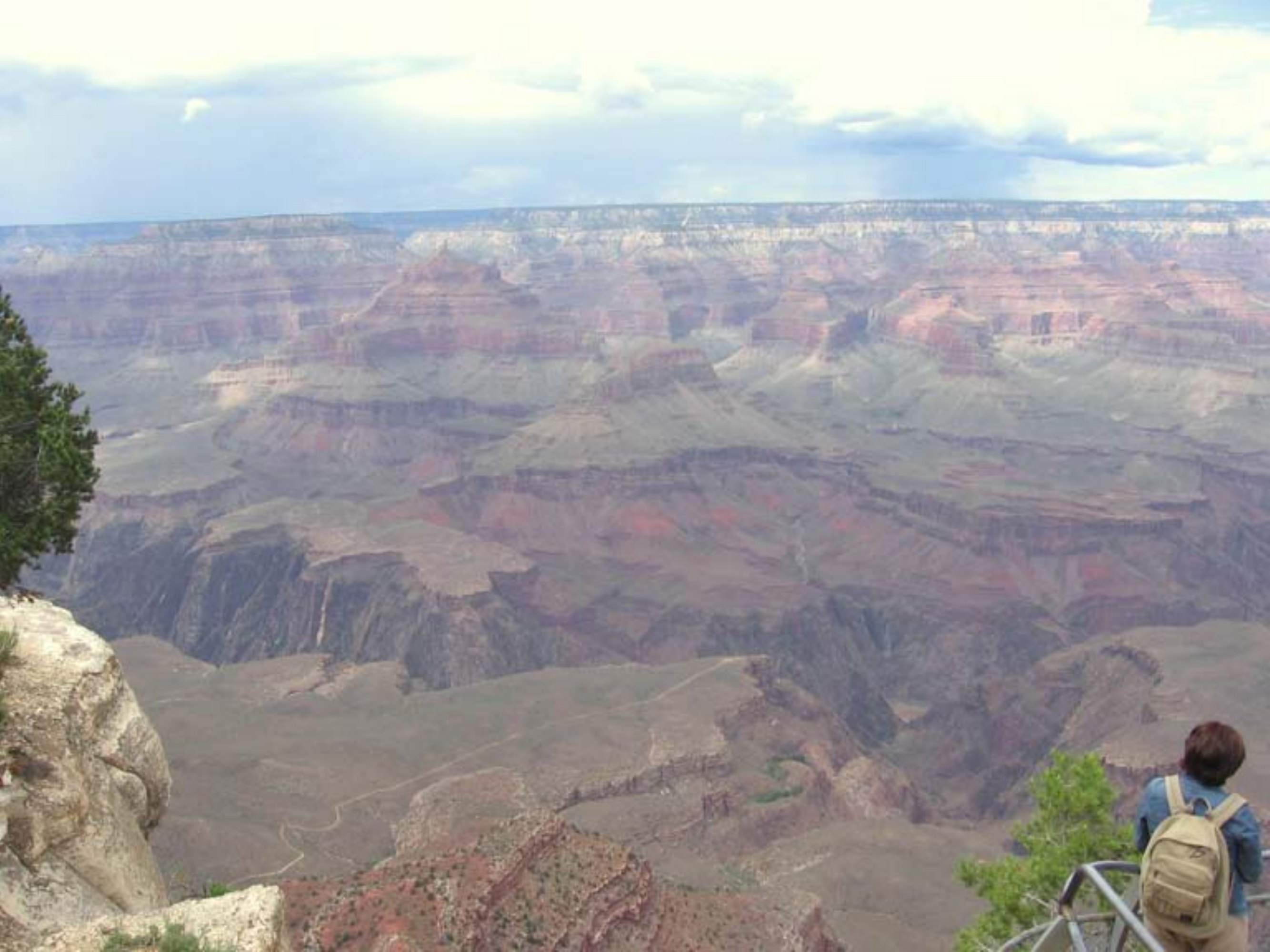} & \hspace{-0.4cm}
			\includegraphics[width = 0.11\textwidth]{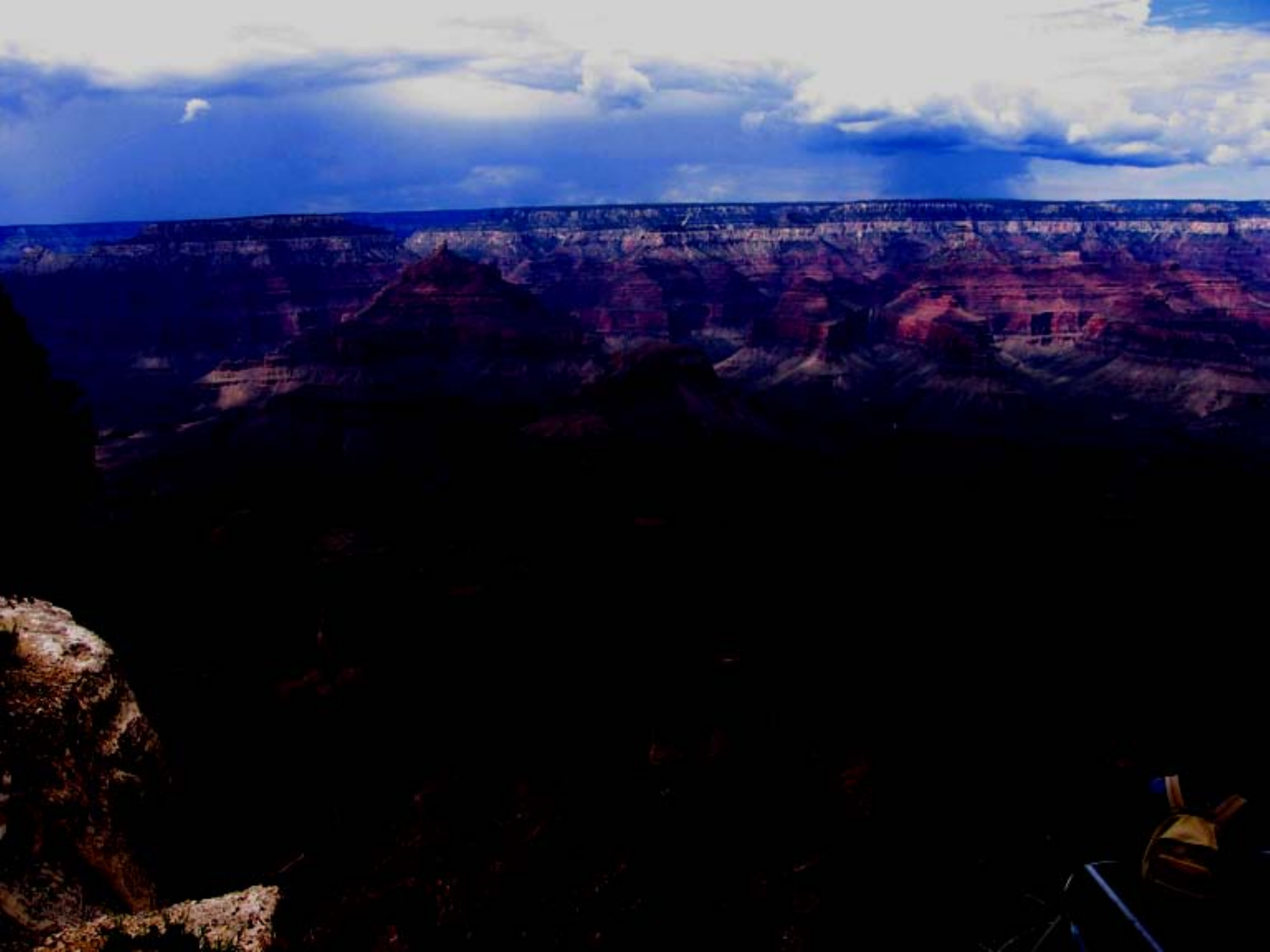} & \hspace{-0.4cm}
			\includegraphics[width = 0.11\textwidth]{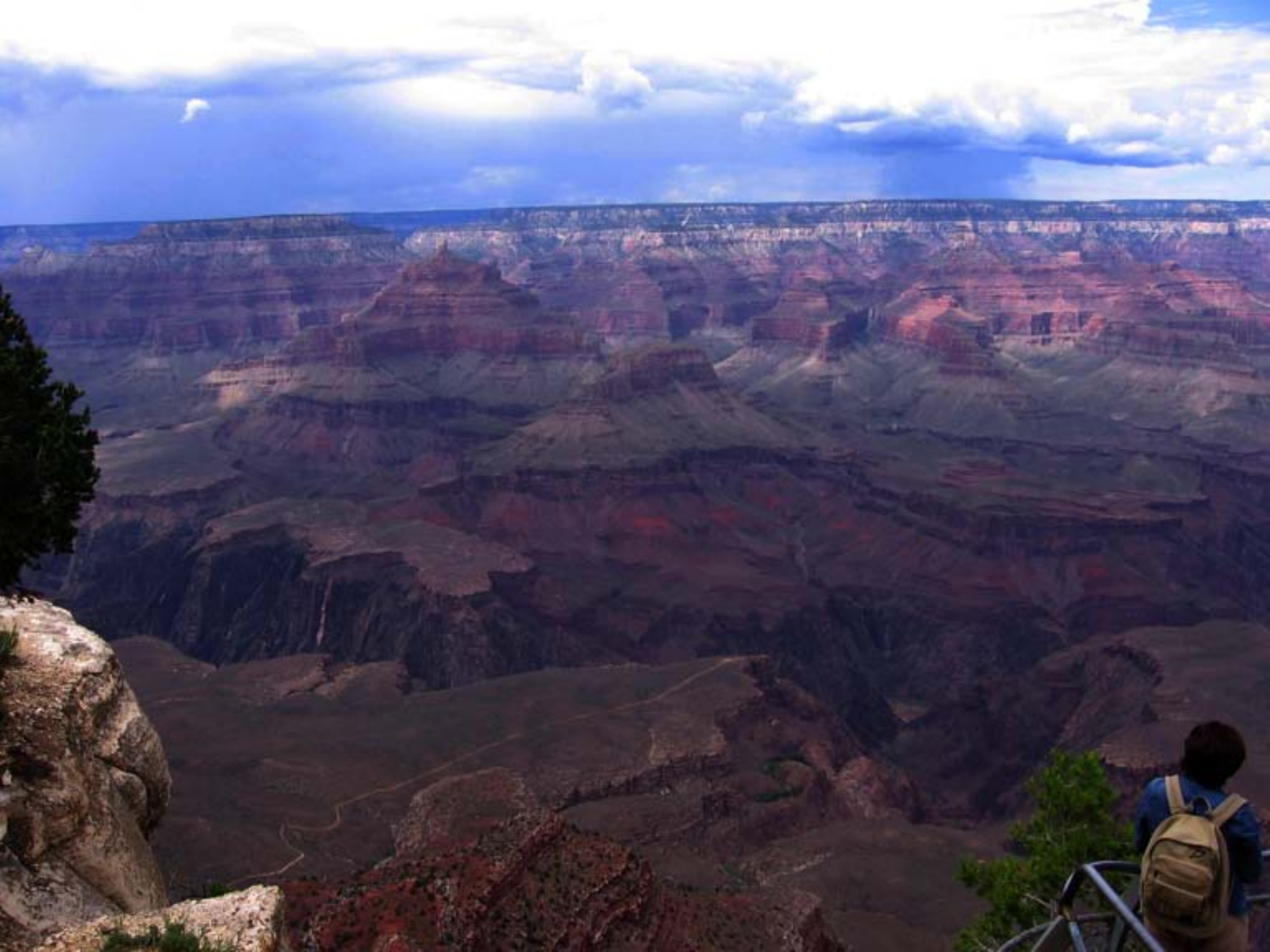} \\
			\includegraphics[width = 0.11\textwidth]{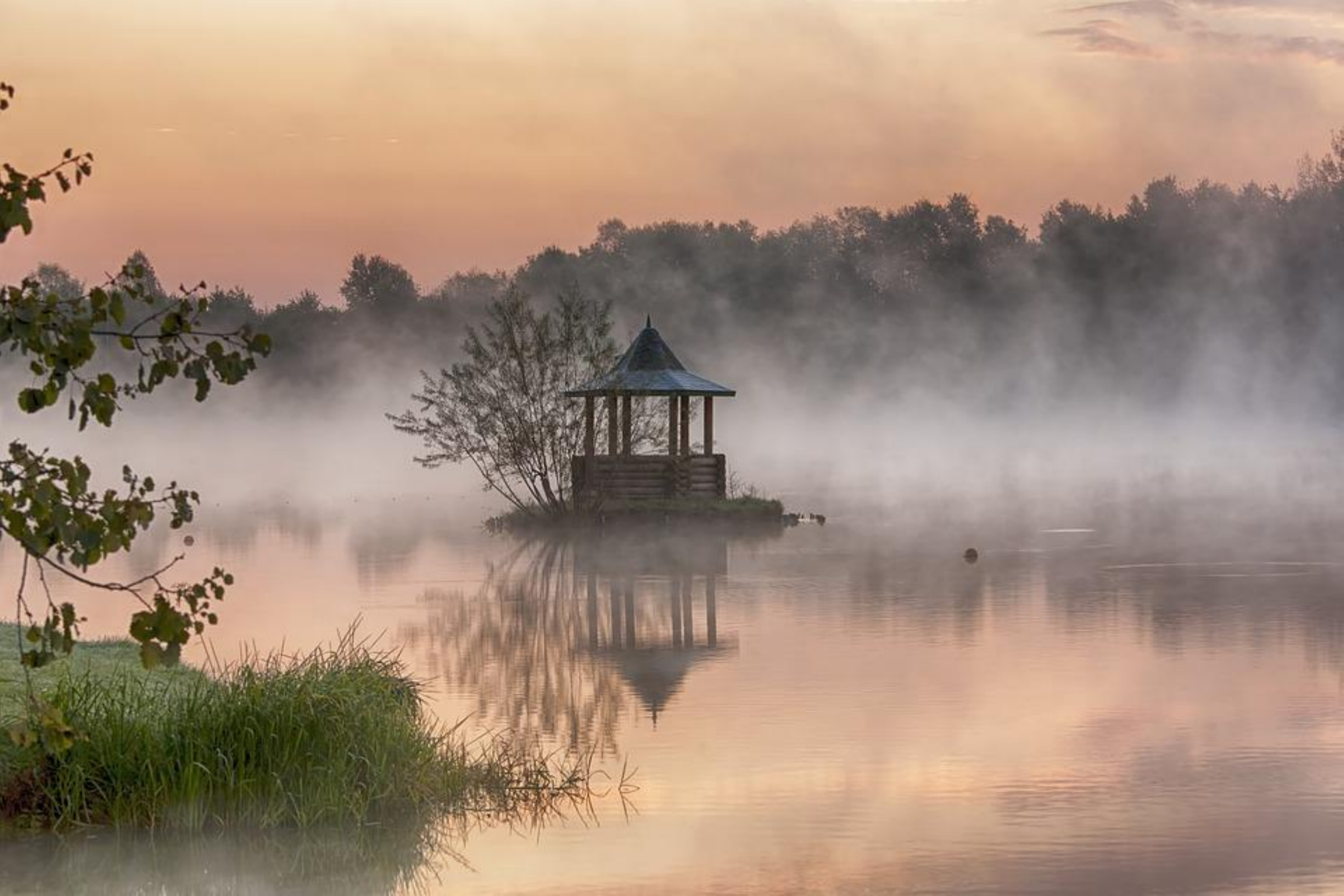} & \hspace{-0.4cm}
			\includegraphics[width = 0.11\textwidth]{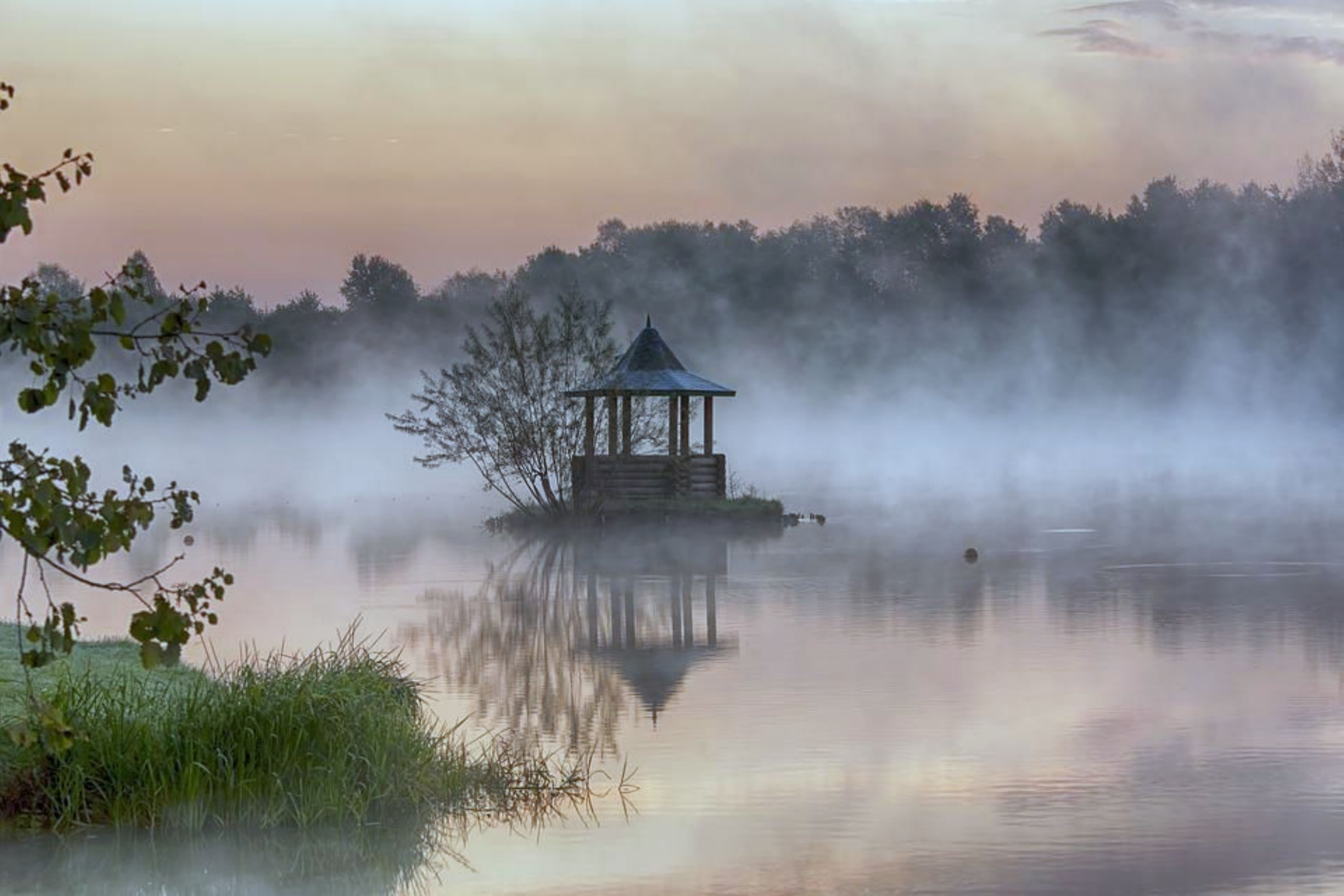} & \hspace{-0.4cm}
			\includegraphics[width = 0.11\textwidth]{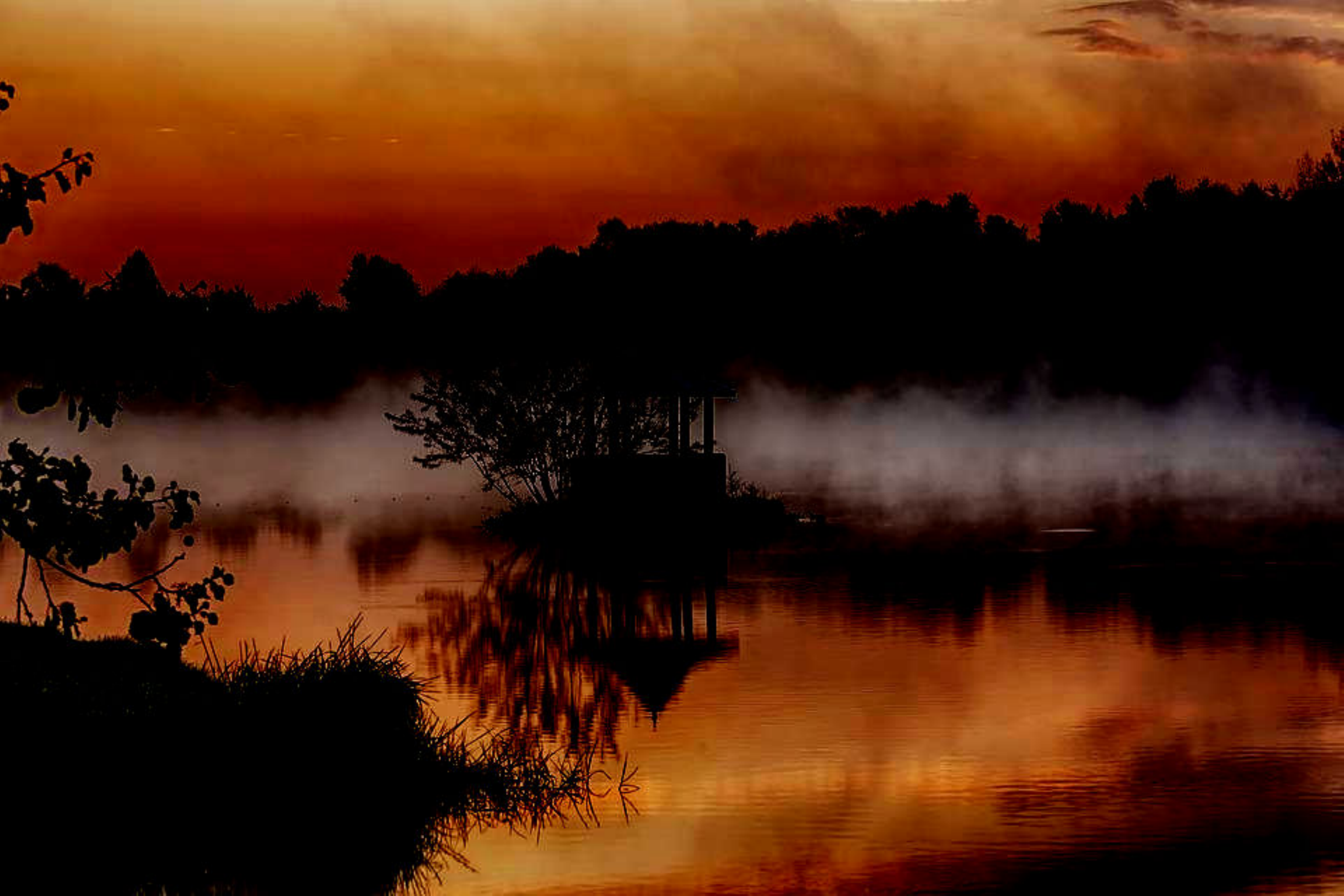} & \hspace{-0.4cm}
			\includegraphics[width = 0.11\textwidth]{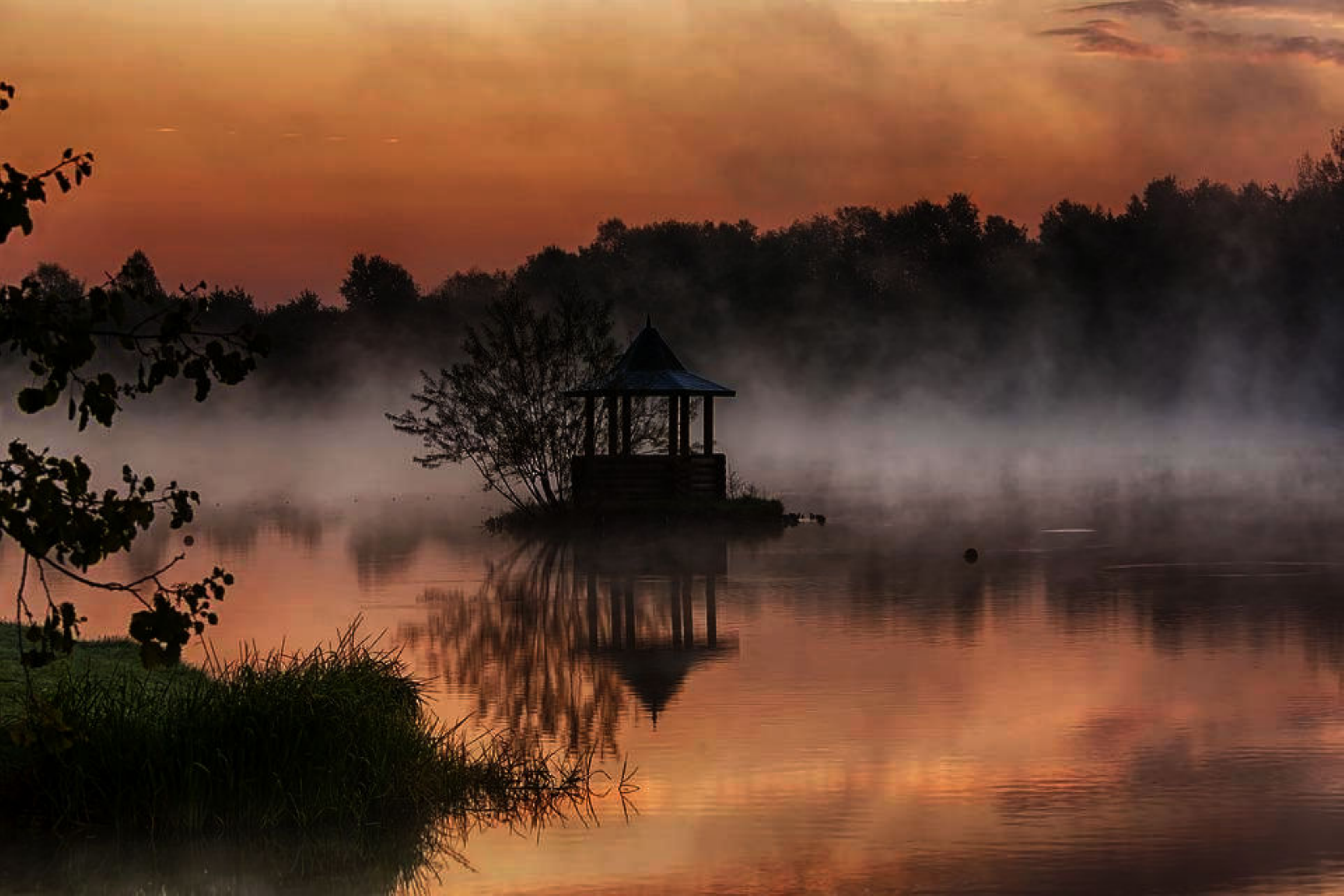} \\
			\includegraphics[width = 0.11\textwidth]{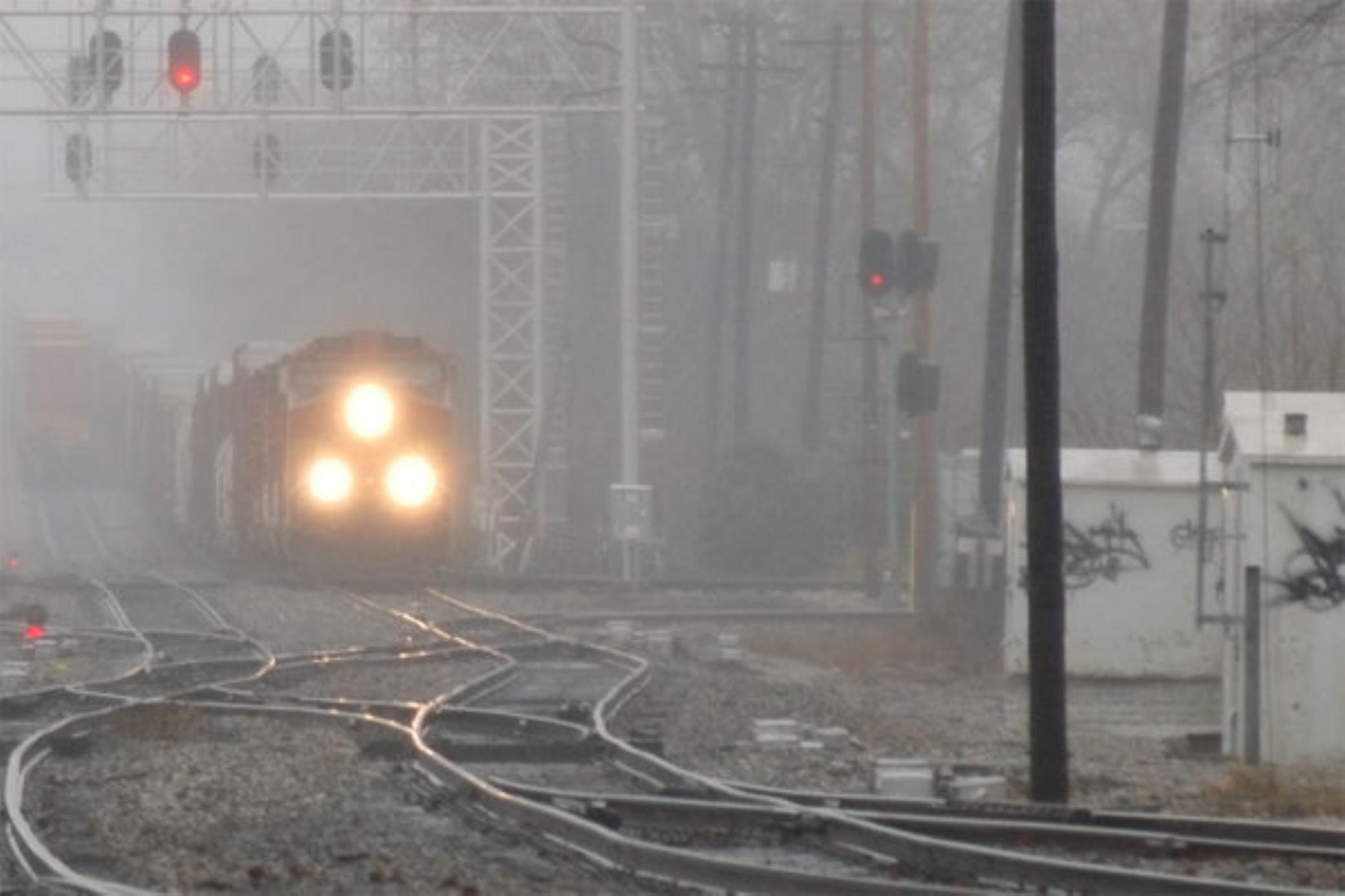} & \hspace{-0.4cm}
			\includegraphics[width = 0.11\textwidth]{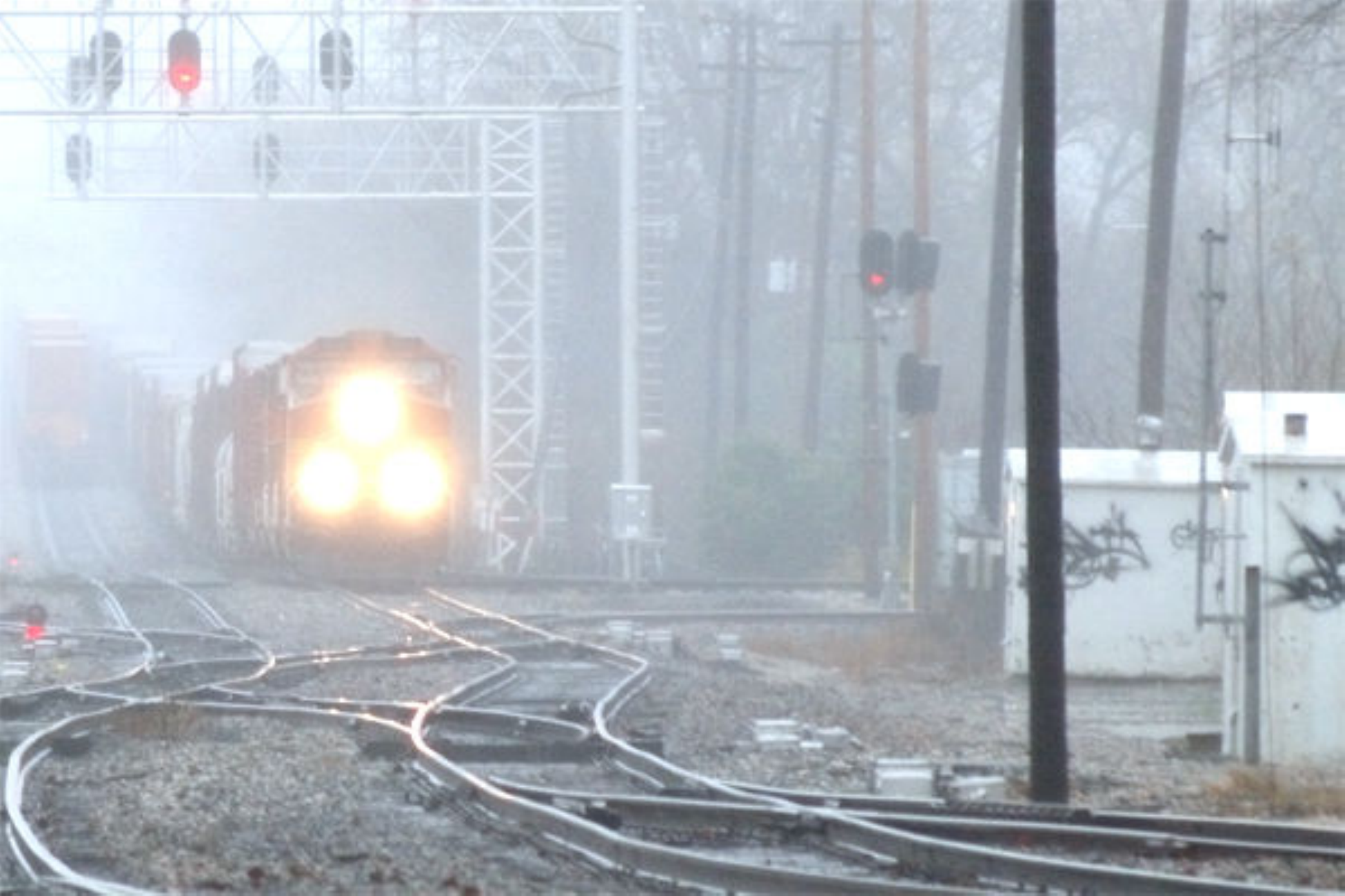} & \hspace{-0.4cm}
			\includegraphics[width = 0.11\textwidth]{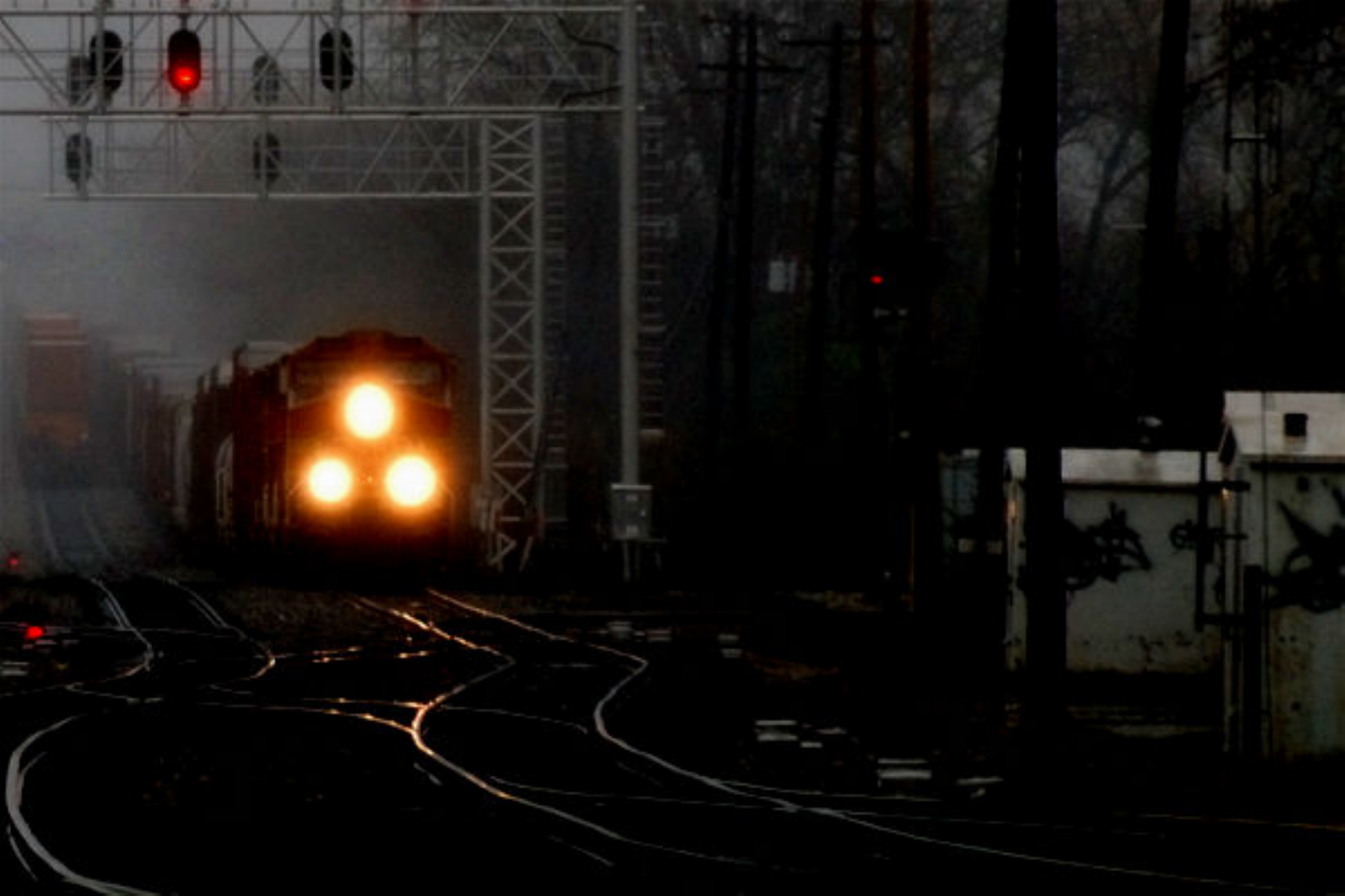} & \hspace{-0.4cm}
			\includegraphics[width = 0.11\textwidth]{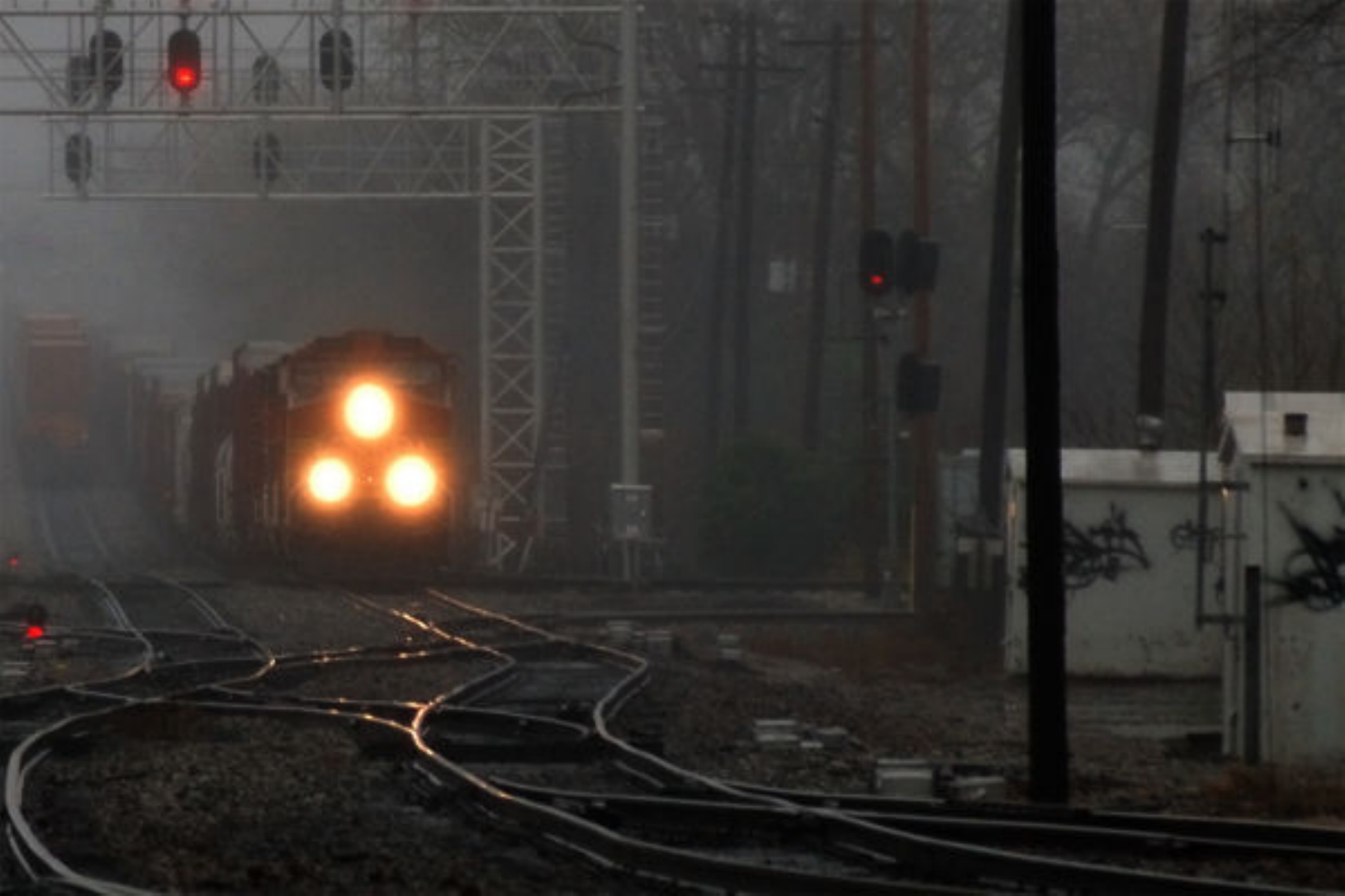} \\
			\includegraphics[width = 0.11\textwidth]{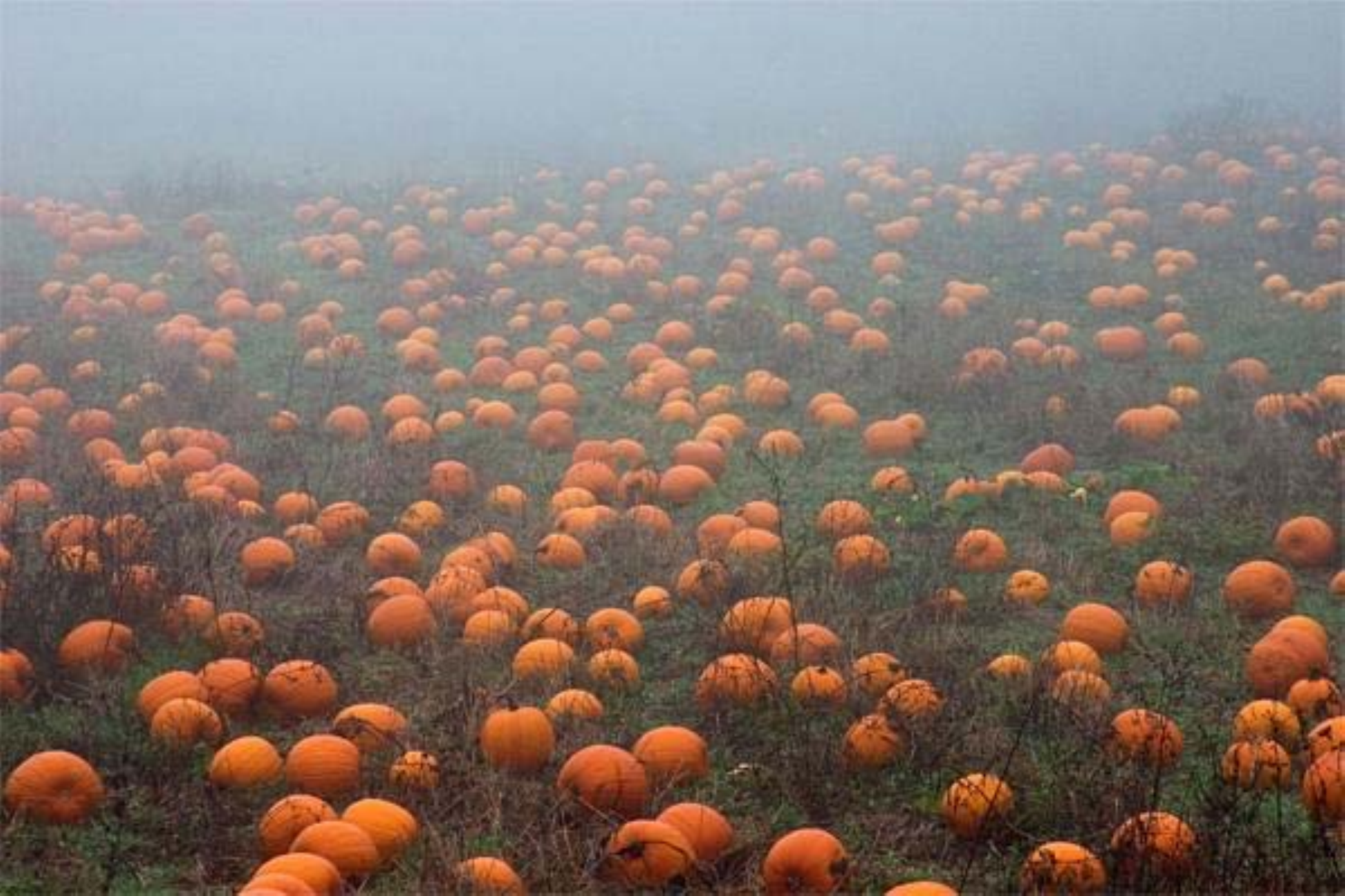} & \hspace{-0.4cm}
			\includegraphics[width = 0.11\textwidth]{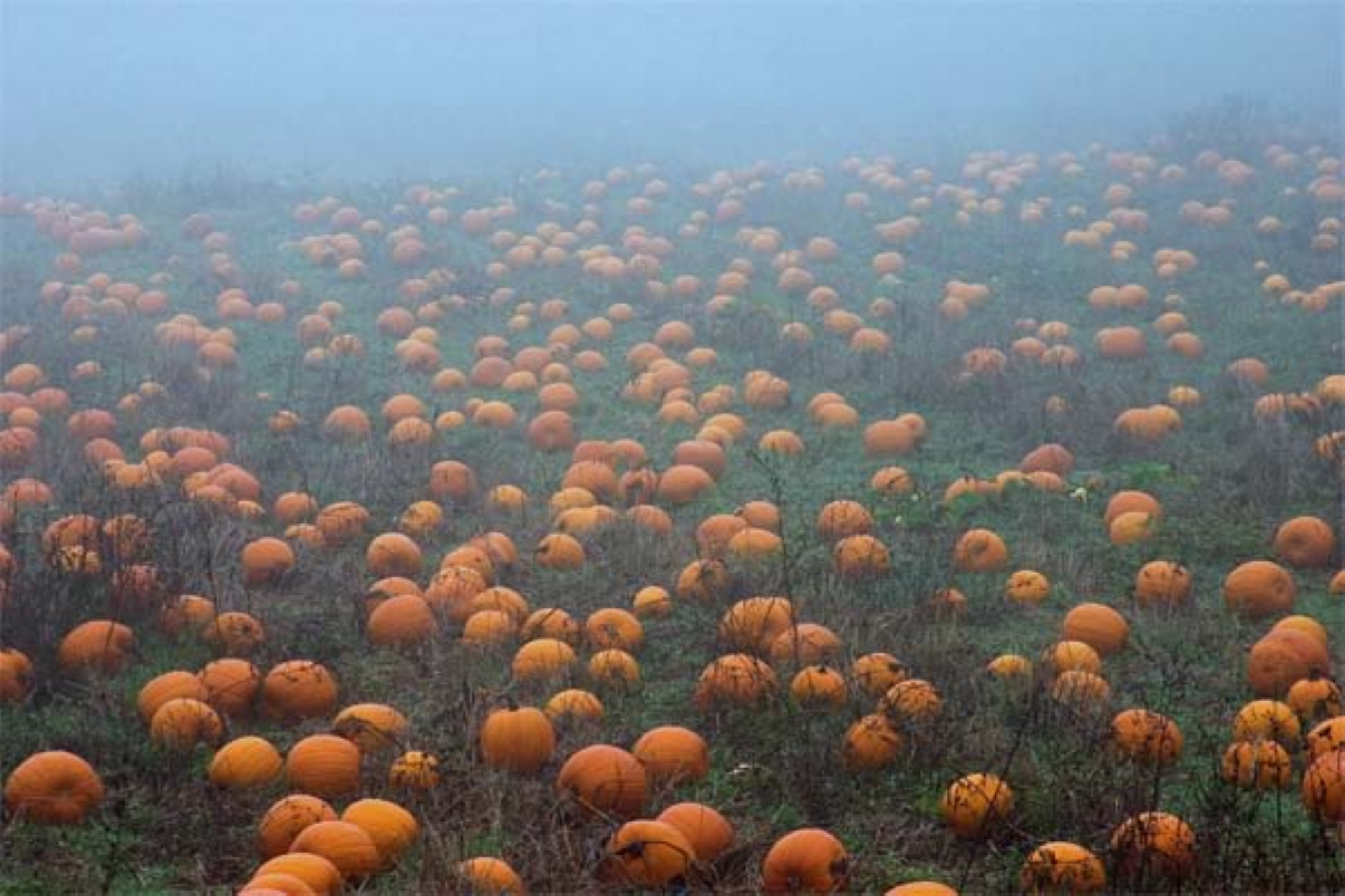} & \hspace{-0.4cm}
			\includegraphics[width = 0.11\textwidth]{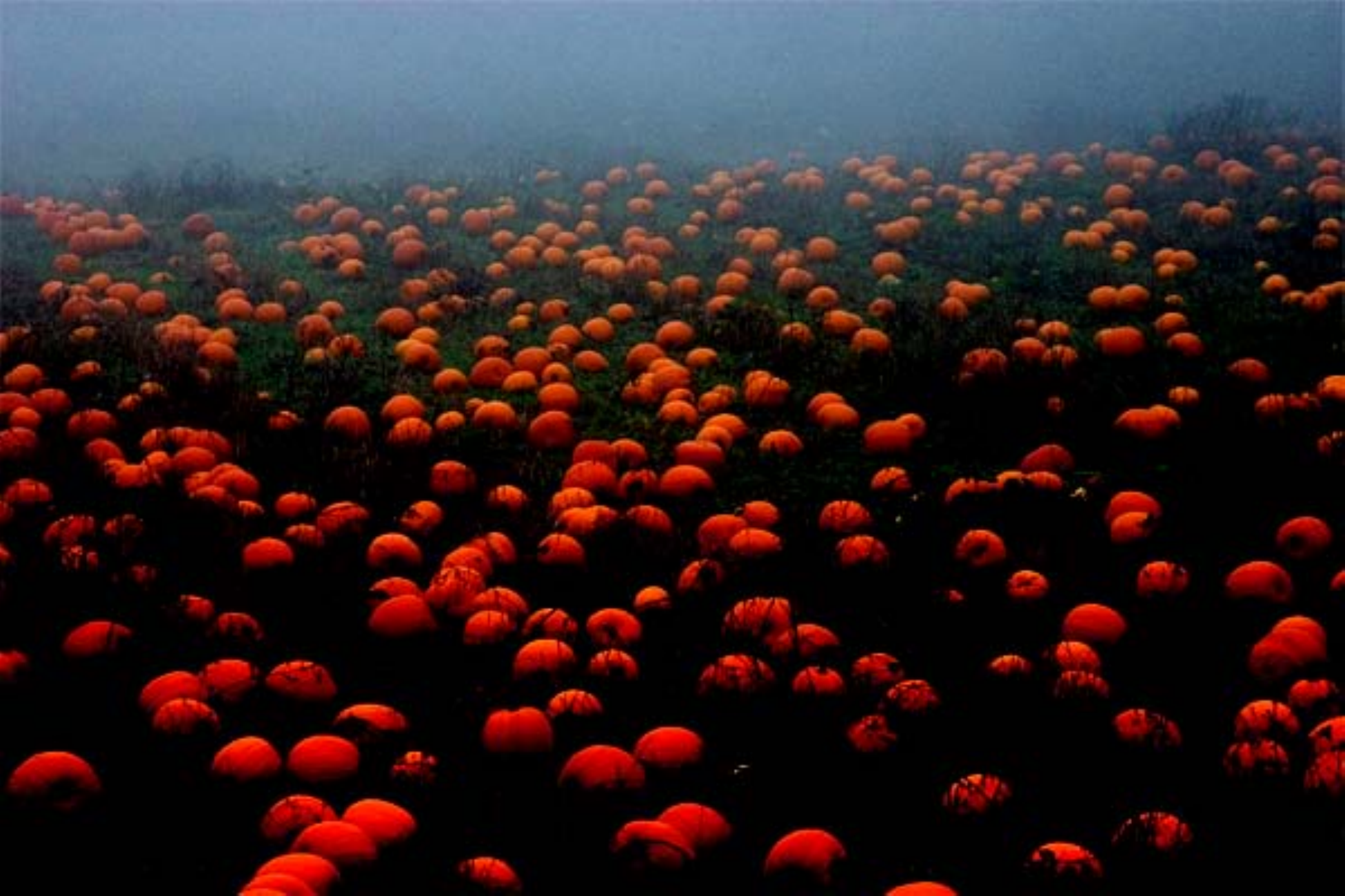} & \hspace{-0.4cm}
			\includegraphics[width = 0.11\textwidth]{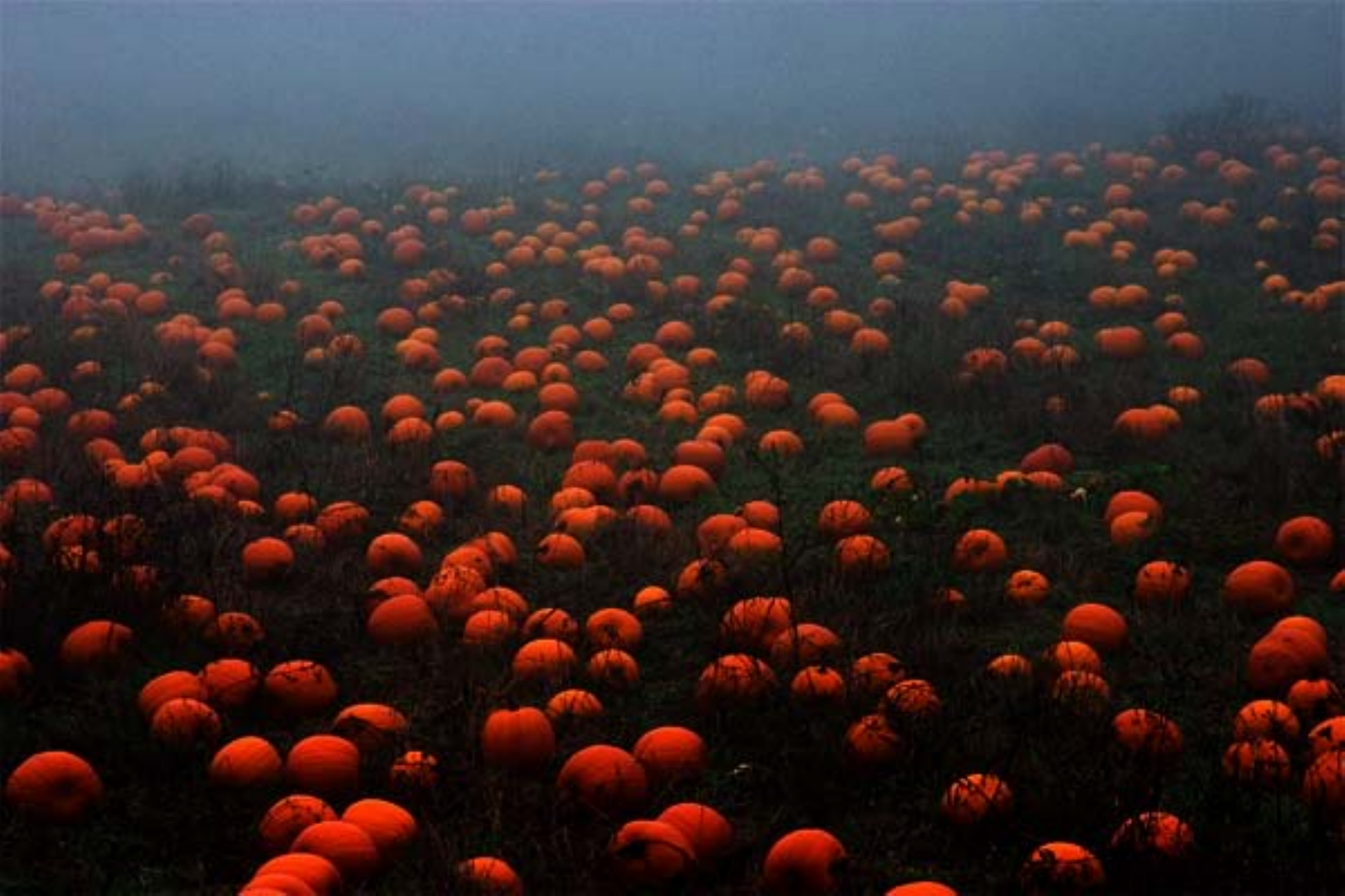} \\
			(a) Hazy inputs & \hspace{-0.4cm}
			(b) WB & \hspace{-0.4cm}
			(c) CE & \hspace{-0.4cm}
			(d) GC
		\end{tabular}
	\end{center}
	\vspace{-0.1cm}
	\caption{We derive three enhanced versions from an input hazy image.
		These three derived inputs contain different important visual cues of the input hazy image.
	}
	\vspace{-0.3cm}
	\label{fig-inputs}
\end{figure}

\vspace{-2mm}
{\flushleft \textbf{White balanced input.}}
Our first input is a white balanced image which aims to eliminate chromatic casts
caused by the atmospheric color.
In the past decades, a number of white balancing approaches~\cite{kawakami2013camera}
have been proposed.
In this paper, we use the gray world assumption~\cite{reinhard2001color} based technique.
Despite its simplicity, this low-level approach has shown to
yield comparable results to those of more complex white
balance methods~\cite{li2016haze}.
The gray world assumption is that given an image with a sufficient quantity of color variations, the average value of the Red, Green and Blue components of the image should average out to a common gray value.
This assumption is in generally valid in any given real-world scene since the variations in colors are random and independent. It would be safe to say that given a large number of samples, the average should tend to converge to the mean value, which is gray.
White balancing algorithms can make use of this gray world assumption by forcing images to have a uniform average gray value for the R, G, and B channels.
For example, if an image is shot under a hazy weather condition, the captured image will have an atmospheric light $\mathbf{A}$ cast over the entire image. The effect of this atmospheric light cast disturbs the gray world assumption of the original image.
By imposing the assumption on the captured image, we would be able to remove the atmospheric light cast and re-acquire the colors of our original scene.
Figure~\ref{fig-inputs}(b) demonstrates such an effect.

Although white balancing could discard the color shifting caused by the atmospheric light, the results still present low contrast. To enhance the contrast, we introduce the following two derived inputs.

\vspace{-2mm}
{\flushleft \textbf{Contrast enhanced input.}}
Inspired by the previous dehazing approaches~\cite{ancuti2013single} and \cite{choi2015referenceless},
our second input is a contrast
enhanced image of the original hazy input.
Ancuti and Ancuti~\cite{ancuti2013single} derived a contrast
enhanced image by subtracting the average luminance value $\tilde{I}$
of the entire image $\mathbf{I}$ from the hazy input, and then using a
factor $\mu$ to linearly increase the luminance in
the recovered hazy regions as follows:
\begin{equation}
\mathbf{I}_{ce} = \mu \big(\mathbf{I} - \tilde{I}\big),
\label{equ-contrast}
\end{equation}
where $\mu=2(0.5 + \tilde{I})$.
Although $\tilde{I}$ is a good indicator of image brightness,
there is a problem in this input, especially in
denser haze regions.
The main reason is that the negative values of $(\mathbf{I} - \tilde{I})$
may dominate the contrast enhanced input as $\tilde{I}$ increases.
As shown in Figure~\ref{fig-inputs}(c), the dark image regions tend to be black after contrast enhancing.

\vspace{-2mm}
{\flushleft \textbf{Gamma corrected input.}}
To overcome the dark limitation in $\mathbf{I}_{ce}$, we create another type of
contrast enhanced image using gamma correction:
\begin{equation}
\mathbf{I}_{gc} = \alpha \mathbf{I}^\gamma.
\label{equ-gamma}
\end{equation}
Gamma correction is a nonlinear operation which is used to encode ($\gamma<1$) and decode ($\gamma>1$) luminance or tristimulus values in image content,
In this paper, we use $\alpha=1$ and a decoding gamma correction $\gamma=2.5$.
We find that using these parameters achieves satisfactory
results, as shown in Figure~\ref{fig-inputs}(d).
The derived inputs by decoding gamma correction effectively
remove the severe dark aspects of $\mathbf{I}_{ce}$ and
enhance the visibility of the original image $\mathbf{I}$.

\subsection{Network Architecture}
We use an encoder-decoder network, which has
been shown to produce good results for a number of generative tasks such as image denoising~\cite{mao2016image}, image harmonization~\cite{tsai2017deep}, time-lapse video generation \cite{wei2018learn}.
In particular, we choose a variation of the residual
encoder-decoder network model for image dehazing.
We use skip connections between encoder and decoder halves of the
network, where features from the encoder side are concatenated
to be fed to the decoder. This significantly accelerates the convergence~\cite{mao2016image} and helps generate a much clear dehazed image.

We perform an early fusion by concatenating the original hazy image and three derived inputs in the input layer.
The network is of a multi-scale style in order to prevent halo artifacts, which
will be discussed in more details in Section~\ref{sec-multiscale}.
We show a diagram of GFN in Figure~\ref{fig-net}. Note that we only show the coarsest level network of GFN in Figure~\ref{fig-net}.
%
%
To leverage more context without losing local details, we use dilation network to enlarge the receptive field in the convolutional layers.
Rectification layers are added after each convolutional or deconvolutional layer.
The convolutional layers act as a feature extractor, which preserve
the primary information of scene colors in the input layer,
meanwhile eliminating the unimportant colors from the inputs.
The deconvolutional layers are then combined to recover the weight maps of three derived inputs. In other words, the outputs of the deconvolutional layers are the \textit{confidence maps} of the derived input images $\mathbf{I}_{wb}$, $\mathbf{I}_{ce}$ and $\mathbf{I}_{gc}$.
%
%

We use 3 convolutional blocks and 3 deconvolutional blocks with stride 1 in each scale. Each layer is of the same type: 32 filters of the size $3 \times 3 \times 32$ except the first and last layers.
The first layer operates on the input image with kernel size $5\times5$, and the last layer is used for confidence map reconstruction.
In this work, we demonstrate that explicitly modeling
confidence maps has several advantages. These are discussed later in Section~\ref{sec-fusion}.
Once the confidence maps for the derived inputs are predicted, they are multiplied
by the three derived inputs to give the final dehazed image in each scale:
\begin{equation}
J = C_{wb}\circ\mathbf{I}_{wb} + C_{ce}\circ\mathbf{I}_{ce} + C_{gc}\circ\mathbf{I}_{gc},
\label{equ-dehaze}
\end{equation}
where $\circ$ denotes element-wise multiplication, and $C_{wb}, C_{ce}$, and $C_{gc}$ are the \textit{confidence maps} for gating $\mathbf{I}_{wb}$, $\mathbf{I}_{ce}$, and $\mathbf{I}_{gc}$, respectively.
\begin{figure}[t]\footnotesize
	\begin{center}
		\includegraphics[width = 0.45\textwidth]{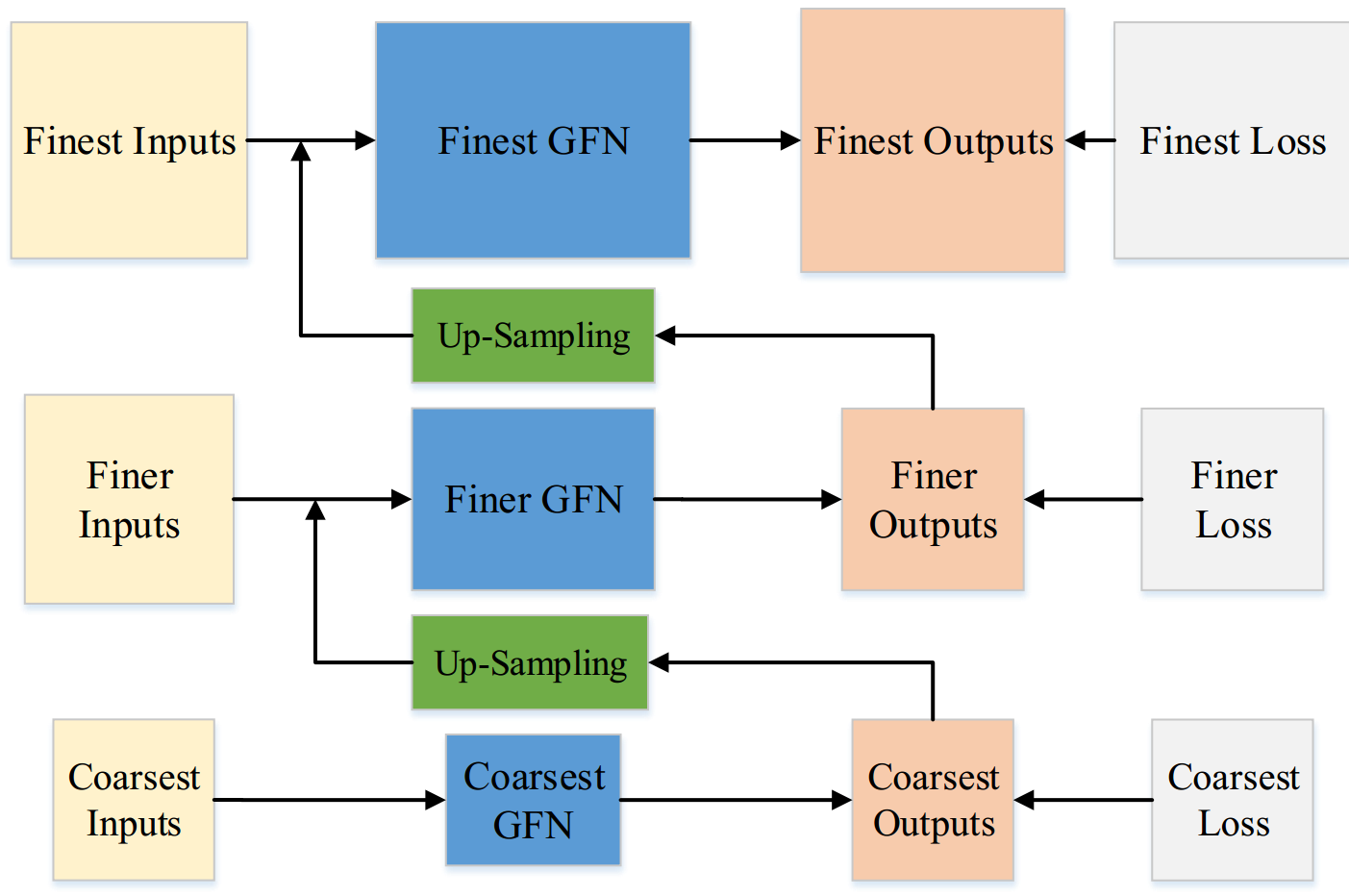}
	\end{center}
	\vspace{-0.1cm}
	\caption{Multi-scale GFN structure.
	}
	\vspace{-4mm}
	\label{fig-multiscale-net}
\end{figure}
\subsection{The multi-Scale Refinement}
\label{sec-multiscale}
The network described in the previous subsection is subject to halo artifacts,
particularly for strong transitions within the confidence maps~\cite{ancuti2013single,choi2015referenceless}.
Hence, we perform estimation by varying the image resolution in a coarse-to-fine manner to prevent halo artifacts.
The multi-scale approach is motivated by the fact that the
human visual system is sensitive to local changes (\eg, edges)
over a wide range of scales. As a merit, the multi-scale approach
provides a convenient way to incorporate local image details
over varying resolutions.
Figure~\ref{fig-multiscale-net} shows the proposed multi-scale fusion network, in which
the coarsest level network is shown in Figure~\ref{fig-net}.
Finer level networks basically have the same structure as the coarsest network. However, the first convolutional layer takes the sharp image from a previous stage as well as its
own hazy image and derived inputs, in a concatenated form.
Each input size is twice the size of its coarser scale network.
There is an up-sampling layer before the next stage. At the finest scale, the original high-resolution image is restored.

The multi-scale approach desires that each scale output is a clear image of the corresponding scale. Thus, we train our network so that all intermediate dehazed images should form a pyramid of the sharp image.
The MSE criterion is applied to every level of the pyramid.
In specific, given a collection of $N$ training pairs $\mathbf{I}_i$ and $\mathbf{J}_i$, where $\mathbf{I}_i$ is a hazy image and $\mathbf{J}_i$ is the clean version as the ground truth, the loss function at the $k$-th scale is defined as follows:
\begin{equation}
\mathcal{L}_{cont}(\Theta, k) = \frac{1}{N}\sum_{i=1}^{N}\left\|\mathcal{F}(\mathbf{I}_{i,k}, \Theta, k) - \mathbf{J}_{i,k}\right\|^2,
\label{equ-loss}
\end{equation}
where $\Theta$ keeps the weights of the convolutional and deconvolutional kernels.

\subsection{Adversarial Loss}
Recently, generative adversarial networks (GANs) are reported to generate
sharp realistic images \cite{nah2017deep}. Therefore, we follow the architecture introduced in \cite{nah2017deep}, and build a discriminator to take the output of the finest scale or the ground-truth sharp image as input. The adversarial loss is defined as follows:
\begin{eqnarray}
\begin{split}
\mathcal{L}_{adv} = & \underset{\mathbf{J}\backsim p_{\text{clear}}(\mathbf{J})}{\mathbb{E}}\big[\log D(\mathbf{J})\big] \\ & +\underset{\mathbf{I}\backsim p_{\text{hazy}}(\mathbf{I})}{\mathbb{E}}\left[\log\big(1-D(\mathcal{F}(\mathbf{I}))\big)\right],
\end{split}
\label{equ-advloss}
\end{eqnarray}
where $\mathcal{F}$ is our multi-scale network in Figure \ref{fig-multiscale-net}, and $D$ is the discriminator.
Finally, by combining the multi-scale content loss and
adversarial loss, our final loss function is
\begin{equation}
\mathcal{L}_{total} = \mathcal{L}_{cont} + 0.001\mathcal{L}_{adv}.
\label{equ-totalloss}
\end{equation}
Through optimizing the network parameters, we train the
model in the combination of two losses, multi-scale content
loss \eqref{equ-loss} and adversarial loss~\eqref{equ-advloss}.

\begin{figure*}[t]\scriptsize
	\begin{center}
		\begin{tabular}{@{}cccccccccc@{}}
			\includegraphics[width = 0.095\textwidth]{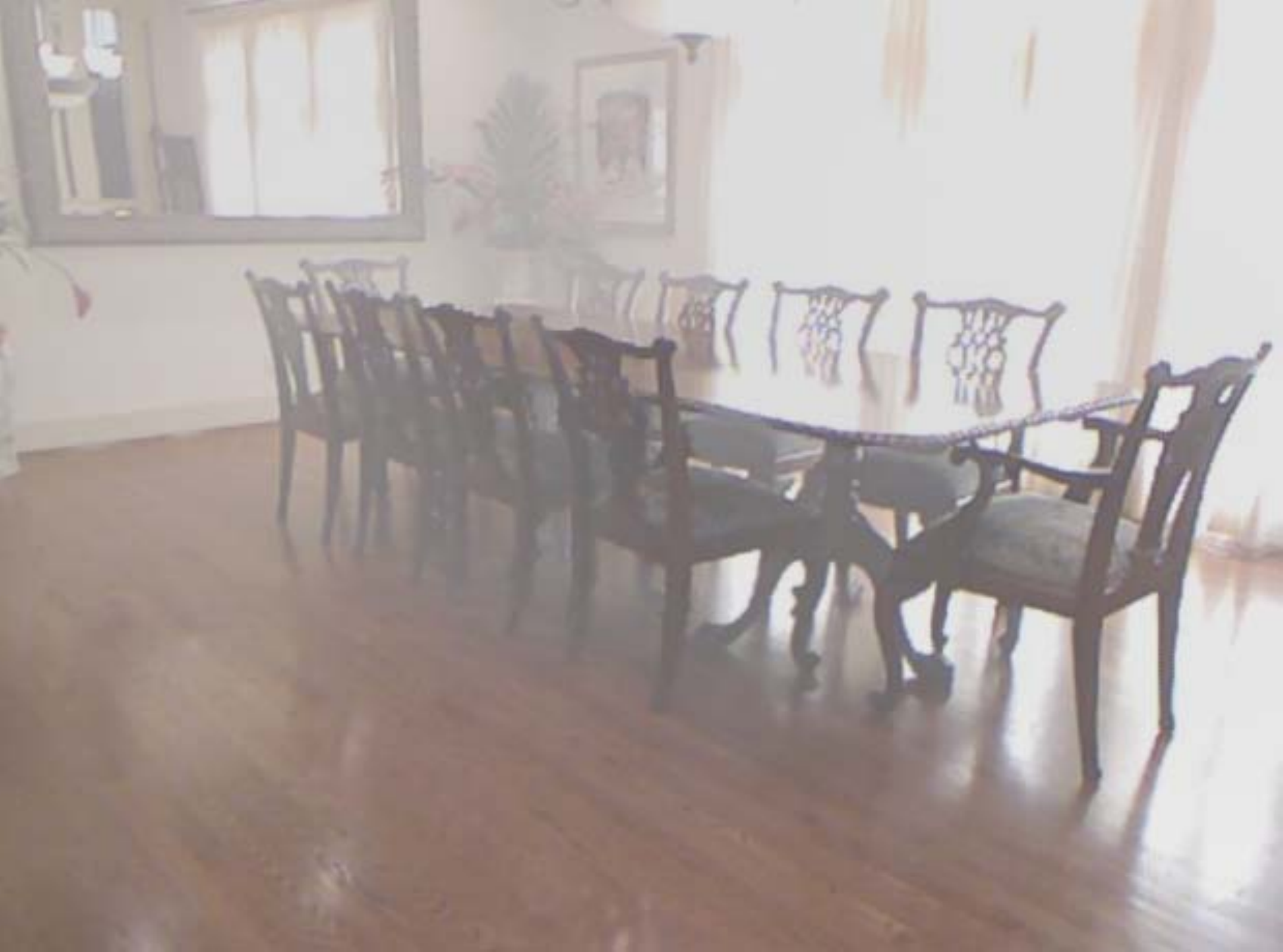} & \hspace{-0.4cm}
			\includegraphics[width = 0.095\textwidth]{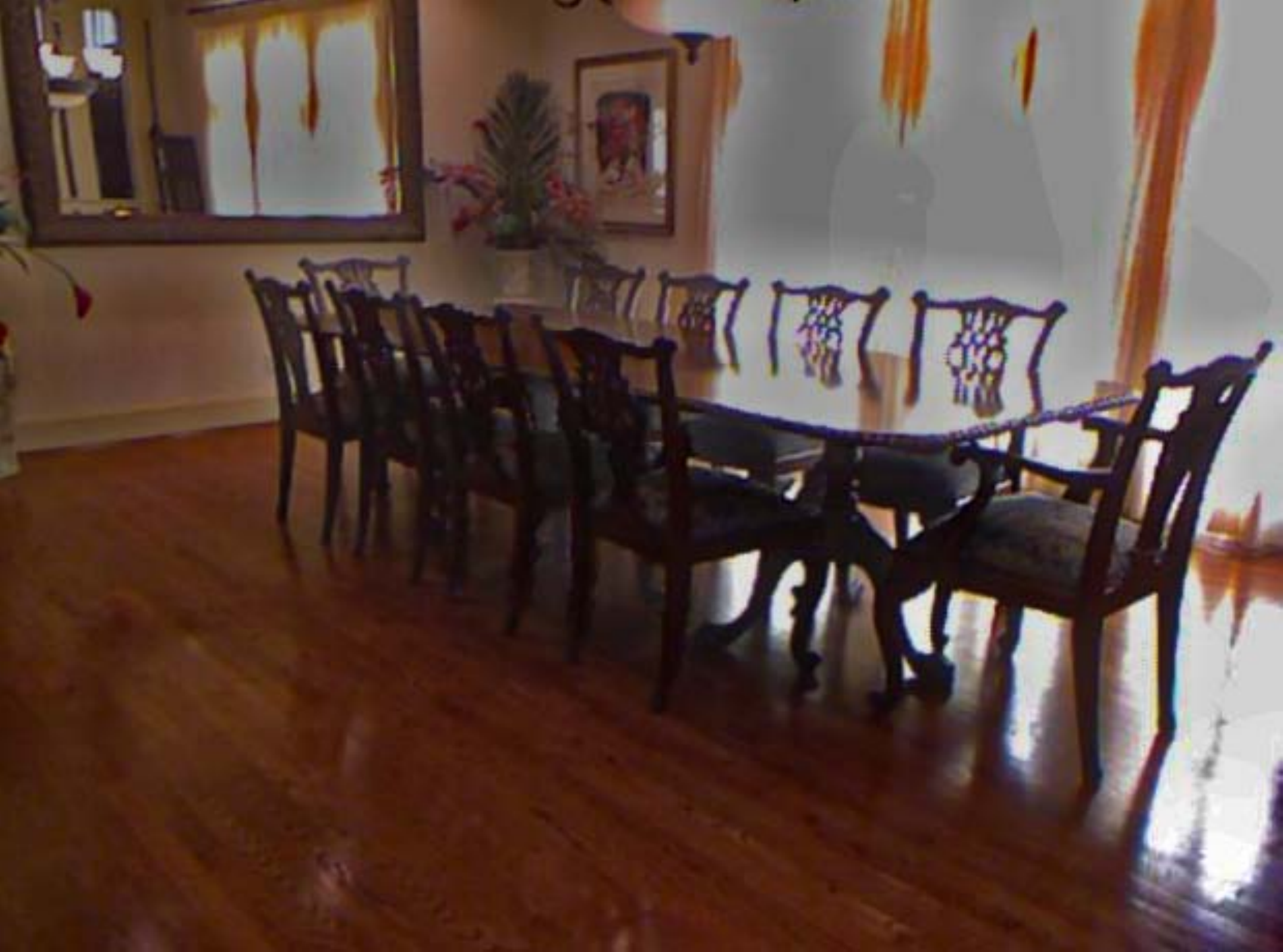} & \hspace{-0.4cm}
			\includegraphics[width = 0.095\textwidth]{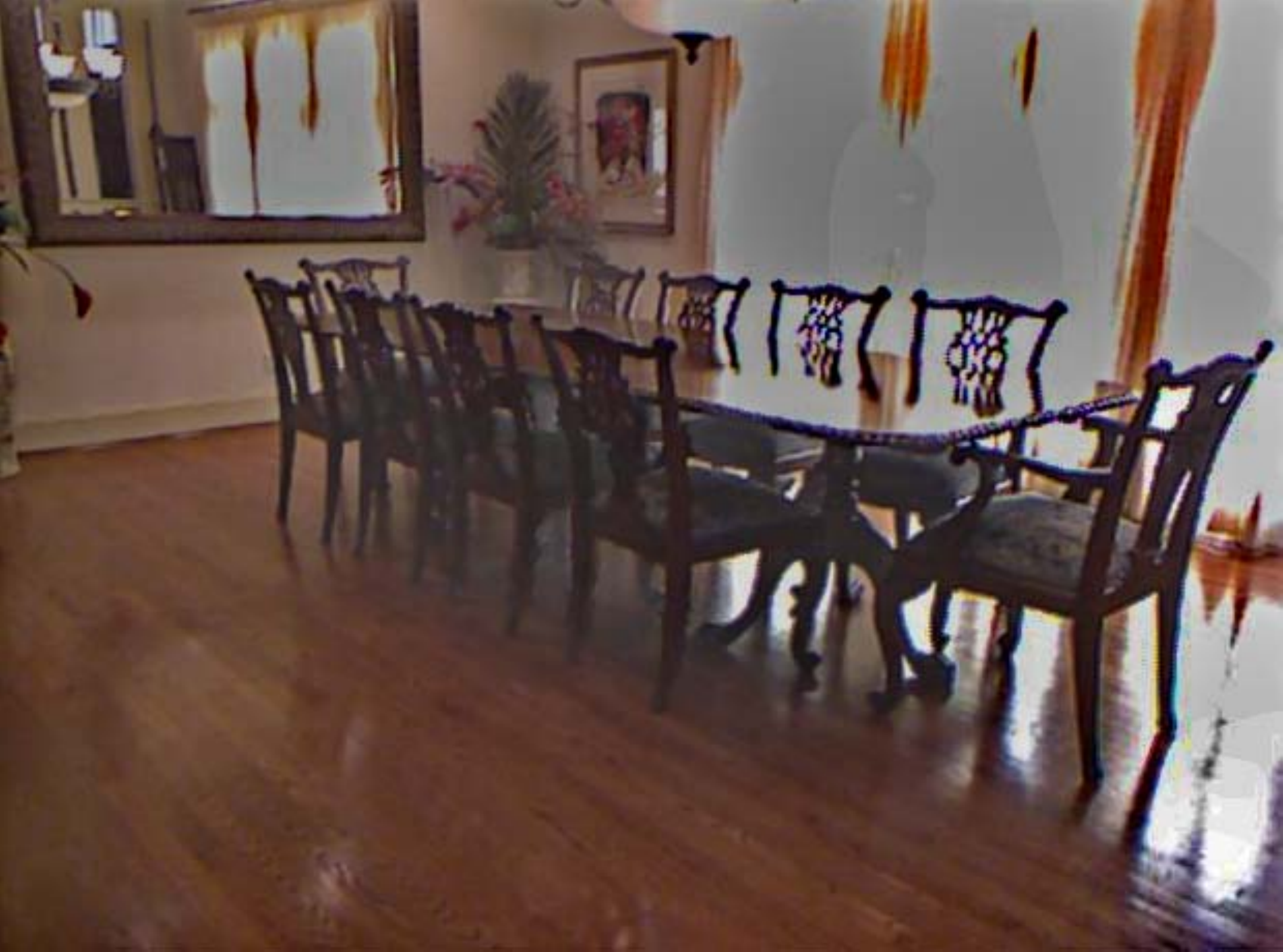} & \hspace{-0.4cm}
			\includegraphics[width = 0.095\textwidth]{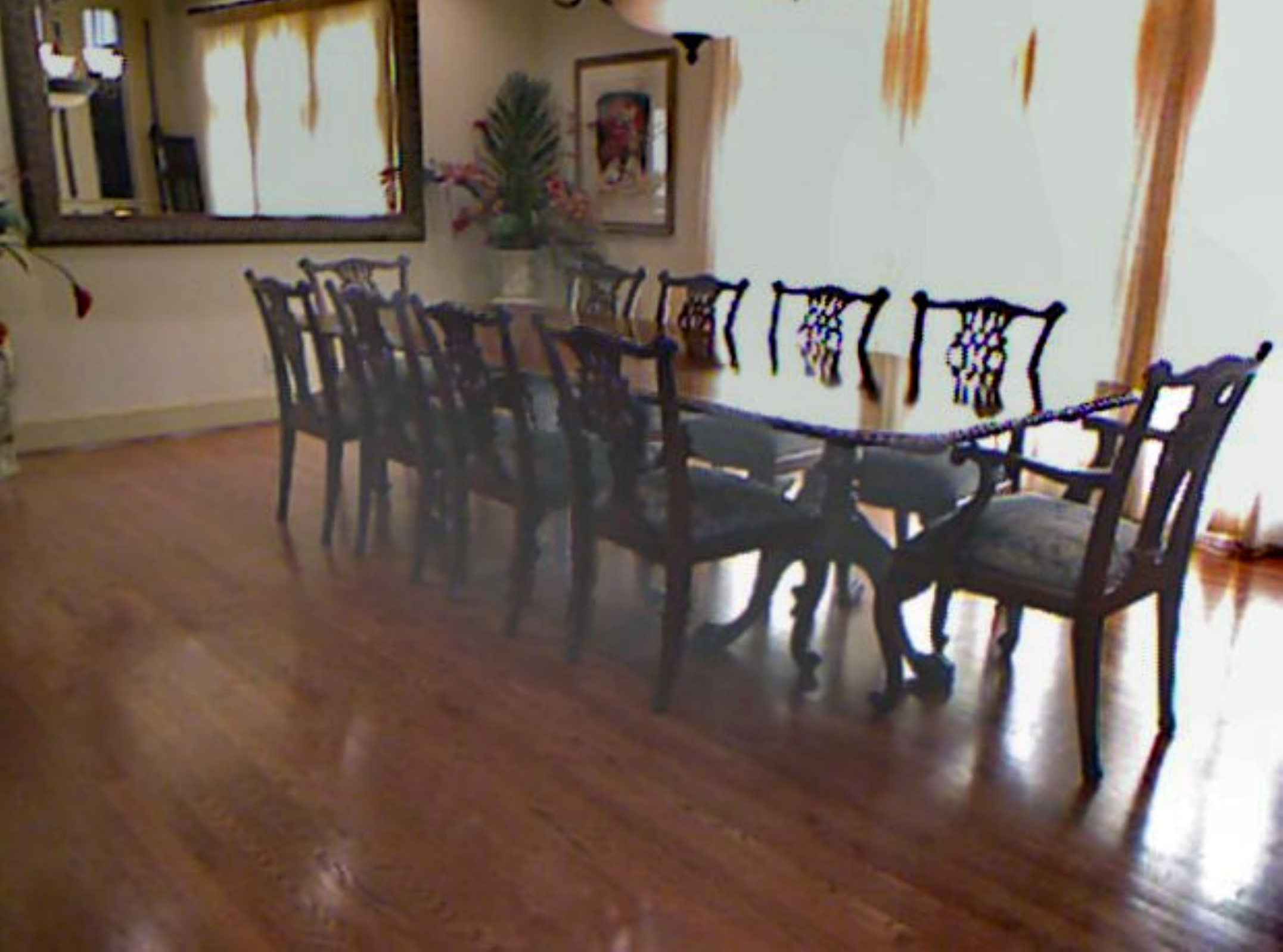} & \hspace{-0.4cm}
			\includegraphics[width = 0.095\textwidth]{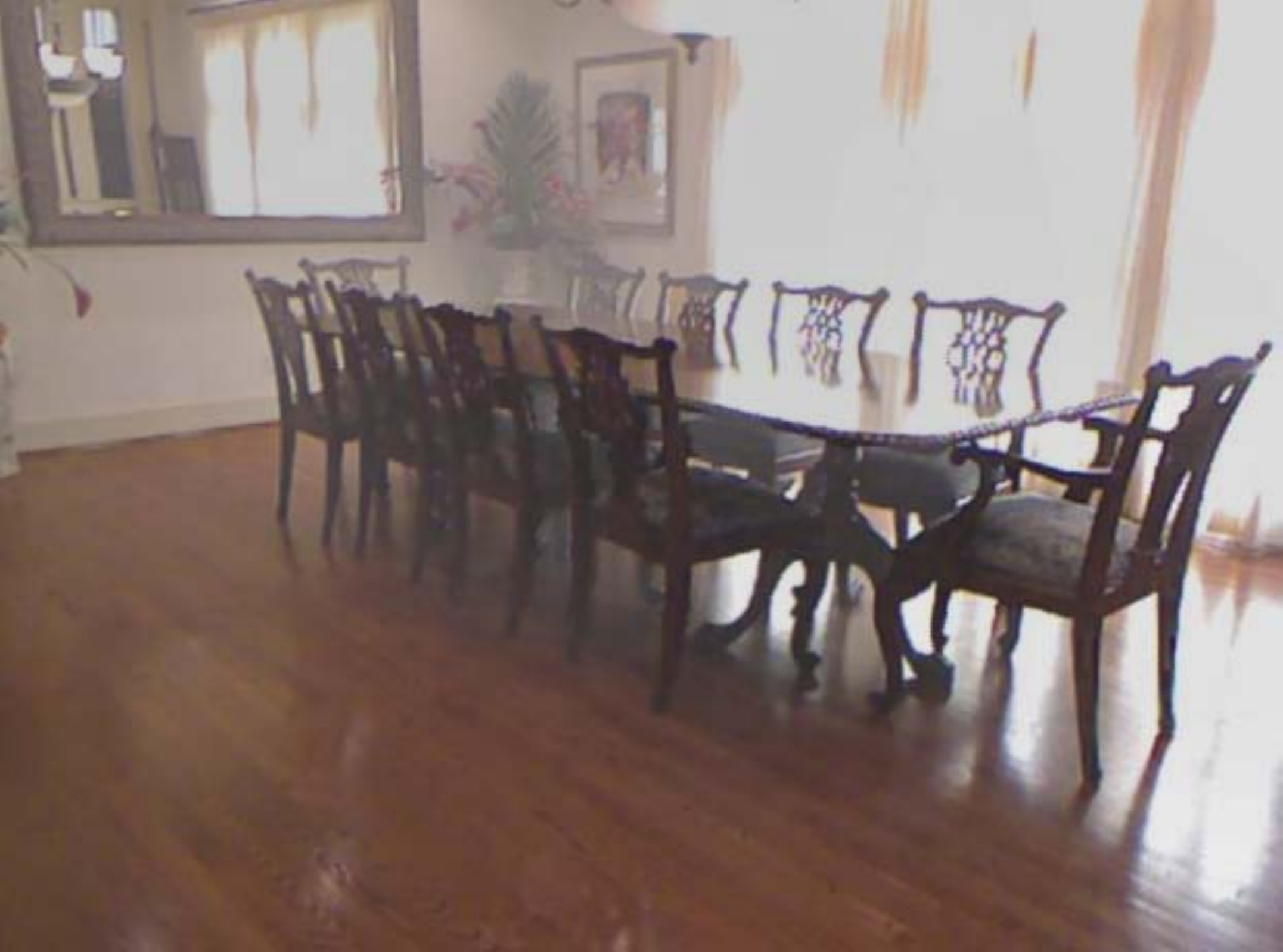} & \hspace{-0.4cm}
			\includegraphics[width = 0.095\textwidth]{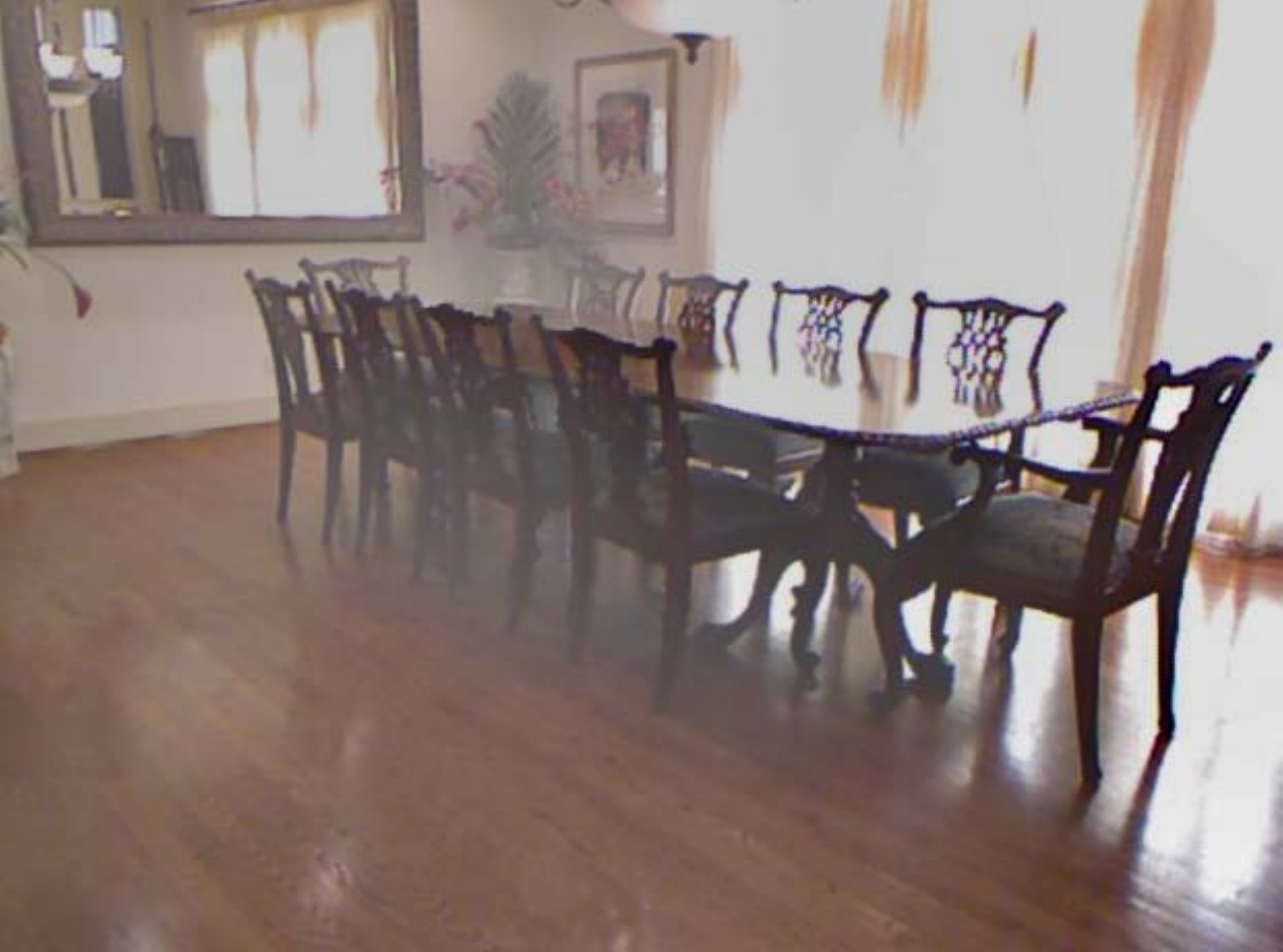} & \hspace{-0.4cm}
			\includegraphics[width = 0.095\textwidth]{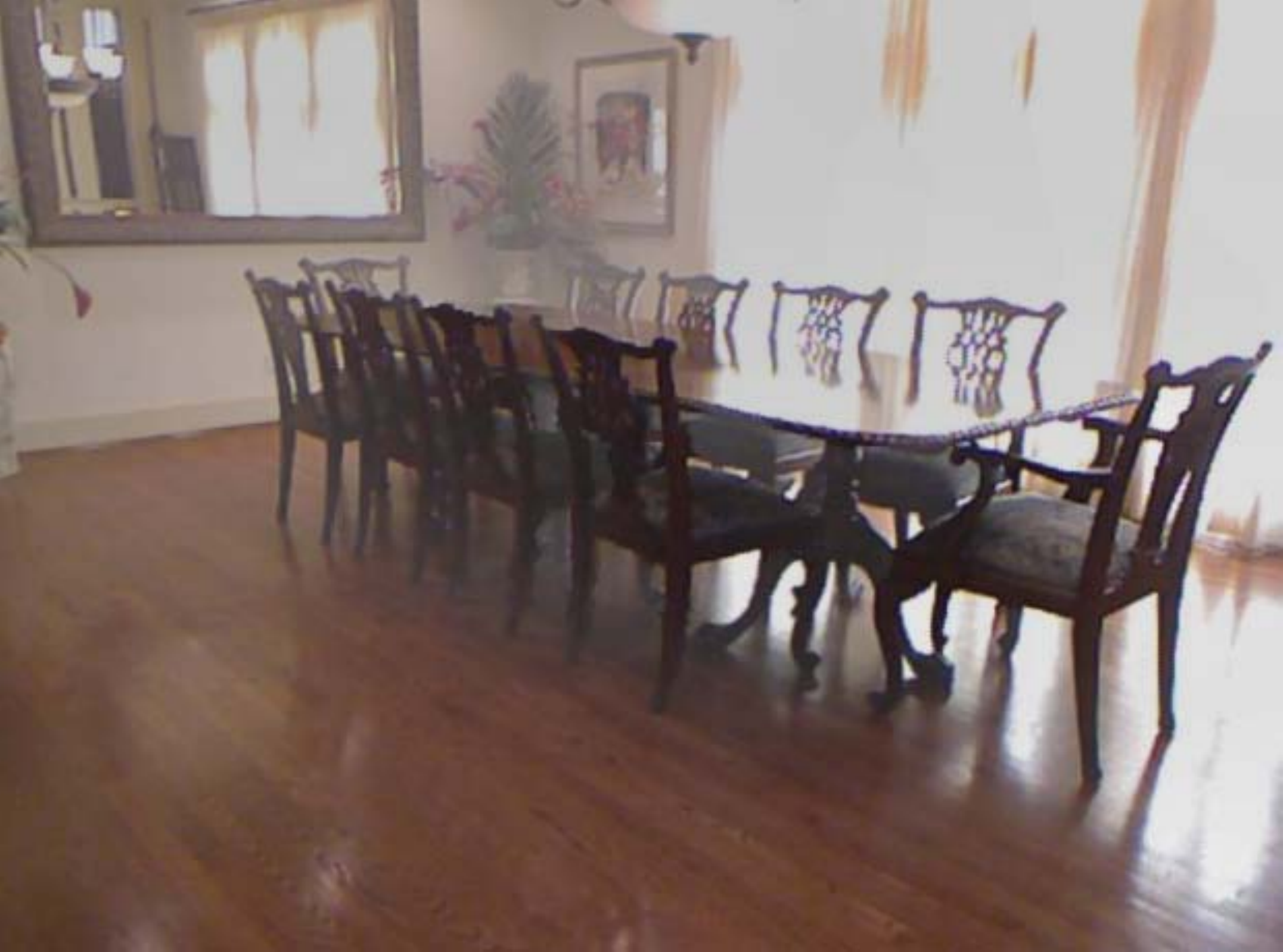} & \hspace{-0.4cm}
			\includegraphics[width = 0.095\textwidth]{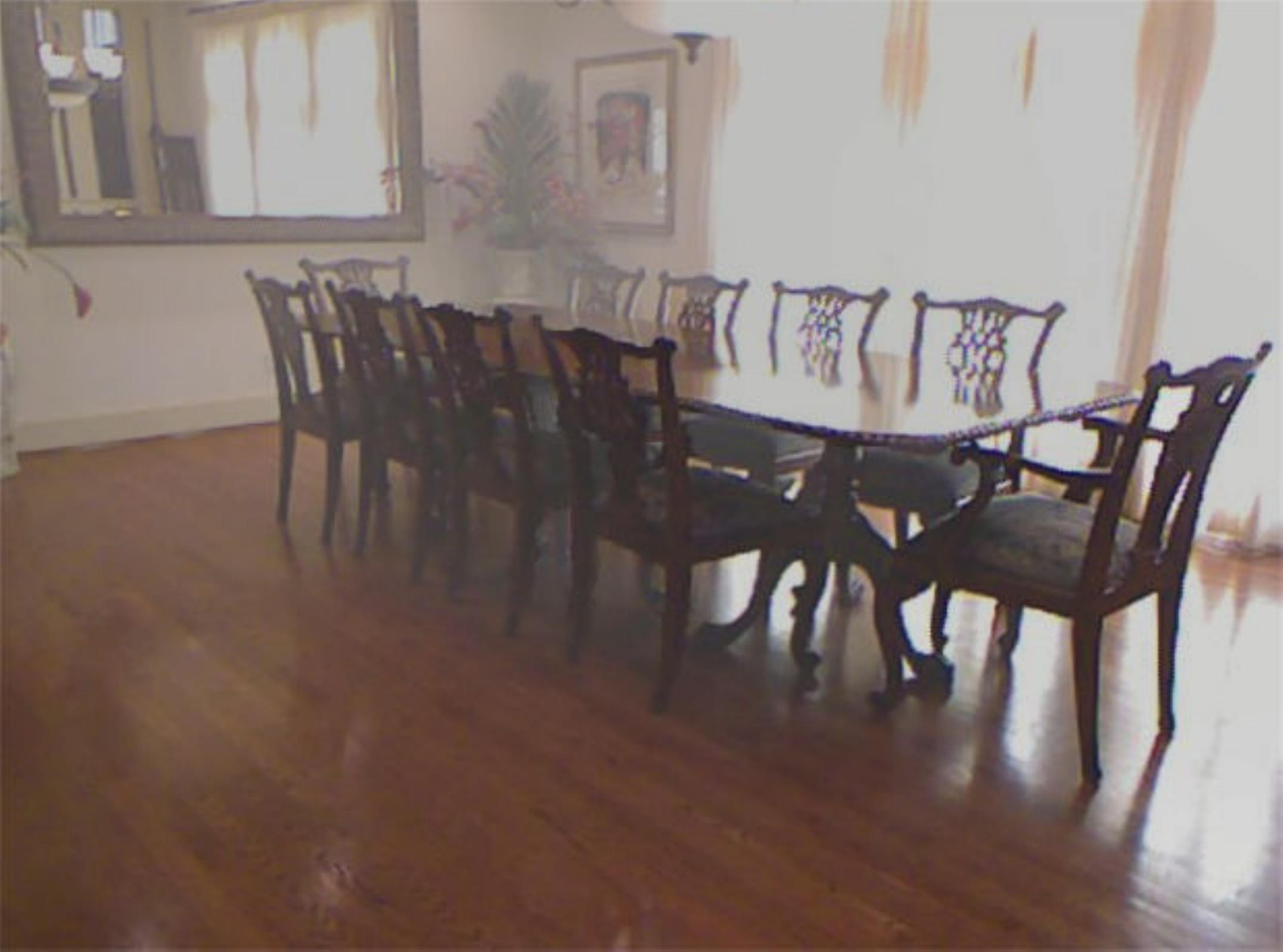} & \hspace{-0.4cm}
			\includegraphics[width = 0.095\textwidth]{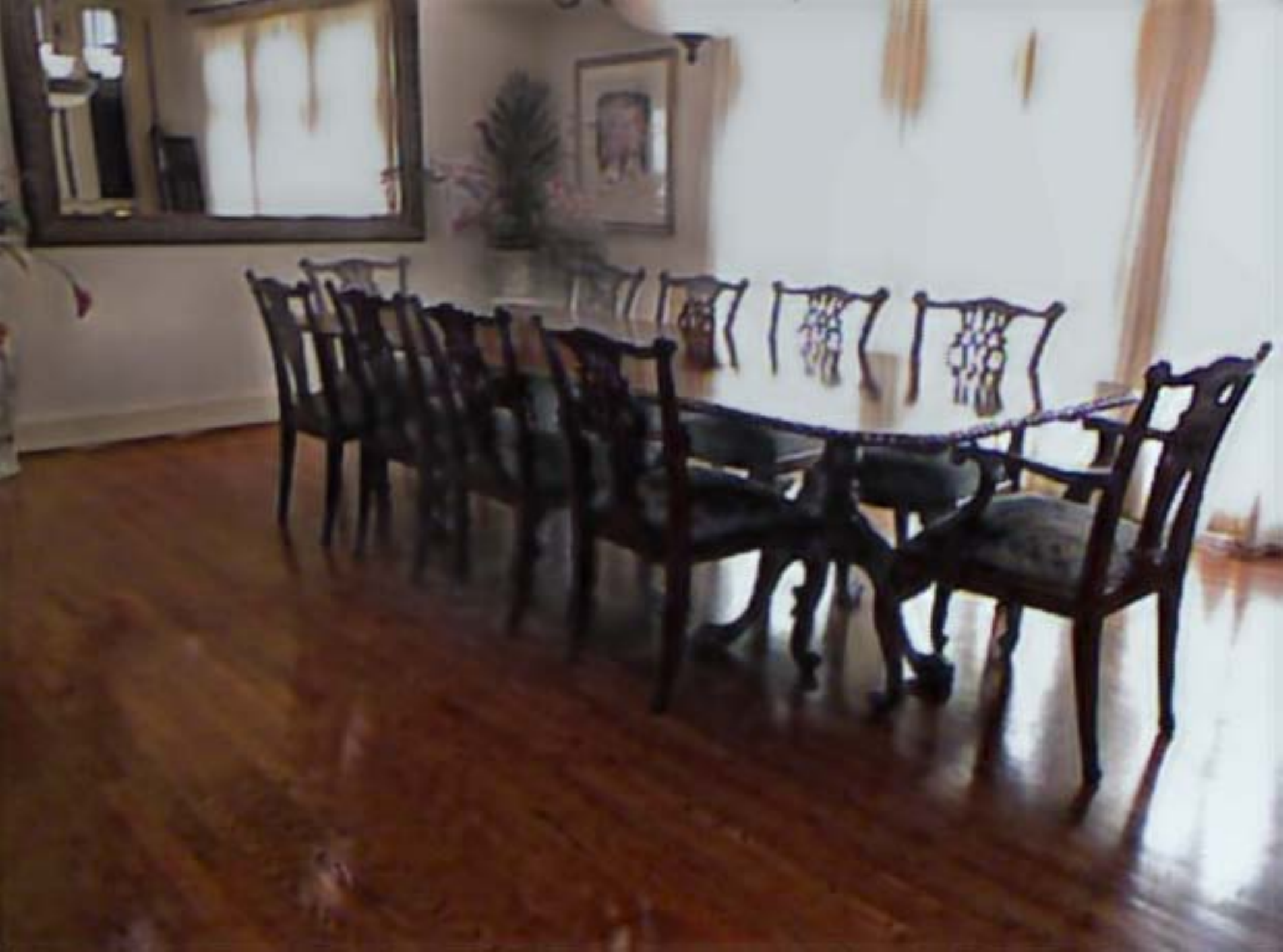} & \hspace{-0.4cm}
			\includegraphics[width = 0.095\textwidth]{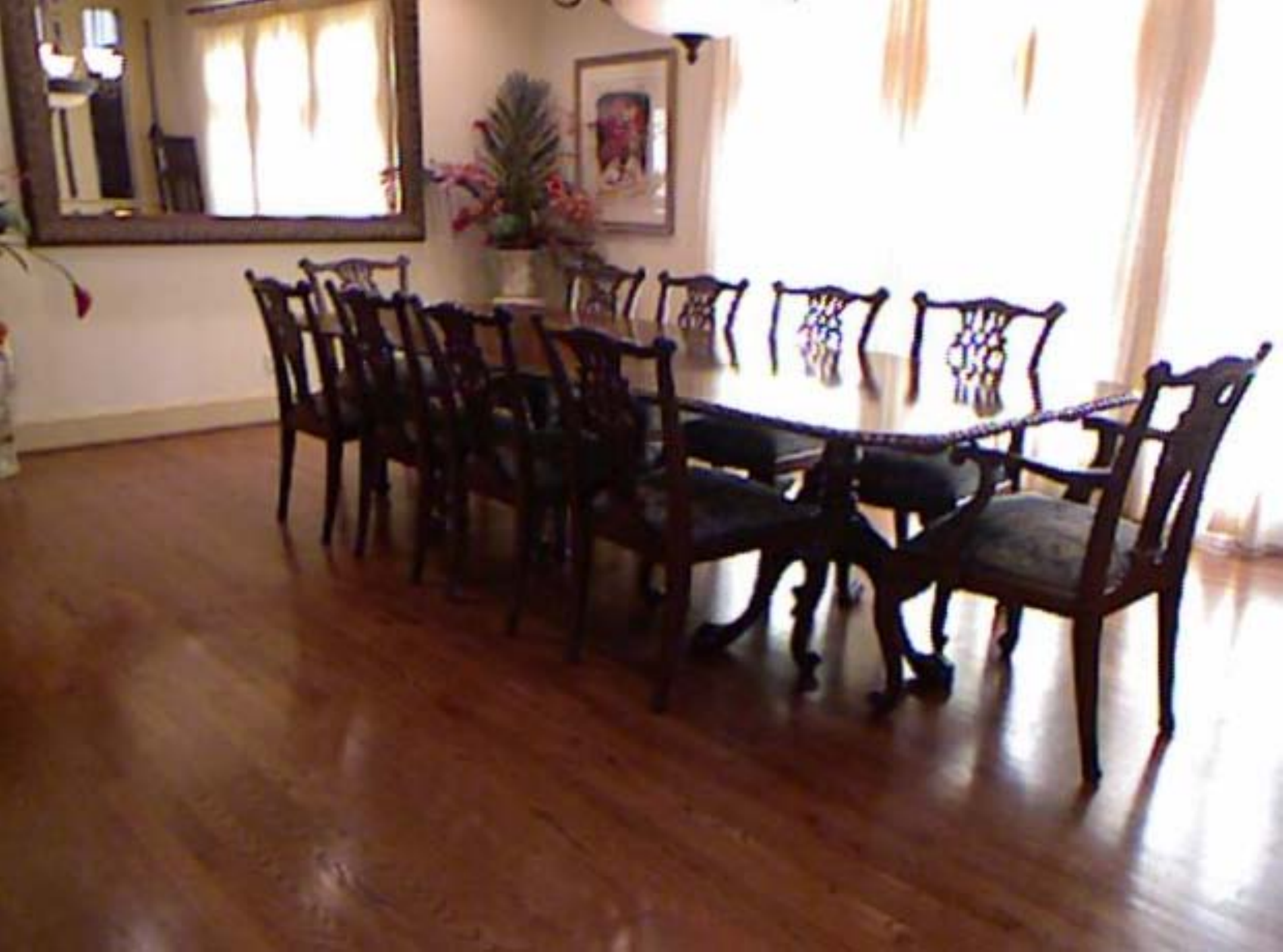}\\
			\includegraphics[width = 0.095\textwidth]{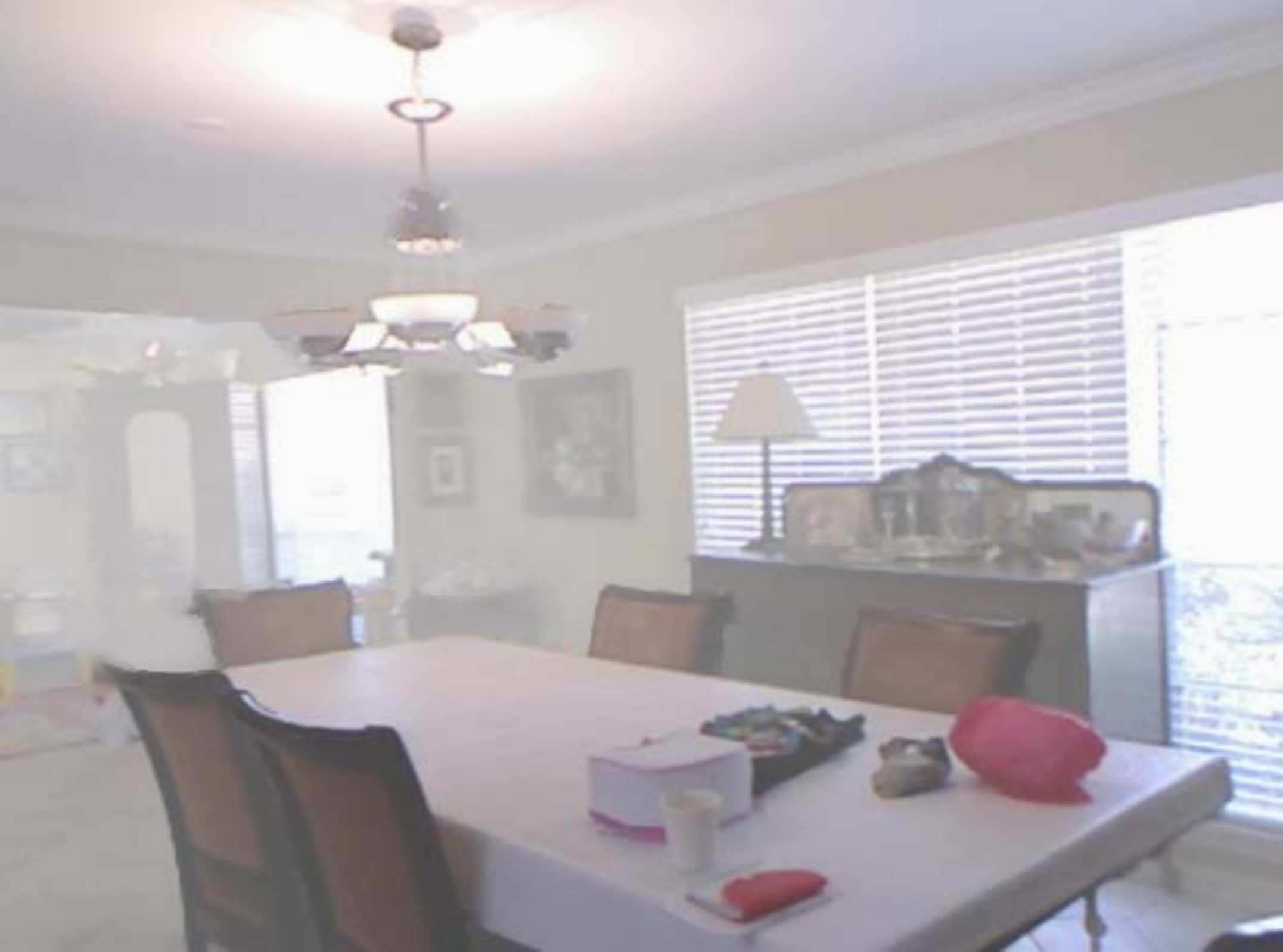} & \hspace{-0.4cm}
			\includegraphics[width = 0.095\textwidth]{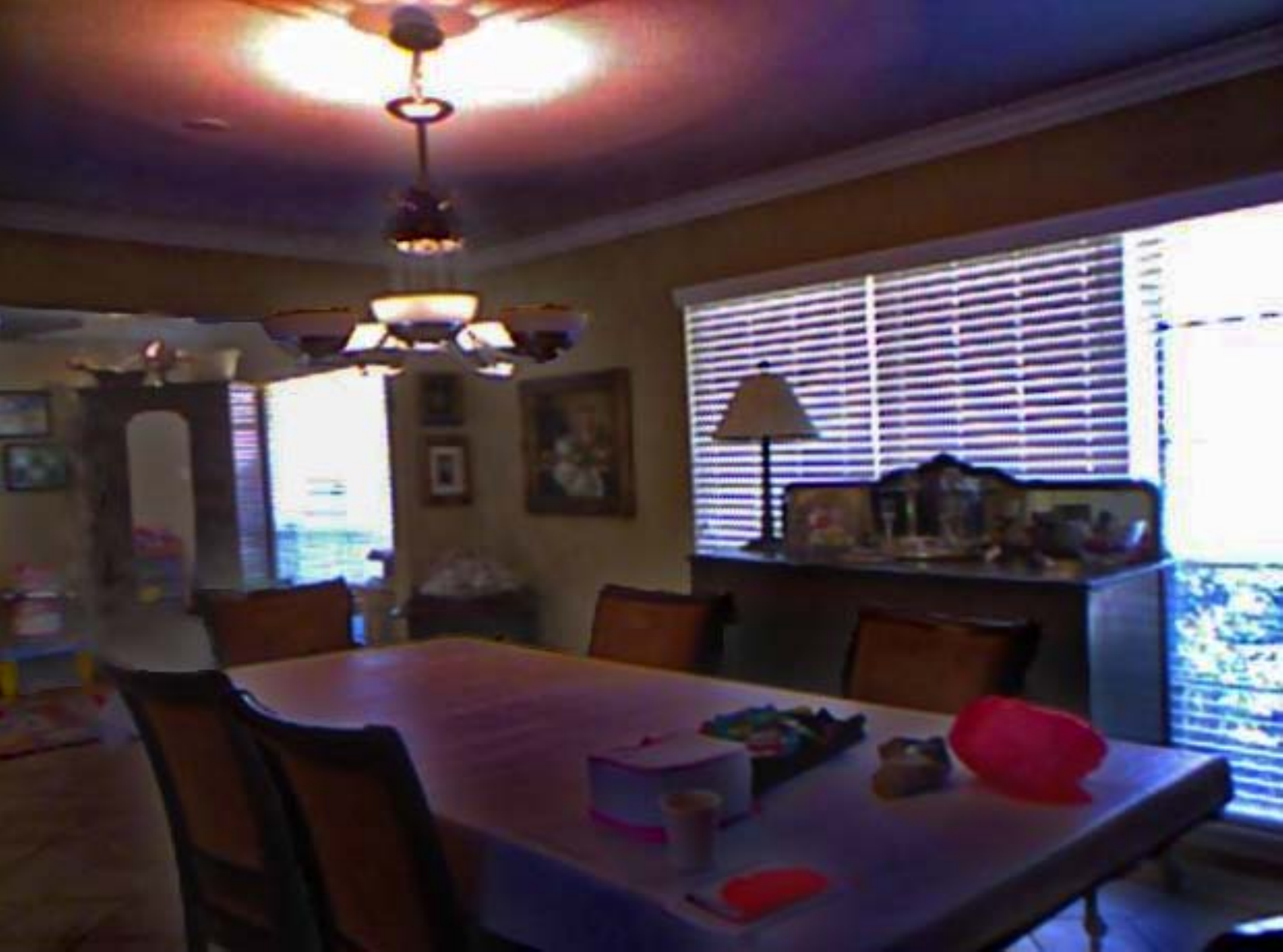} & \hspace{-0.4cm}
			\includegraphics[width = 0.095\textwidth]{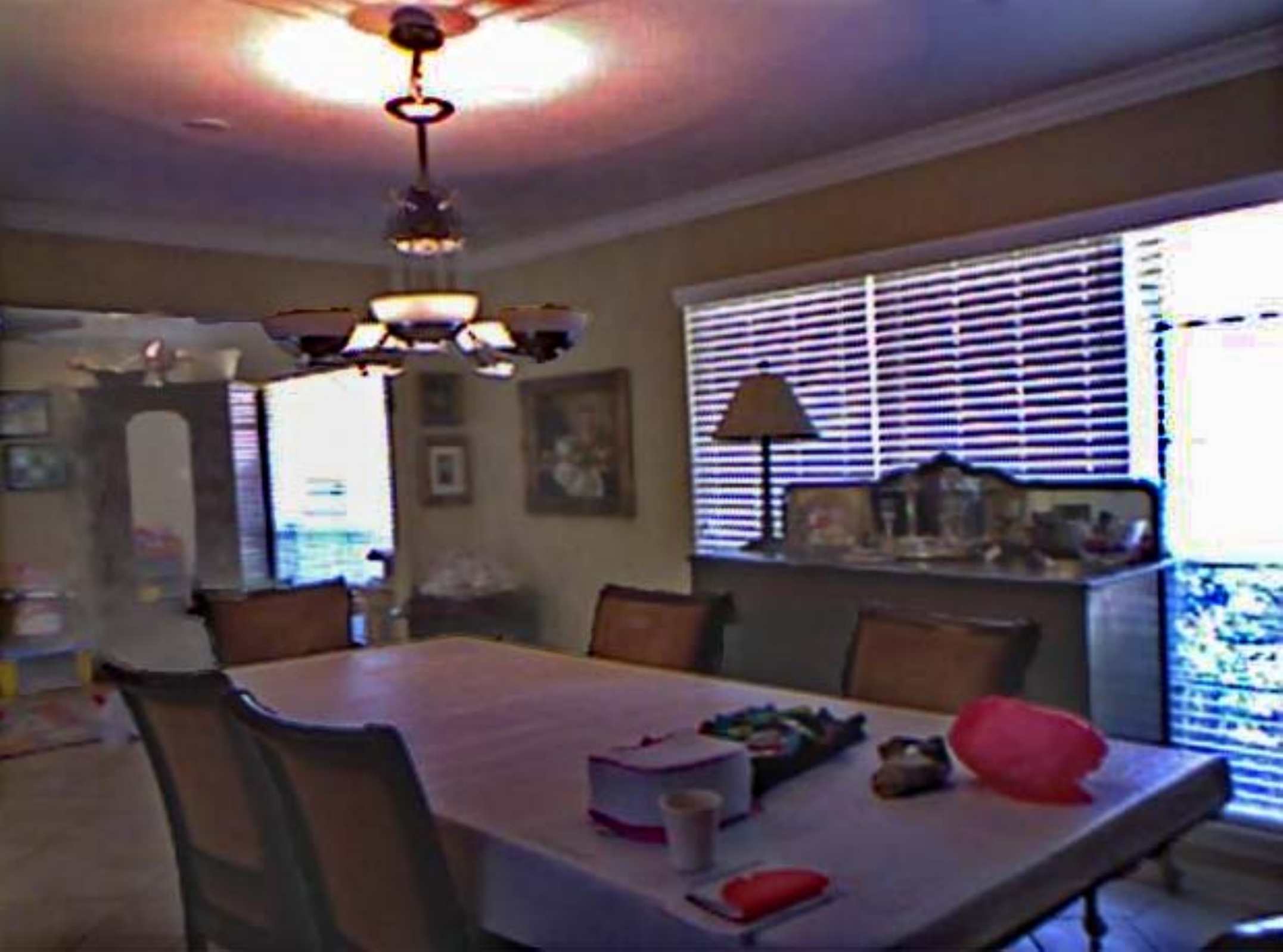} & \hspace{-0.4cm}
			\includegraphics[width = 0.095\textwidth]{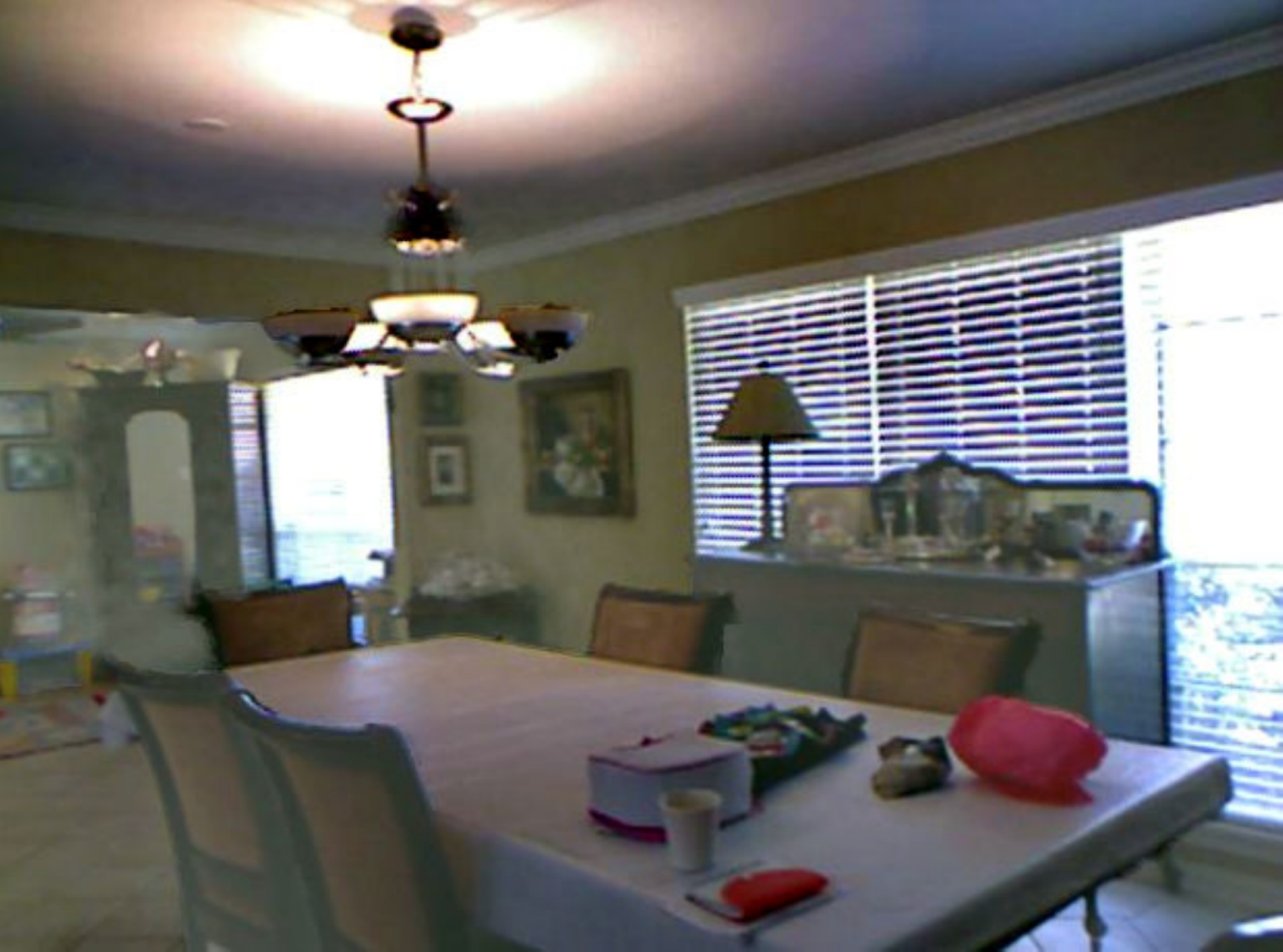} & \hspace{-0.4cm}
			\includegraphics[width = 0.095\textwidth]{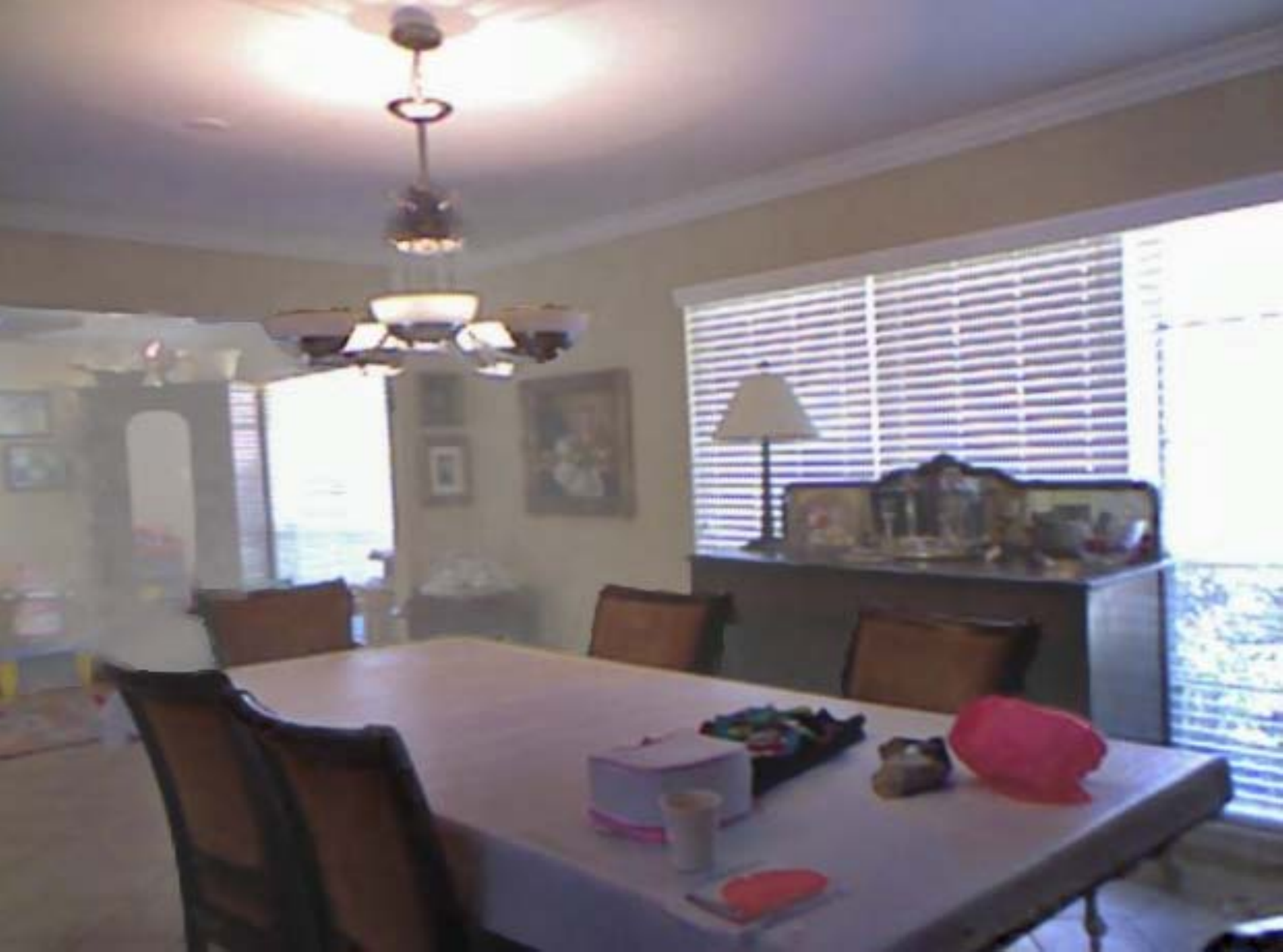} & \hspace{-0.4cm}
			\includegraphics[width = 0.095\textwidth]{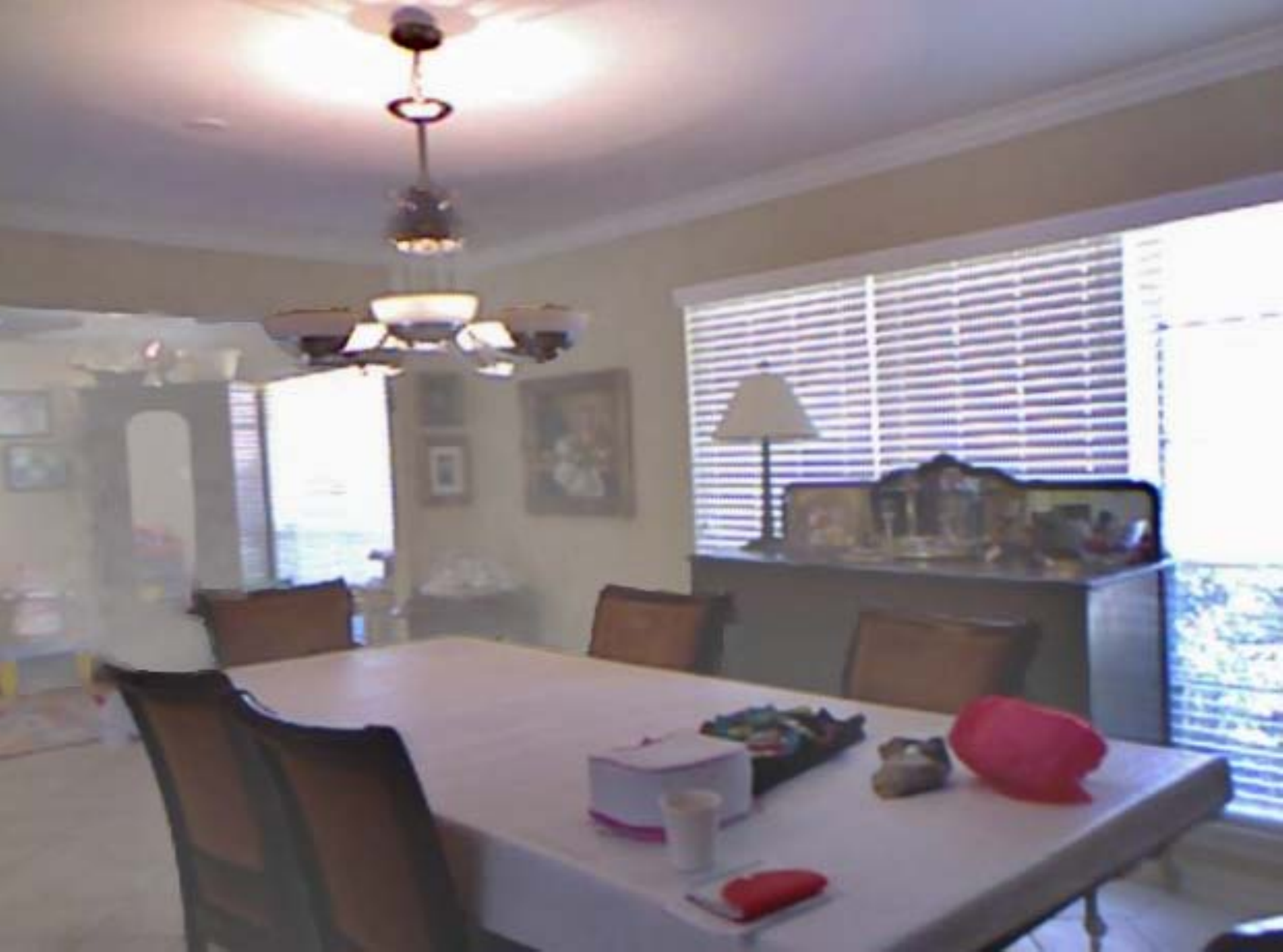} & \hspace{-0.4cm}
			\includegraphics[width = 0.095\textwidth]{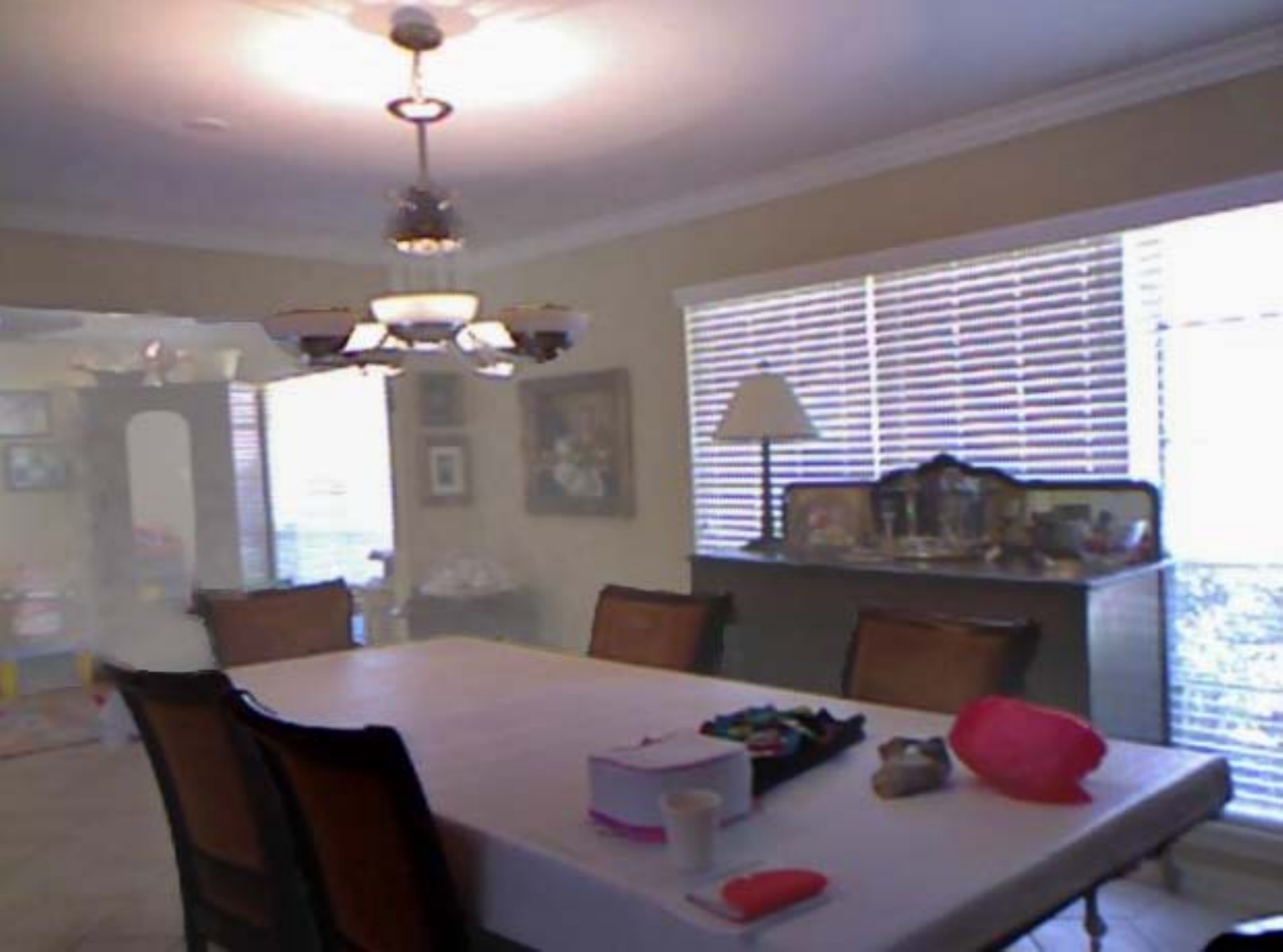} & \hspace{-0.4cm}
			\includegraphics[width = 0.095\textwidth]{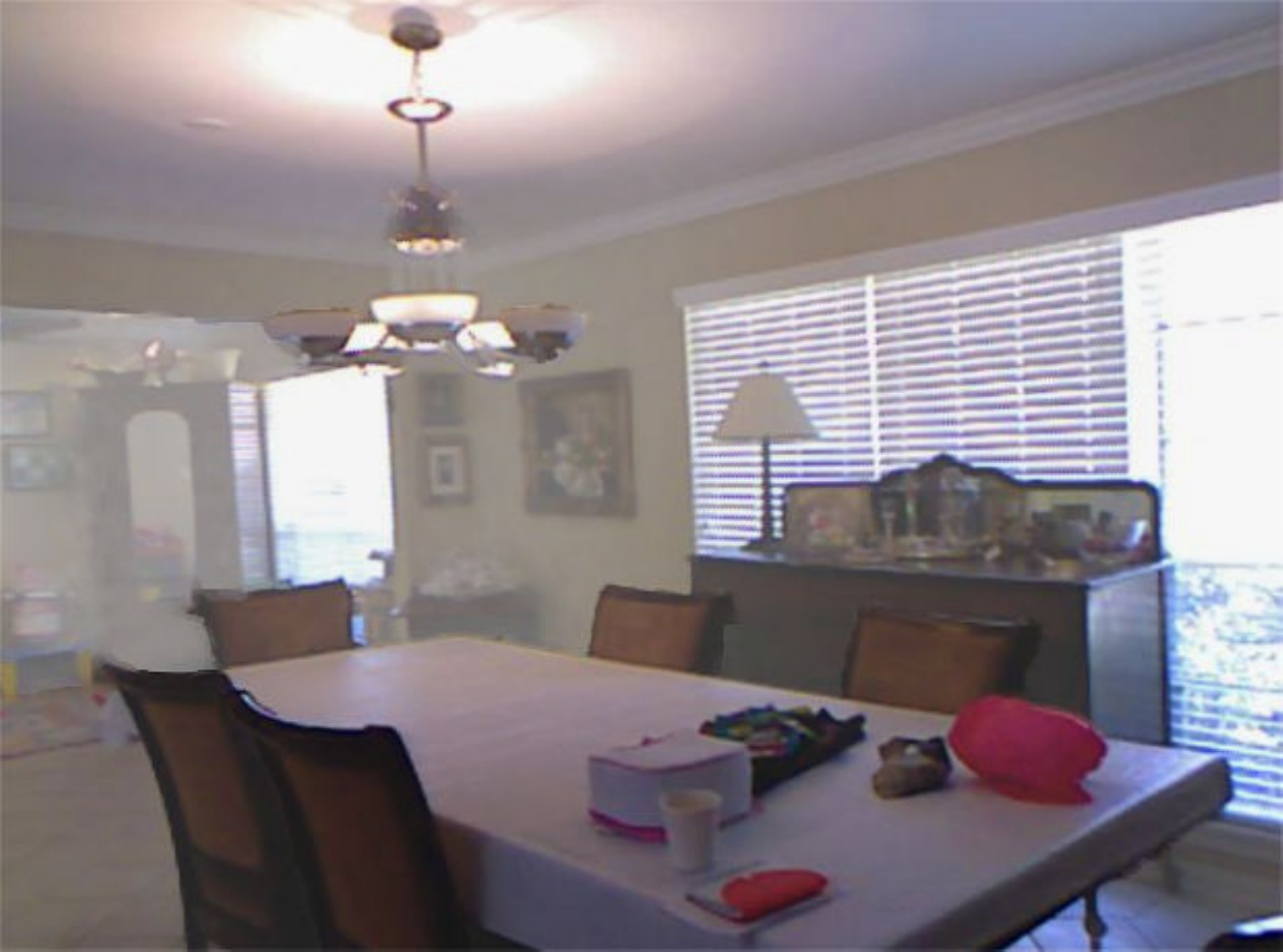} & \hspace{-0.4cm}
			\includegraphics[width = 0.095\textwidth]{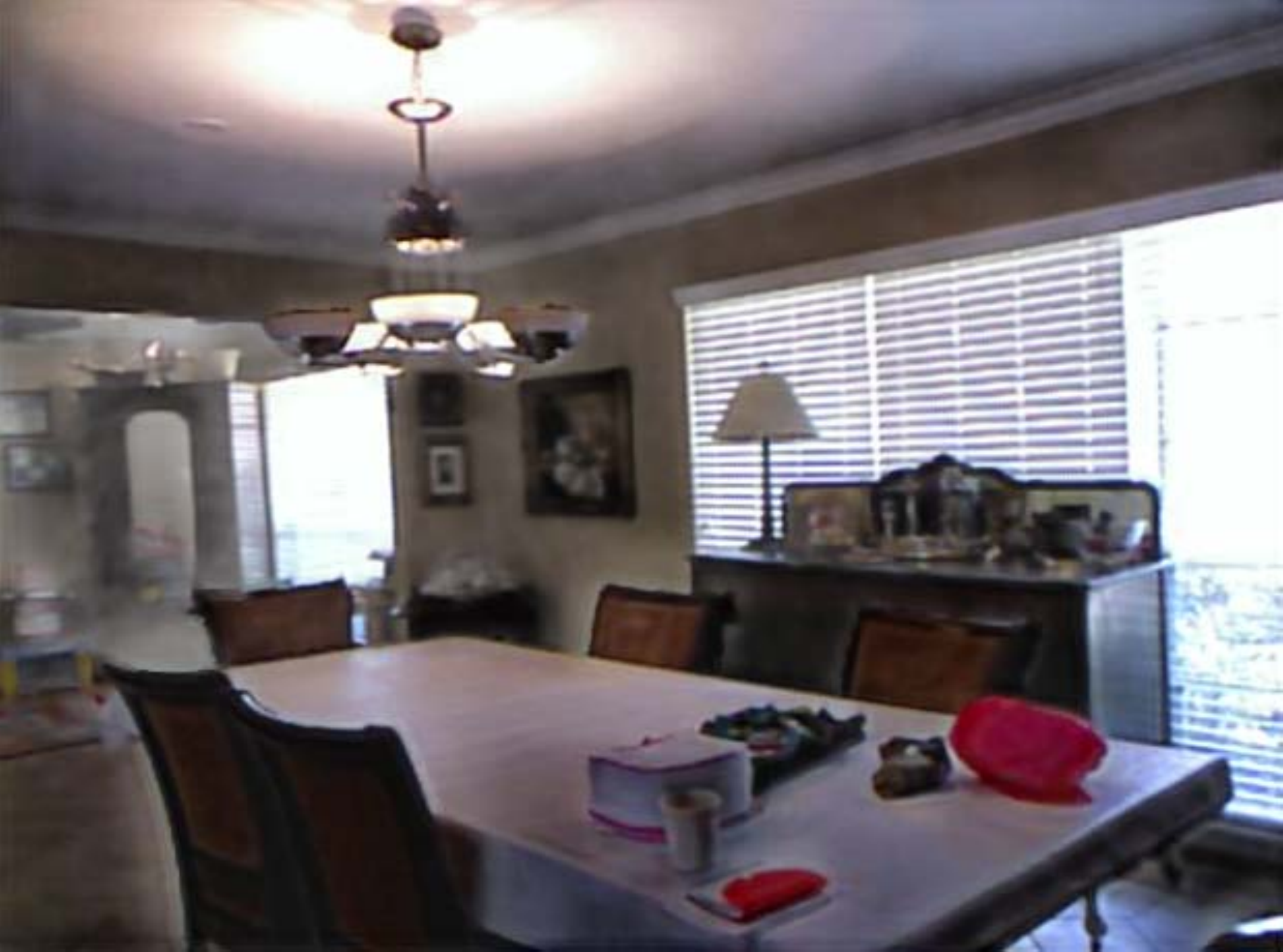} & \hspace{-0.4cm}
			\includegraphics[width = 0.095\textwidth]{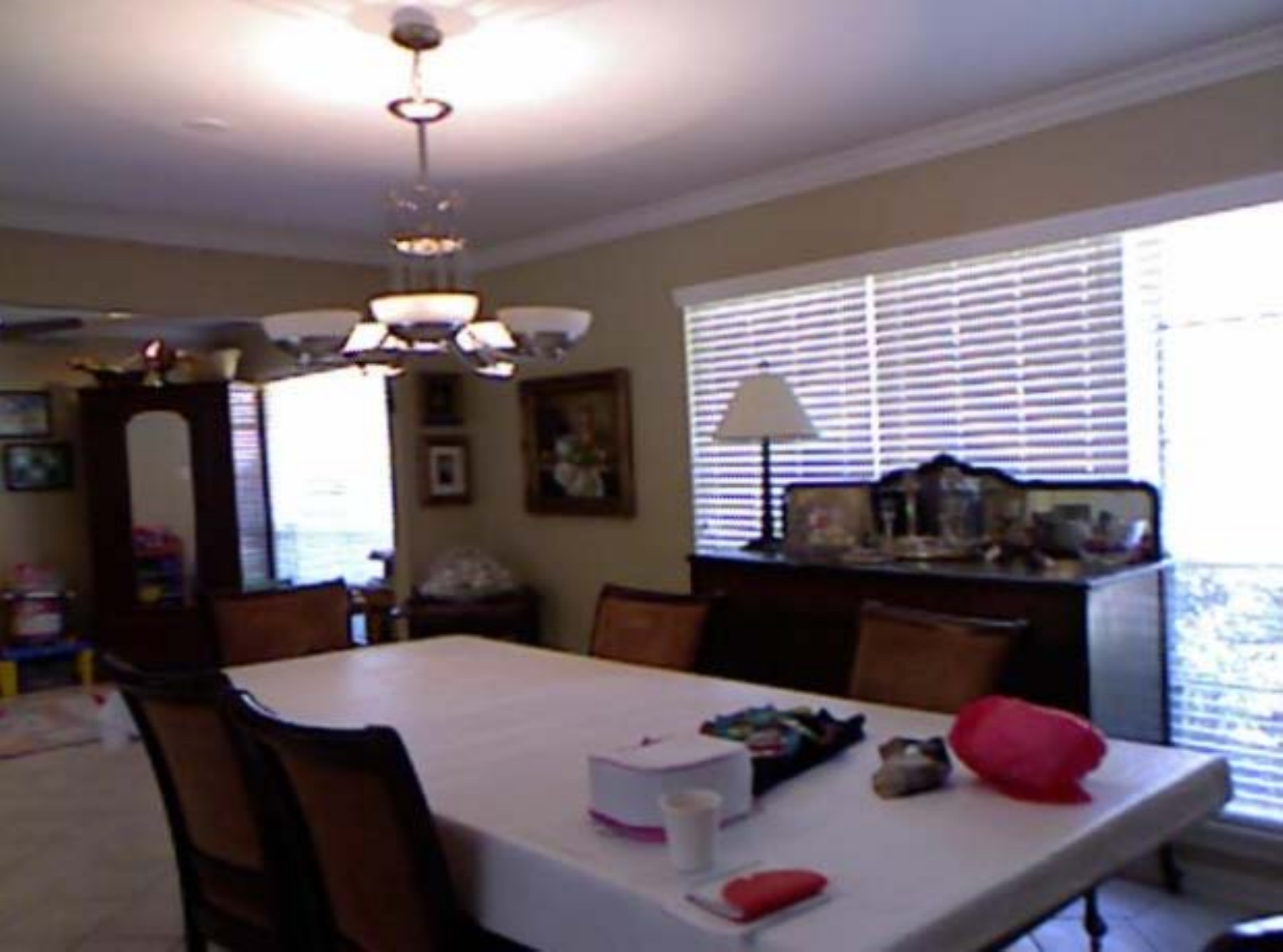}\\
			\includegraphics[width = 0.095\textwidth]{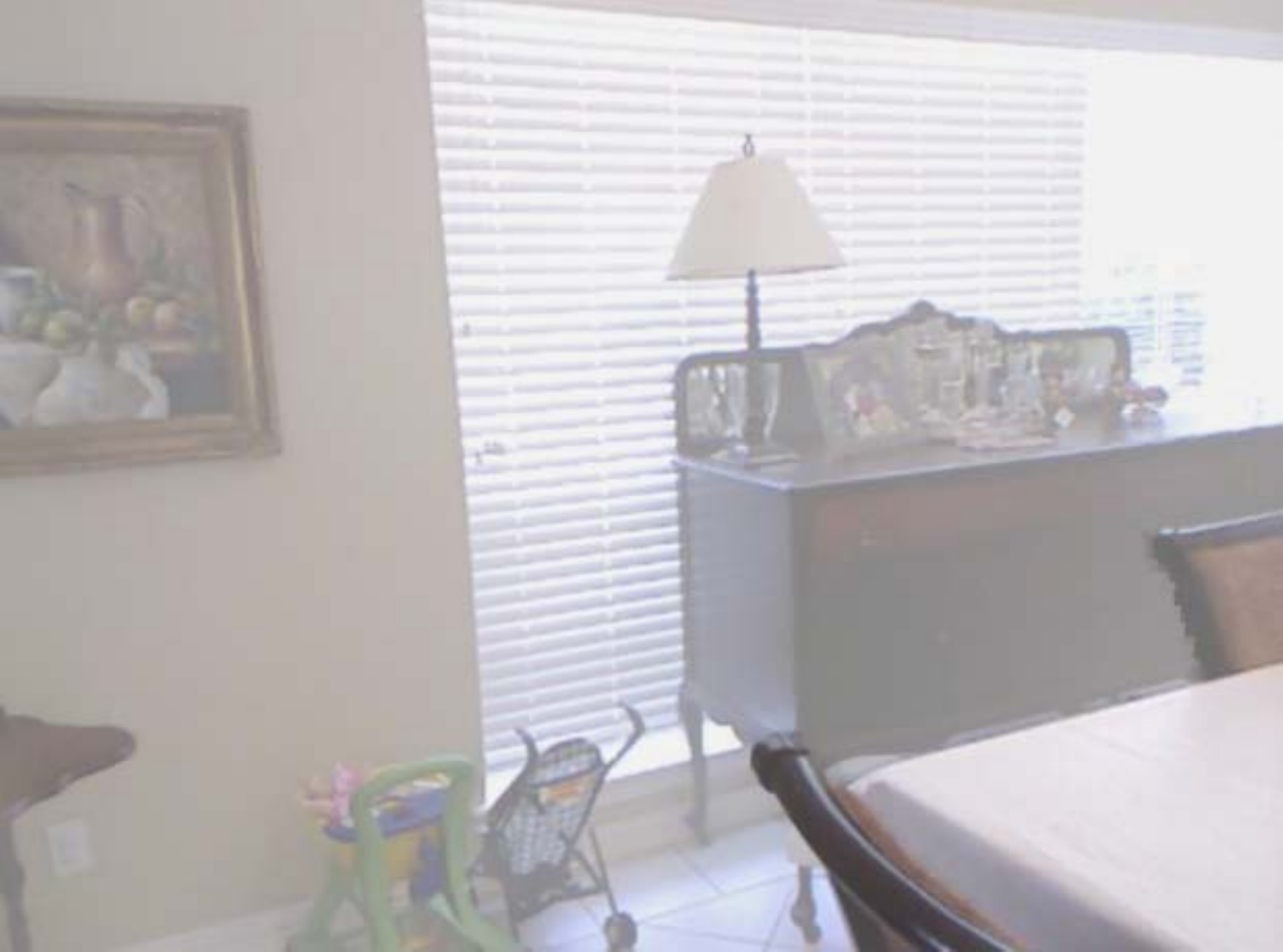} & \hspace{-0.4cm}
			\includegraphics[width = 0.095\textwidth]{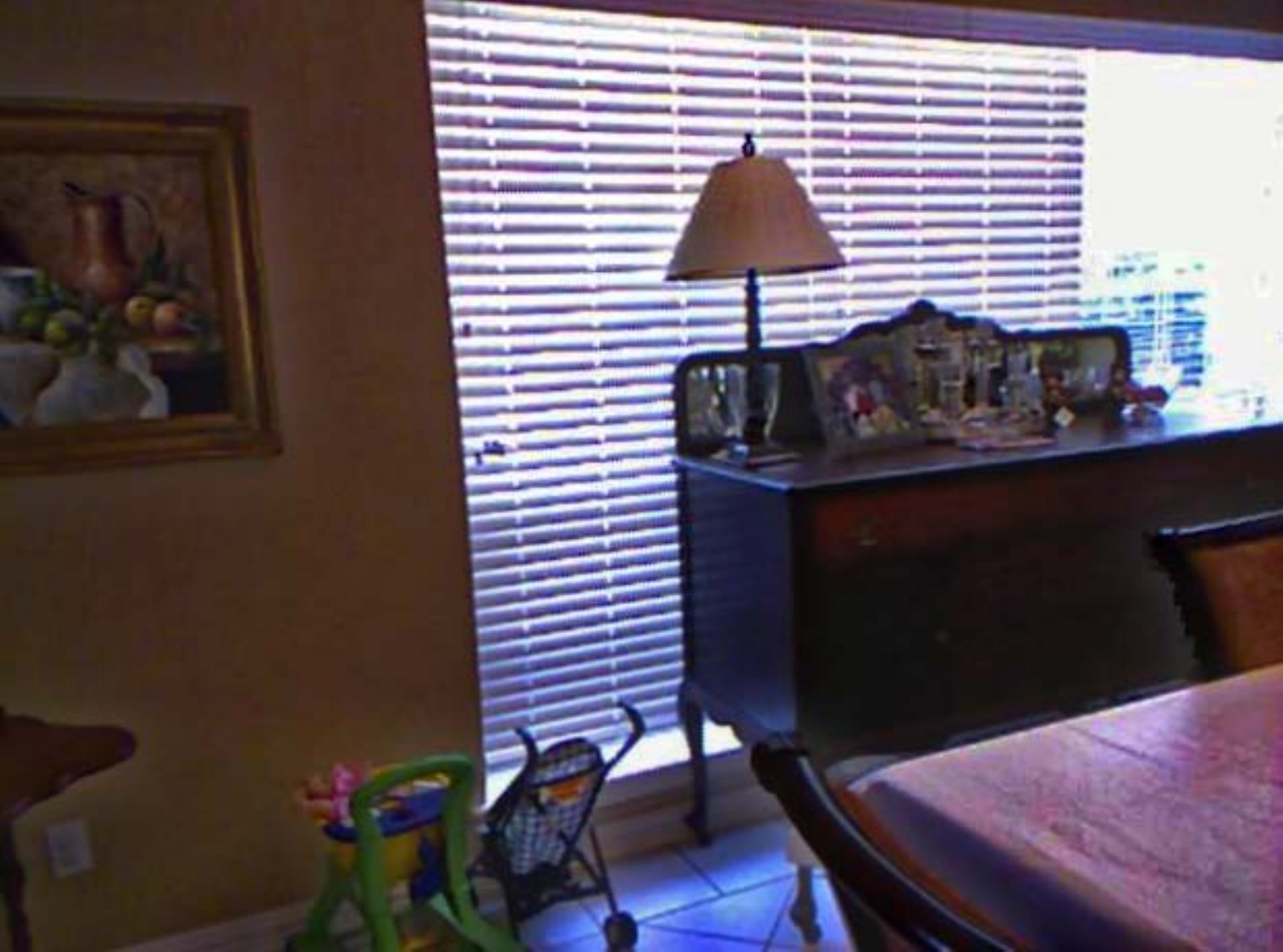} & \hspace{-0.4cm}
			\includegraphics[width = 0.095\textwidth]{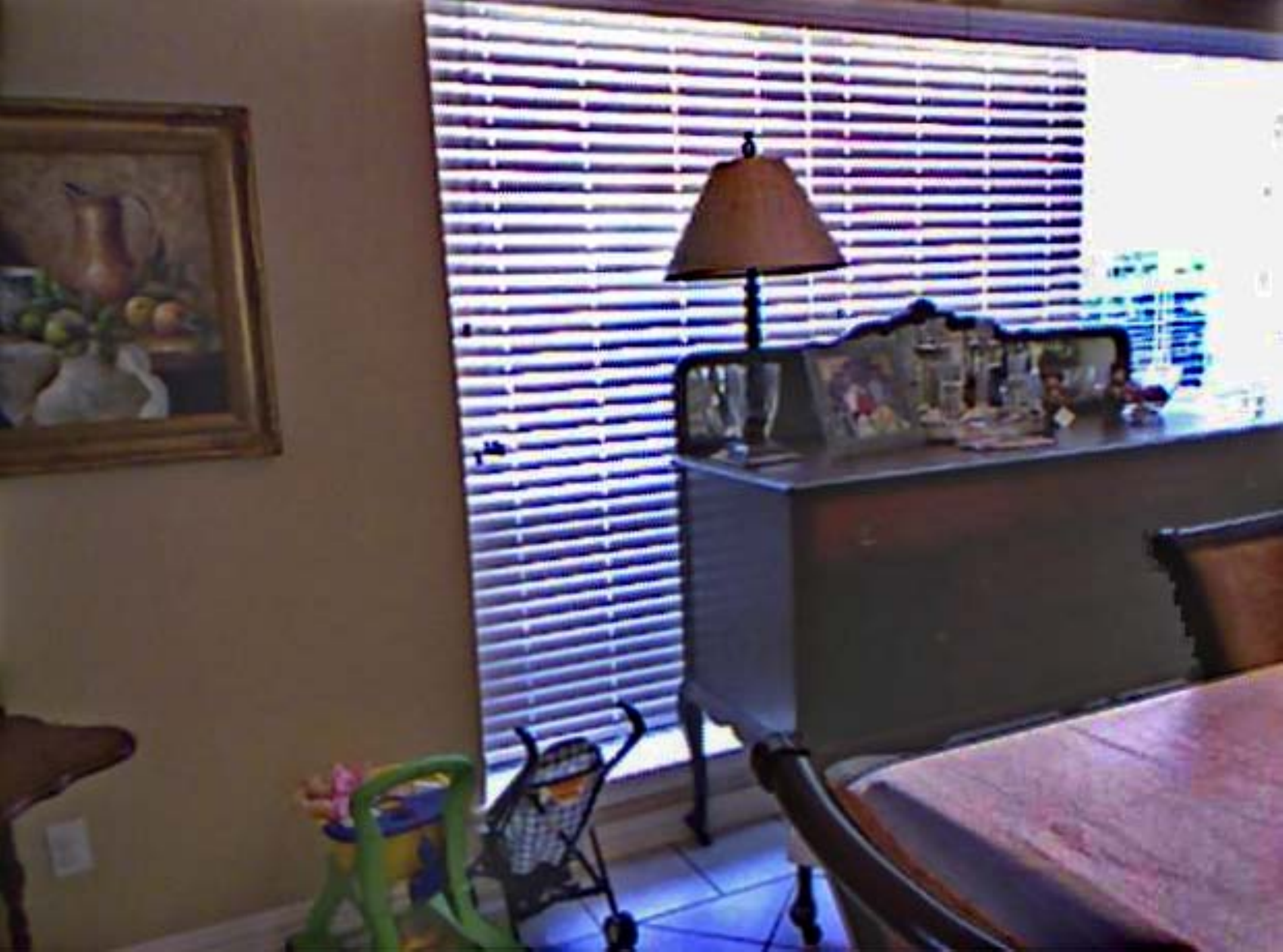} & \hspace{-0.4cm}
			\includegraphics[width = 0.095\textwidth]{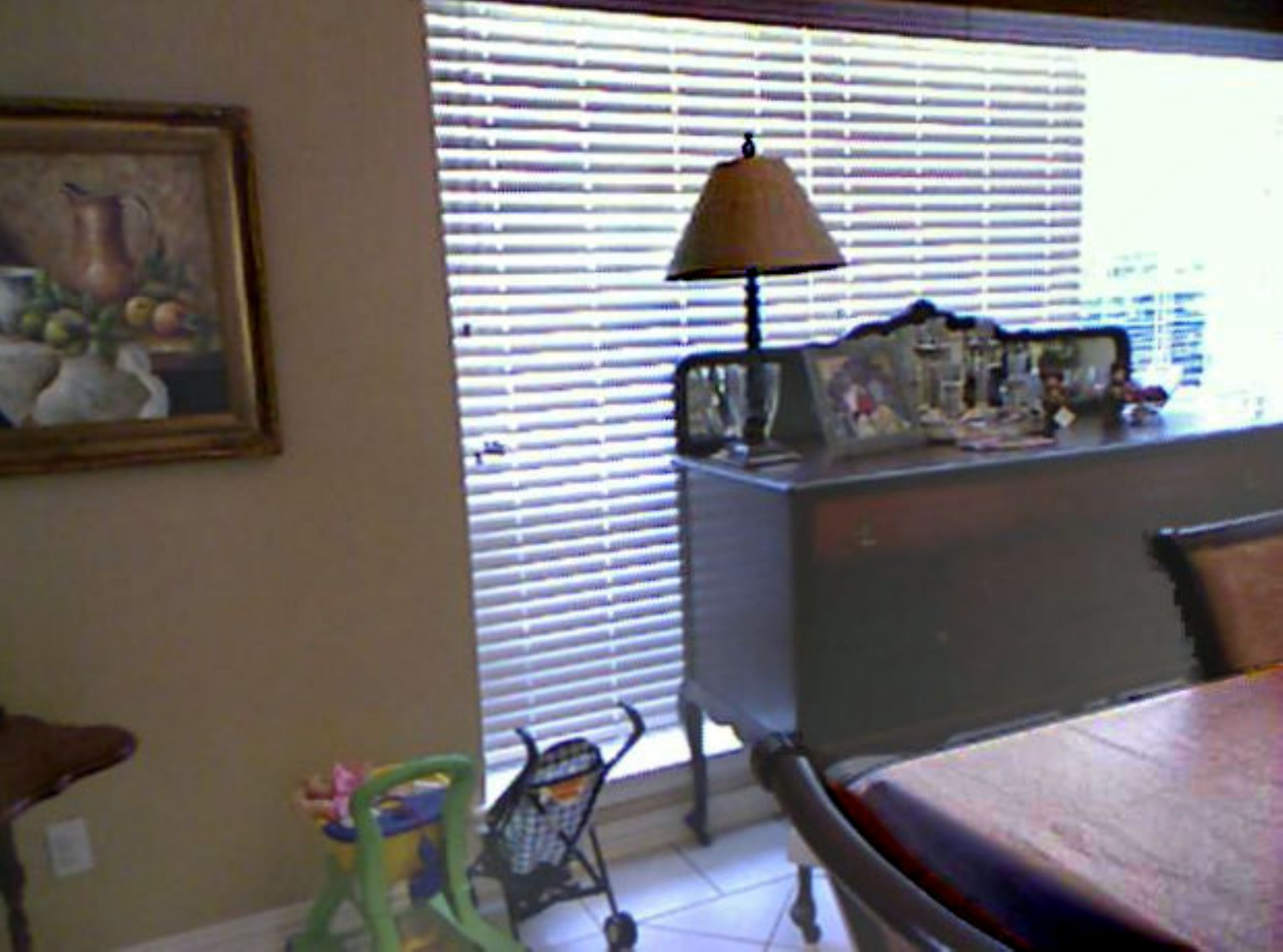} & \hspace{-0.4cm}
			\includegraphics[width = 0.095\textwidth]{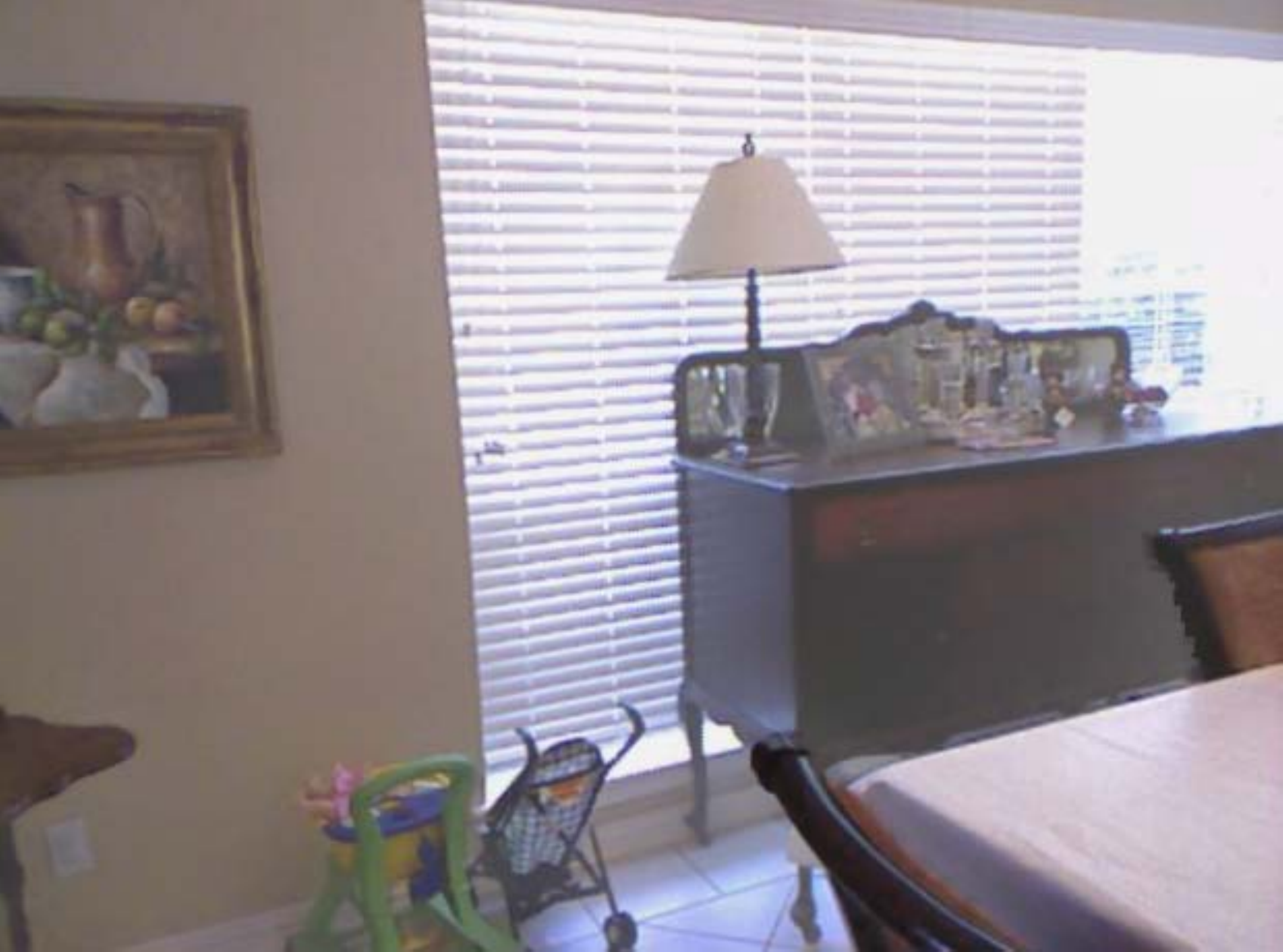} & \hspace{-0.4cm}
			\includegraphics[width = 0.095\textwidth]{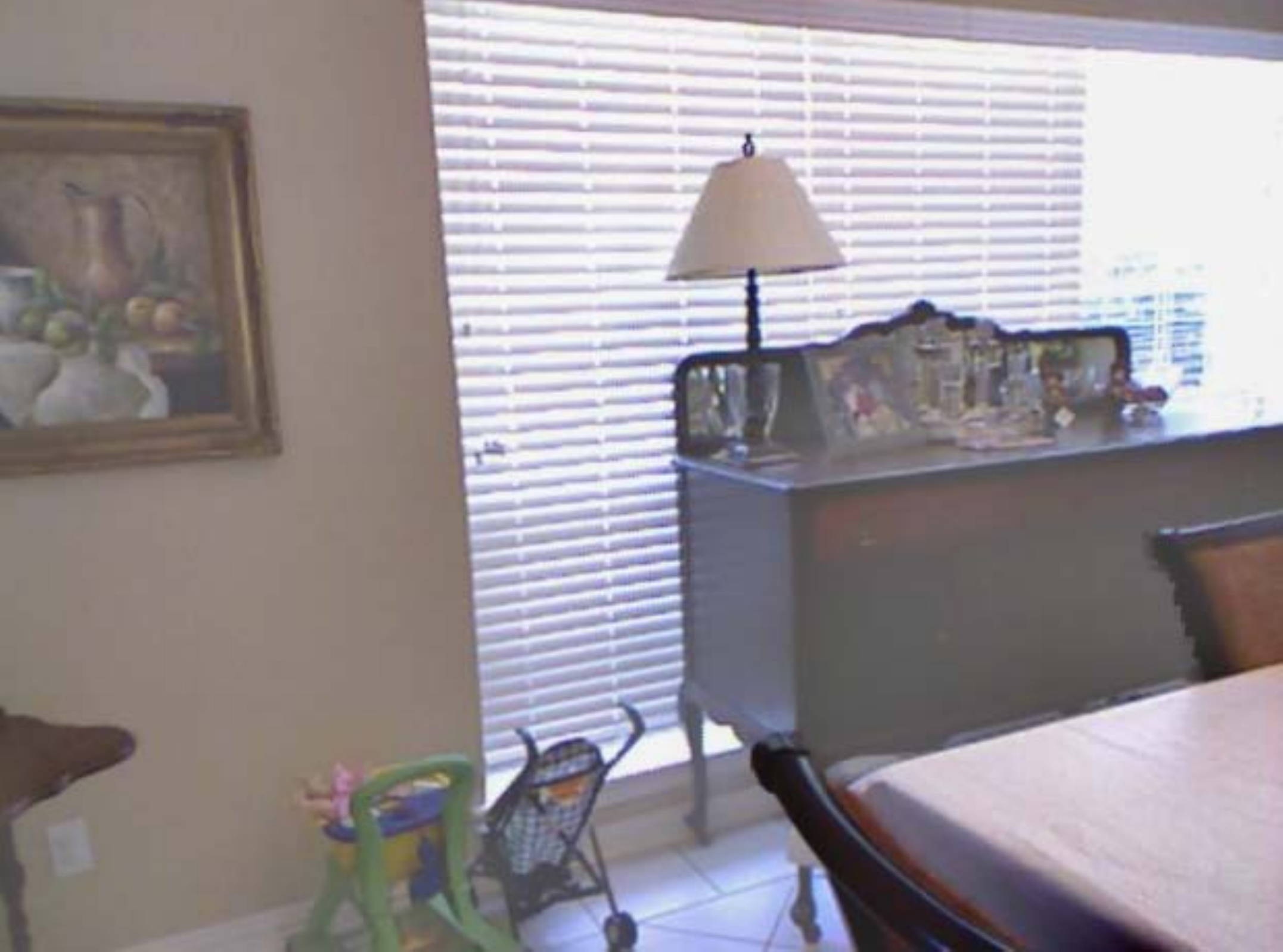} & \hspace{-0.4cm}
			\includegraphics[width = 0.095\textwidth]{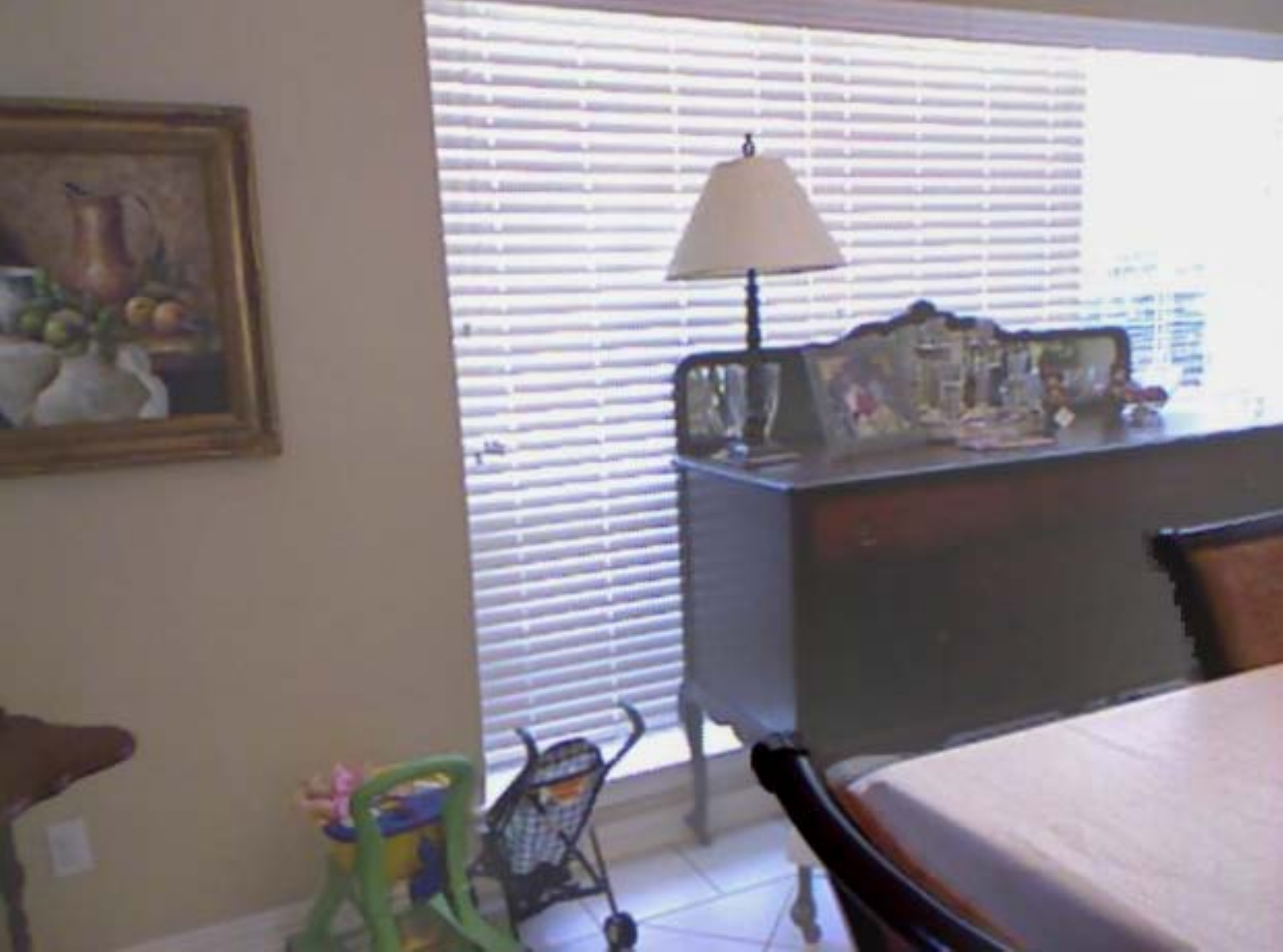} & \hspace{-0.4cm}
			\includegraphics[width = 0.095\textwidth]{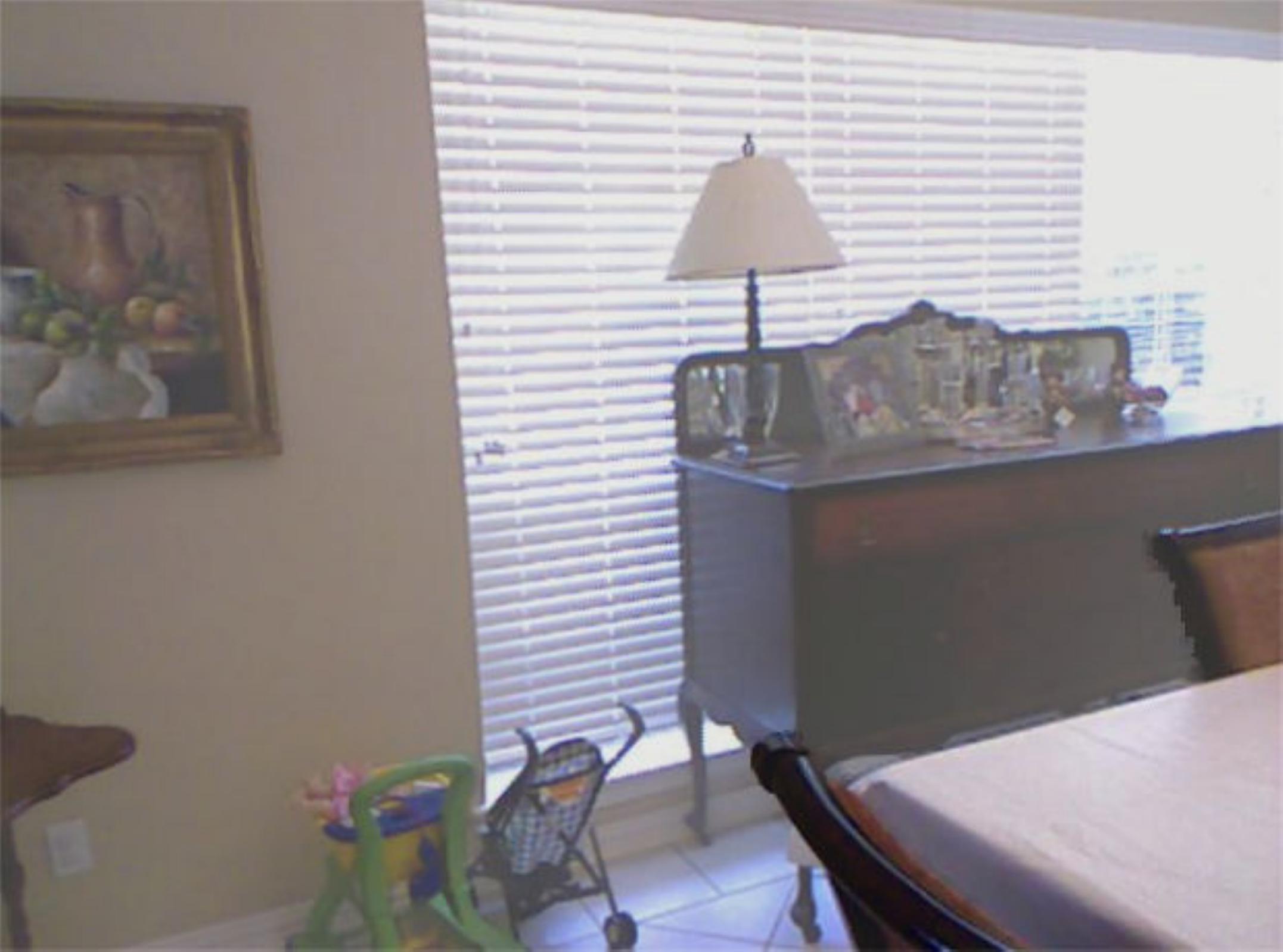} & \hspace{-0.4cm}
			\includegraphics[width = 0.095\textwidth]{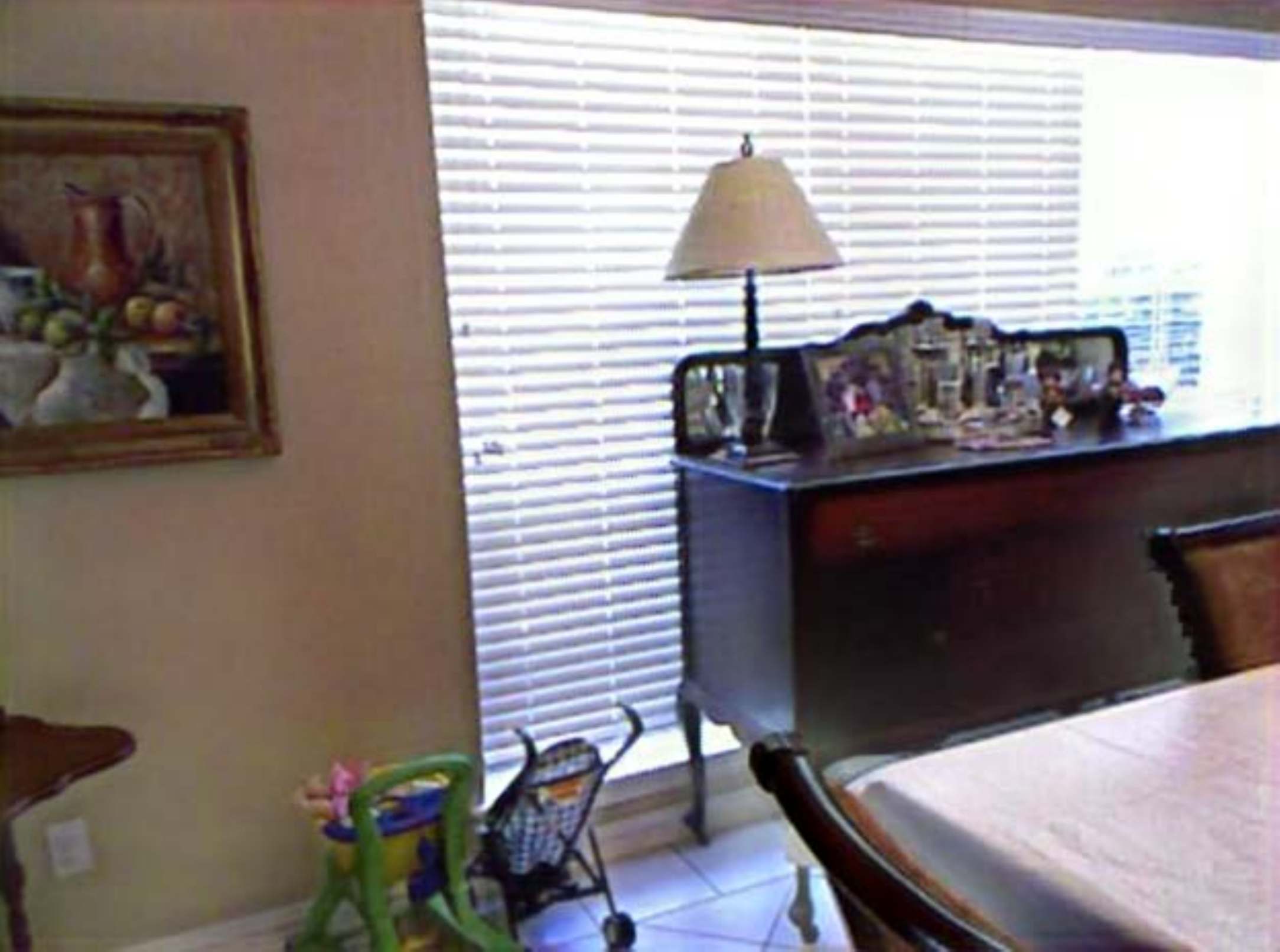} & \hspace{-0.4cm}
			\includegraphics[width = 0.095\textwidth]{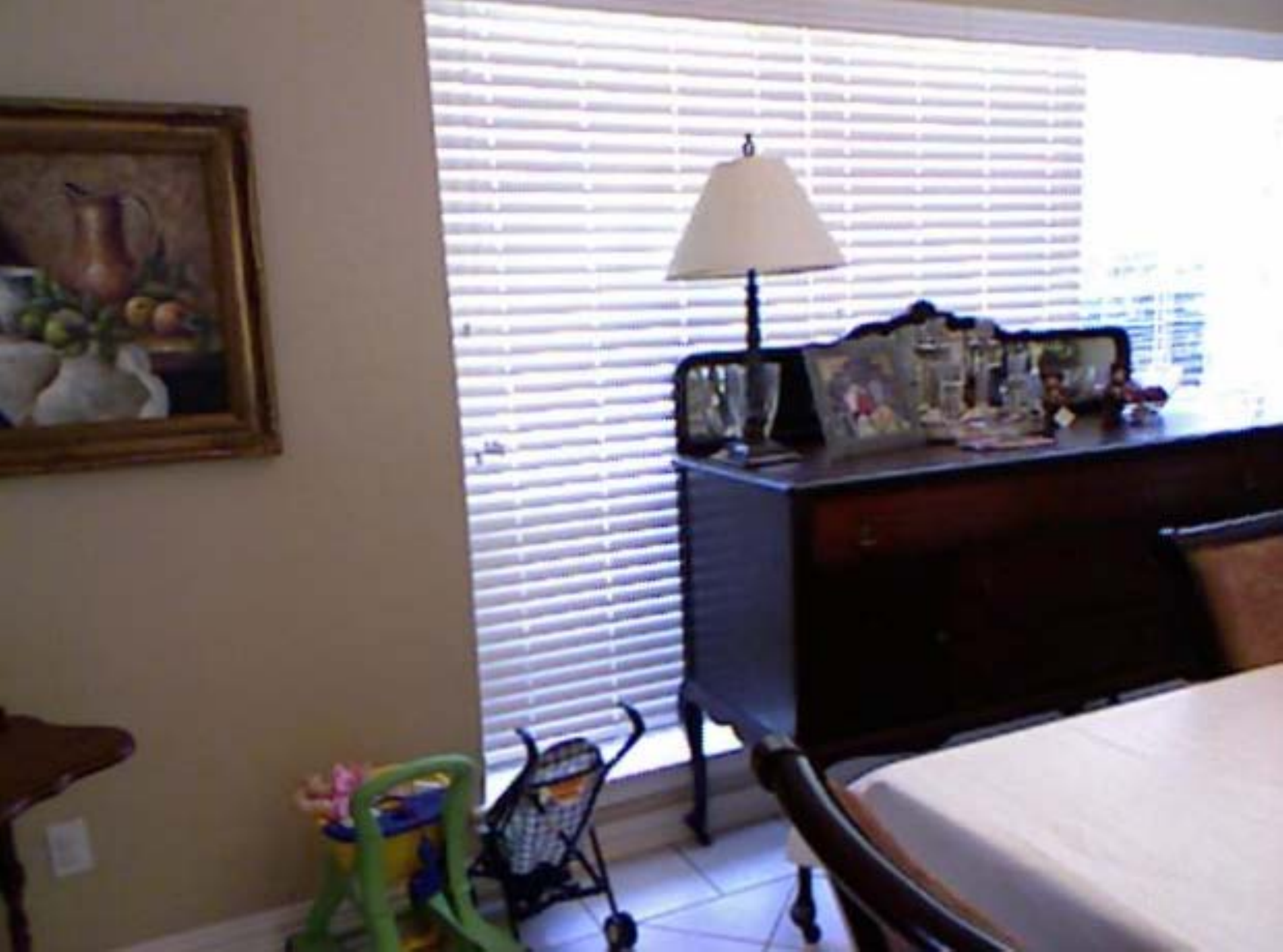}\\
			\includegraphics[width = 0.095\textwidth]{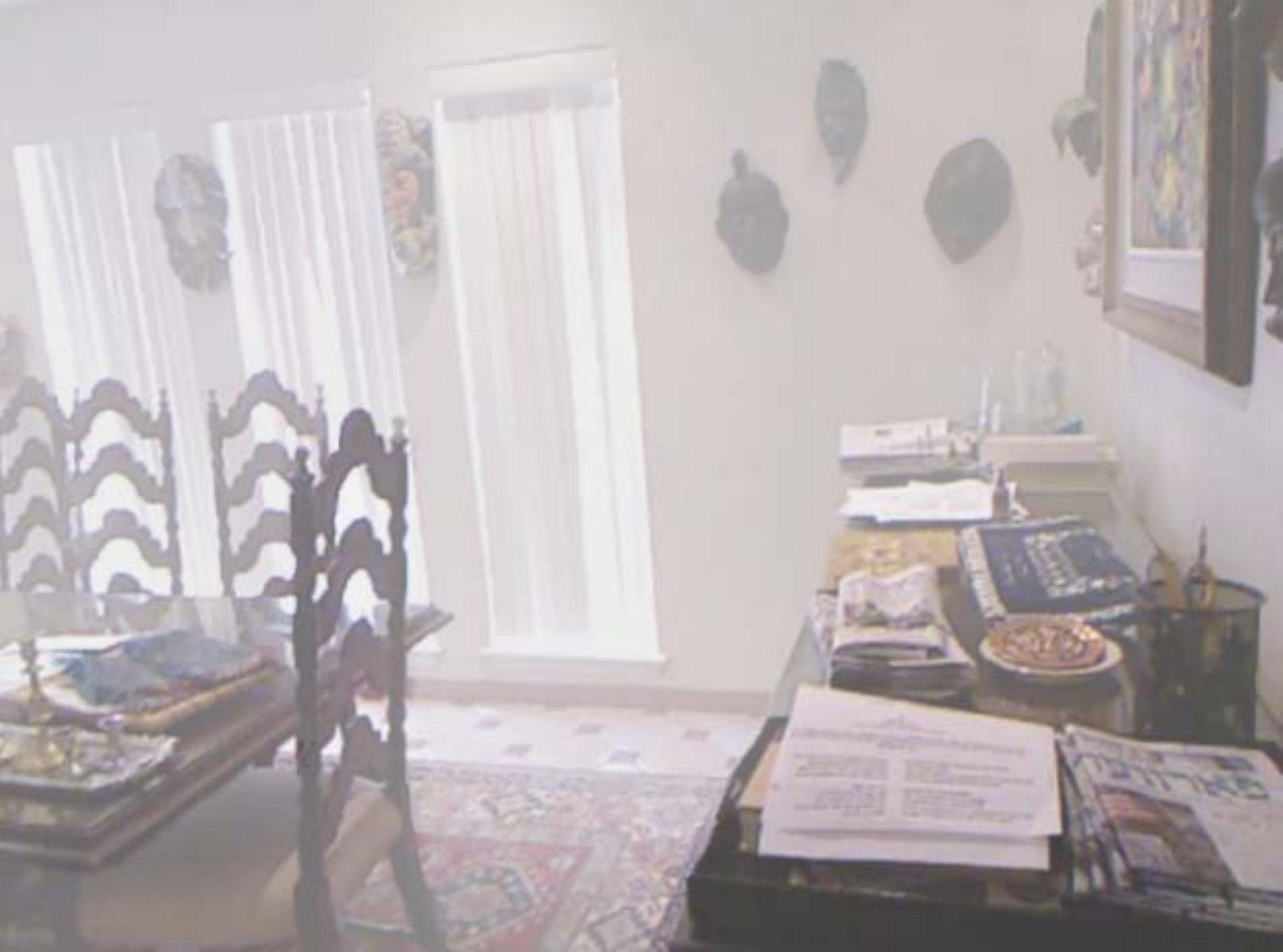} & \hspace{-0.4cm}
			\includegraphics[width = 0.095\textwidth]{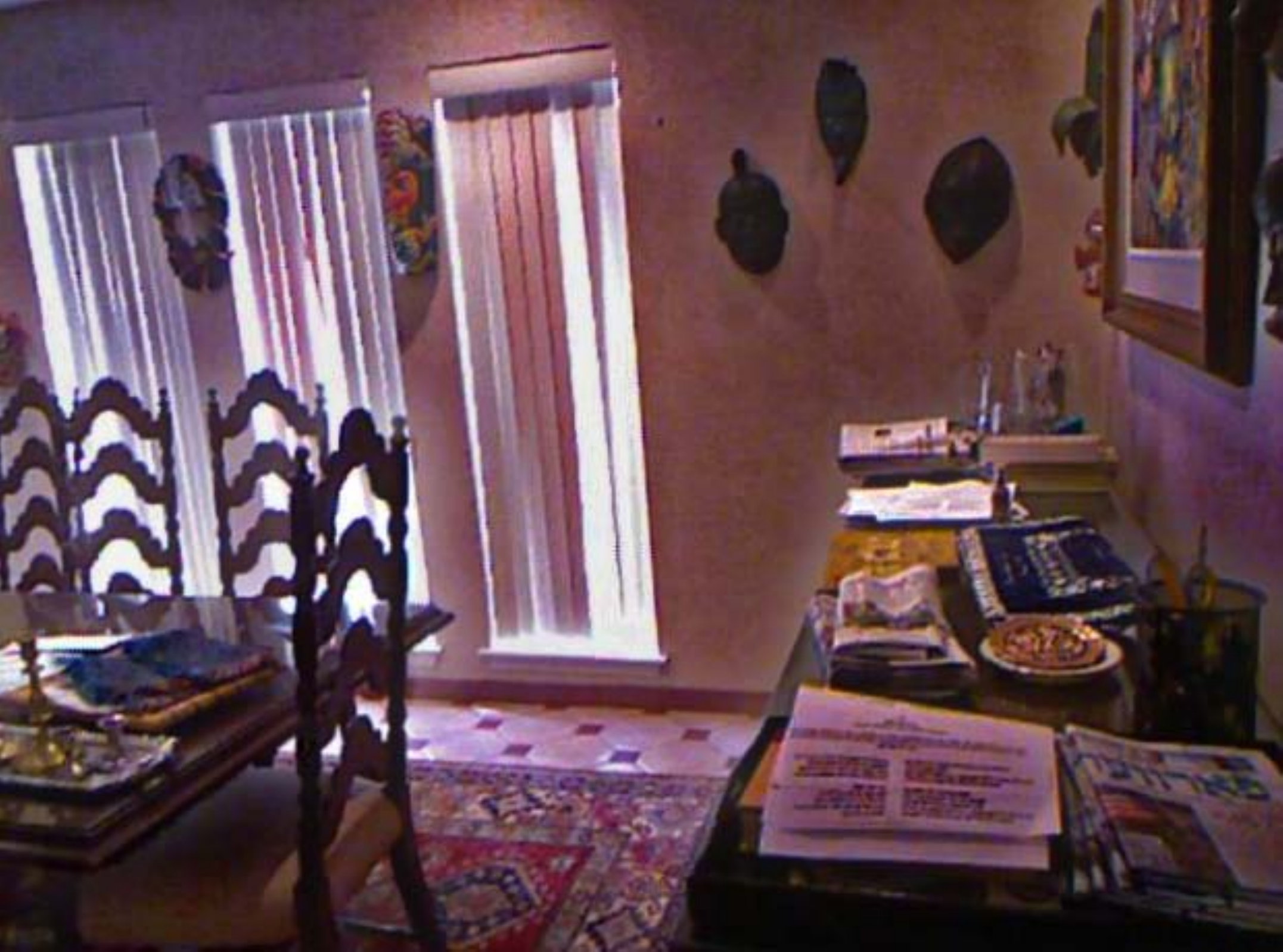} & \hspace{-0.4cm}
			\includegraphics[width = 0.095\textwidth]{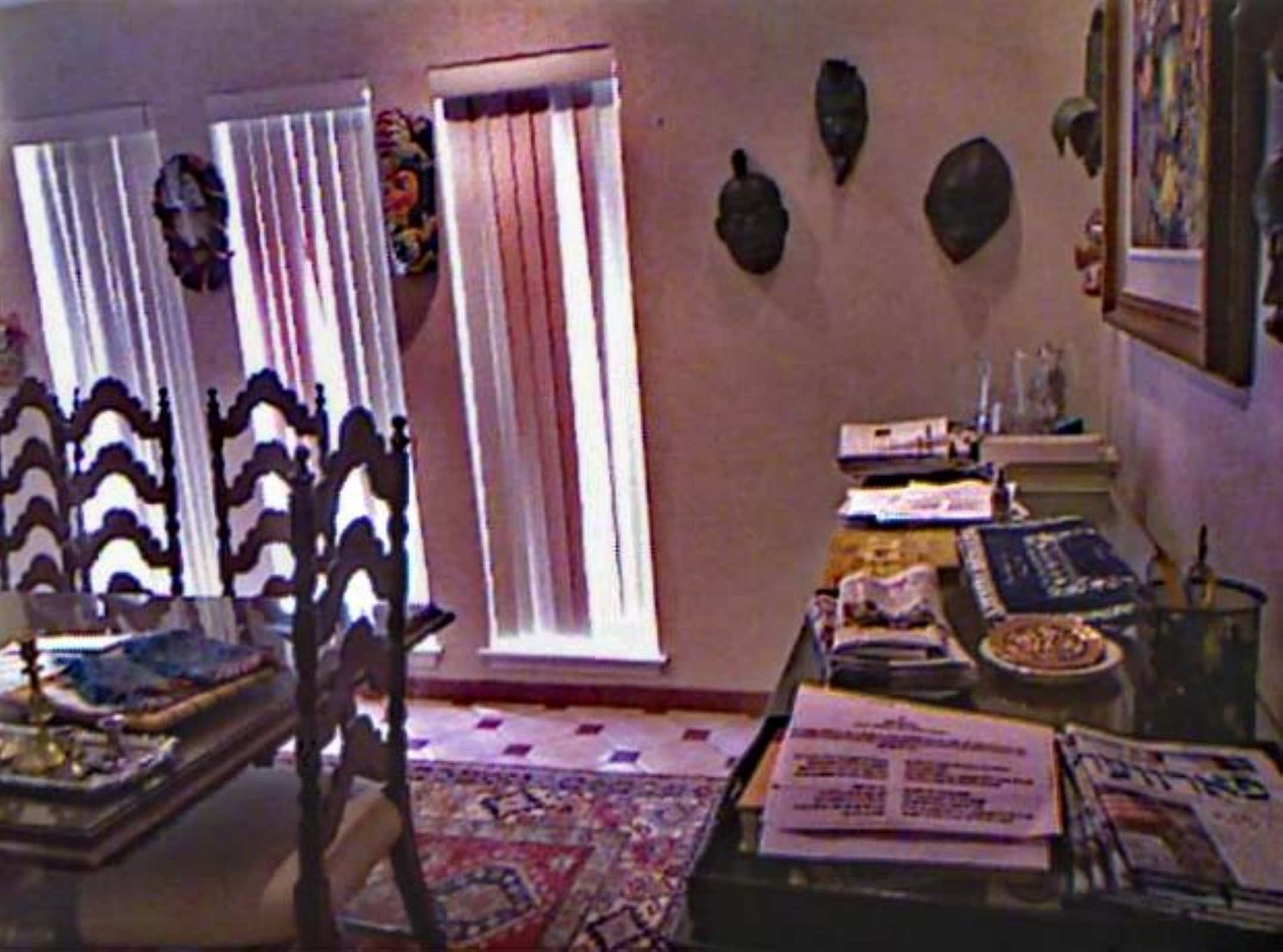} & \hspace{-0.4cm}
			\includegraphics[width = 0.095\textwidth]{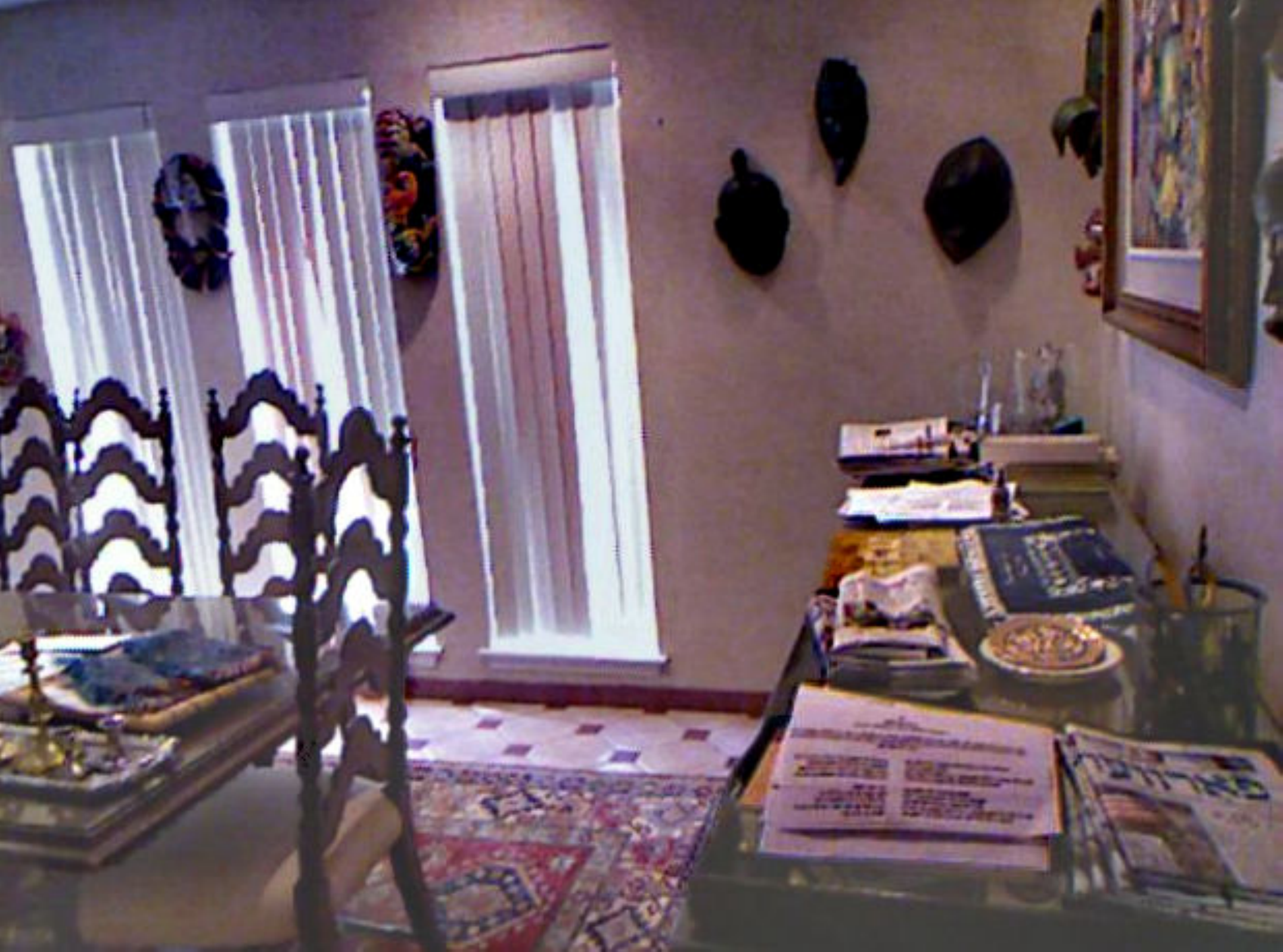} & \hspace{-0.4cm}
			\includegraphics[width = 0.095\textwidth]{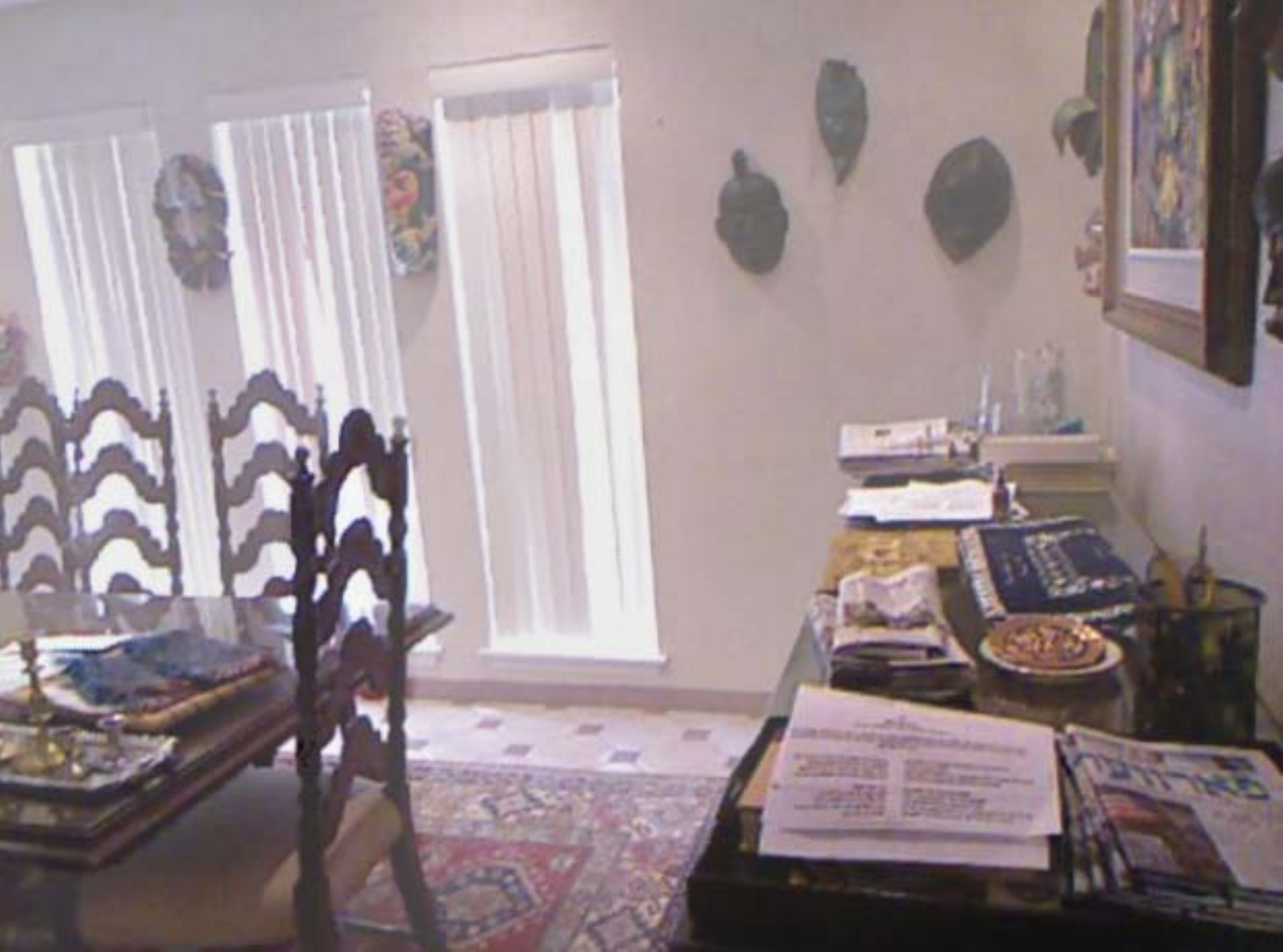} & \hspace{-0.4cm}
			\includegraphics[width = 0.095\textwidth]{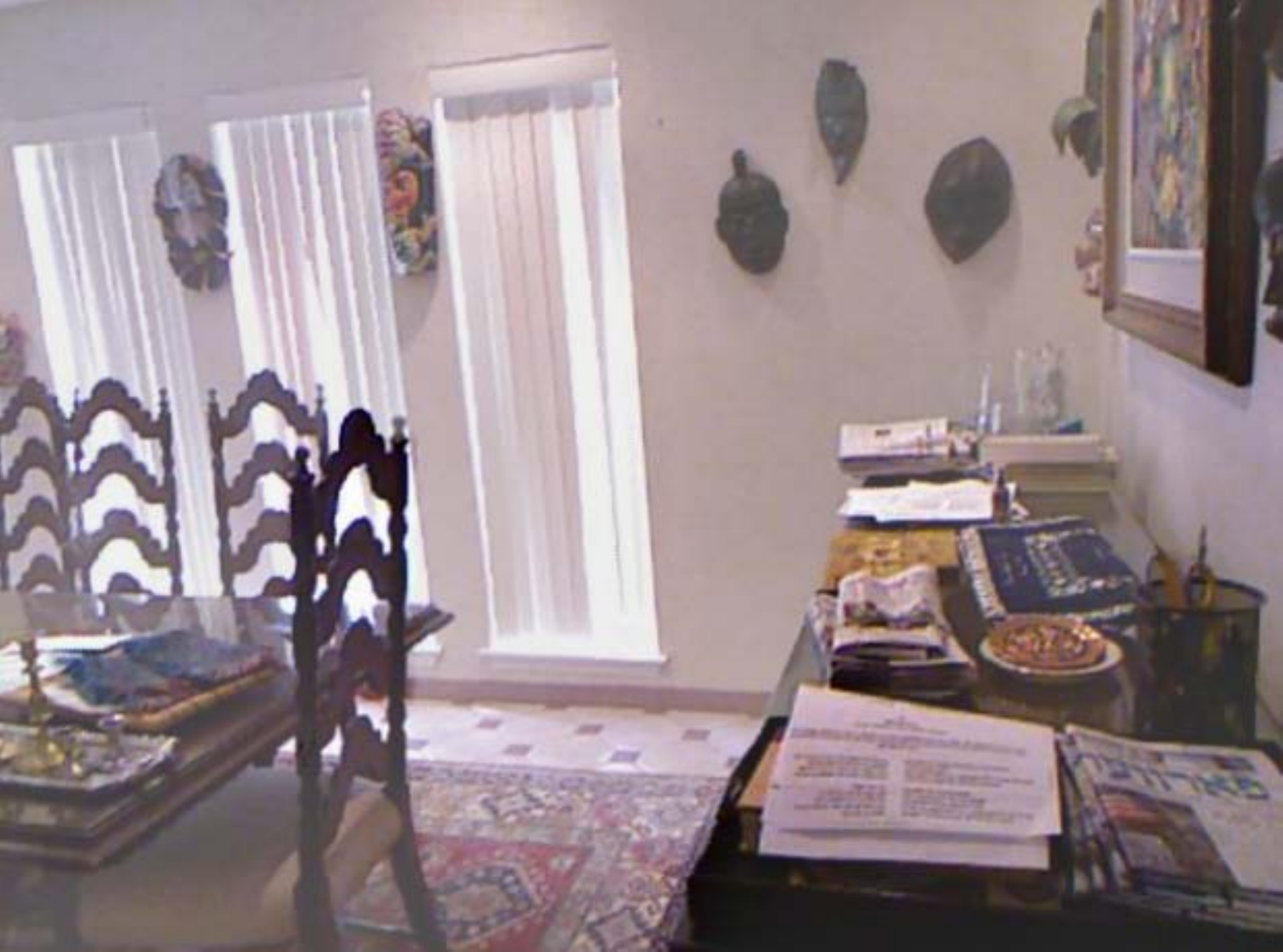} & \hspace{-0.4cm}
			\includegraphics[width = 0.095\textwidth]{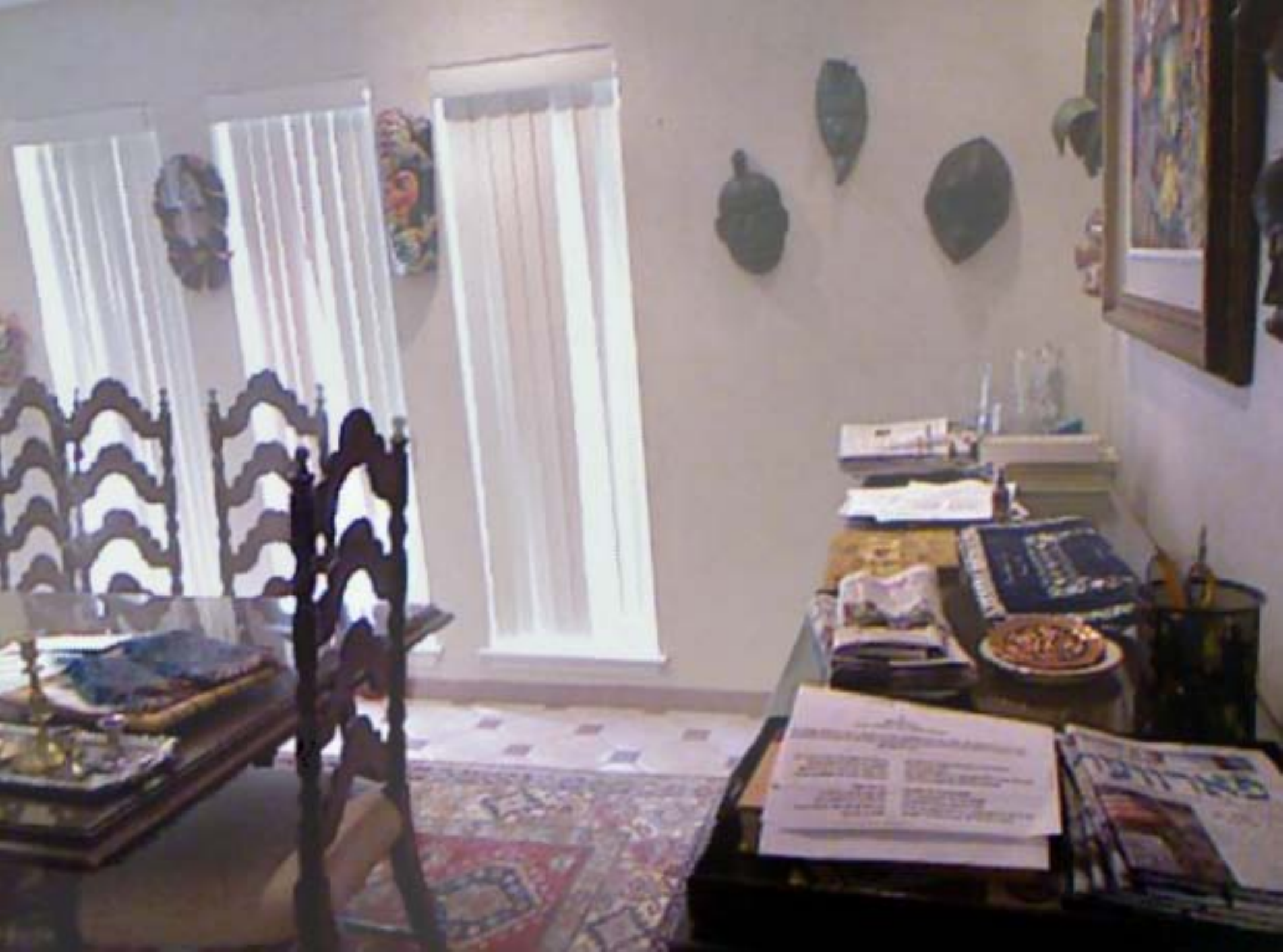} & \hspace{-0.4cm}
			\includegraphics[width = 0.095\textwidth]{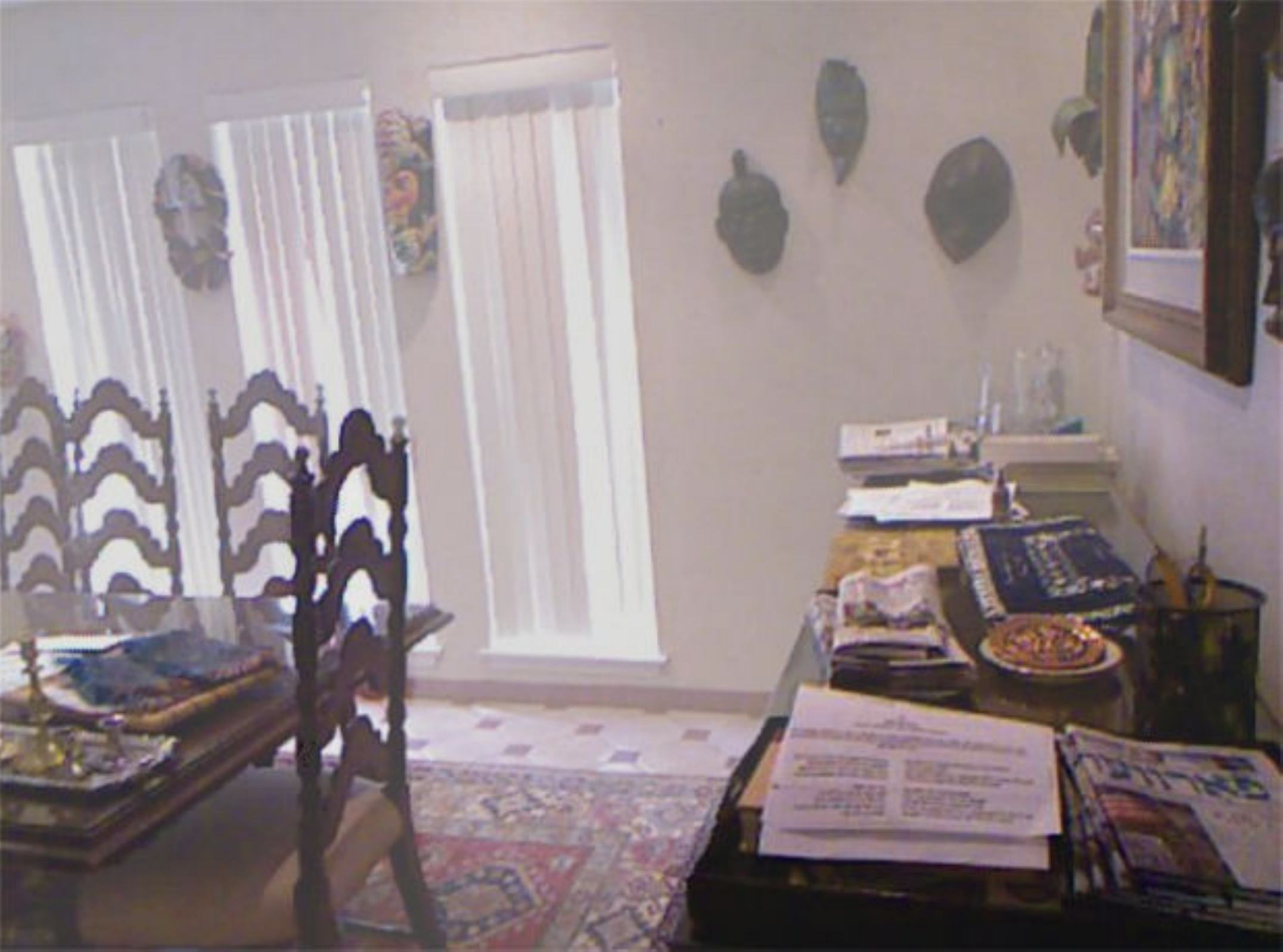} & \hspace{-0.4cm}
			\includegraphics[width = 0.095\textwidth]{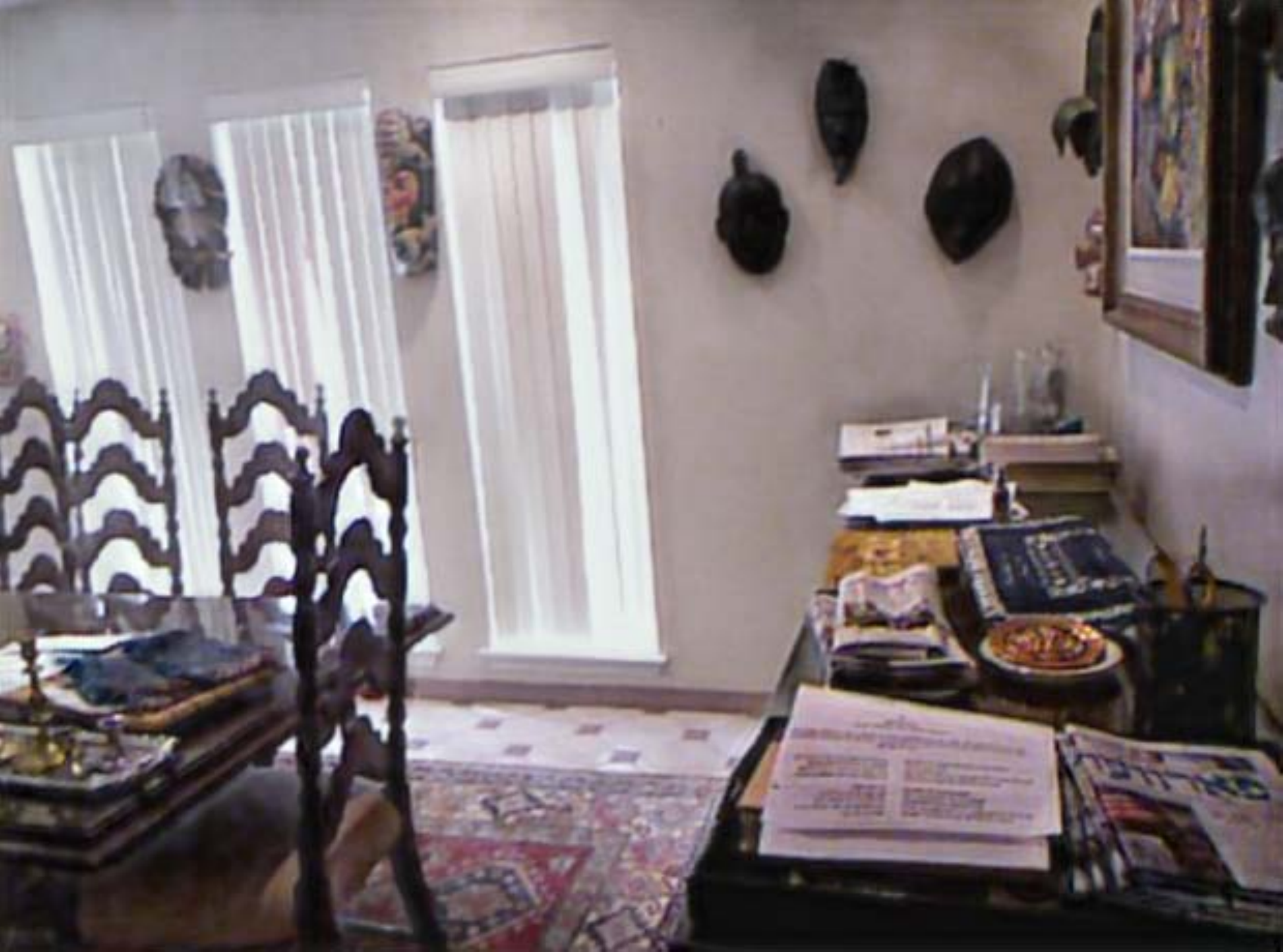} & \hspace{-0.4cm}
			\includegraphics[width = 0.095\textwidth]{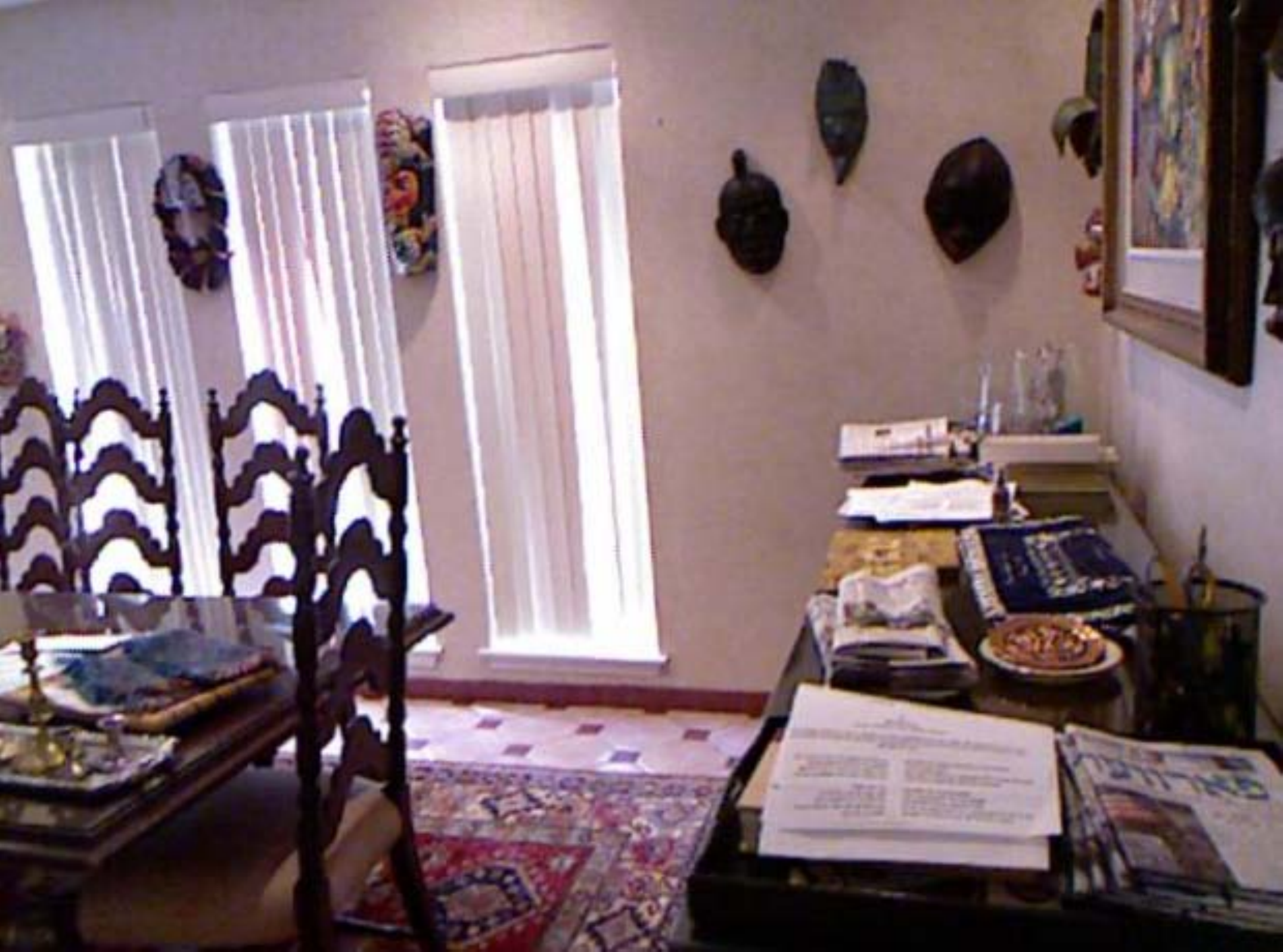}\\
			\includegraphics[width = 0.095\textwidth]{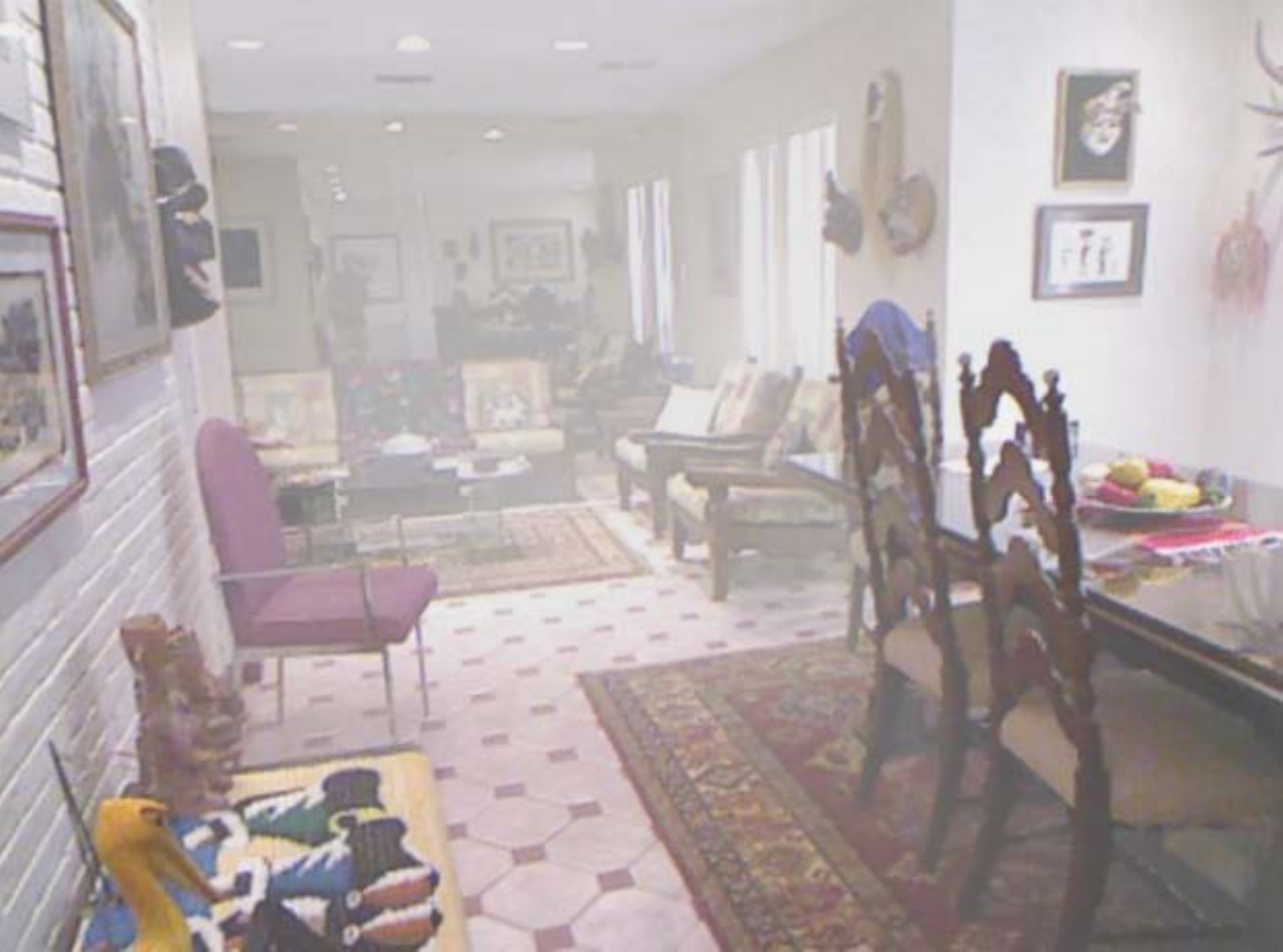} & \hspace{-0.4cm}
			\includegraphics[width = 0.095\textwidth]{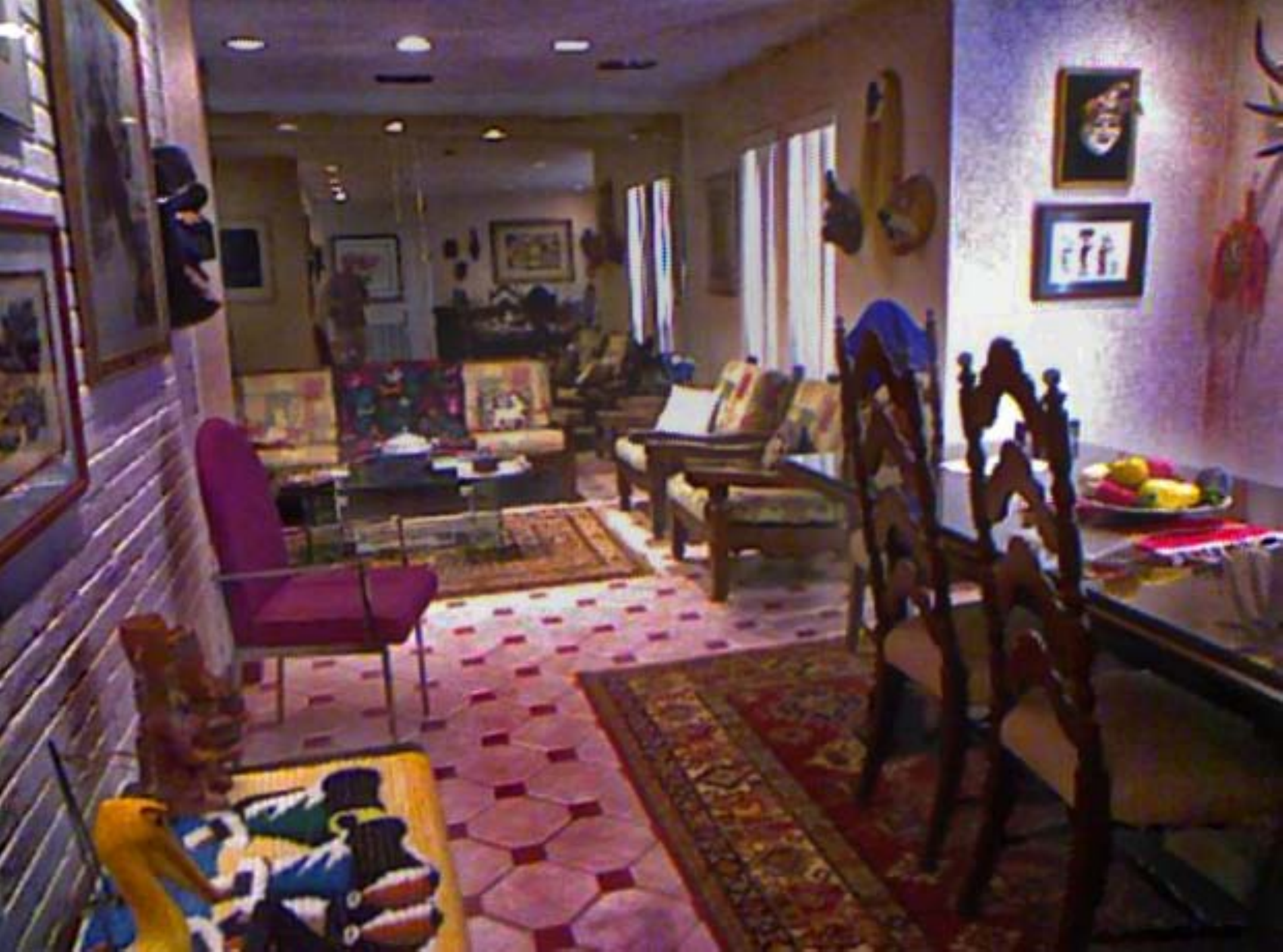} & \hspace{-0.4cm}
			\includegraphics[width = 0.095\textwidth]{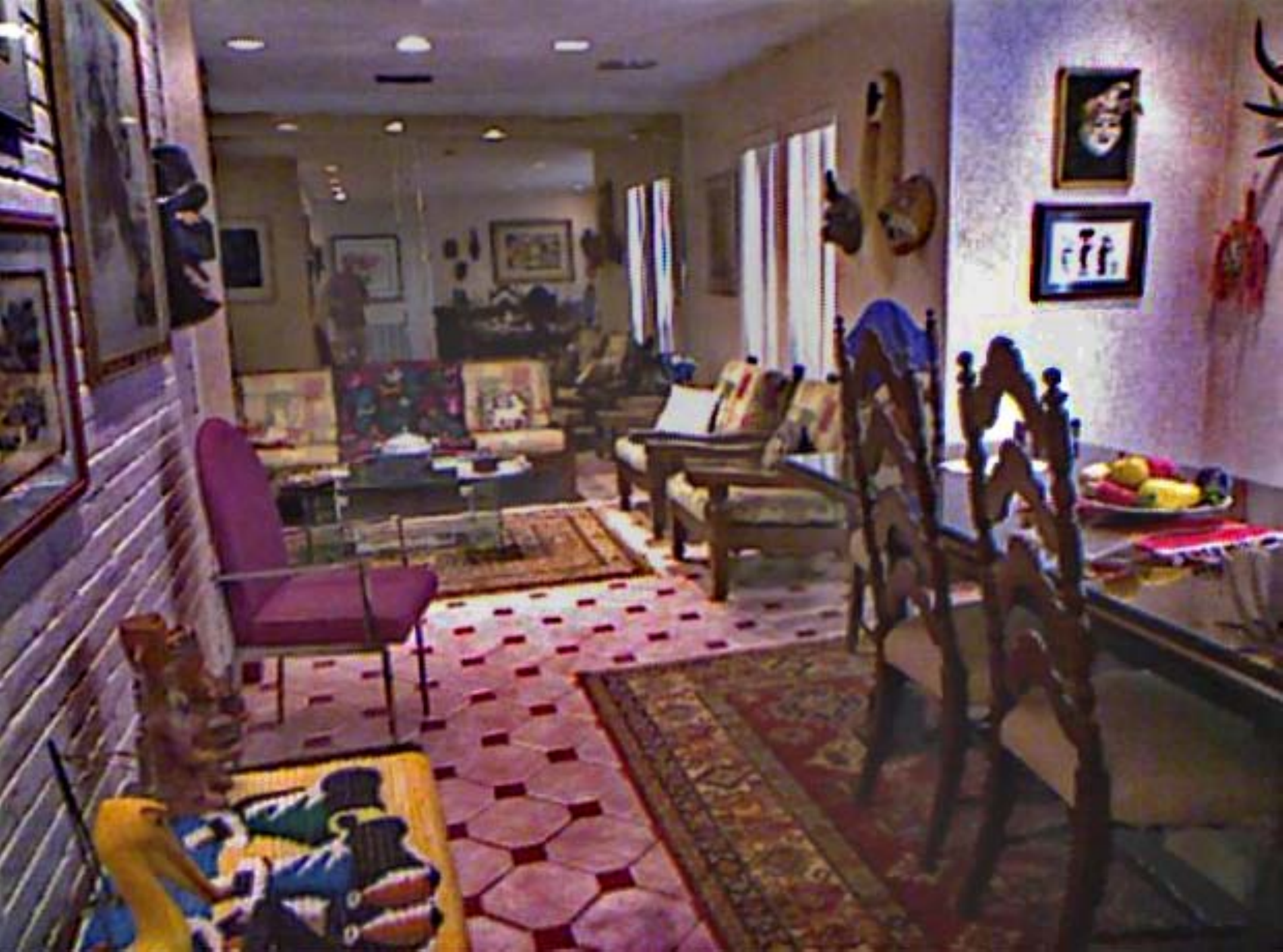} & \hspace{-0.4cm}
			\includegraphics[width = 0.095\textwidth]{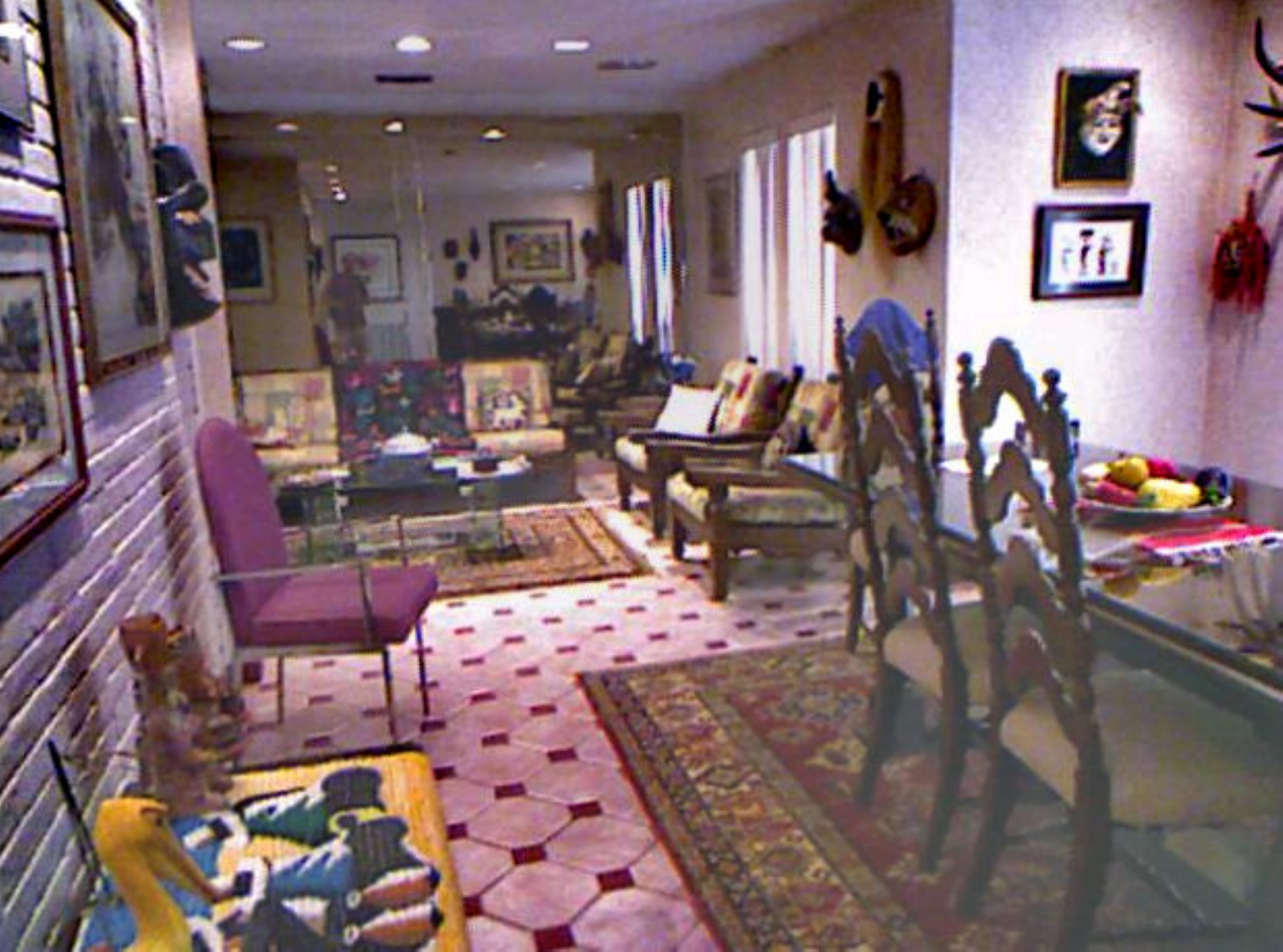} & \hspace{-0.4cm}
			\includegraphics[width = 0.095\textwidth]{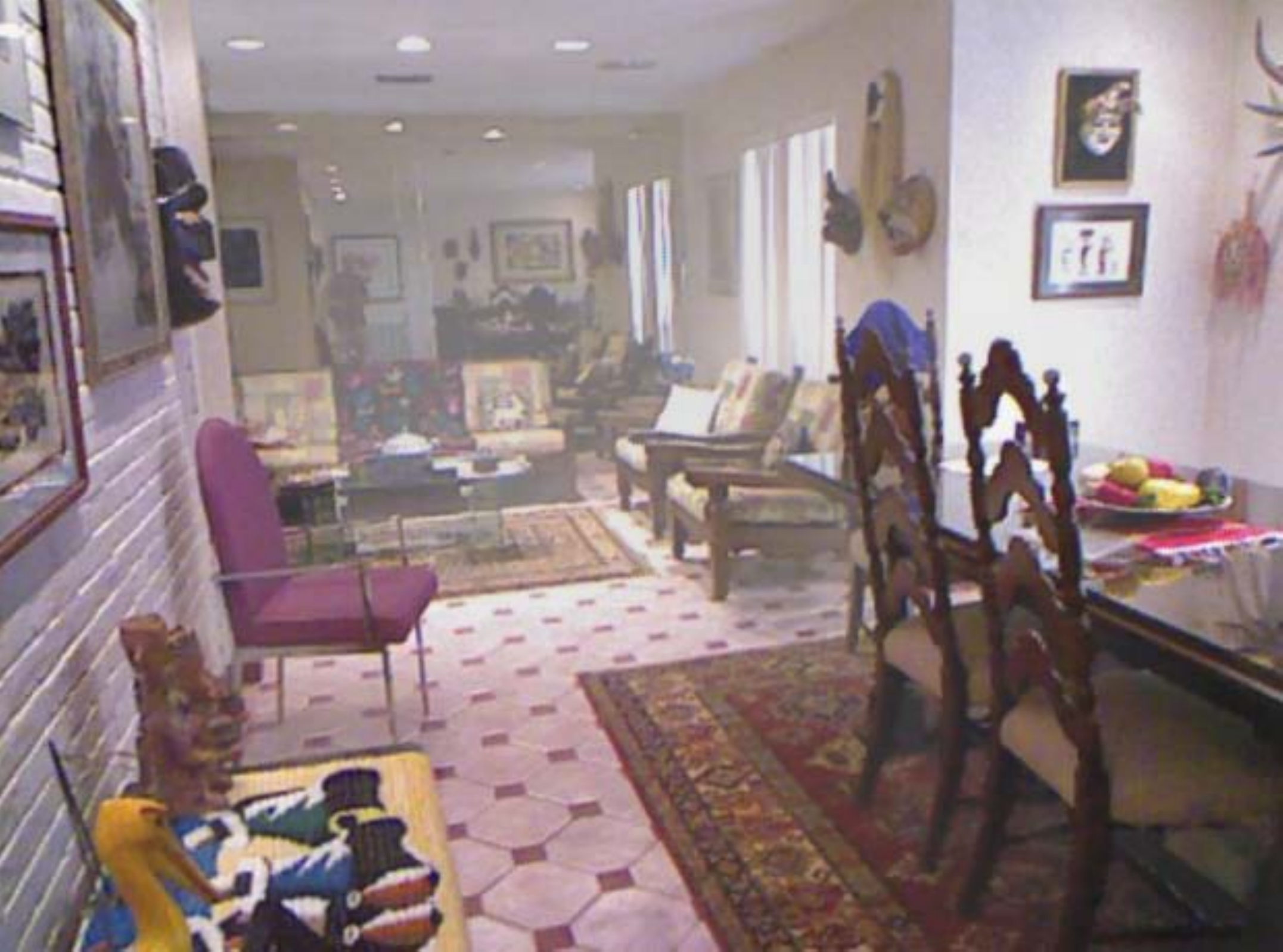} & \hspace{-0.4cm}
			\includegraphics[width = 0.095\textwidth]{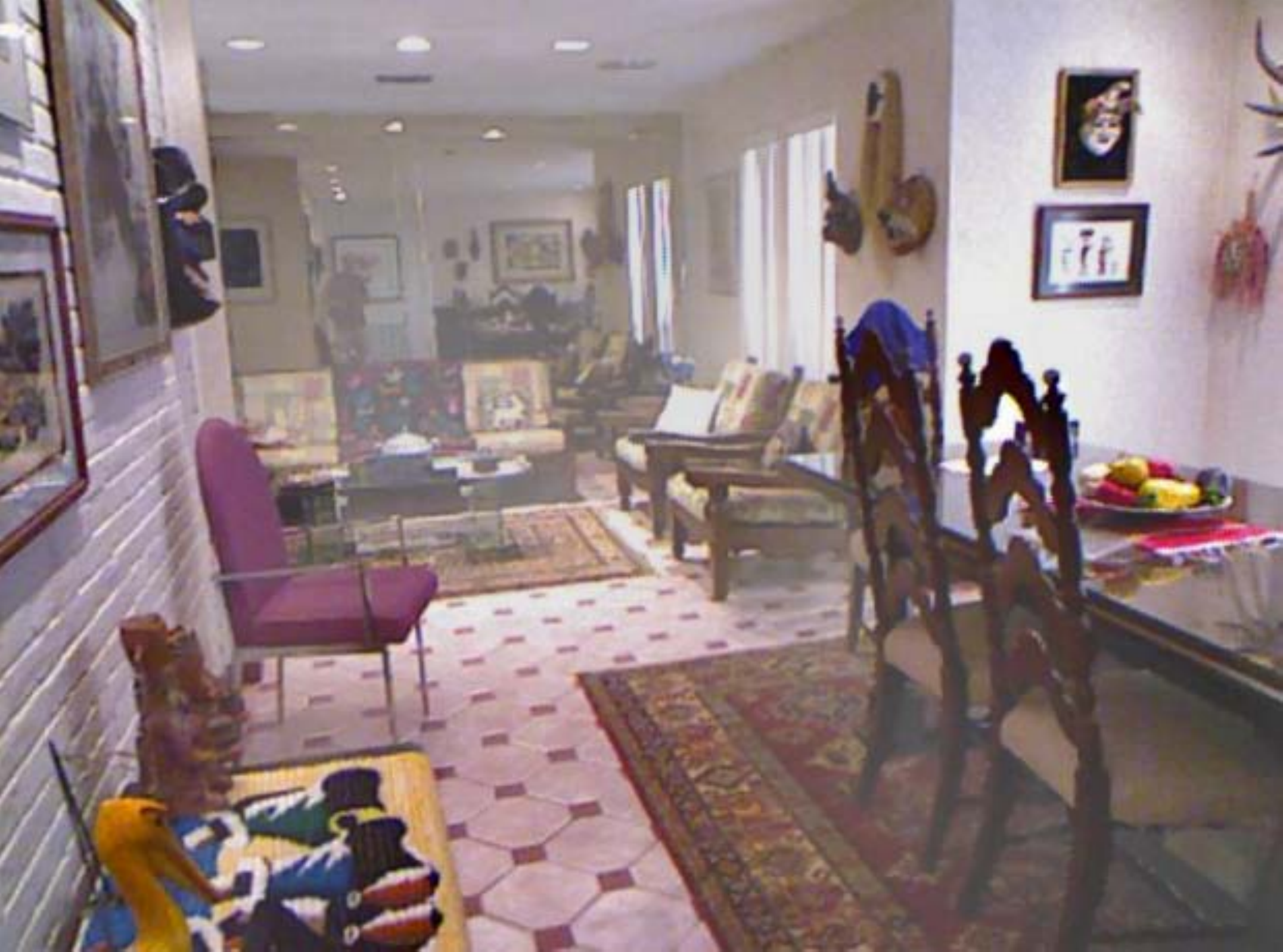} & \hspace{-0.4cm}
			\includegraphics[width = 0.095\textwidth]{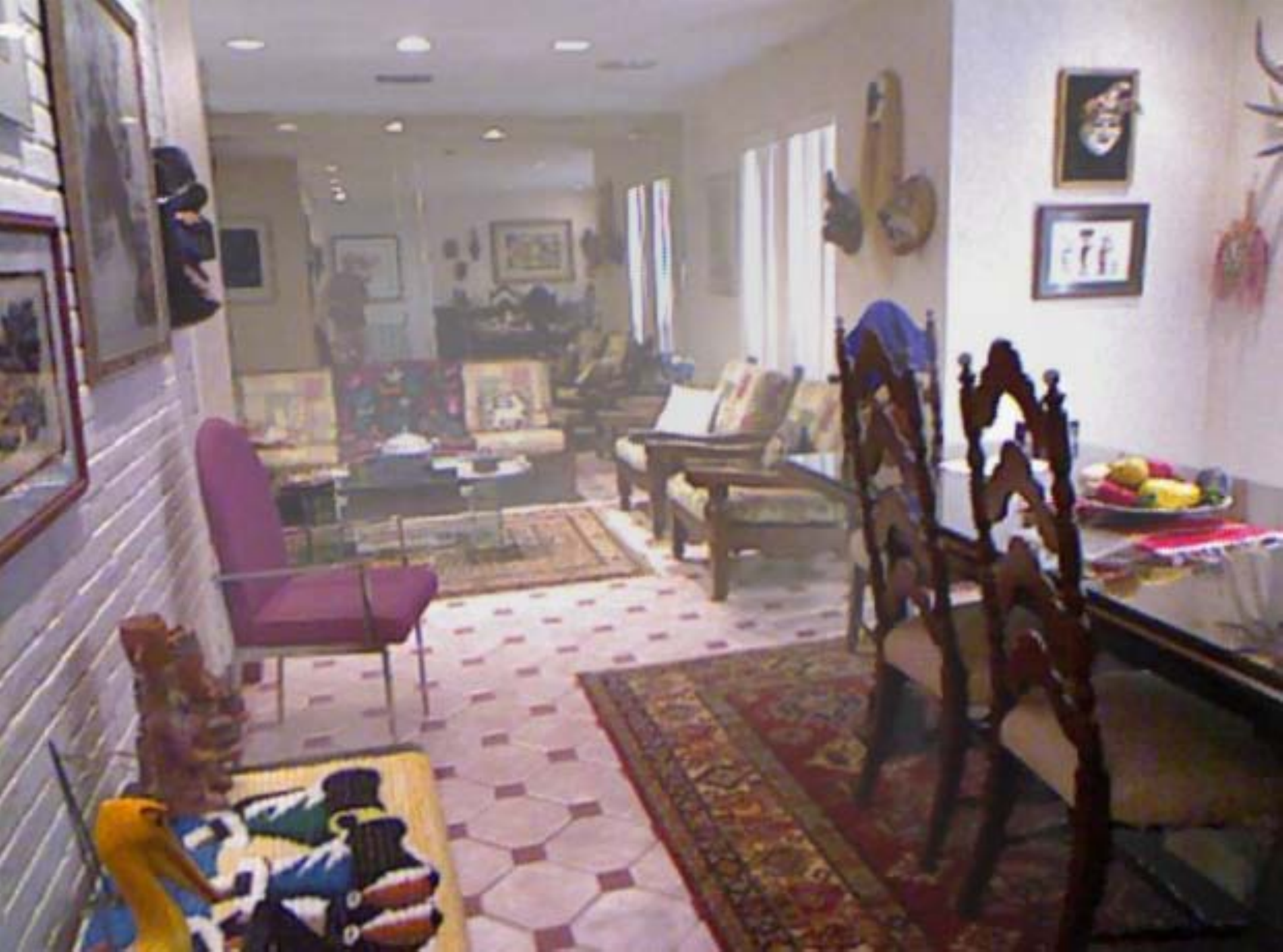} & \hspace{-0.4cm}
			\includegraphics[width = 0.095\textwidth]{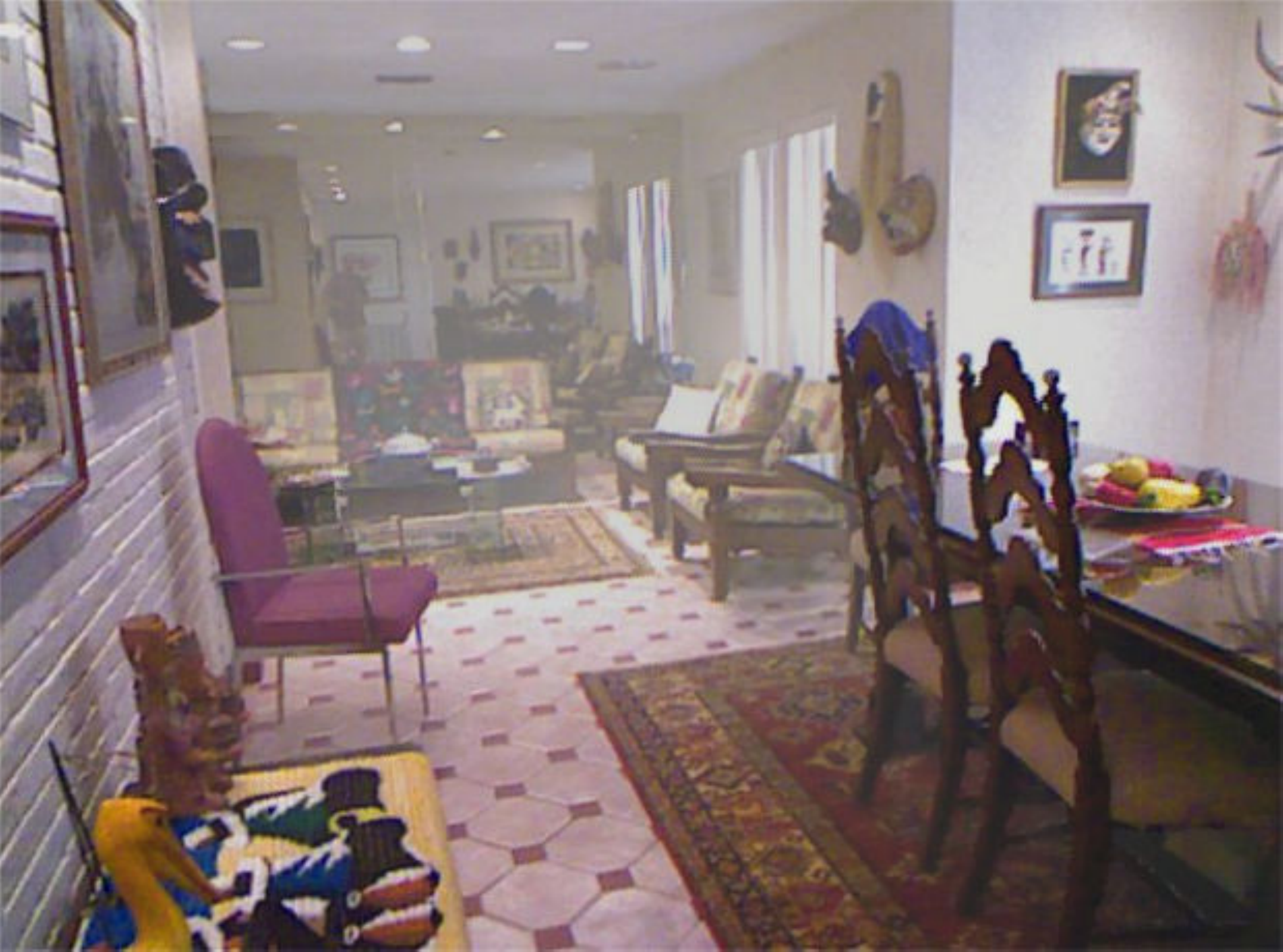} & \hspace{-0.4cm}
			\includegraphics[width = 0.095\textwidth]{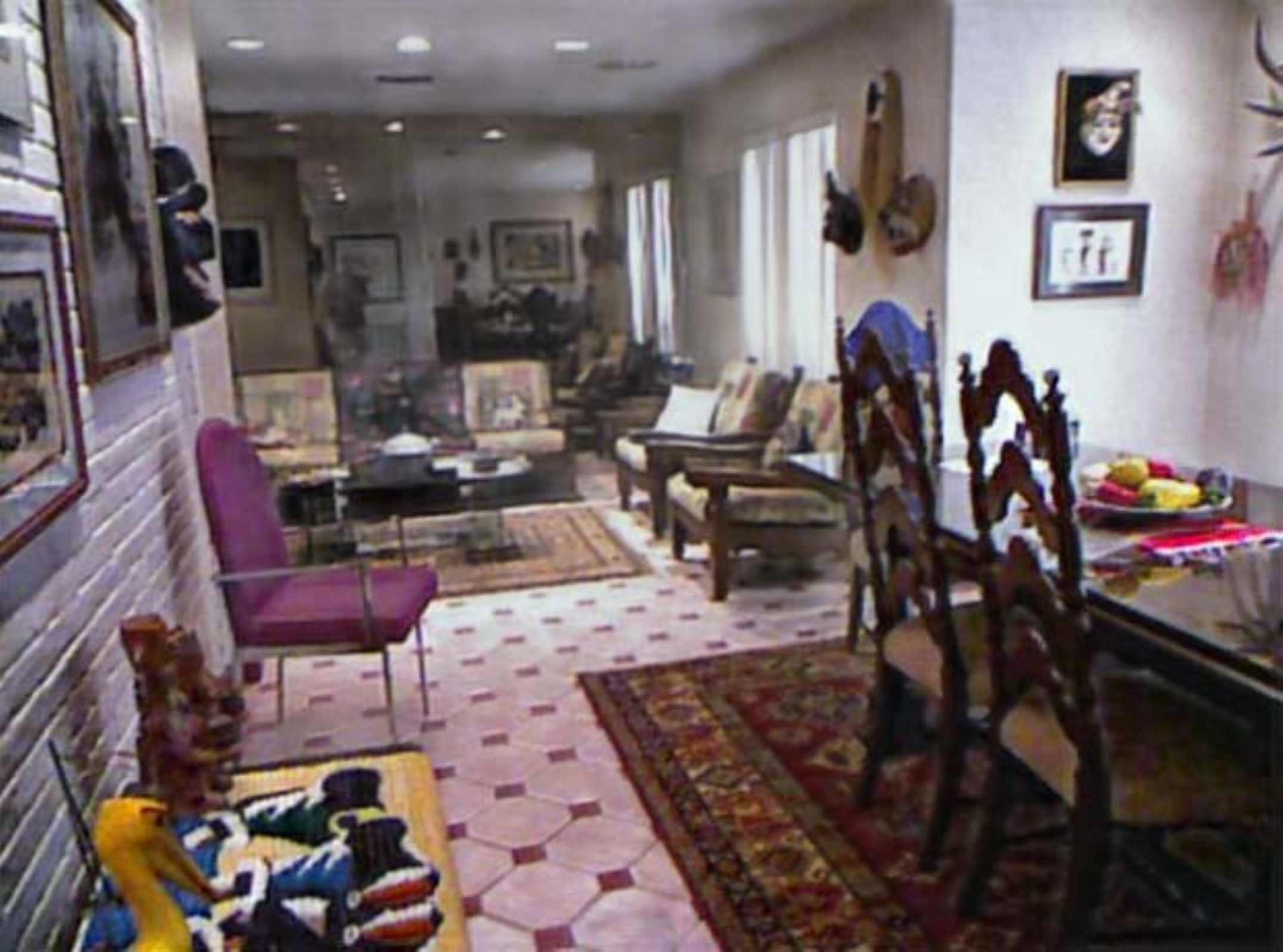} & \hspace{-0.4cm}
			\includegraphics[width = 0.095\textwidth]{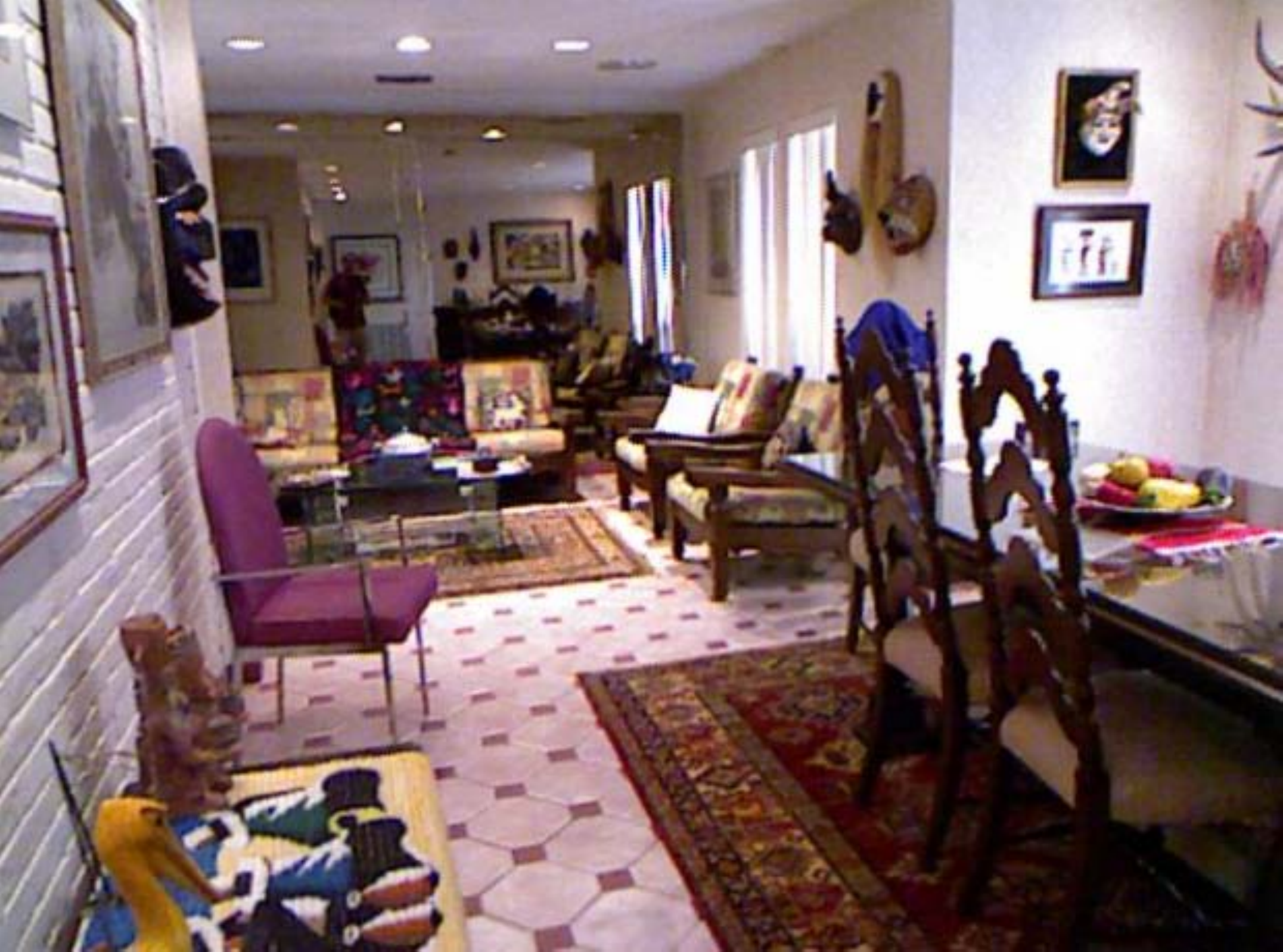}\\
			\includegraphics[width = 0.095\textwidth]{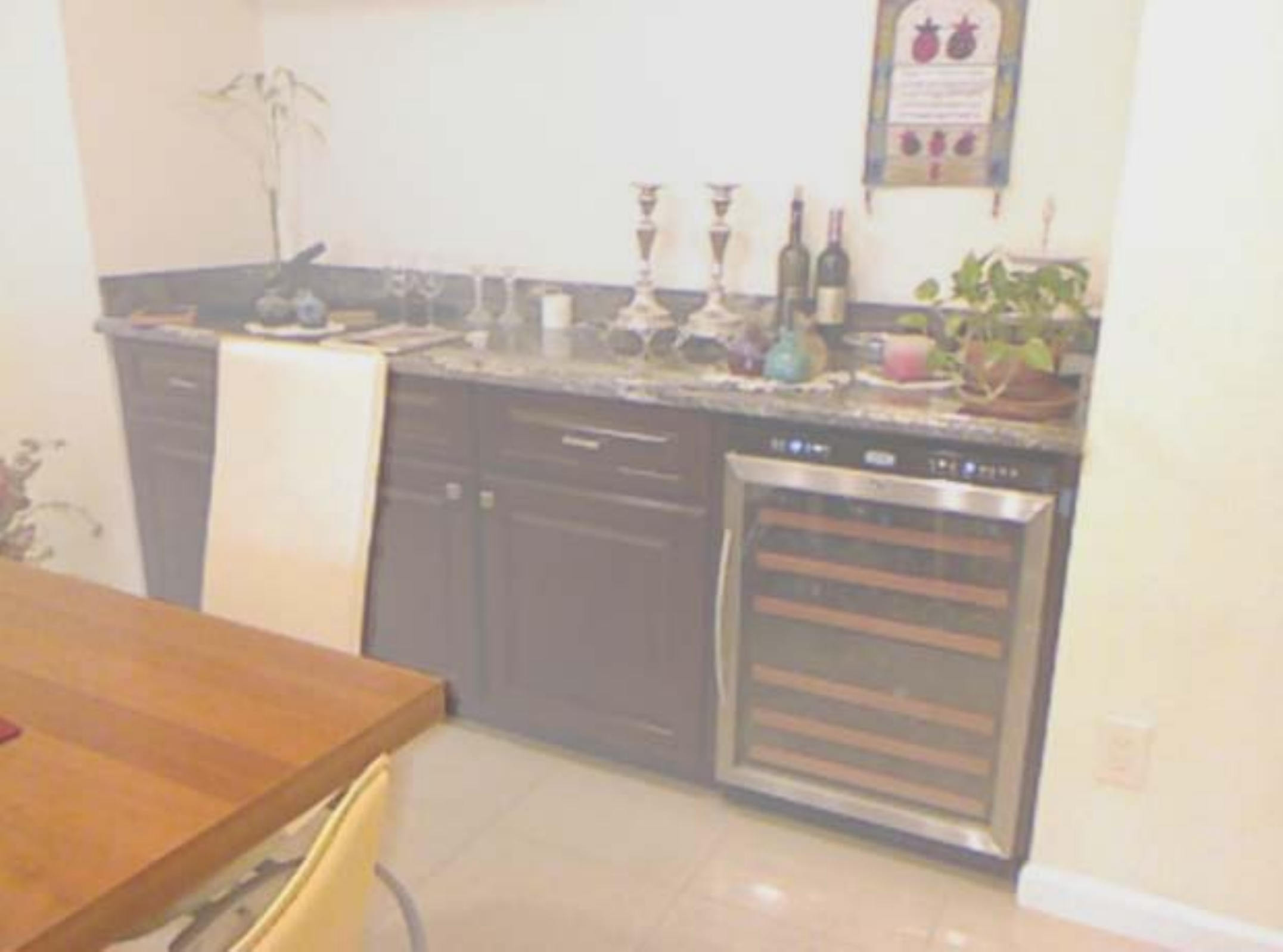} & \hspace{-0.4cm}
			\includegraphics[width = 0.095\textwidth]{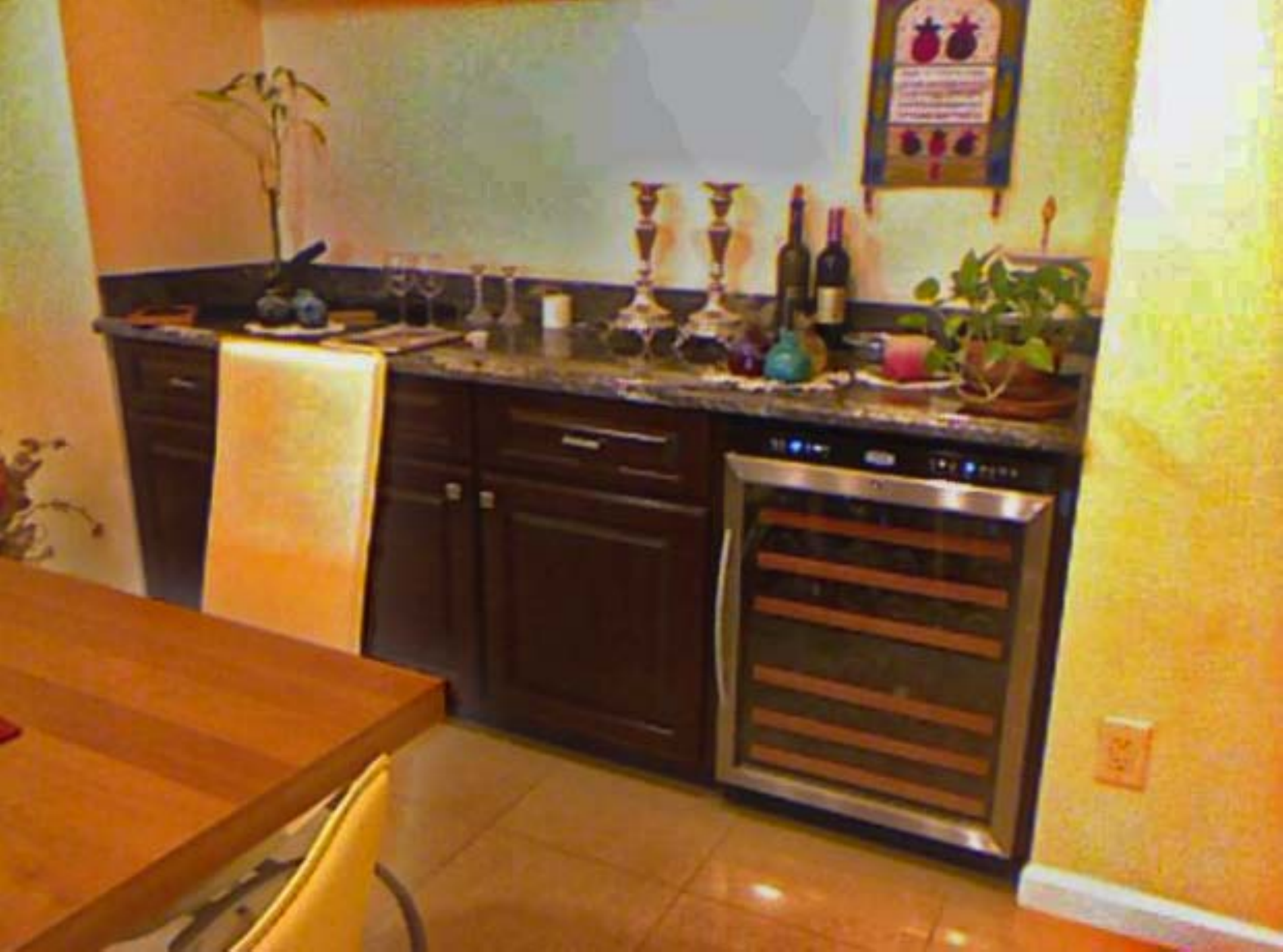} & \hspace{-0.4cm}
			\includegraphics[width = 0.095\textwidth]{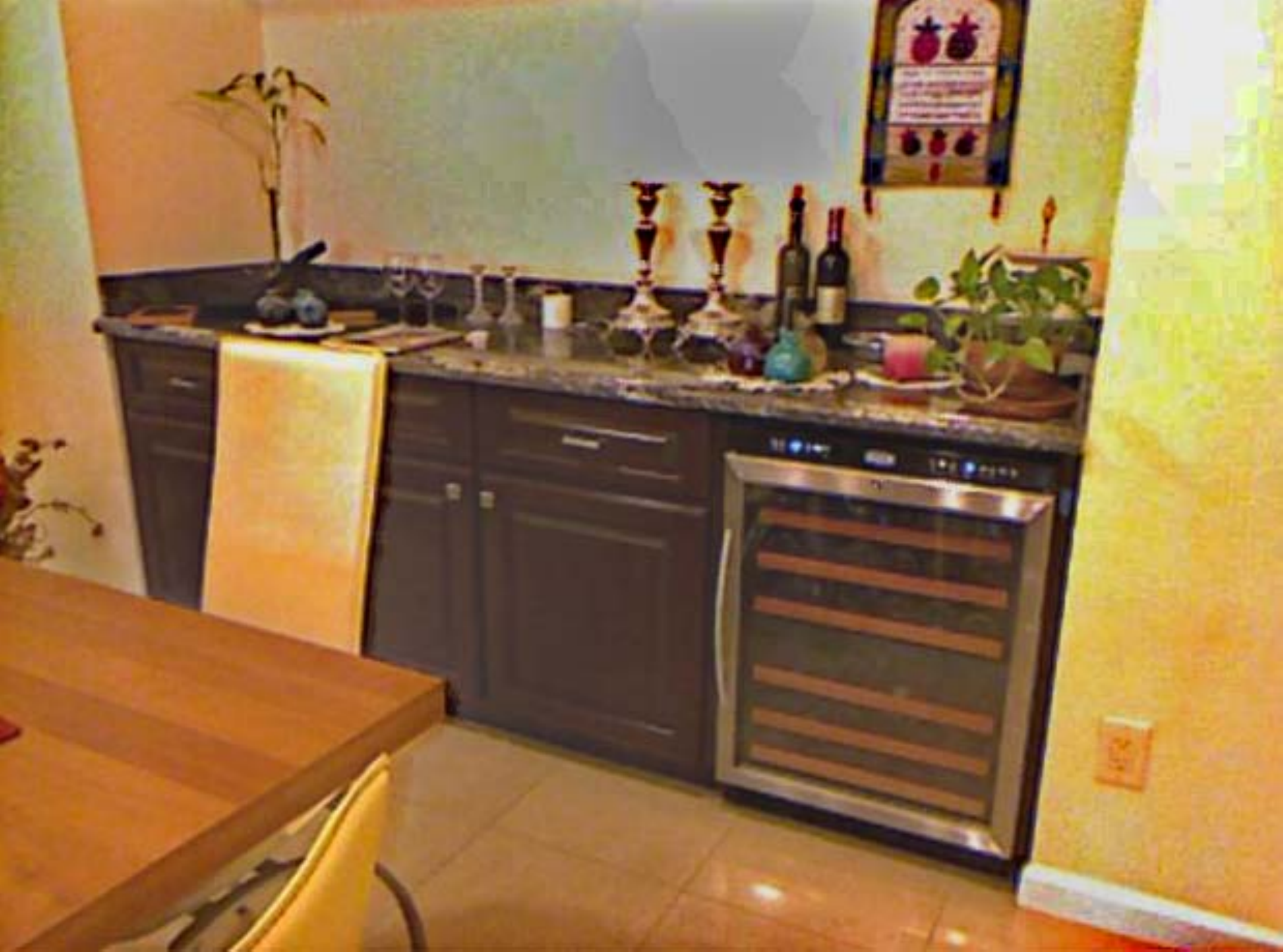} & \hspace{-0.4cm}
			\includegraphics[width = 0.095\textwidth]{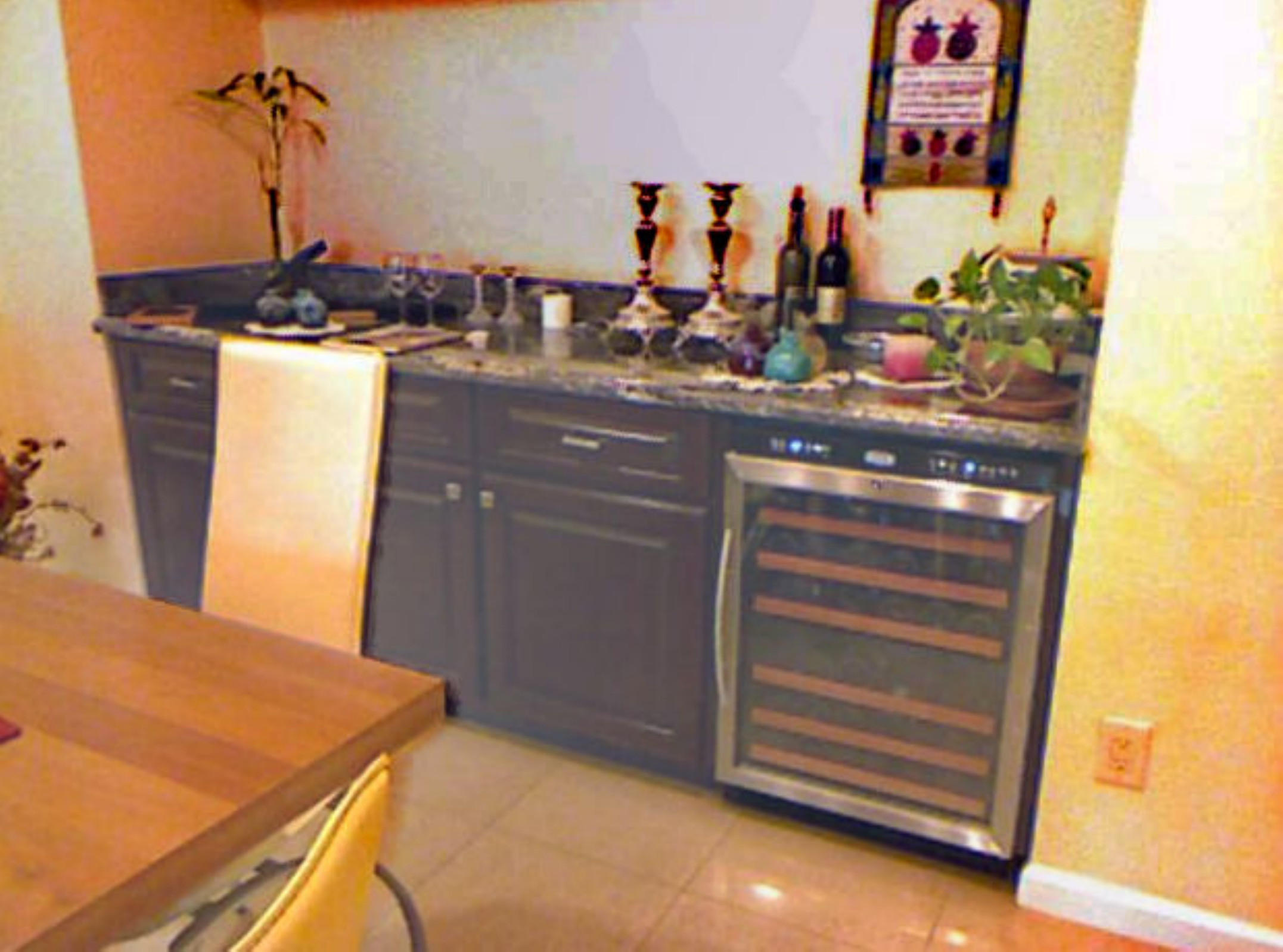} & \hspace{-0.4cm}
			\includegraphics[width = 0.095\textwidth]{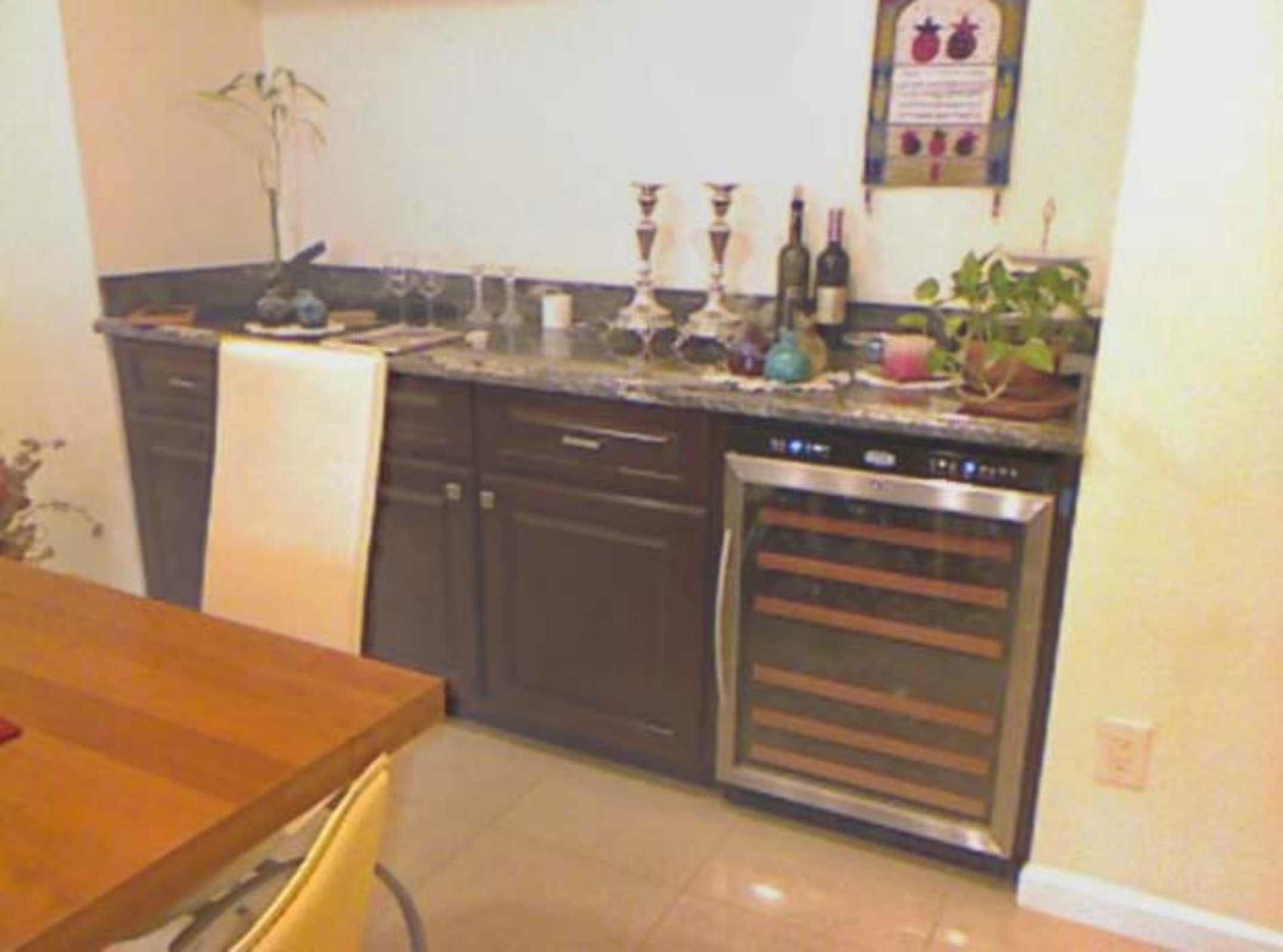} & \hspace{-0.4cm}
			\includegraphics[width = 0.095\textwidth]{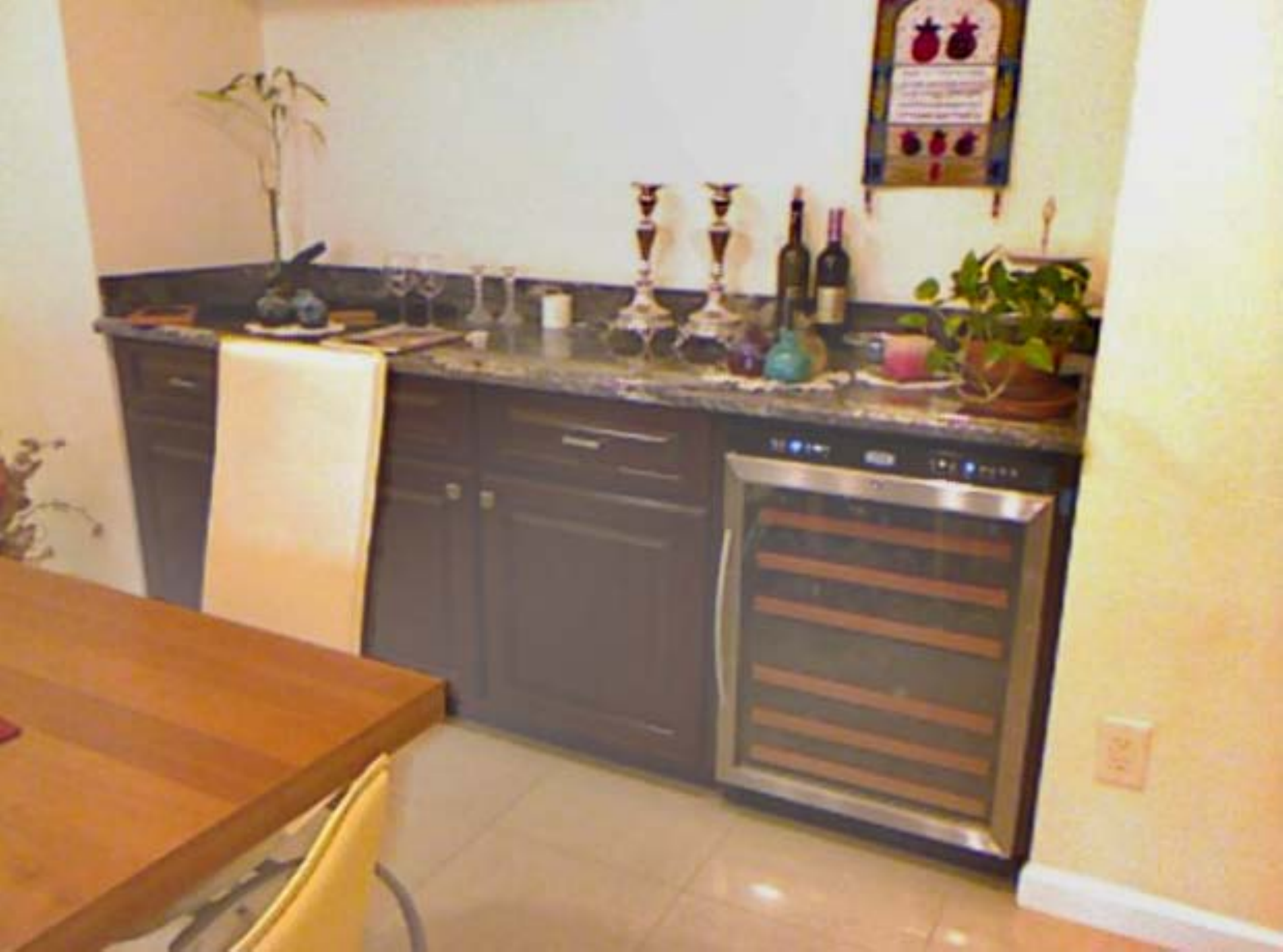} & \hspace{-0.4cm}
			\includegraphics[width = 0.095\textwidth]{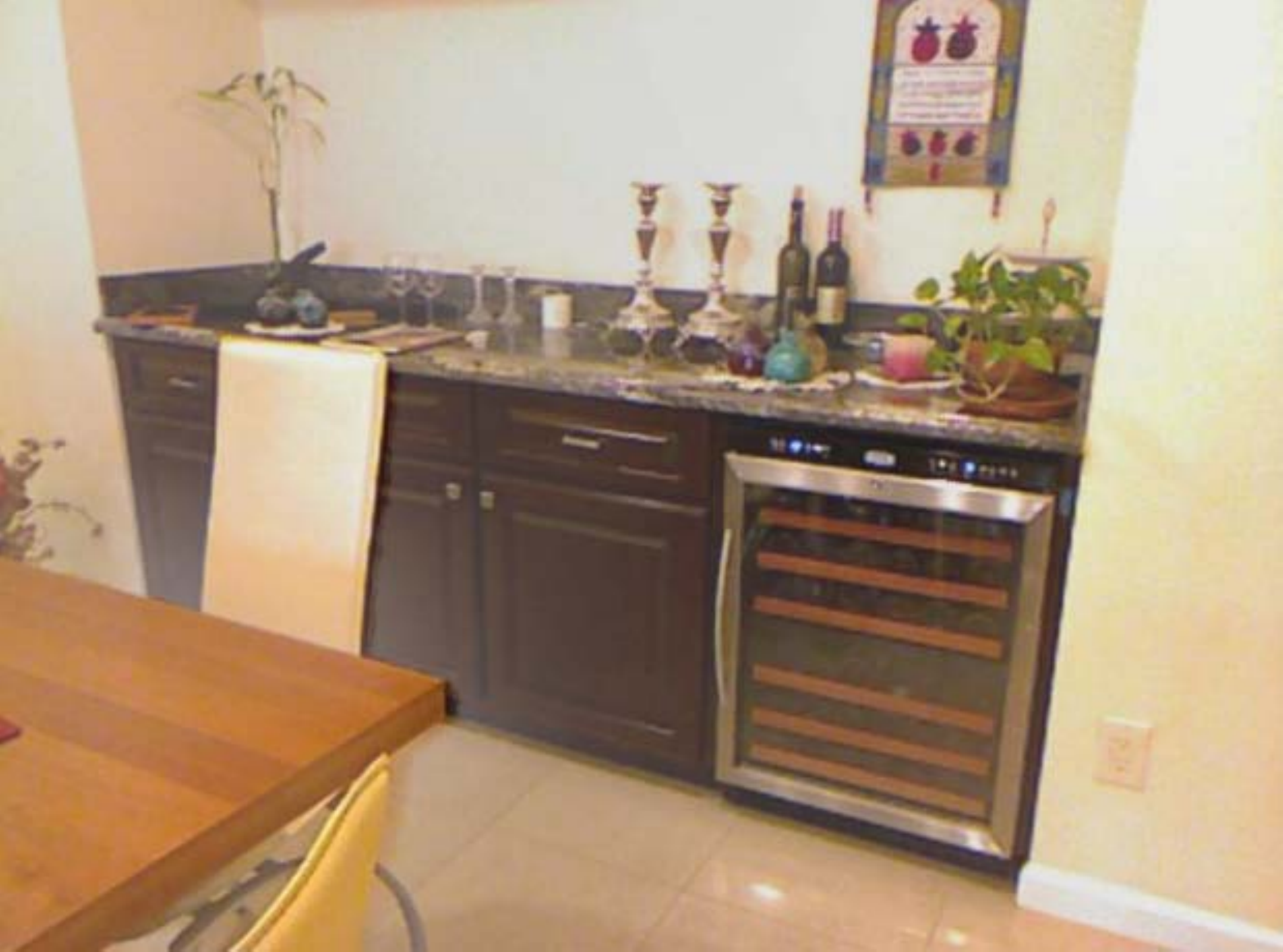} & \hspace{-0.4cm}
			\includegraphics[width = 0.095\textwidth]{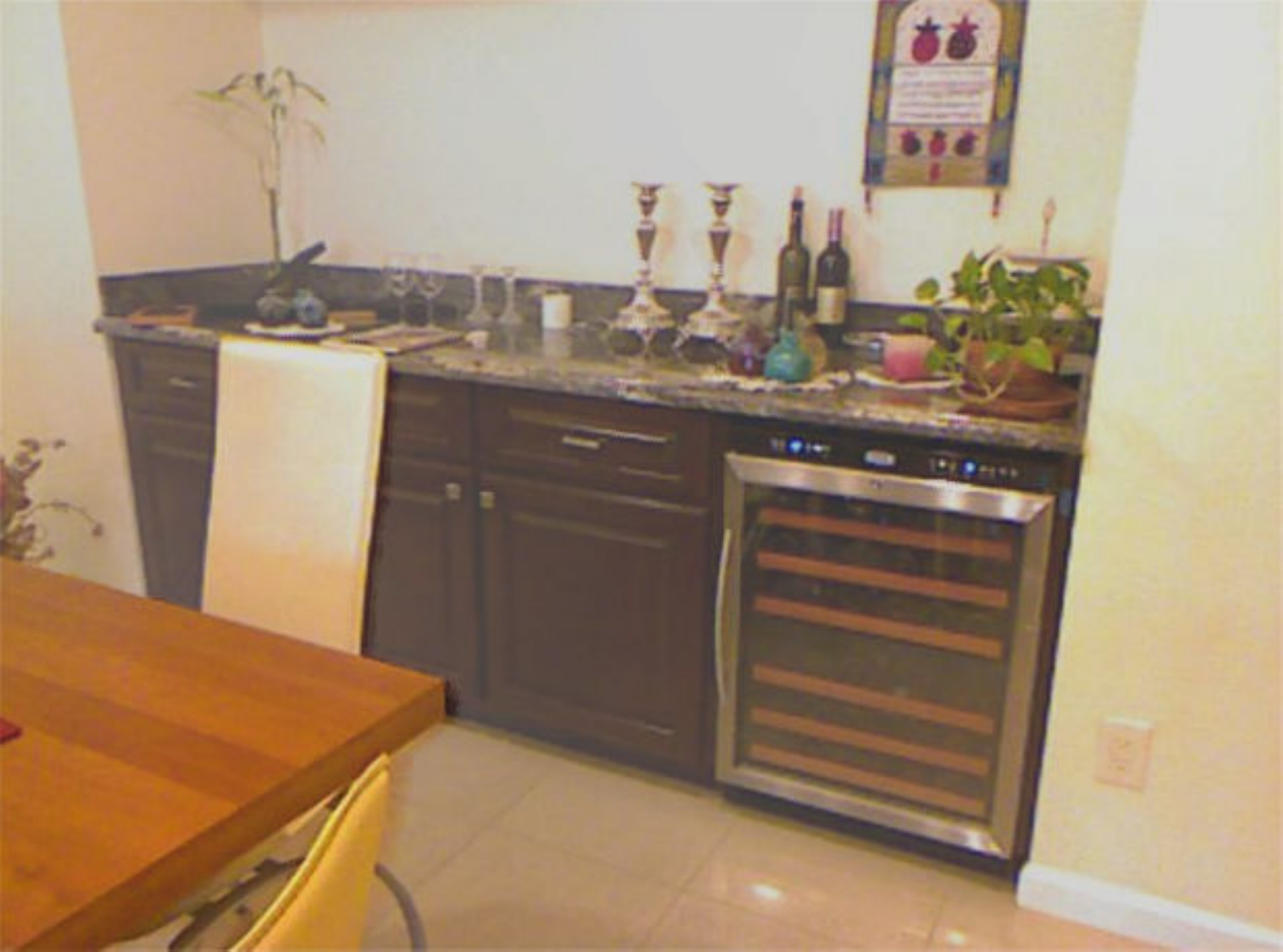} & \hspace{-0.4cm}
			\includegraphics[width = 0.095\textwidth]{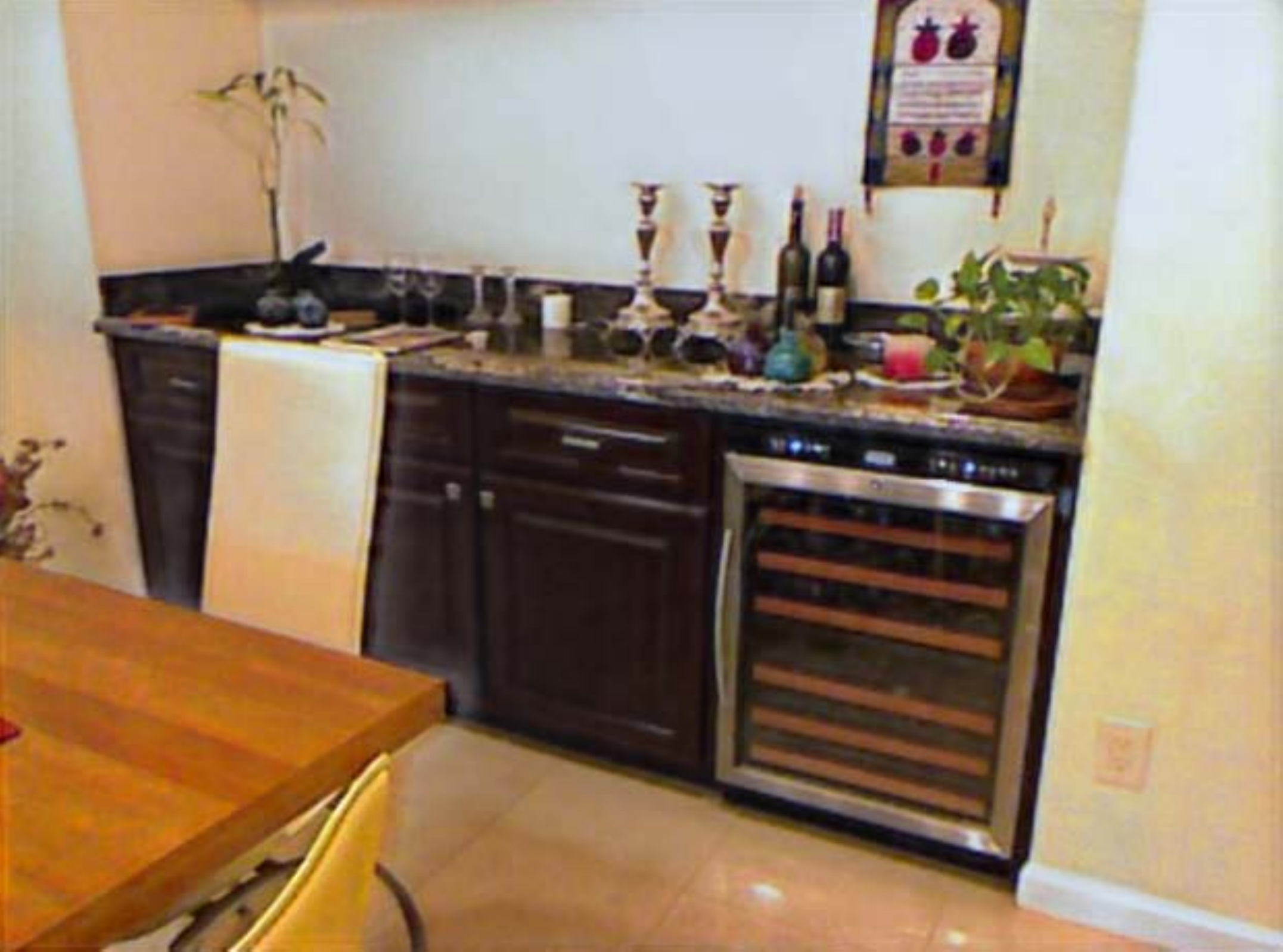} & \hspace{-0.4cm}
			\includegraphics[width = 0.095\textwidth]{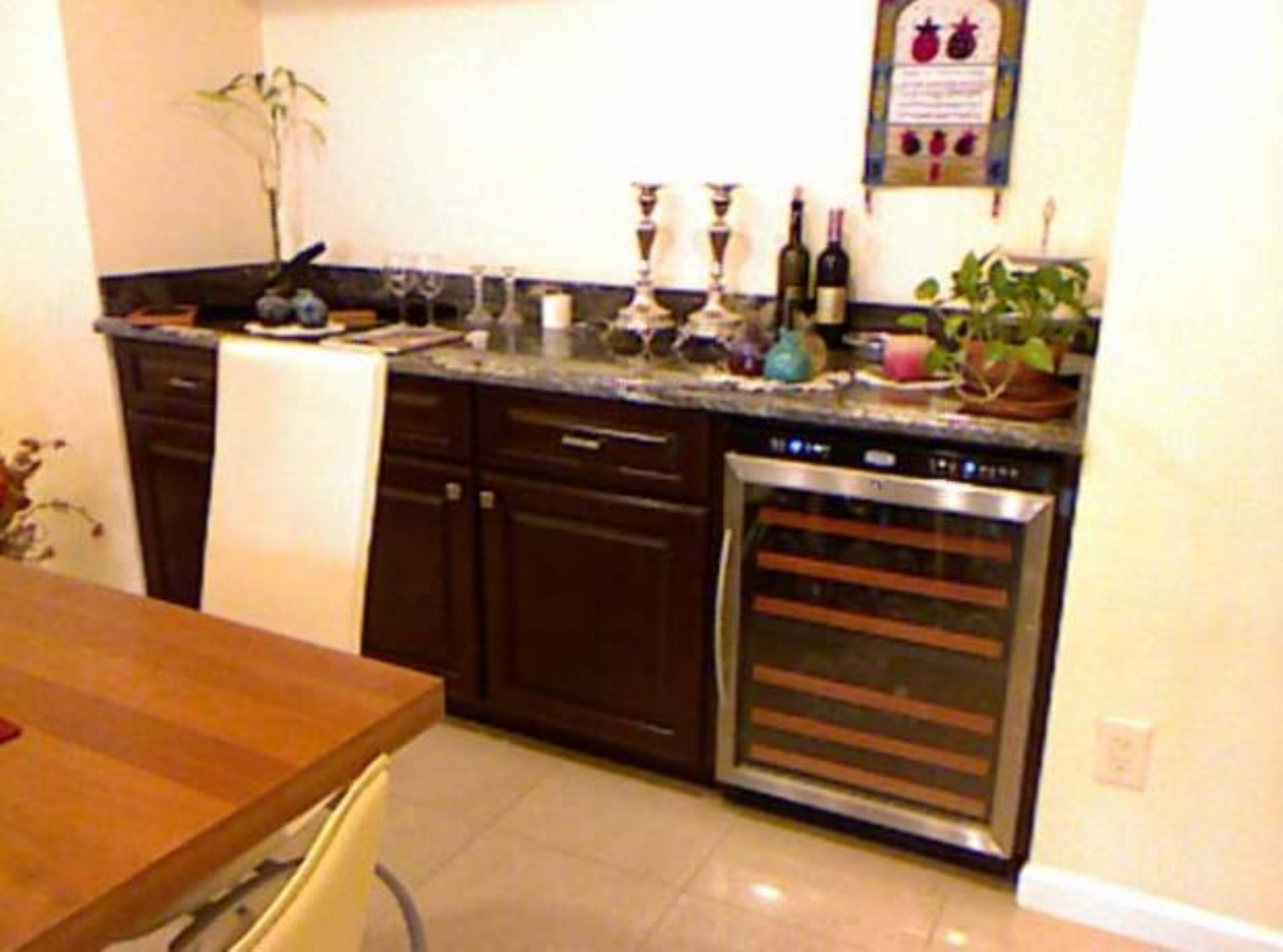}\\
			(a) Hazy inputs & \hspace{-0.4cm}
			(b) DCP~ & \hspace{-0.4cm}
			(c) BCCR~  & \hspace{-0.4cm}
			(d) NLD~  & \hspace{-0.4cm}
			(e) CAP~ & \hspace{-0.4cm}
			(f) MSCNN~  & \hspace{-0.4cm}
			(g) DehazeNet~  & \hspace{-0.4cm}
			(h) AOD-Net~  & \hspace{-0.4cm}
			(i) GFN & \hspace{-0.4cm}
			(j) Ground truths
		\end{tabular}
	\end{center}
	\vspace{-0.3cm}
	\caption{Dehazed results on the synthetic dataset. Dehazed results generated by the priors based methods~\cite{he2011single,meng2013efficient,berman2016non} have some color distortions in some regions. The learning based methods~\cite{zhu2015fast,ren2016single,cai2016dehazenet,li2017aod} tend to underestimate haze concentration so that the dehazed results have some remaining hazes. In contrast, the dehazed results by our method are close to the ground-truth images.
	}
	\label{fig-syn-results}
	\vspace{-0.1cm}
\end{figure*}
\begin{table*}[htbp]
	\caption{Average PSNR and SSIM values of dehazed results on the synthetic dataset.}
	\begin{center}\scriptsize{
			\begin{tabular}{cccccccccc}
				\toprule
				\multirow{2}{*}{} &	\multicolumn{9}{c}{PSNR/SSIM}\\
				\cline{2-10}
				& DCP~\cite{he2011single} & BCCR~\cite{meng2013efficient} & NLD~\cite{berman2016non}& CAP~\cite{zhu2015fast} & MSCNN~\cite{ren2016single} & DehazeNet~\cite{cai2016dehazenet} & AOD-Net~\cite{li2017aod} & GFN ($G$) & GFN ($G+D$) \\
				\midrule
				Light & 18.74/0.77 & 17.72/0.76 & 18.61/0.71 & 21.92/0.83 & 22.34/0.82 & \textbf{24.87}/0.84 & 22.64/\textbf{0.85}  & 24.78/\textbf{0.85} & 24.60/0.83 \\
				\midrule
				Medium & 18.68/0.77 & 17.54/0.75 & 18.47/0.70 & 21.40/0.82 & 21.21/0.80 & 23.37/0.83 & 21.33/\textbf{0.84}  &\textbf{23.68}/\textbf{0.84} & 23.55/\textbf{0.84}\\
				\midrule
				Heavy & 18.67/0.77 & 17.43/0.75 & 18.21/0.70 & 20.21/0.80 & 20.51/0.79 & 21.98 /0.82 & 20.24/0.81 &\textbf{22.32}/\textbf{0.83} & 22.75/0.82\\
				\midrule
				Random & 18.58/0.77 & 17.35/0.75 & 18.28/0.71 & 19.99/0.78 & 20.01/0.78 & 20.97/0.80 & 19.36/0.78  &\textbf{22.41}/0.81 & 22.20/\textbf{0.82}\\
				\bottomrule
		\end{tabular}}
		\label{tab-psnr-ssim}
	\end{center}
	\vspace{-0.8cm}
\end{table*}
%
\vspace{-0.2cm}
\section{Experimental Results}
We quantitatively evaluate the proposed algorithm on both synthetic dataset and
real-world hazy photographs, with comparisons to the state-of-the-art methods
in terms of accuracy and visual effect. The implementation code can be found at our project website. 

\subsection{Implementation Details}
In our network, patch size is set as $128\times128$.
We use ADAM~\cite{kingma2014adam} optimizer with a batch size
10 for training. The initial learning rate is 0.0001 and we decrease the learning rate by 0.75 every 10,000 iterations.
For all the results reported in the paper, we train the
network for 240,000 iterations, which takes about 35 hours
on an Nvidia K80 GPU. Default values of $\beta_1$ and $\beta_2$
are used, which are 0.9 and 0.999, respectively, and
we set weight decay to 0.00001.
Since our approach dehazes images in a single forward pass, it is computationally
very efficient. Using a NVidia K80 GPU, we can process a $640\times480$ image within 0.3s.

\subsection{Training Data}
Generating realistic training data is a major challenge
for tasks where ground truth data cannot be easily collected.
For training our neural network, we adopt the NYU2 dataset~\cite{silberman2012indoor}
and the synthetic method in~\cite{ren2016single} to synthesize the training data.
We use 1400 clean images and the corresponding labeled depth maps from the NYU Depth dataset~\cite{silberman2012indoor} to construct the training set.
Given a clear image $\mathbf{J}$, a random atmospheric light $\mathbf{A}\in (0.8,1.0)$ and the ground truth depth $d$, we use
$t(x)=e^{-\beta d(x)}$
to synthesize transmission first, then generate
hazy image using the physical model~\eqref{eq-math_model}.
For scattering coefficient $\beta$,
we randomly select it from
0.5 to 1.5 as suggested in~\cite{ren2016single}.
We use 7 different $\beta$ for each clean image, so that we can
synthesize different haze concentration images for each input image.
In addition, $1\%$ Gaussian noise is added to each hazy input to increase the robustness of the trained network.

\subsection{Quantitative Evaluation on Synthetic Dataset}
For quantitative evaluation, we use the remaining 49 clean images in the label data except the 1400 training images from the NYU2 dataset~\cite{silberman2012indoor} to synthetic hazy images with known depth map $d$ as like in~\cite{ren2016single}.
We evaluate these methods by two criteria: Structure Similarity (SSIM) and Peak Signal to Noise Ratio (PSNR).
In this section, we compare the proposed algorithm with the following seven
methods on the synthesized datasets.

\vspace{-0.2cm}
{\flushleft \textbf{Priors based methods}}~\cite{he2011single,meng2013efficient,berman2016non}.
We use three prior based methods for comparisons. The first one is the DCP proposed by He~\etal~\cite{he2011singlecvpr,he2011single}. This is a commonly used baseline approach in most dehazing papers.
The second is Boundary Constrained Context Regularization (BCCR) proposed by Meng~\etal~\cite{meng2013efficient} and the third is the Non-local Image Dehazing (NLD) algorithm in~\cite{berman2016non}.
\vspace{-0.2cm}
{\flushleft \textbf{Learning based methods}}~\cite{zhu2015fast,cai2016dehazenet,ren2016single,li2017aod}.
We also use four learning based methods for comparisons.
The first one learns a linear model based on Color Attenuation Prior (CAP).
The second and third are CNNs based methods of DehazeNet~\cite{cai2016dehazenet} and MSCNN~\cite{ren2016single}.
These methods implement image dehazing by learning the map between hazy inputs and their transmission based on convolutional neural networks.
The last AOD-Net~\cite{li2017aod} is also a CNNs based method, but integrates the transmission and atmospheric light into a new variable.
\begin{figure*}[t]\scriptsize
	\begin{center}
		\begin{tabular}{@{}ccccccccc@{}}
			\includegraphics[width = 0.105\textwidth]{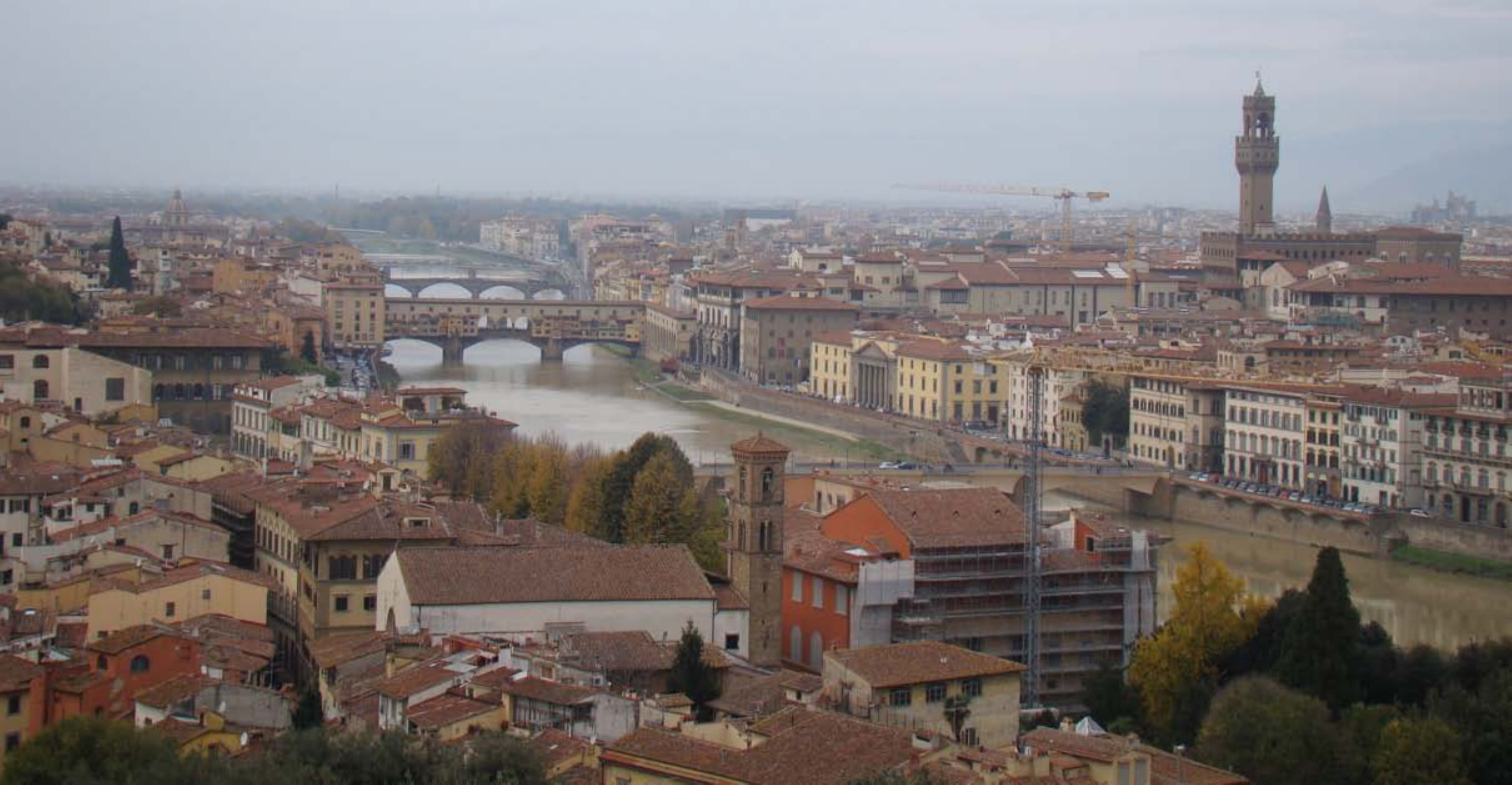} & \hspace{-0.4cm}
			\includegraphics[width = 0.105\textwidth]{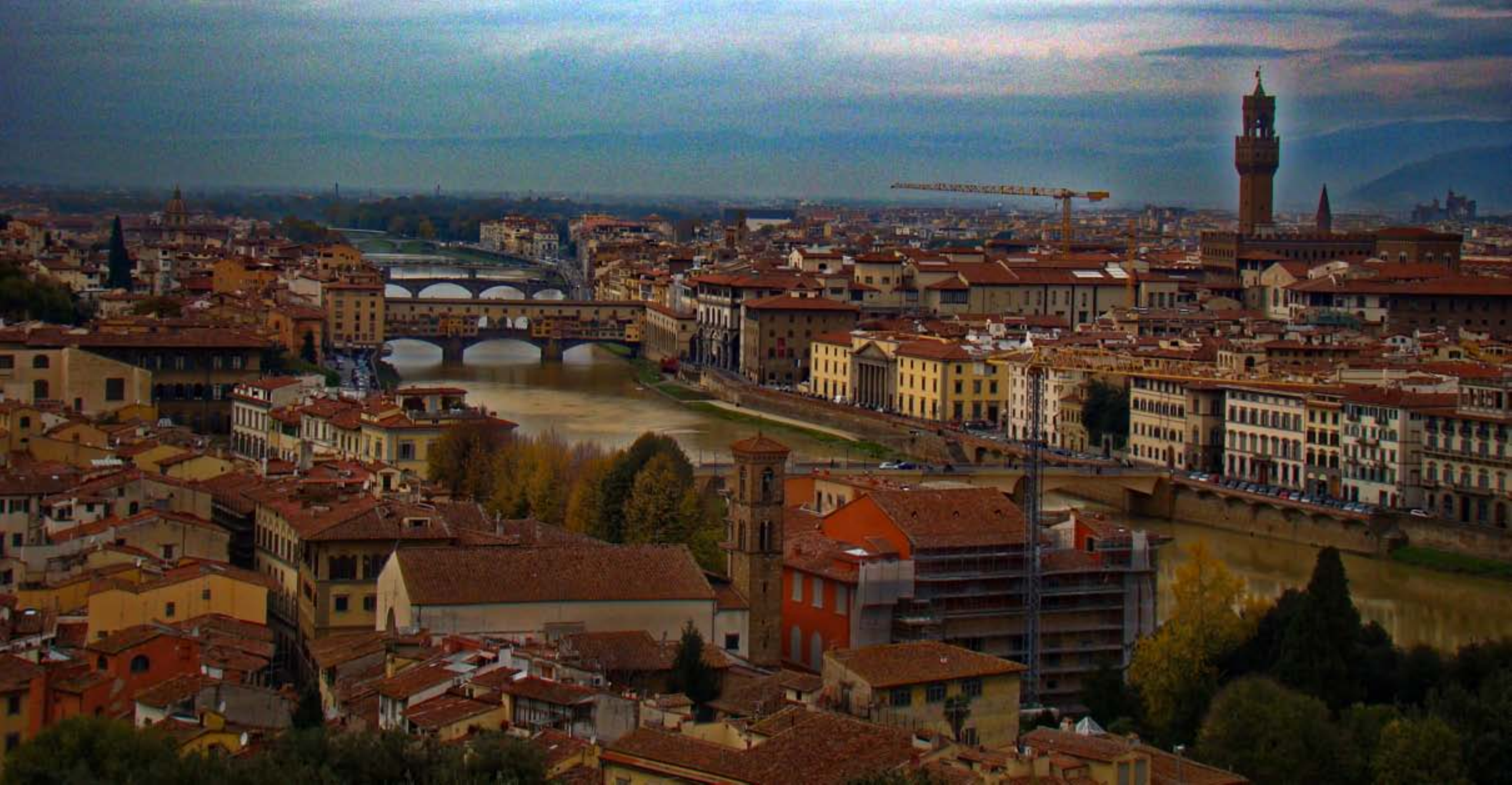} & \hspace{-0.4cm}
			\includegraphics[width = 0.105\textwidth]{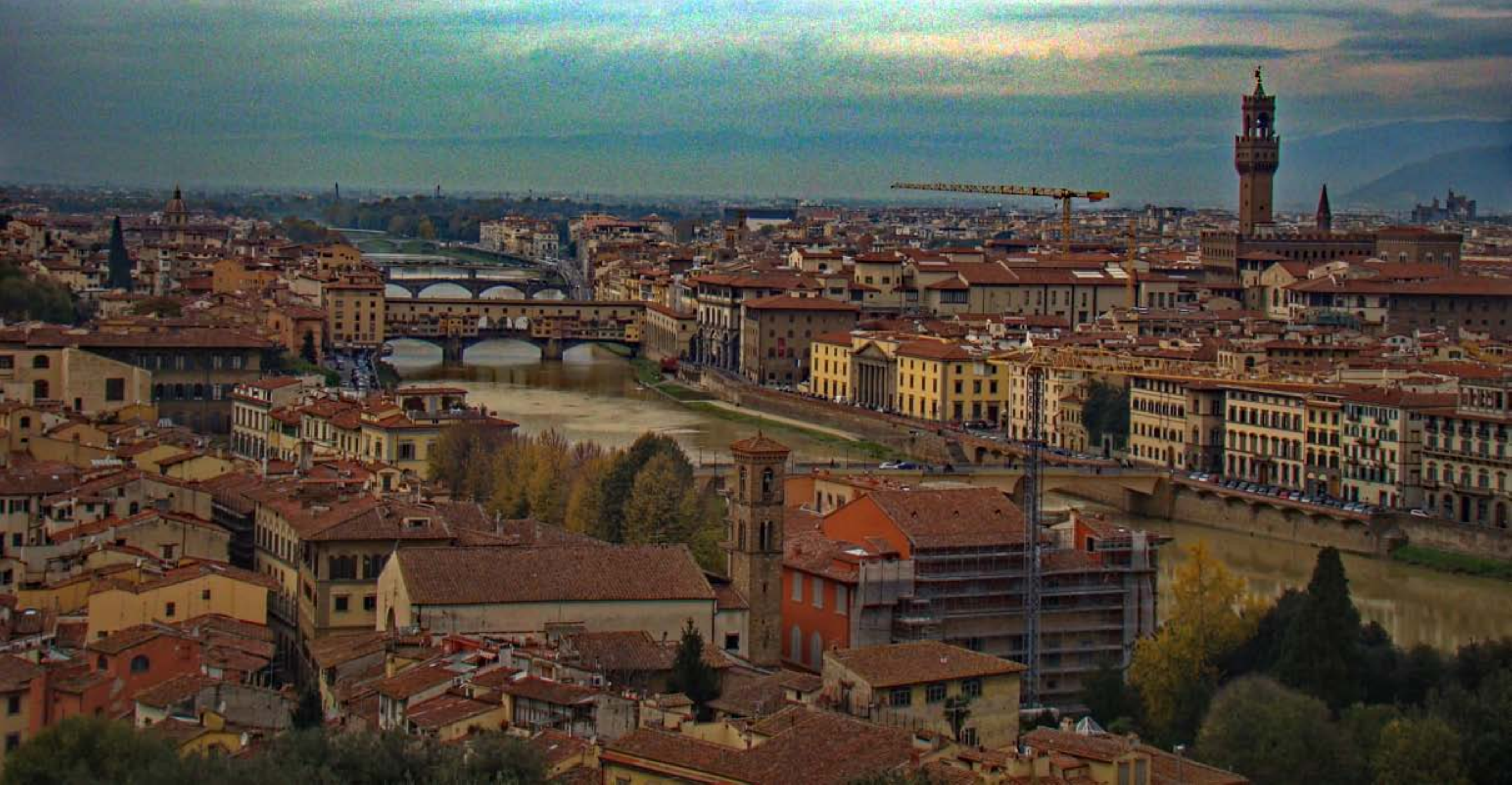} & \hspace{-0.4cm}
			\includegraphics[width = 0.105\textwidth]{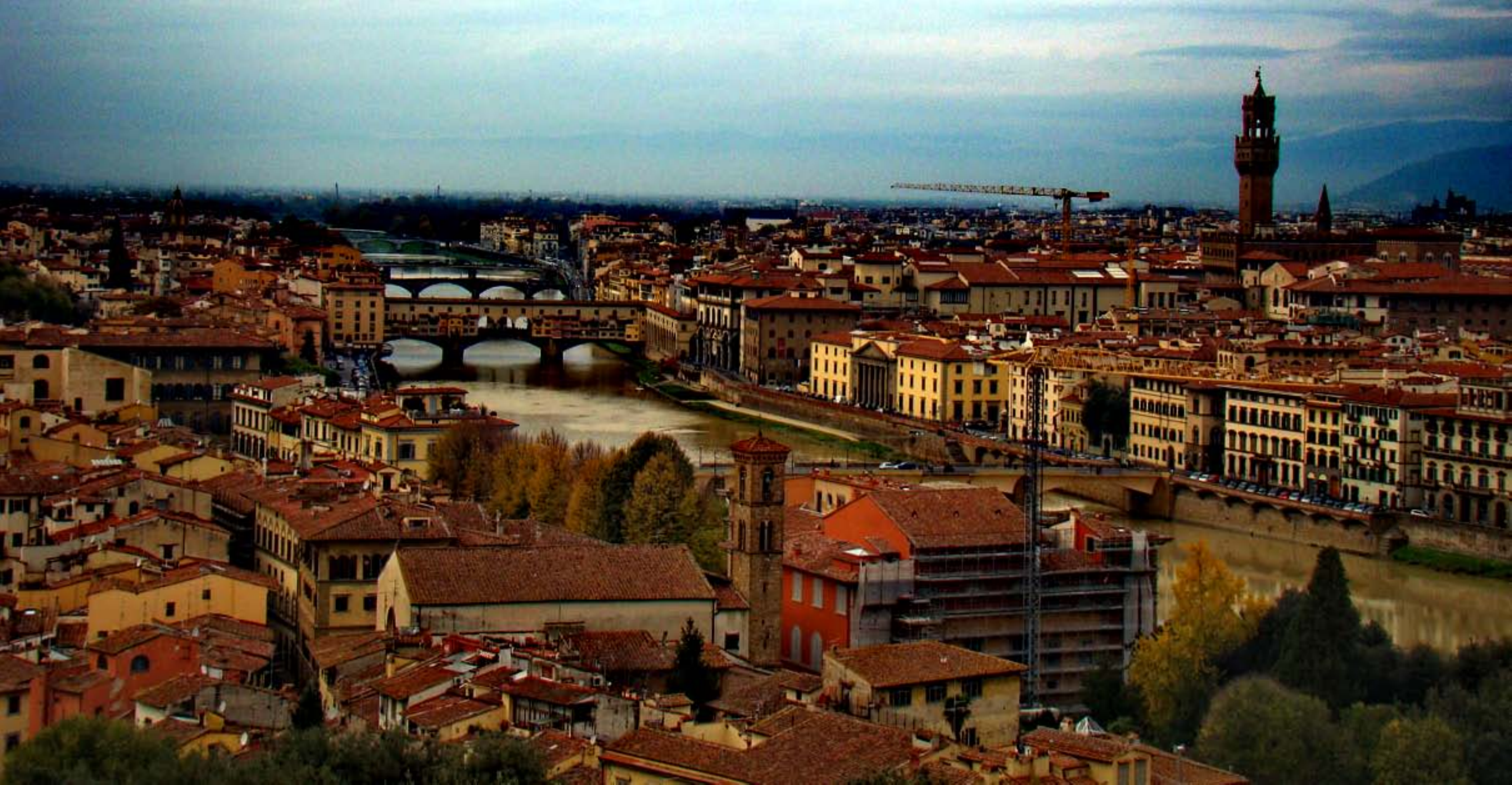} & \hspace{-0.4cm}
			\includegraphics[width = 0.105\textwidth]{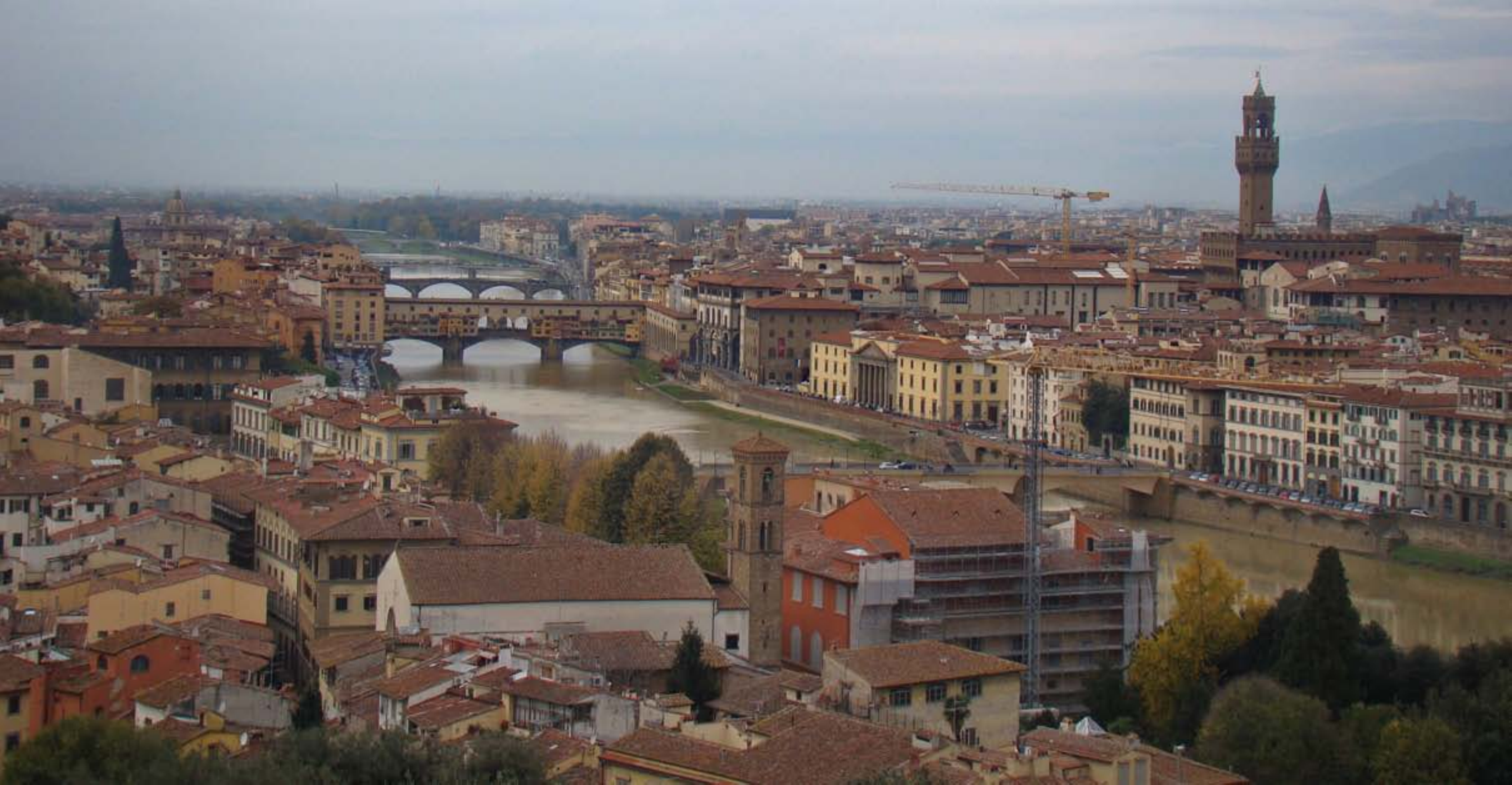} & \hspace{-0.4cm}
			\includegraphics[width = 0.105\textwidth]{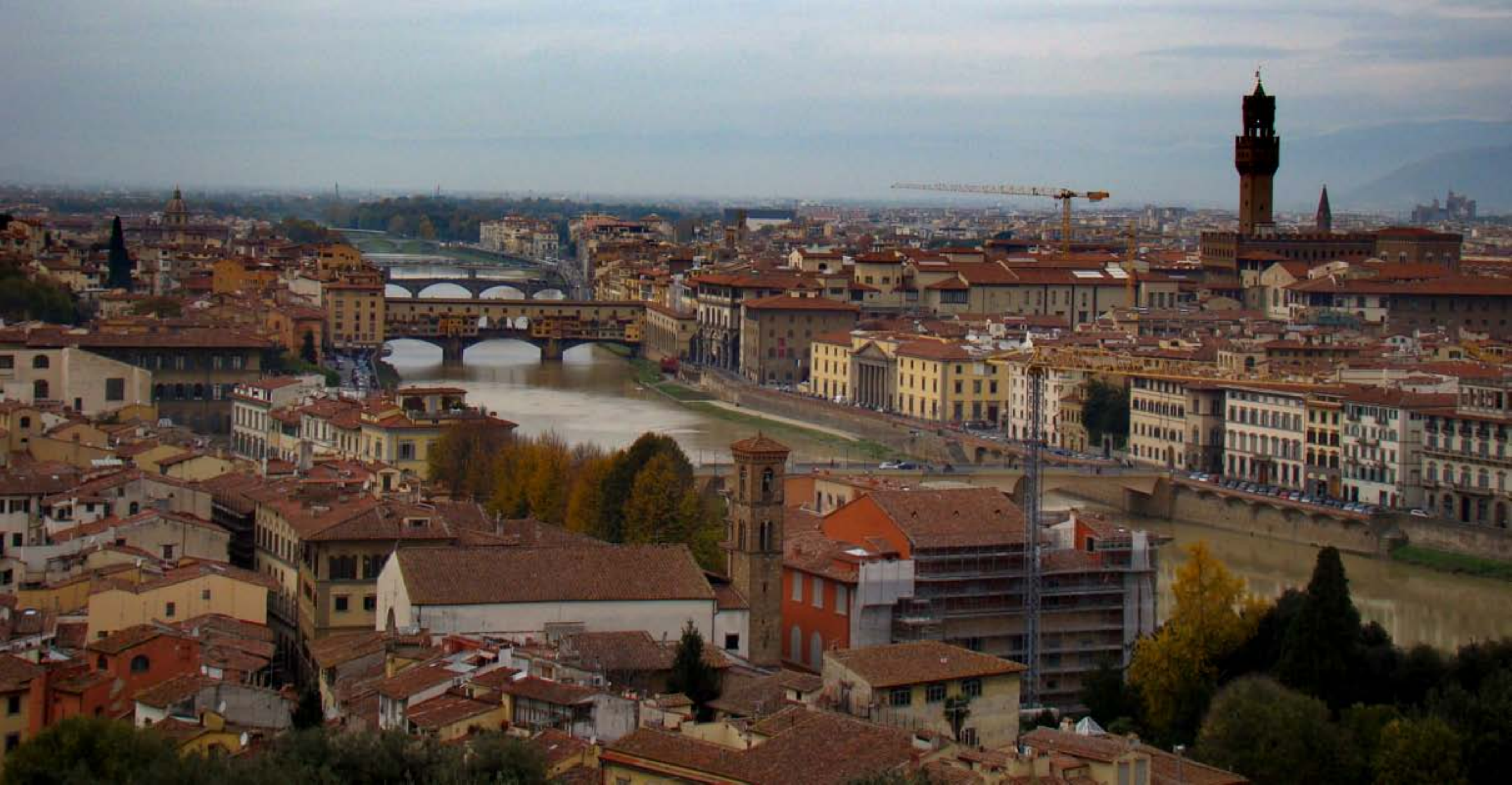} & \hspace{-0.4cm}
			\includegraphics[width = 0.105\textwidth]{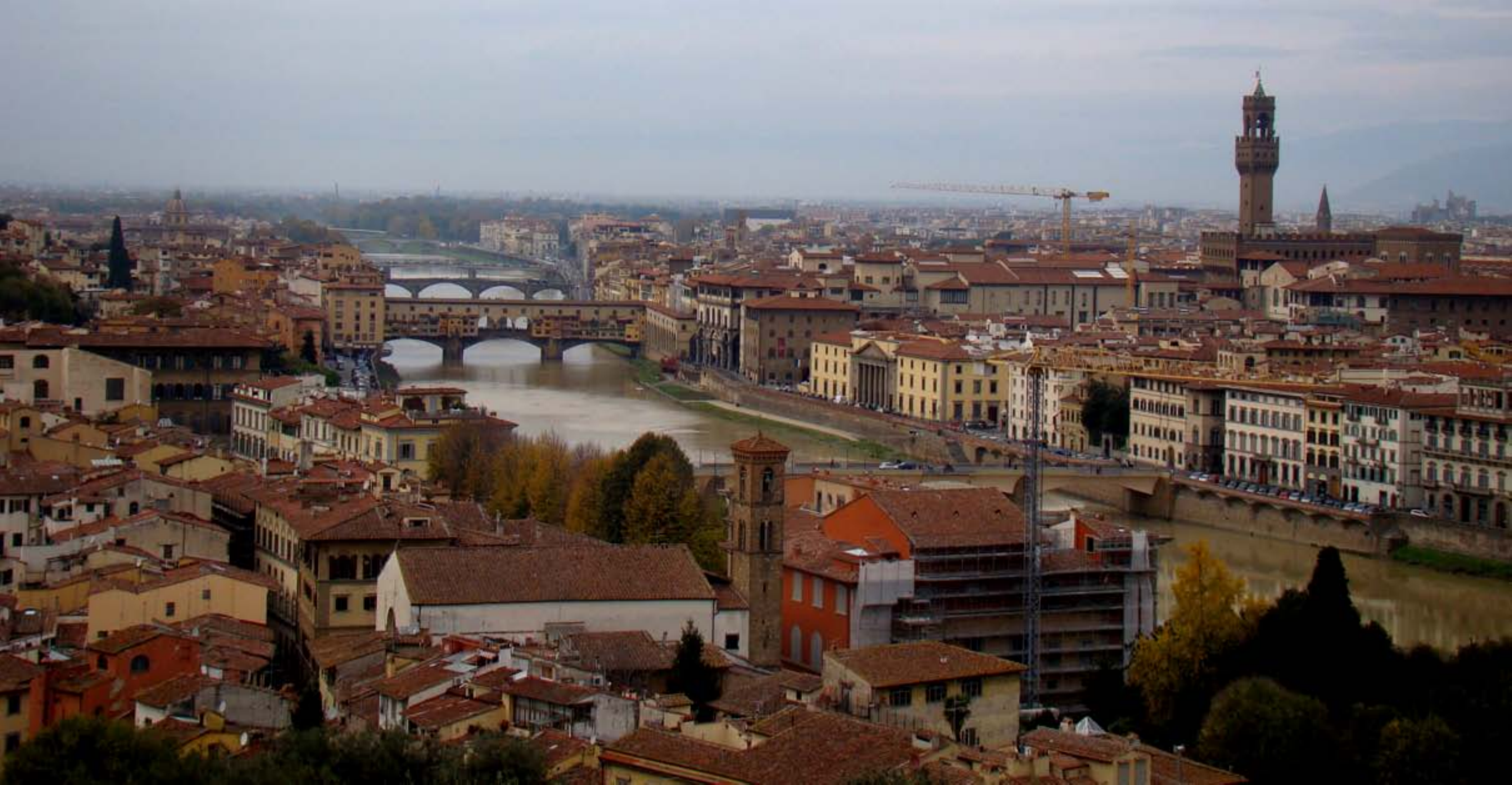} & \hspace{-0.4cm}
			\includegraphics[width = 0.105\textwidth]{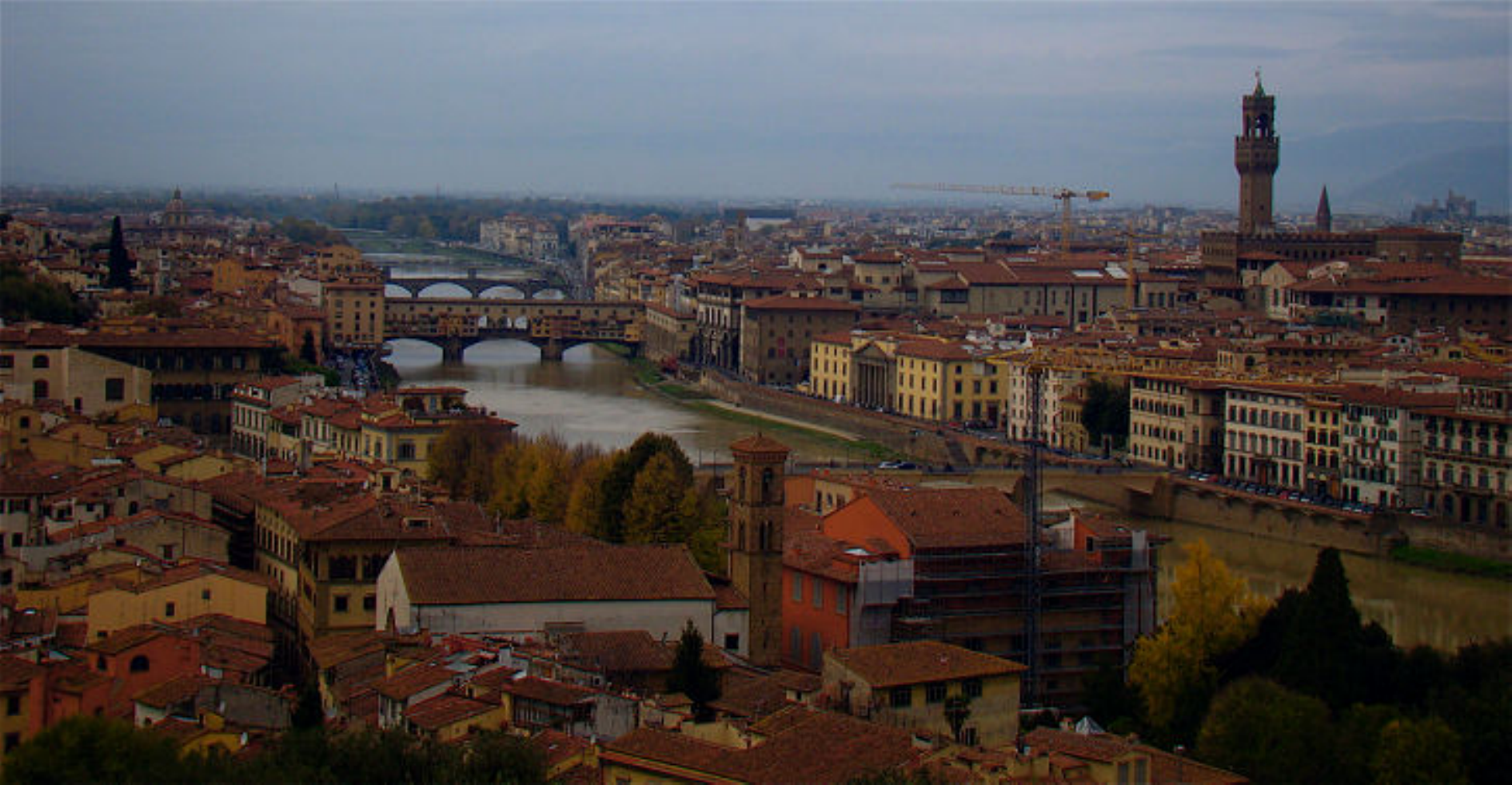} & \hspace{-0.4cm}
			\includegraphics[width = 0.105\textwidth]{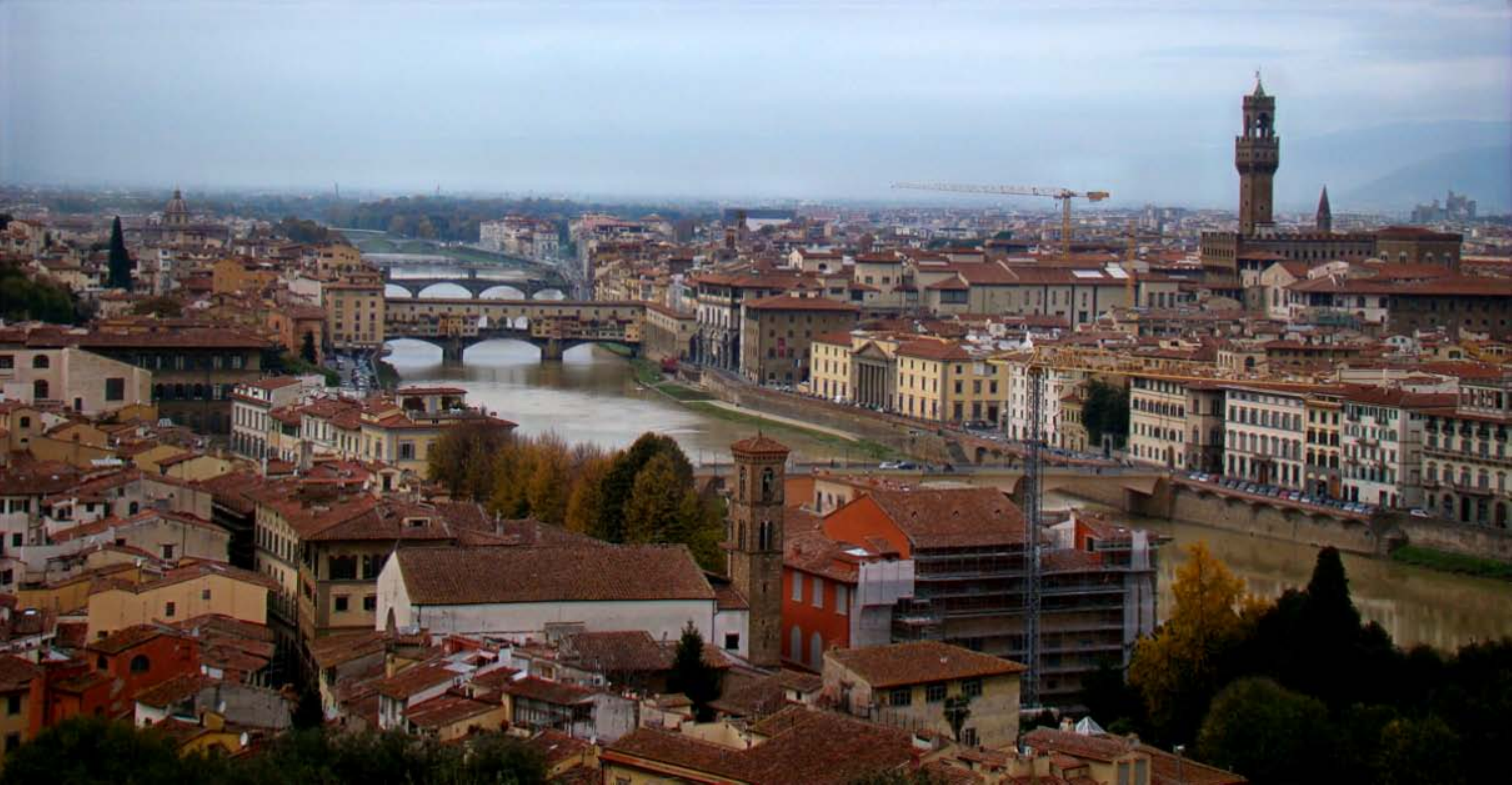} \\
			\includegraphics[width = 0.105\textwidth]{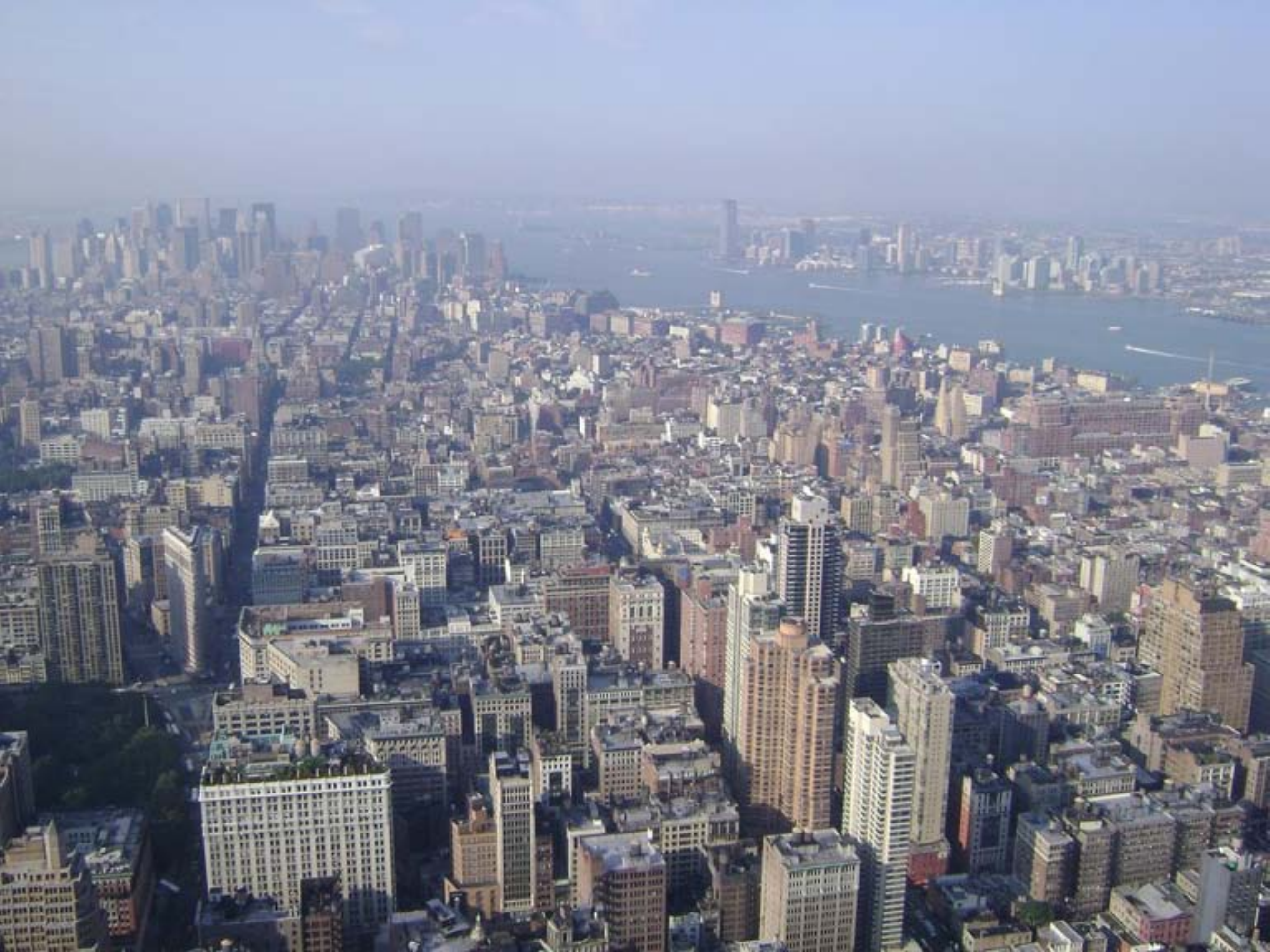} & \hspace{-0.4cm}
			\includegraphics[width = 0.105\textwidth]{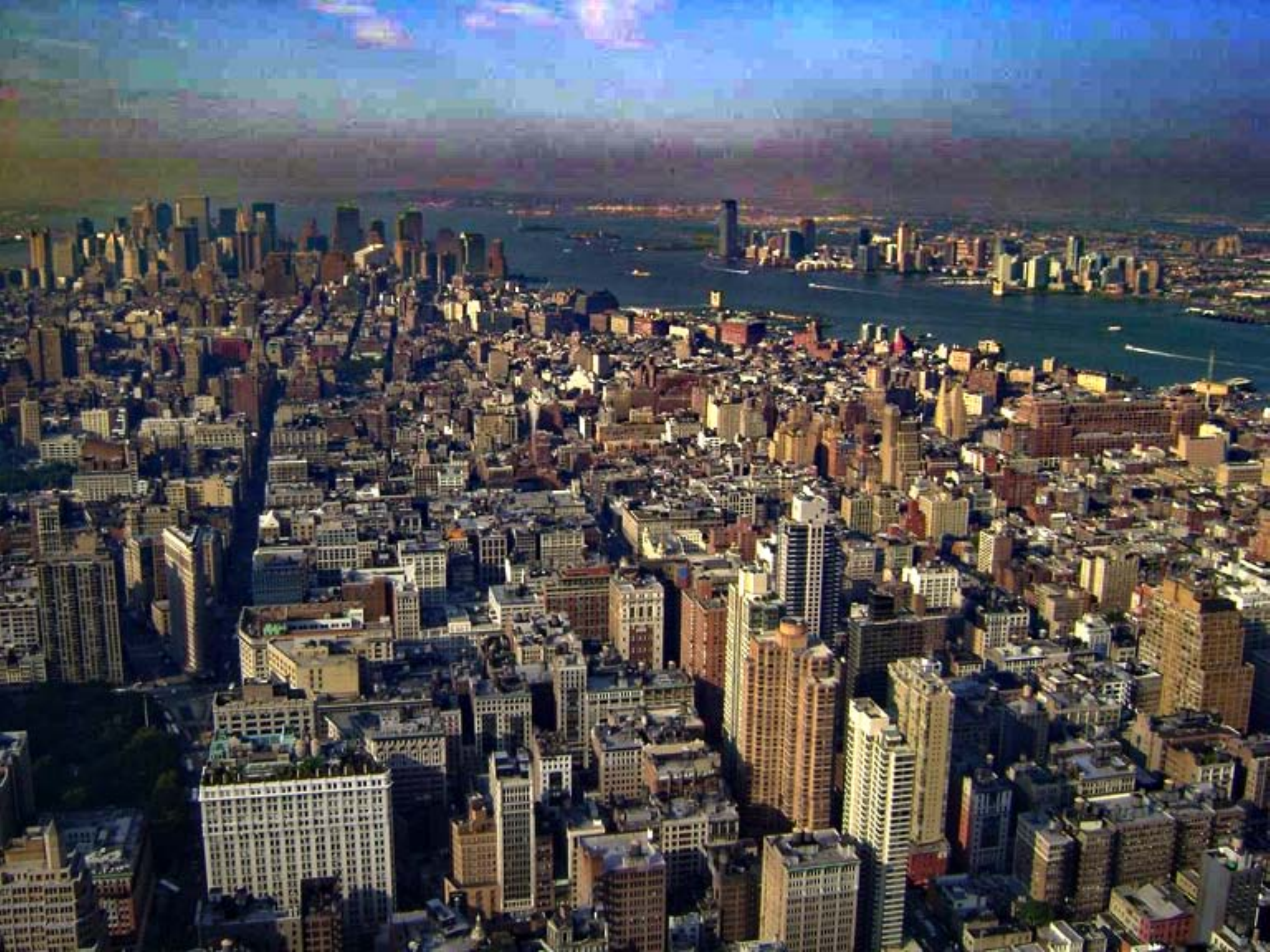} & \hspace{-0.4cm}
			\includegraphics[width = 0.105\textwidth]{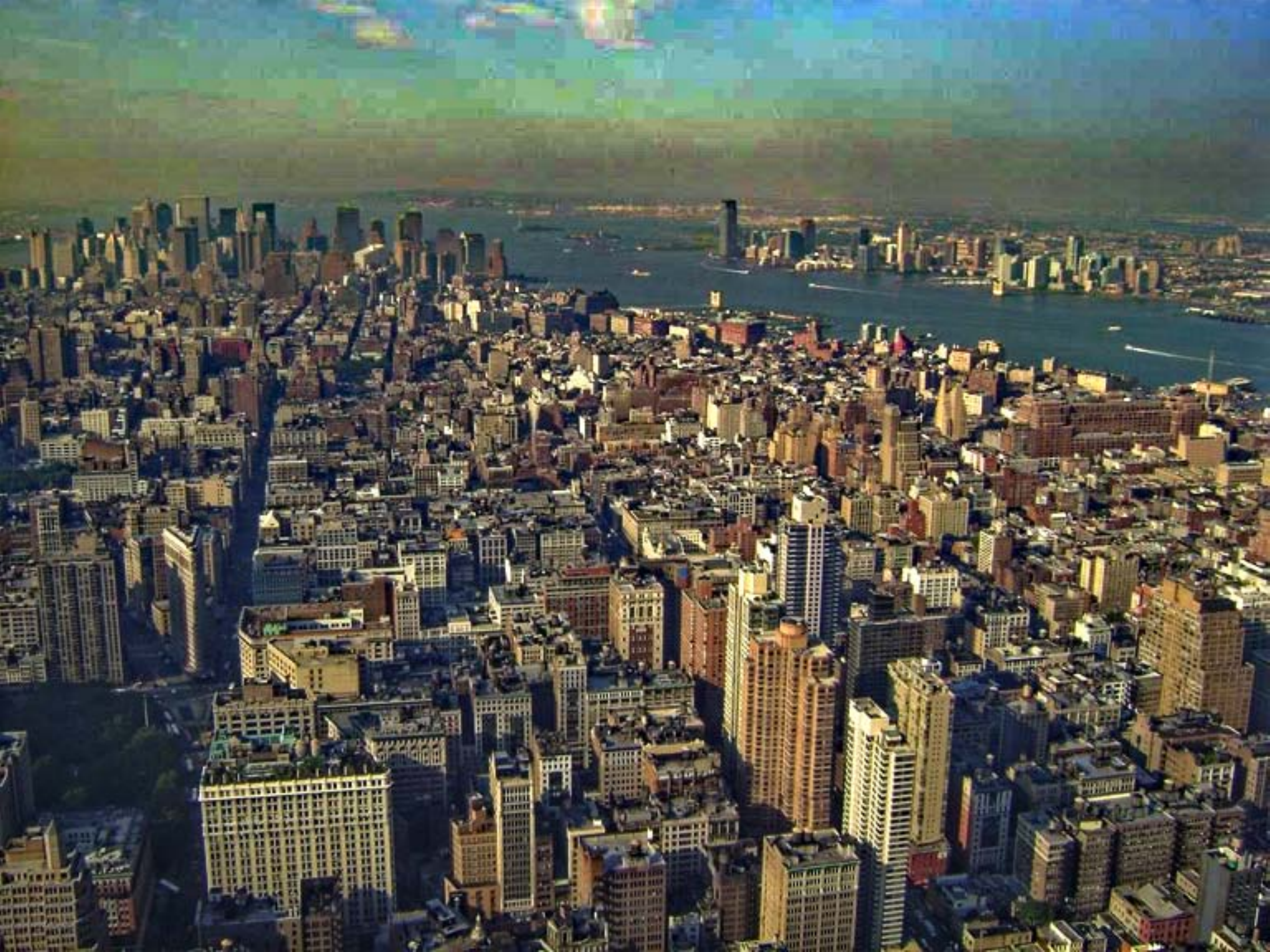} & \hspace{-0.4cm}
			\includegraphics[width = 0.105\textwidth]{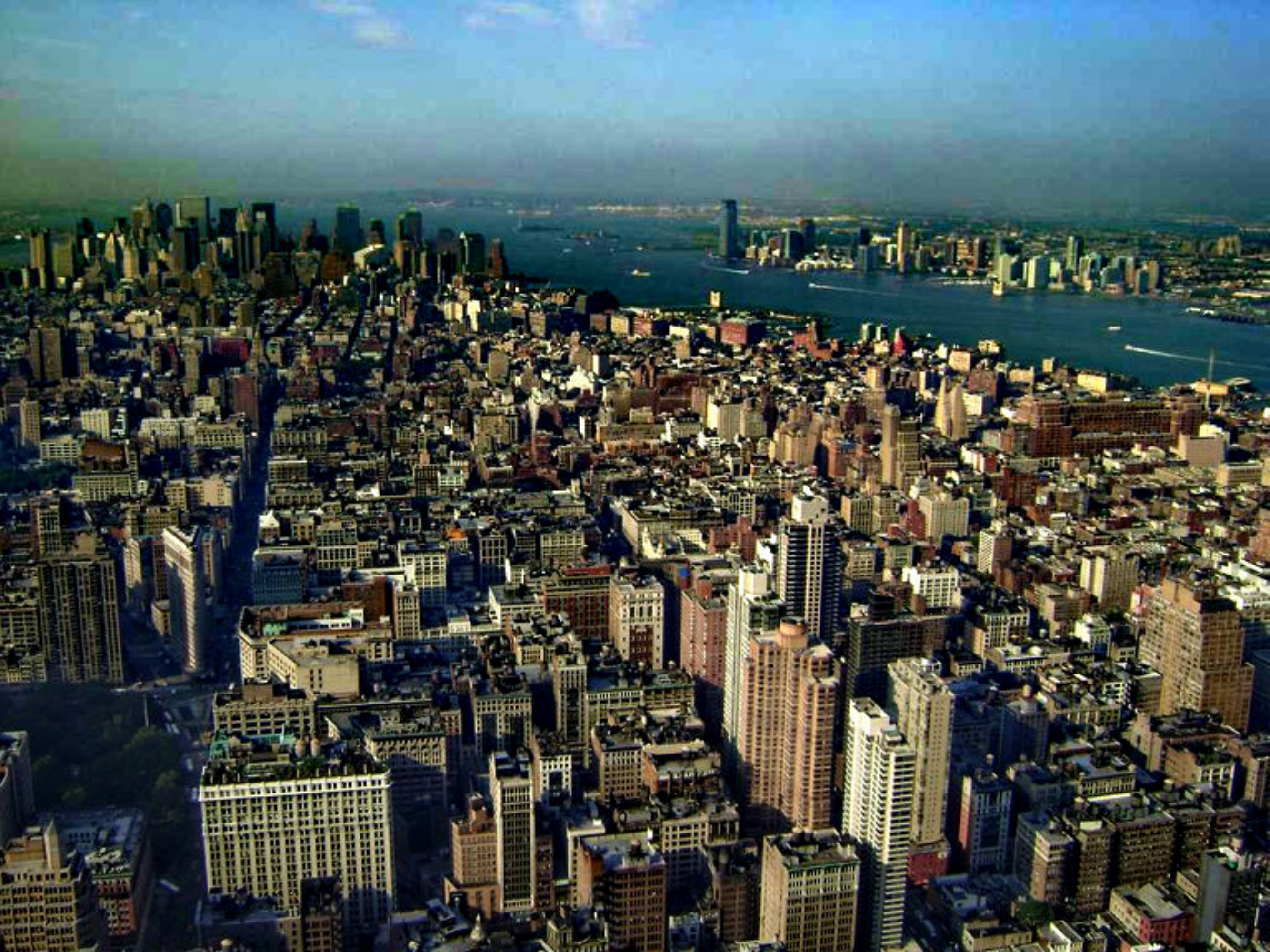} & \hspace{-0.4cm}
			\includegraphics[width = 0.105\textwidth]{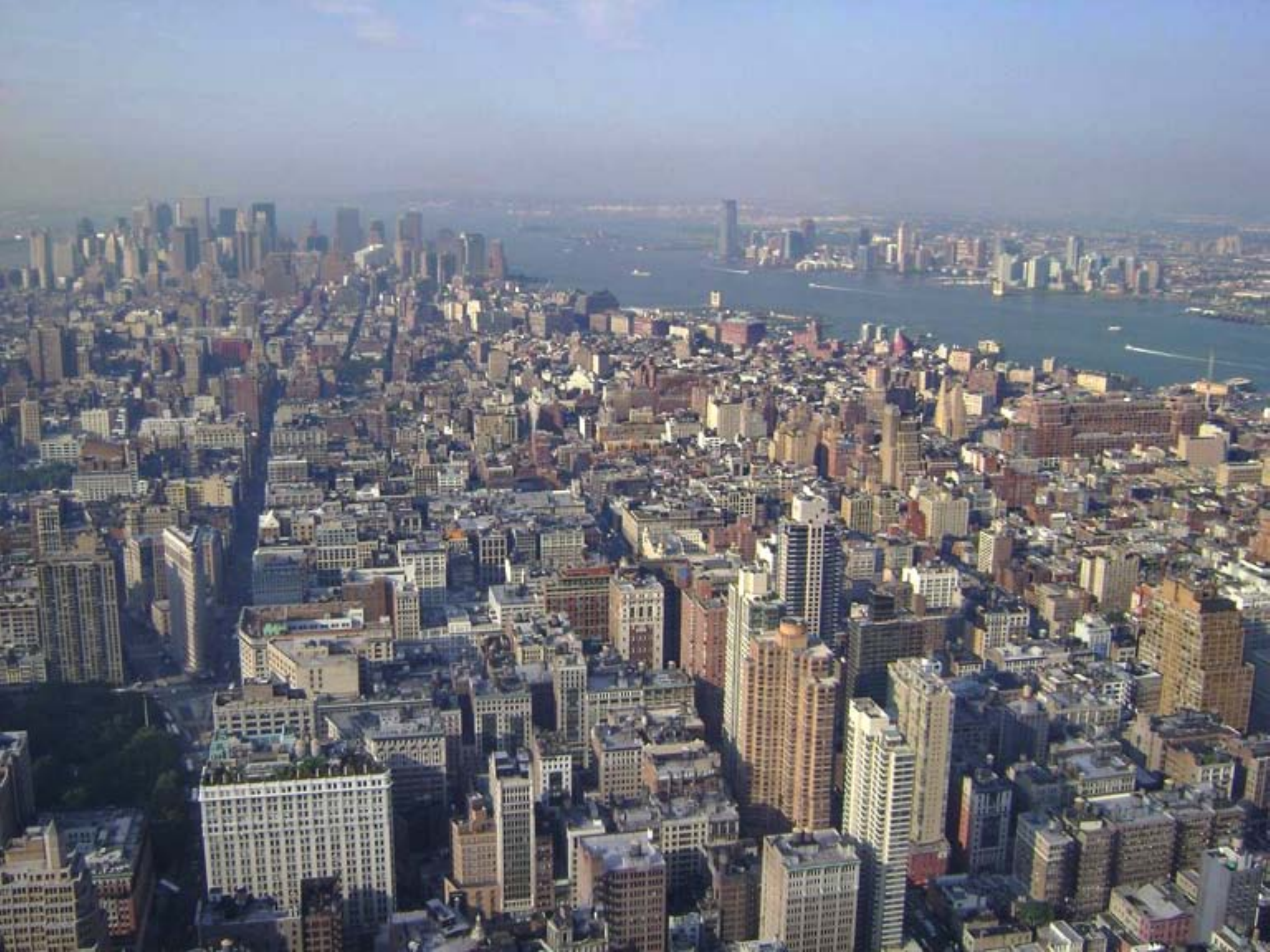} & \hspace{-0.4cm}
			\includegraphics[width = 0.105\textwidth]{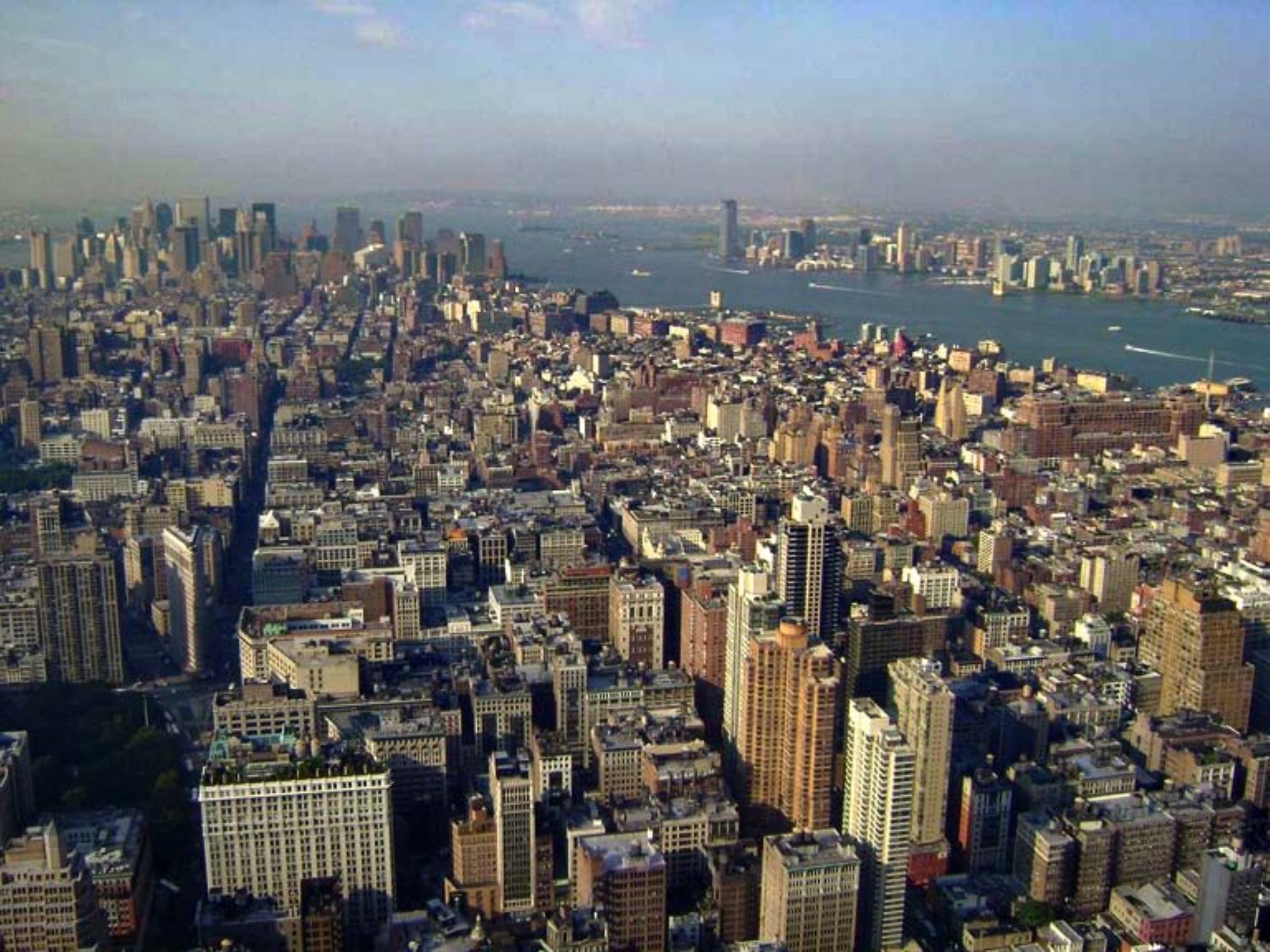} & \hspace{-0.4cm}
			\includegraphics[width = 0.105\textwidth]{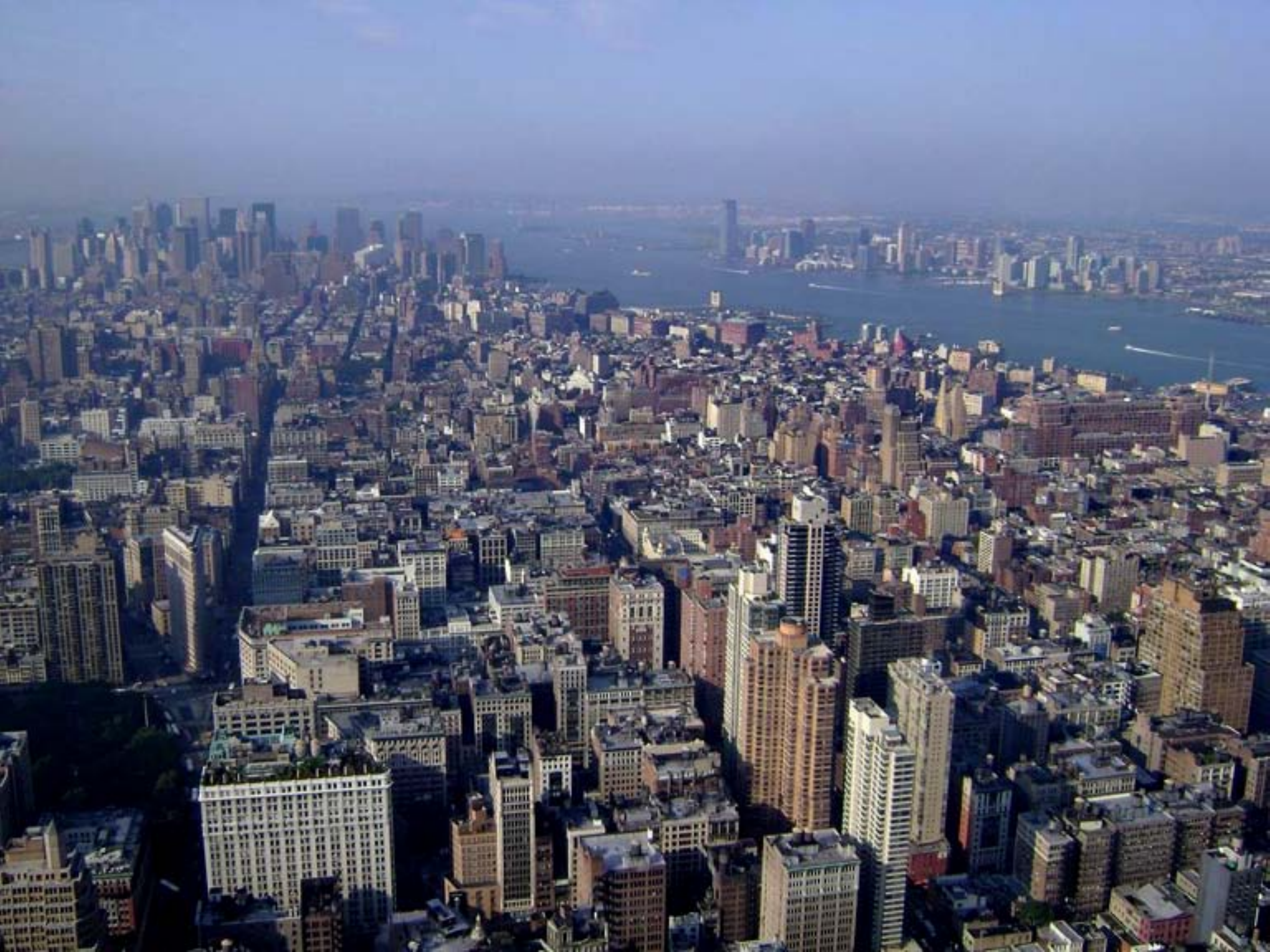} & \hspace{-0.4cm}
			\includegraphics[width = 0.105\textwidth]{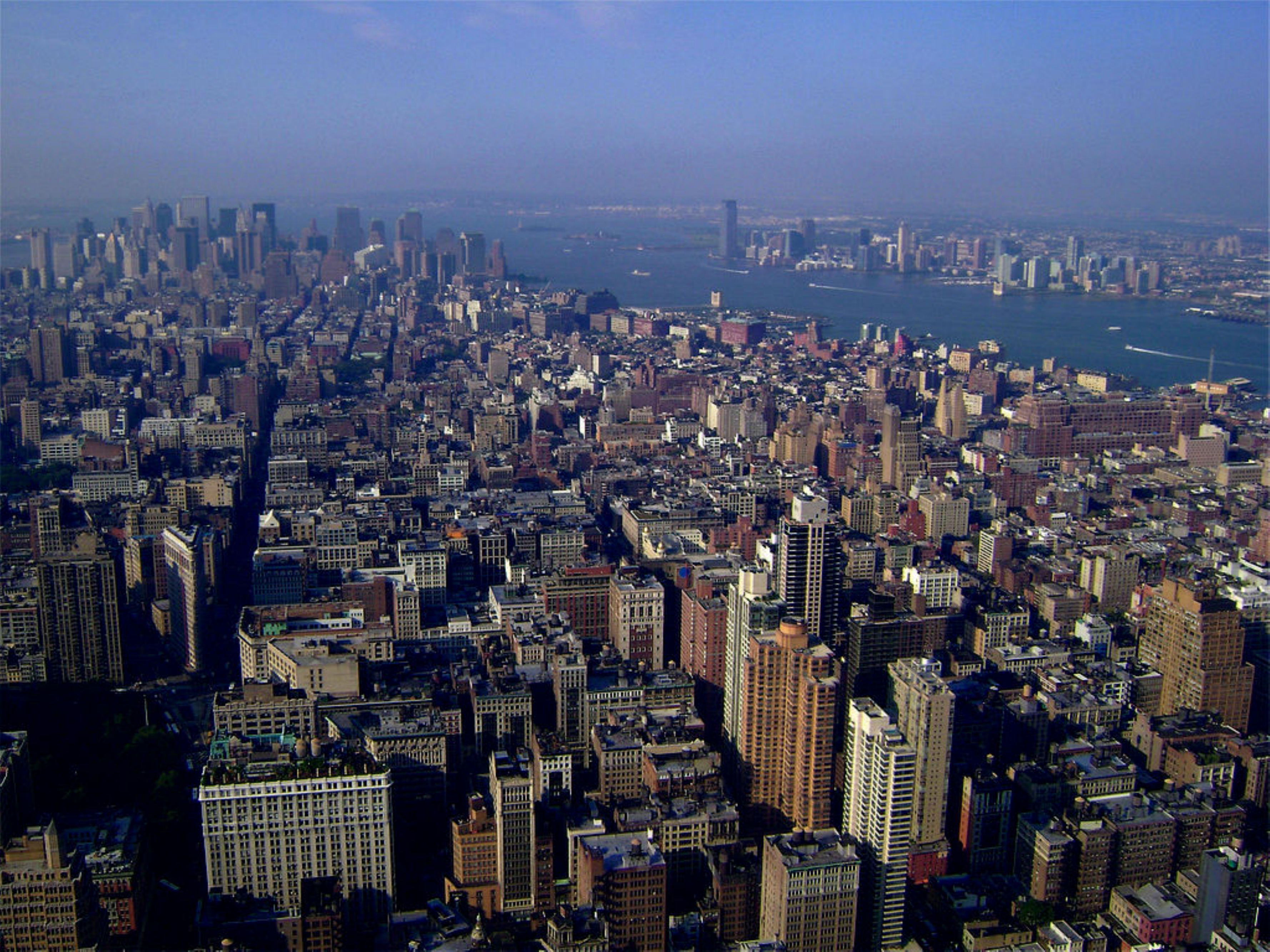} & \hspace{-0.4cm}
			\includegraphics[width = 0.105\textwidth]{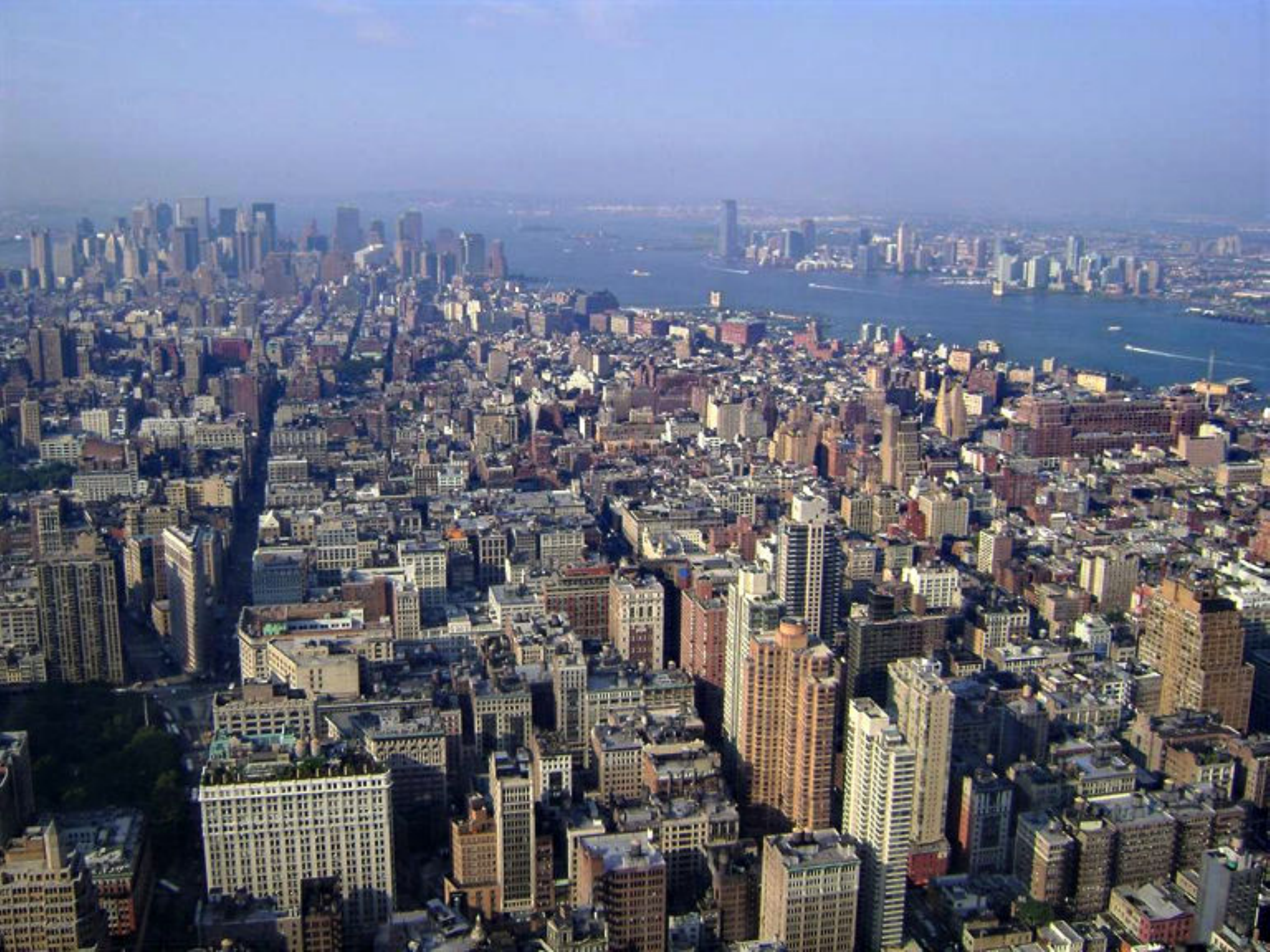}\\
			\includegraphics[width = 0.105\textwidth]{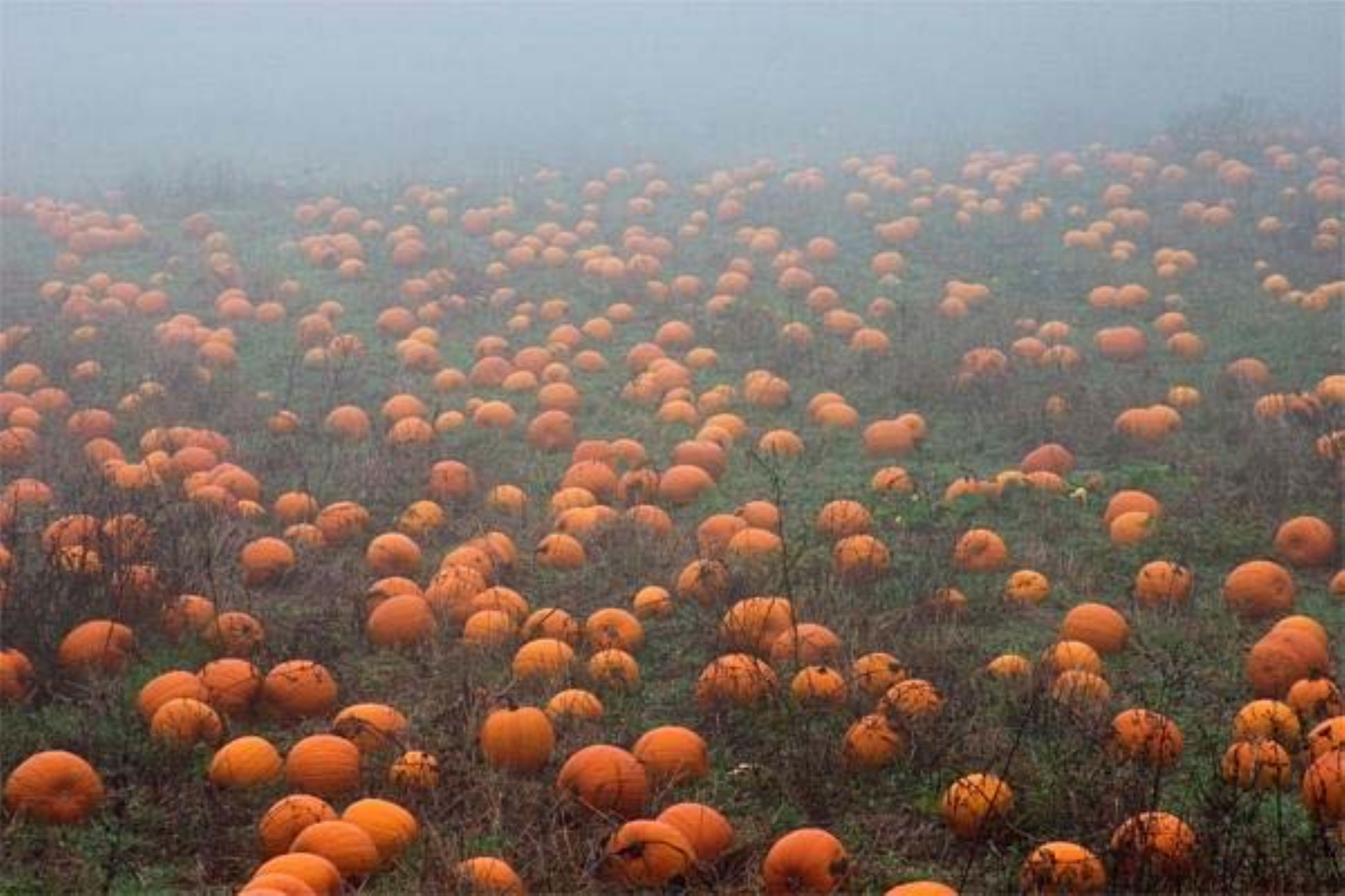} & \hspace{-0.4cm}
			\includegraphics[width = 0.105\textwidth]{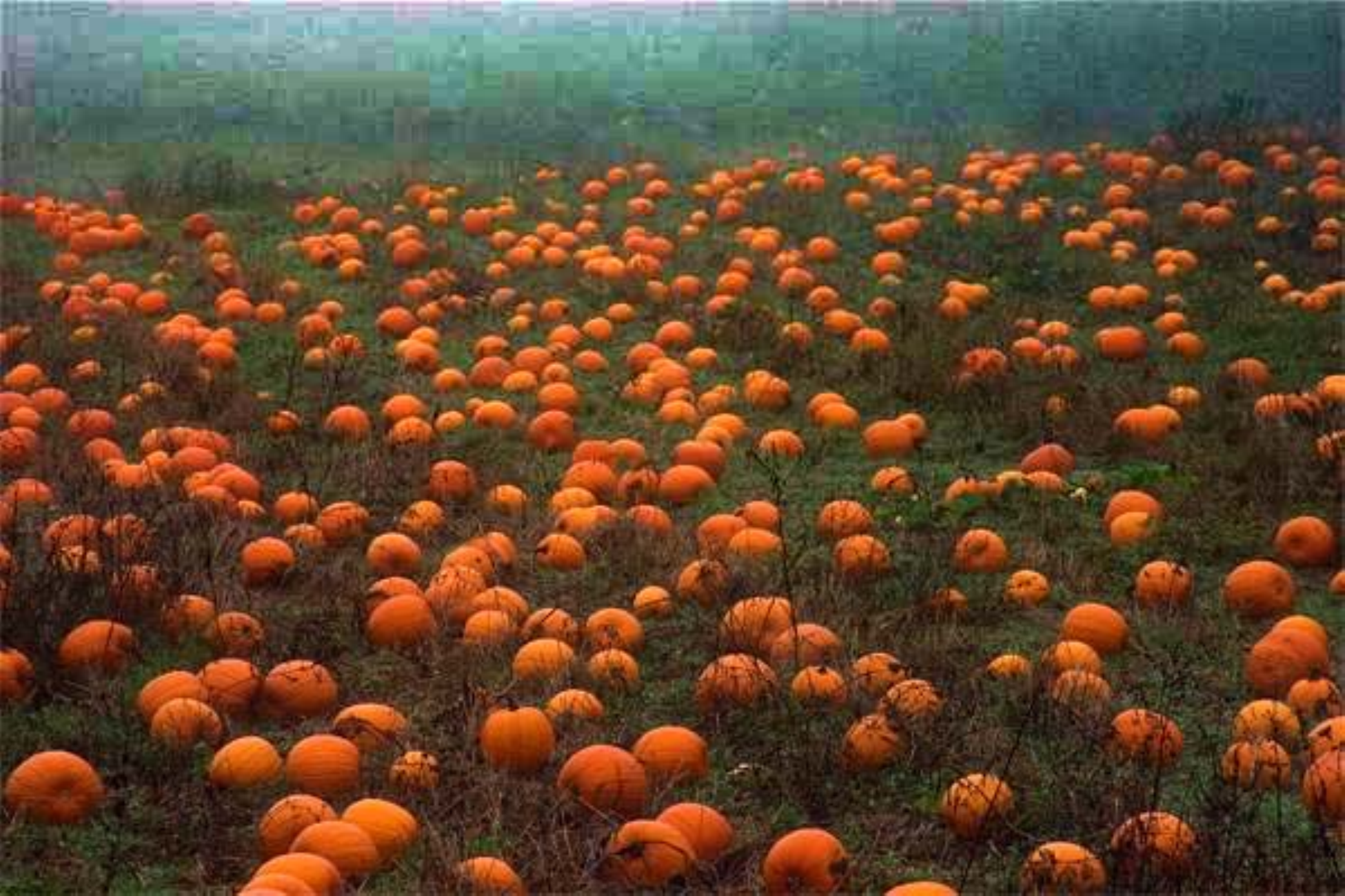} & \hspace{-0.4cm}
			\includegraphics[width = 0.105\textwidth]{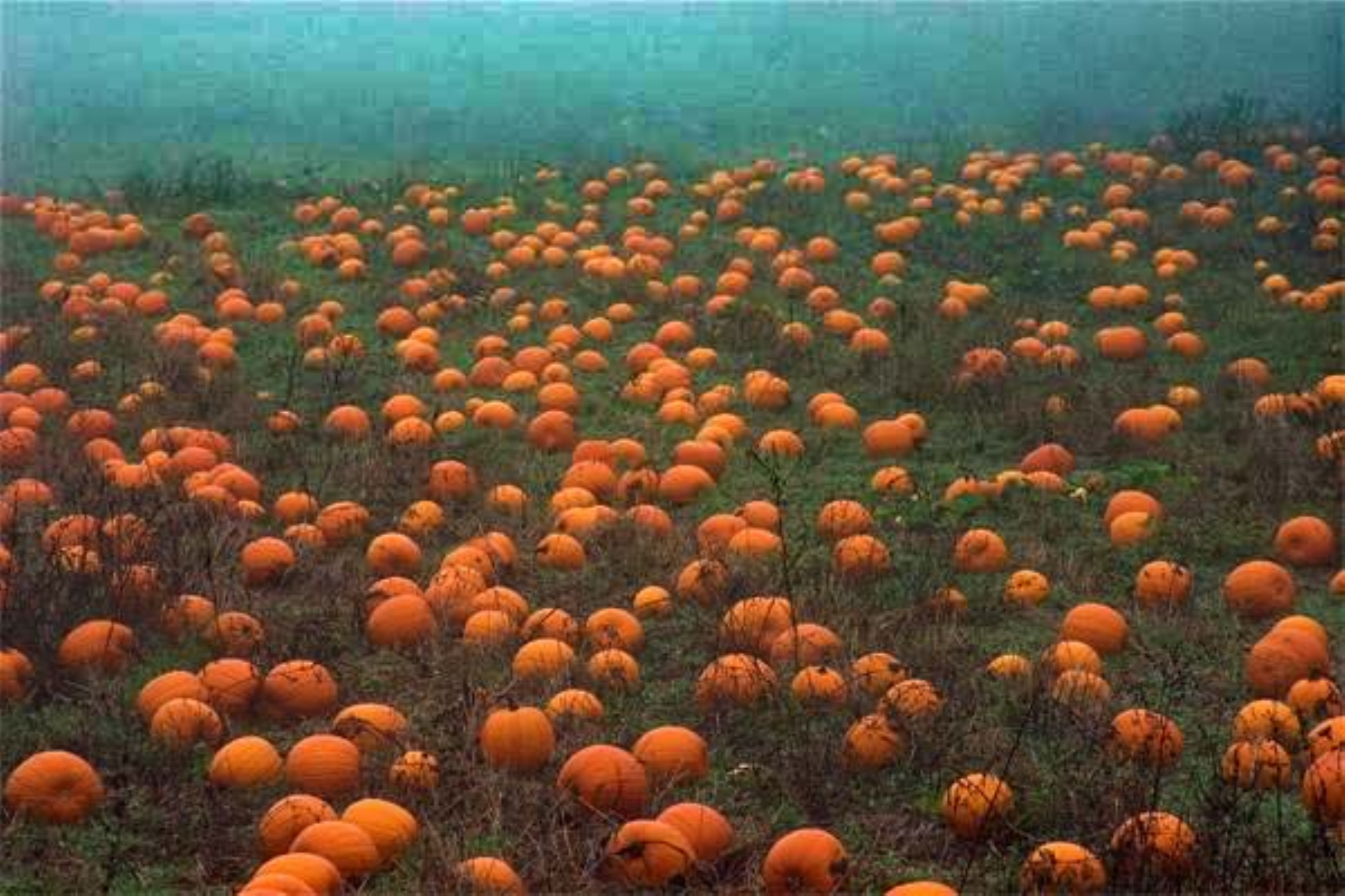} & \hspace{-0.4cm}
			\includegraphics[width = 0.105\textwidth]{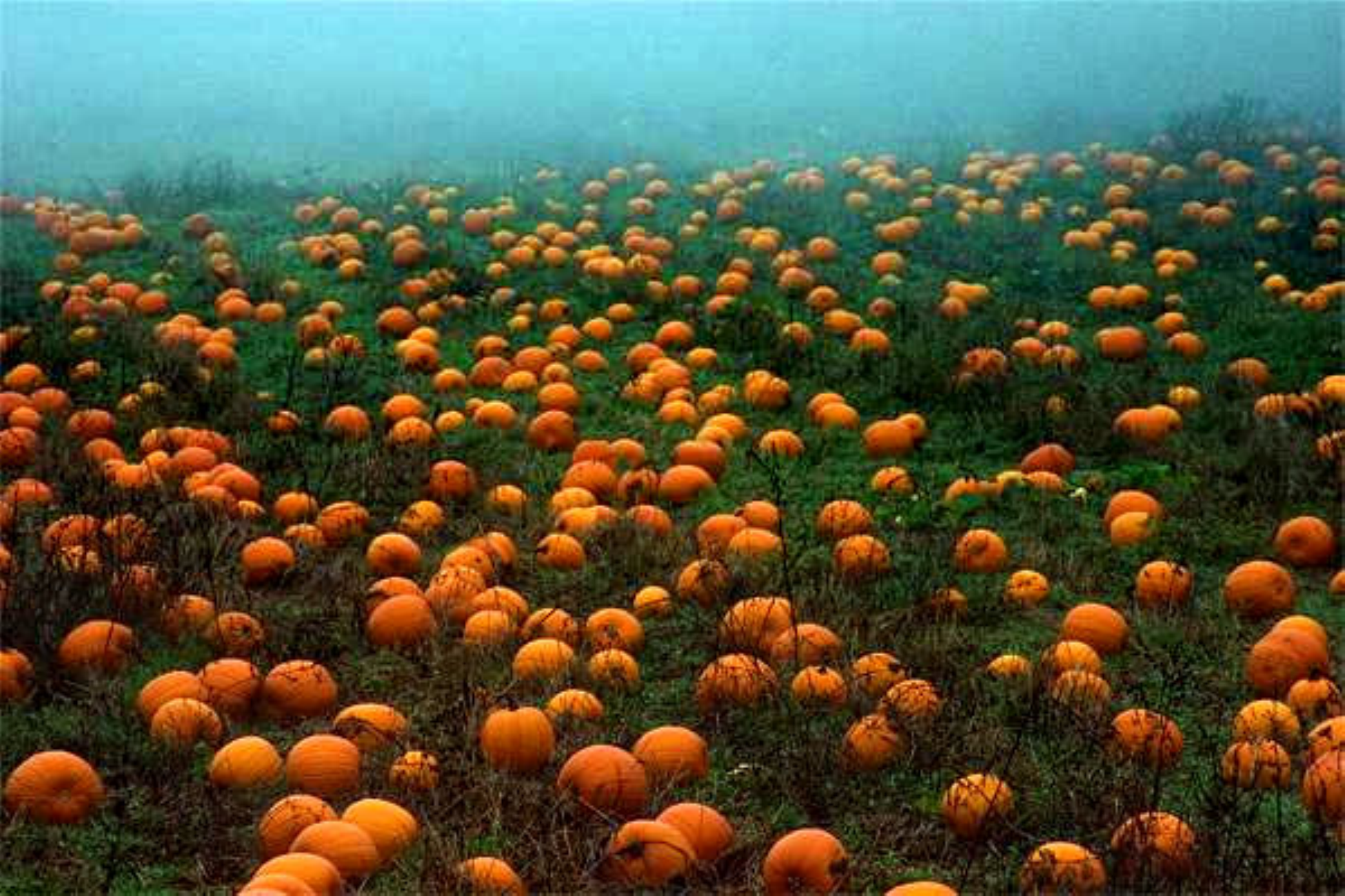} & \hspace{-0.4cm}
			\includegraphics[width = 0.105\textwidth]{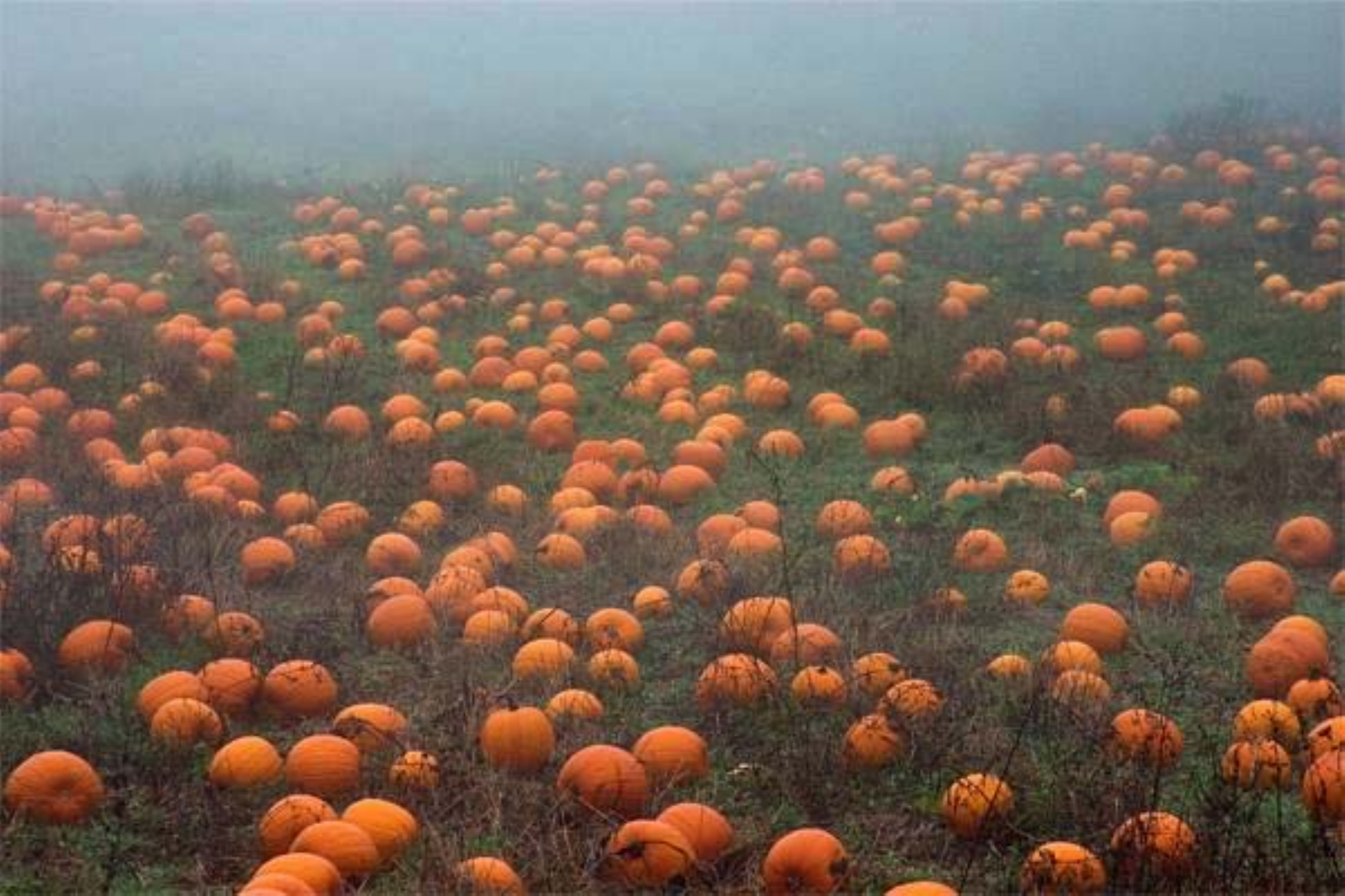} & \hspace{-0.4cm}
			\includegraphics[width = 0.105\textwidth]{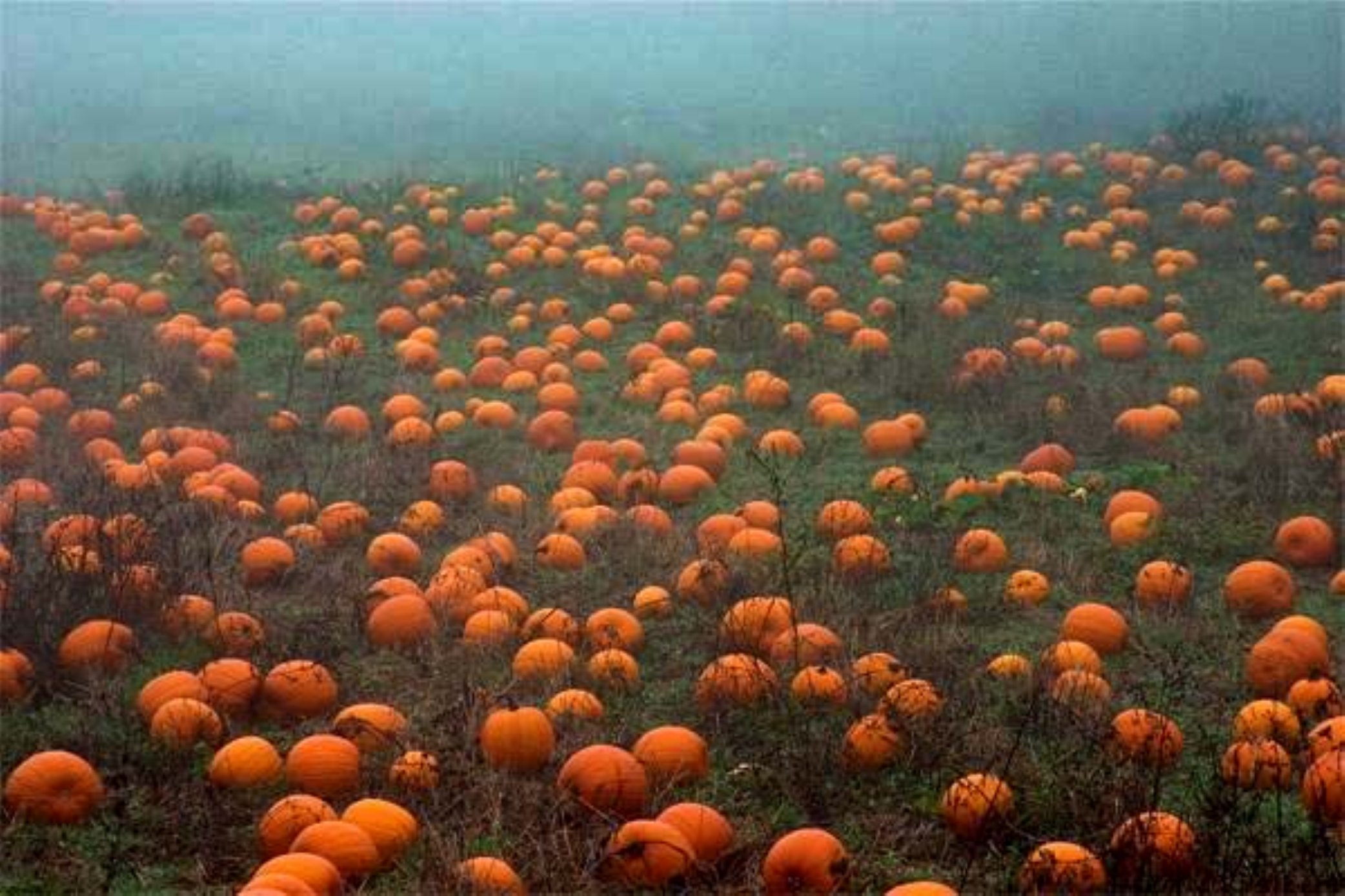} & \hspace{-0.4cm}
			\includegraphics[width = 0.105\textwidth]{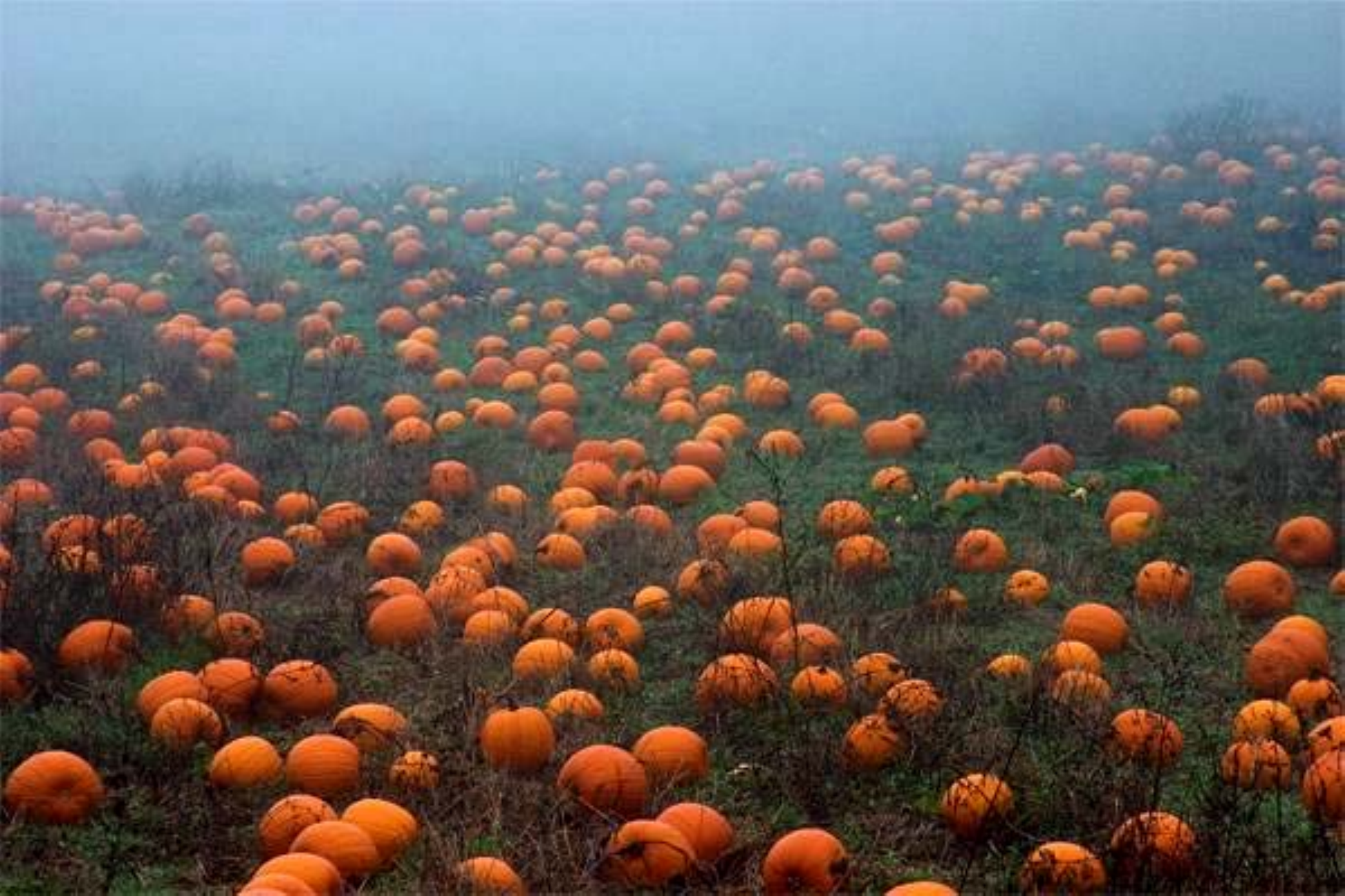} & \hspace{-0.4cm}
			\includegraphics[width = 0.105\textwidth]{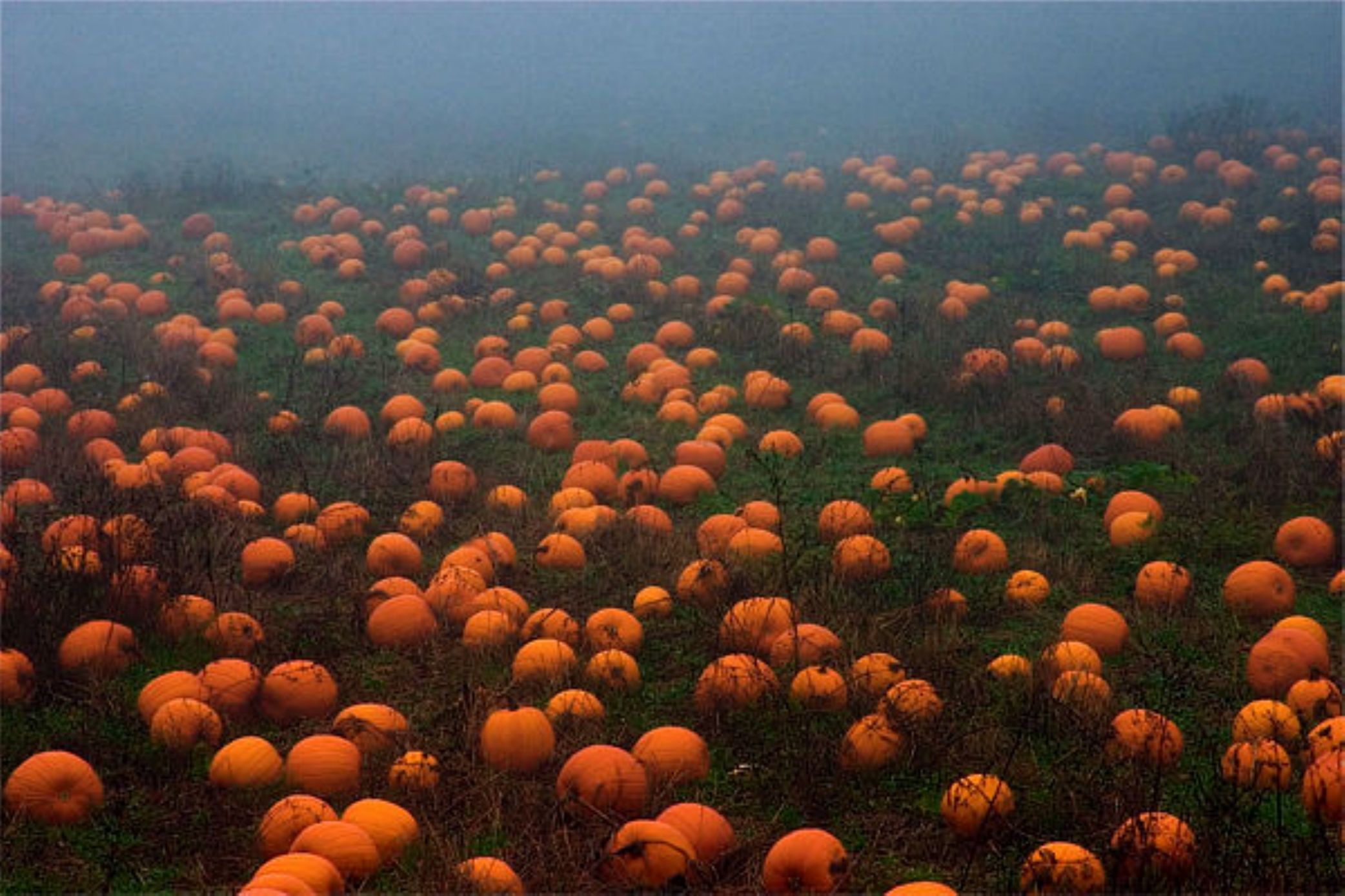} & \hspace{-0.4cm}
			\includegraphics[width = 0.105\textwidth]{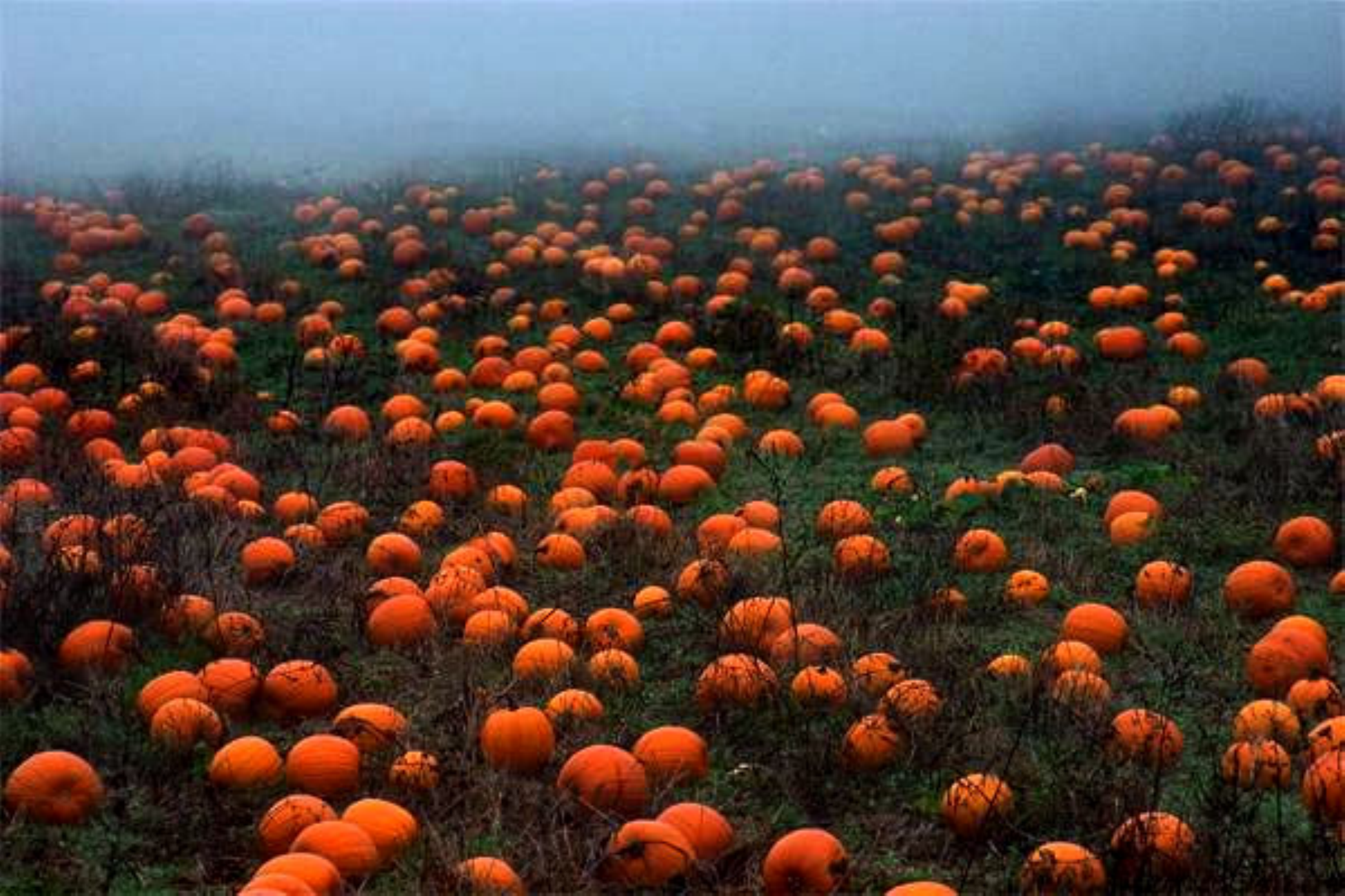}\\
			\includegraphics[width = 0.105\textwidth]{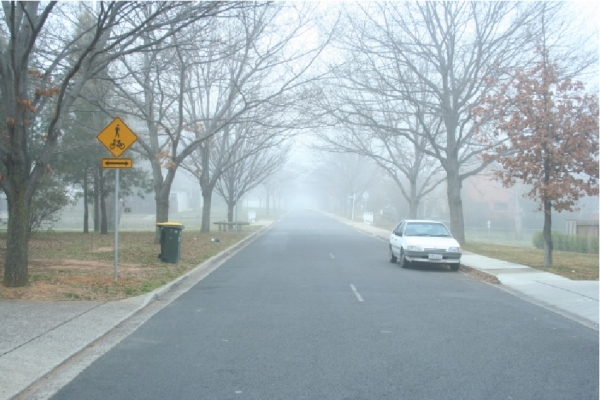} & \hspace{-0.4cm}
			\includegraphics[width = 0.105\textwidth]{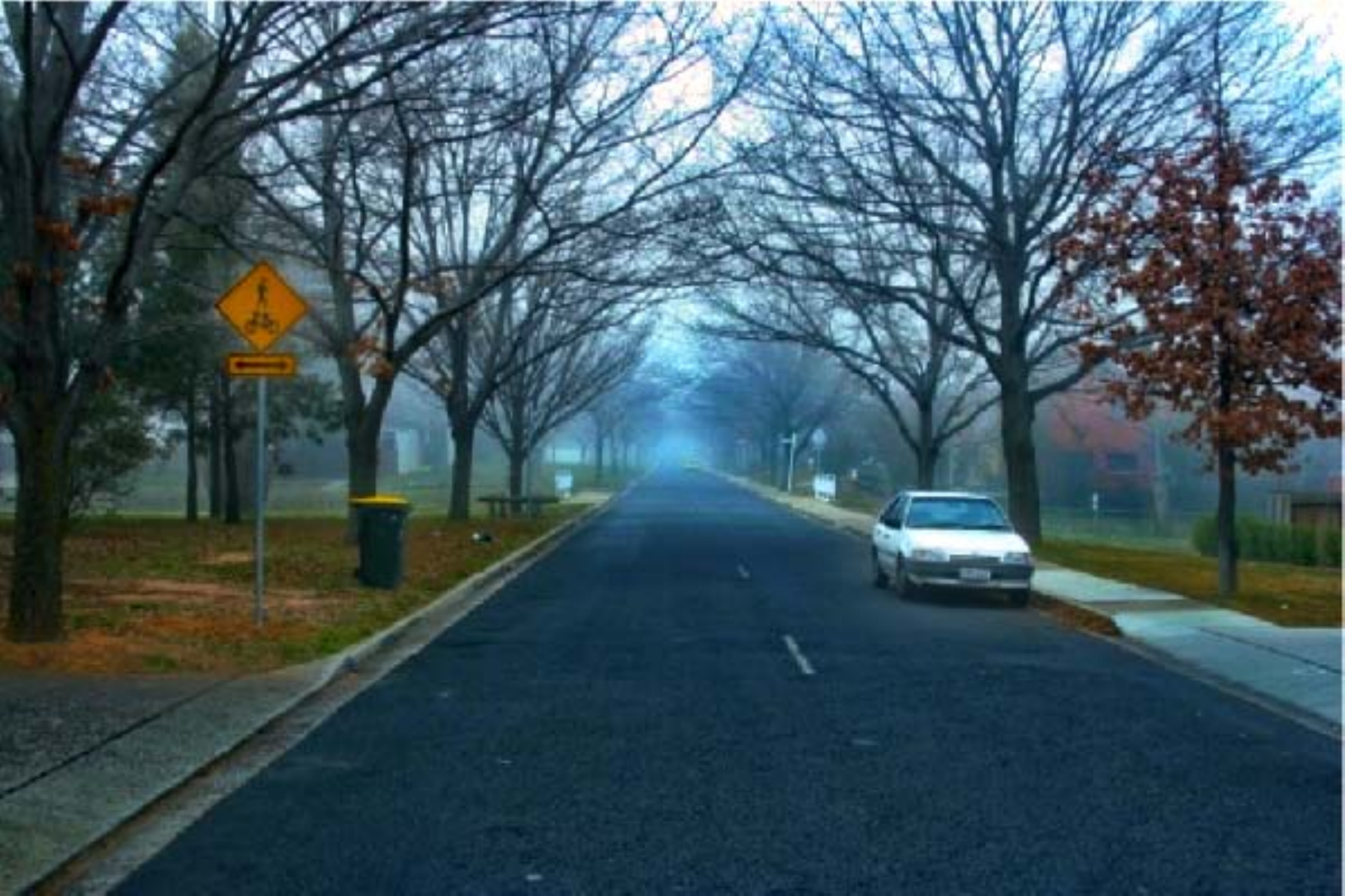} & \hspace{-0.4cm}
			\includegraphics[width = 0.105\textwidth]{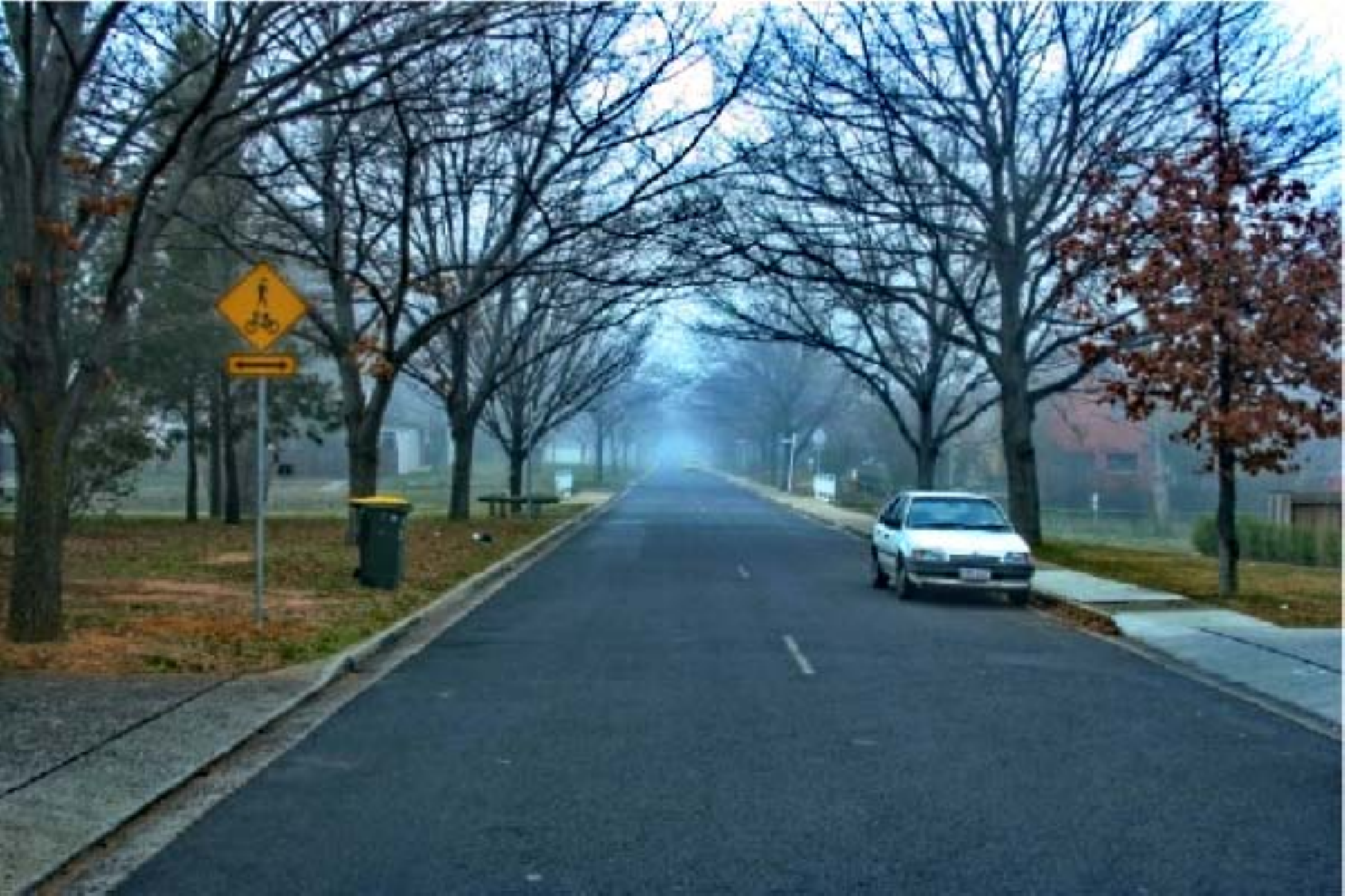} & \hspace{-0.4cm}
			\includegraphics[width = 0.105\textwidth]{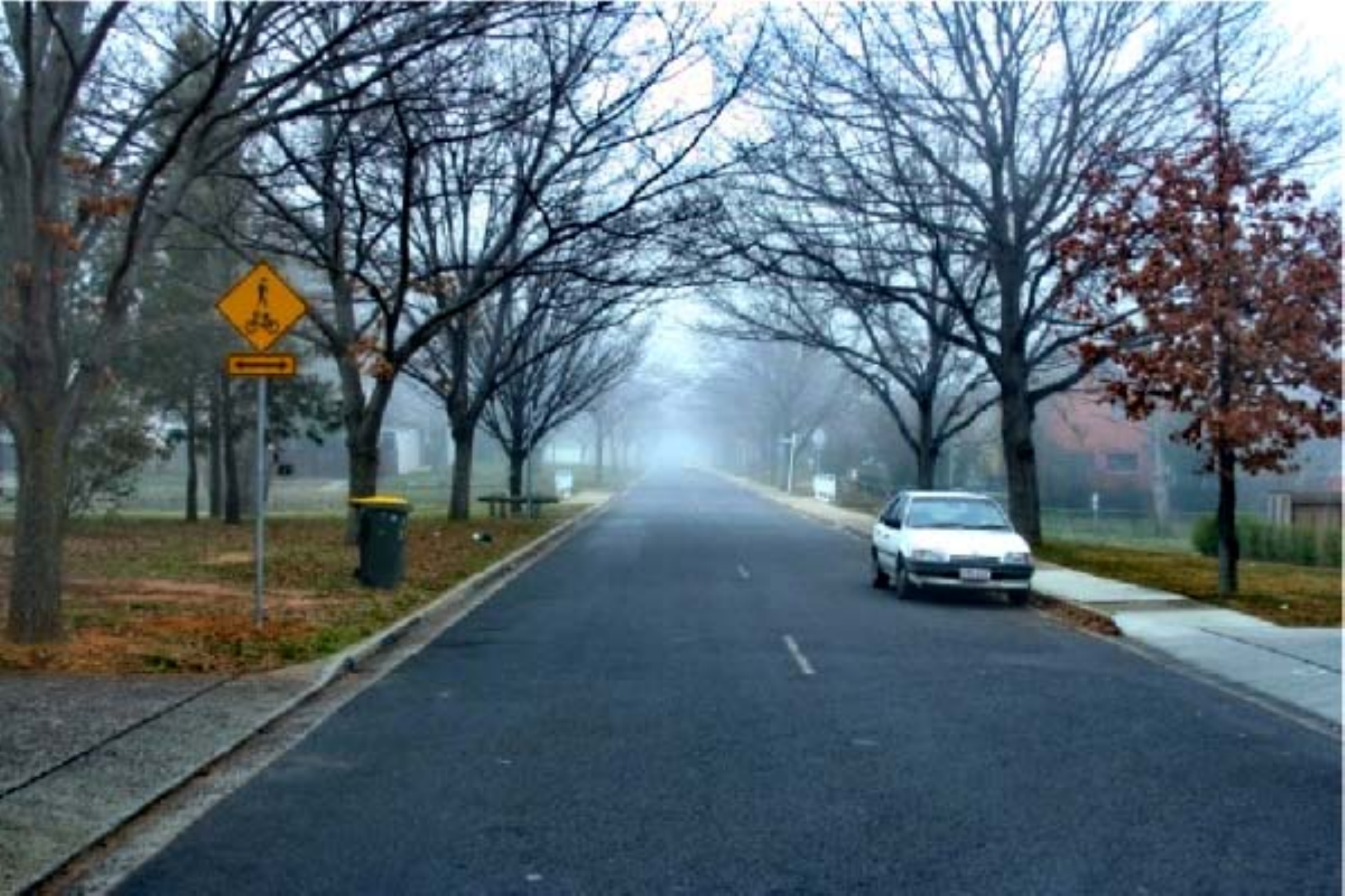} & \hspace{-0.4cm}
			\includegraphics[width = 0.105\textwidth]{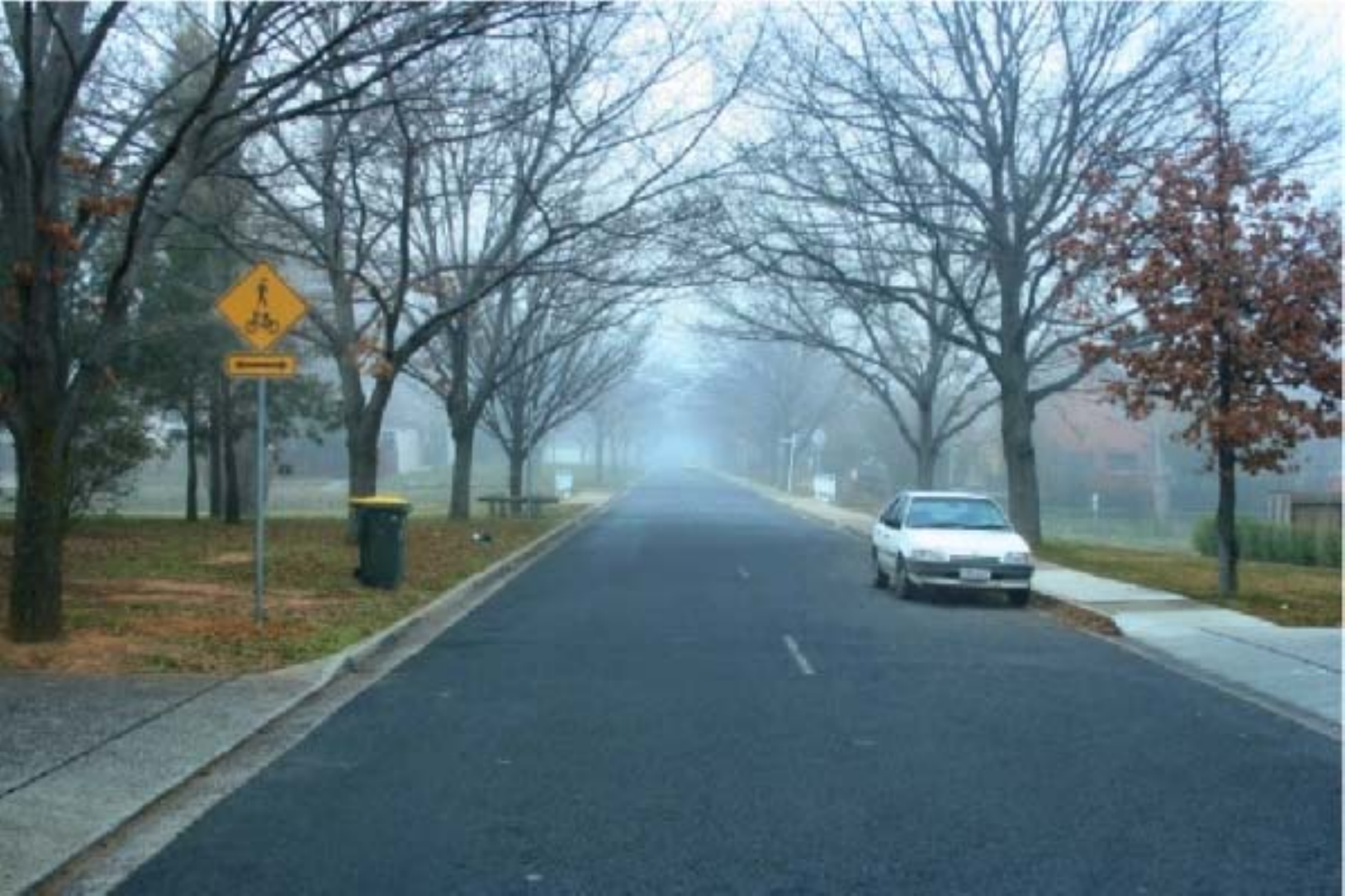} & \hspace{-0.4cm}
			\includegraphics[width = 0.105\textwidth]{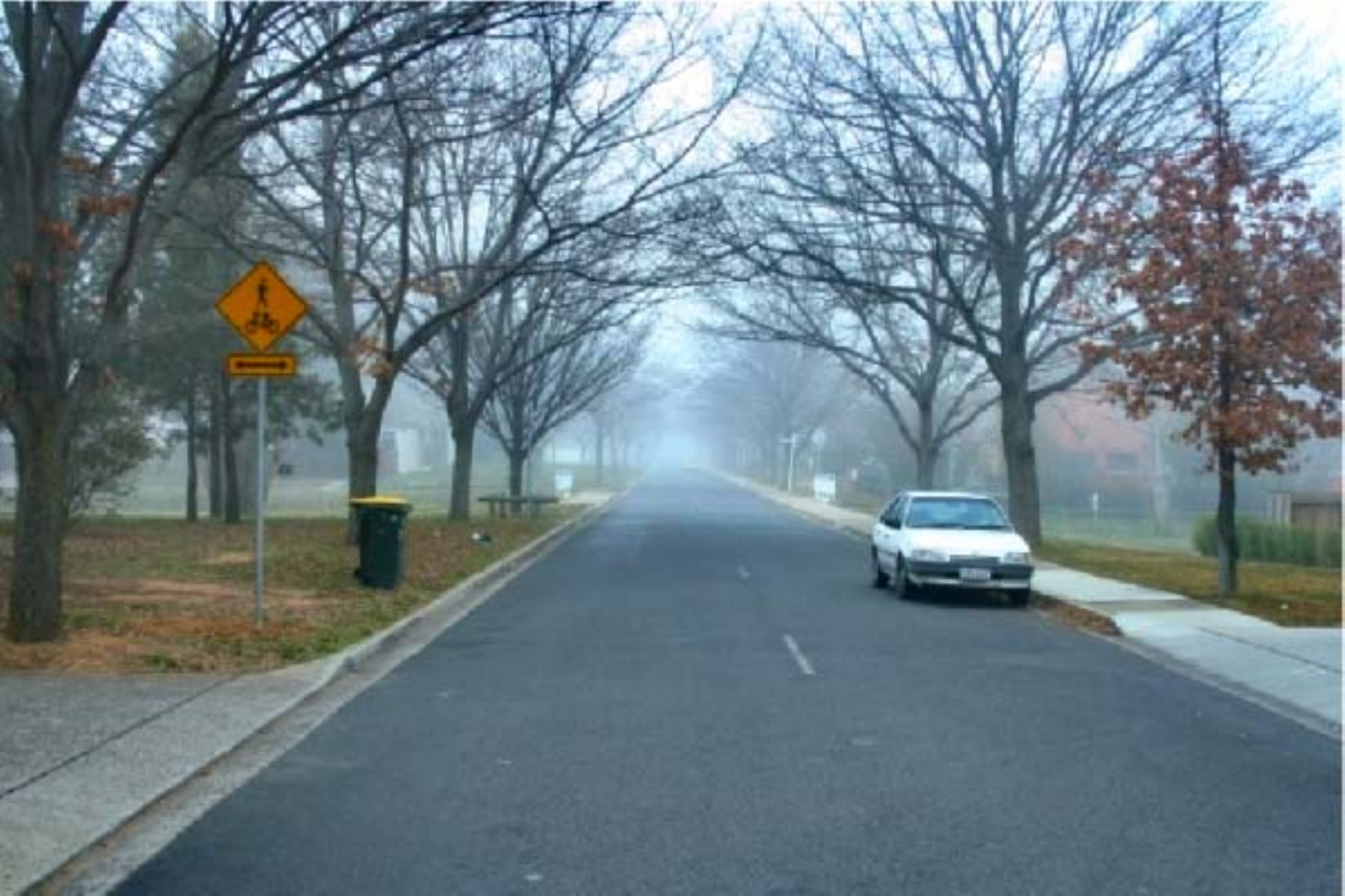} & \hspace{-0.4cm}
			\includegraphics[width = 0.105\textwidth]{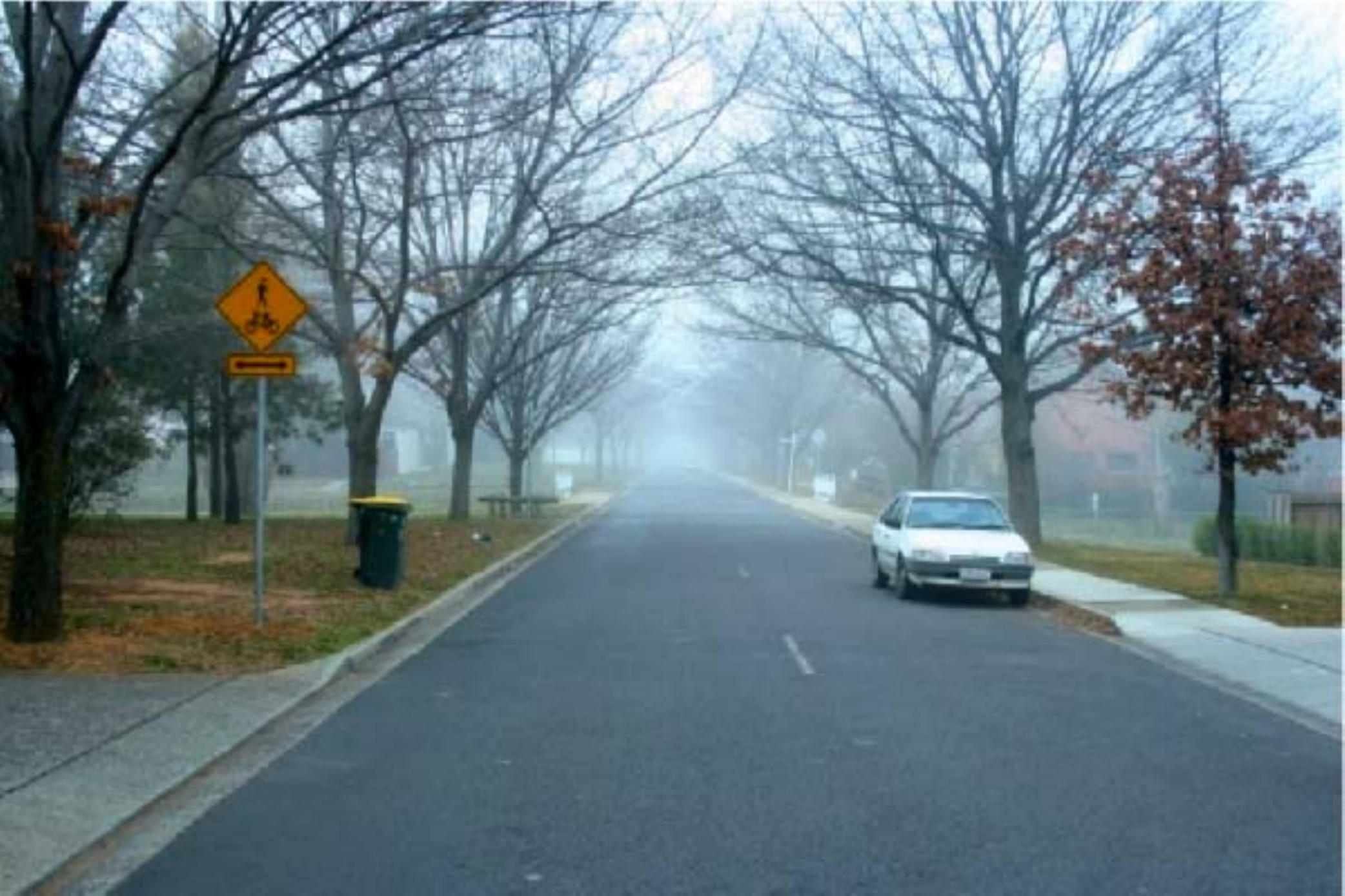} & \hspace{-0.4cm}
			\includegraphics[width = 0.105\textwidth]{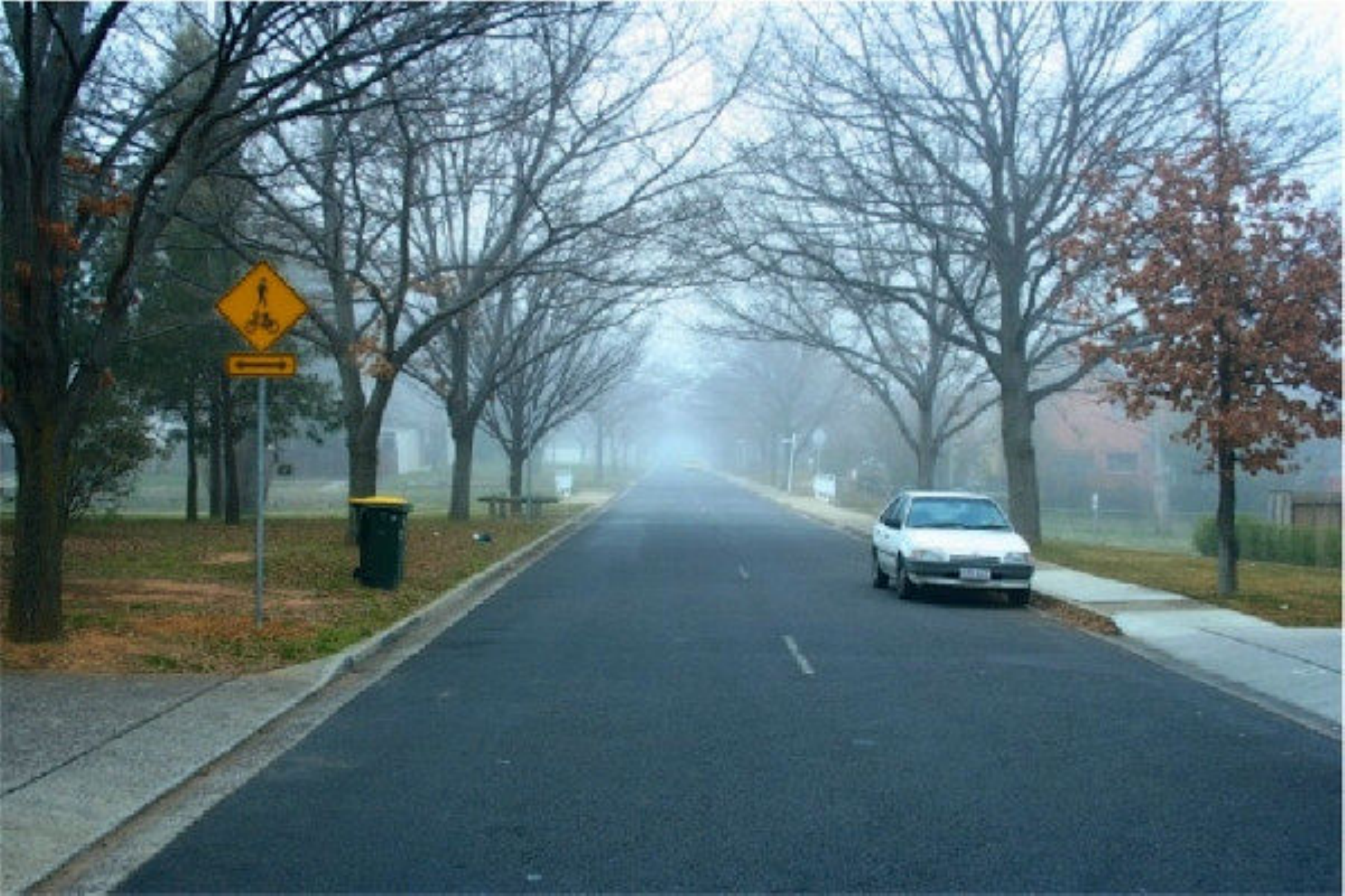} & \hspace{-0.4cm}
			\includegraphics[width = 0.105\textwidth]{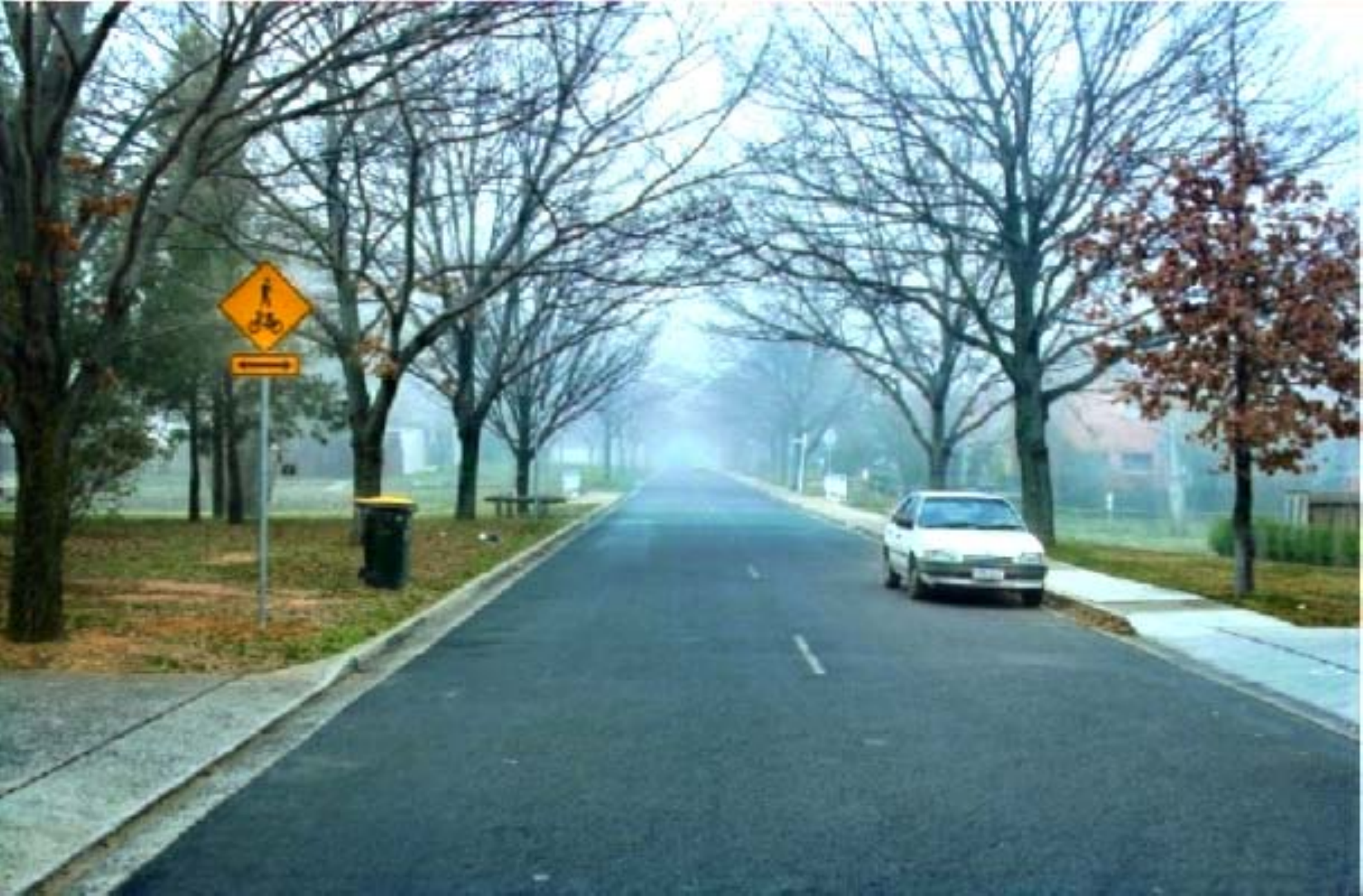}\\
			\includegraphics[width = 0.105\textwidth]{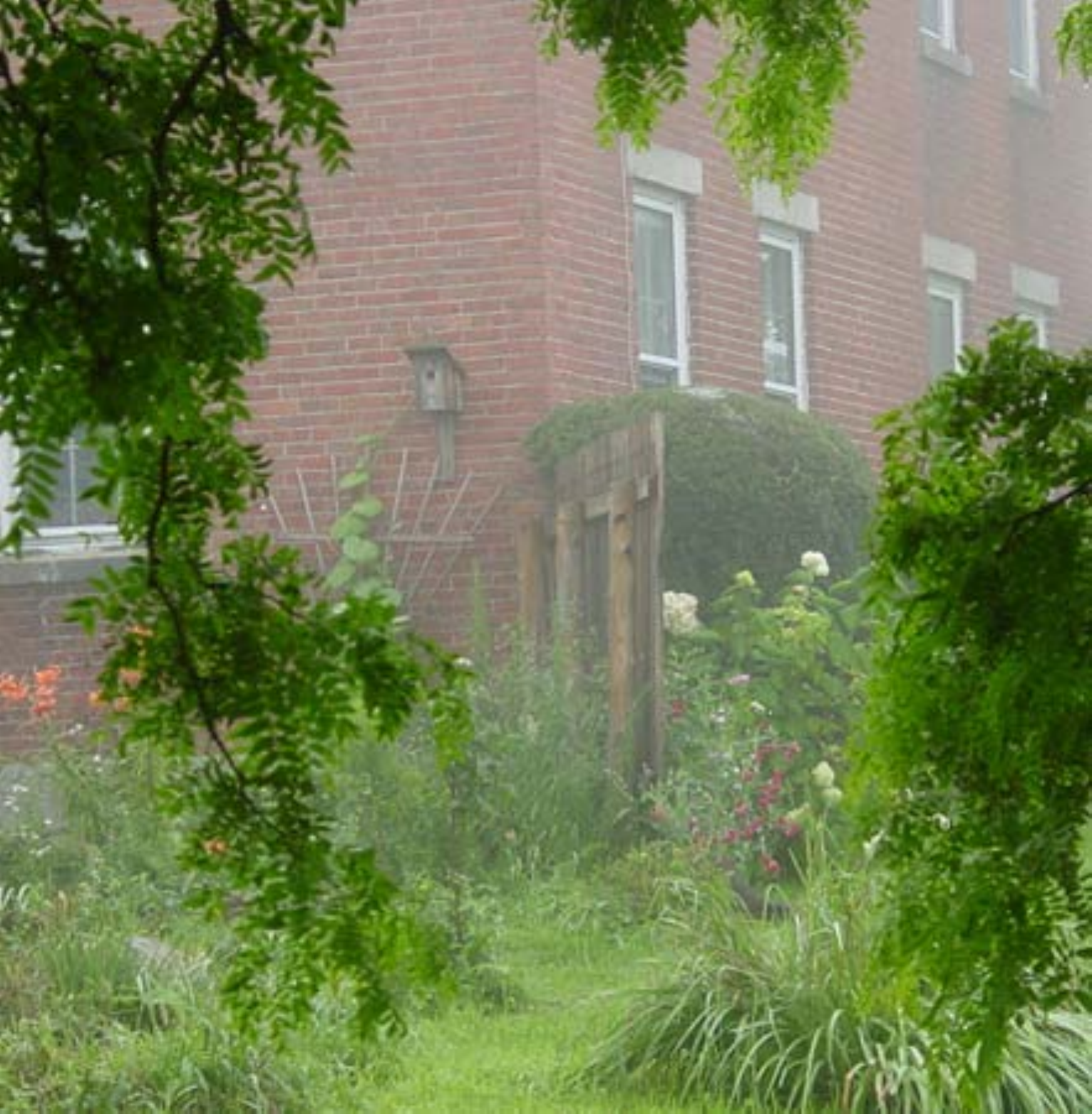} & \hspace{-0.4cm}
			\includegraphics[width = 0.105\textwidth]{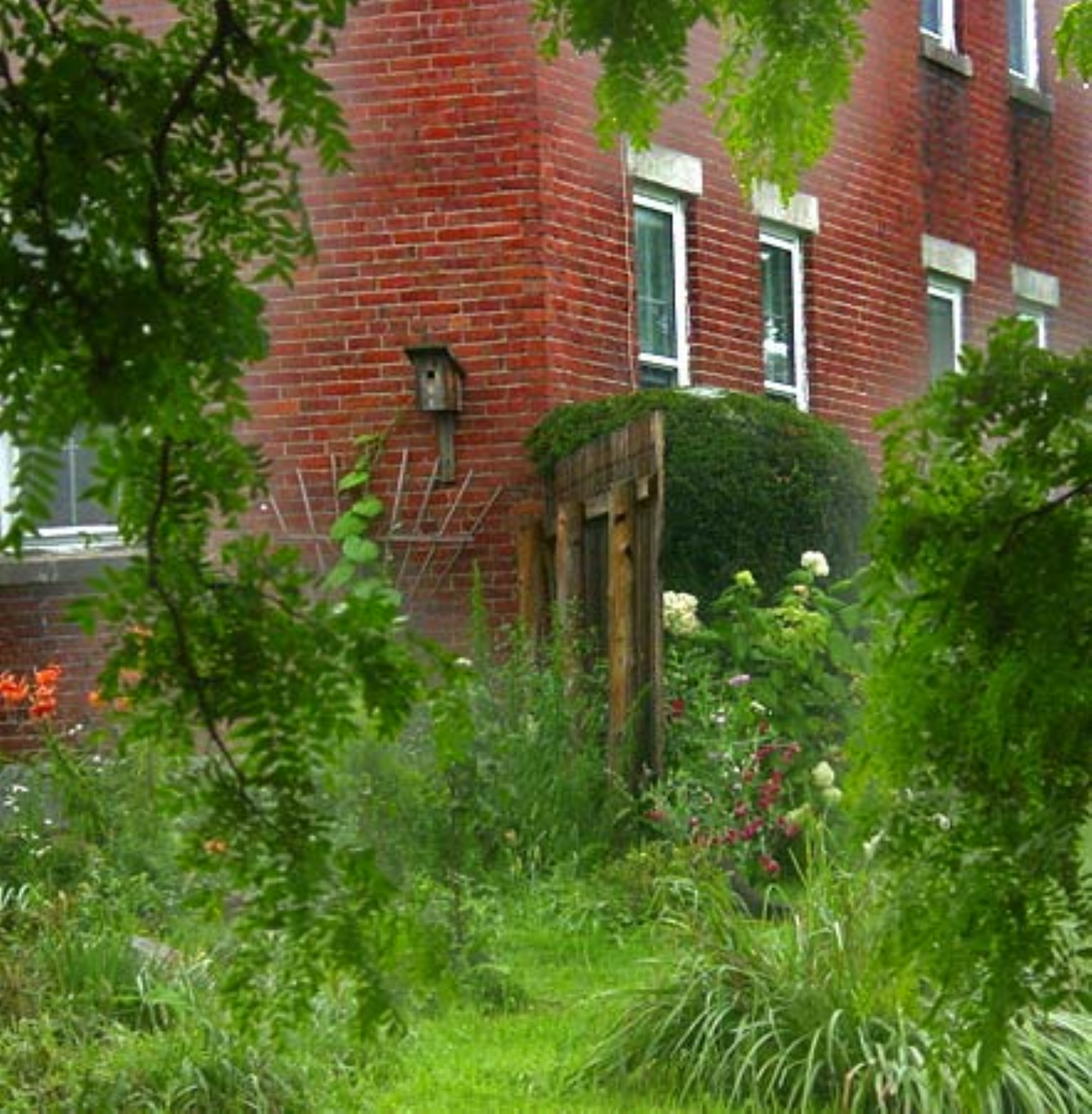} & \hspace{-0.4cm}
			\includegraphics[width = 0.105\textwidth]{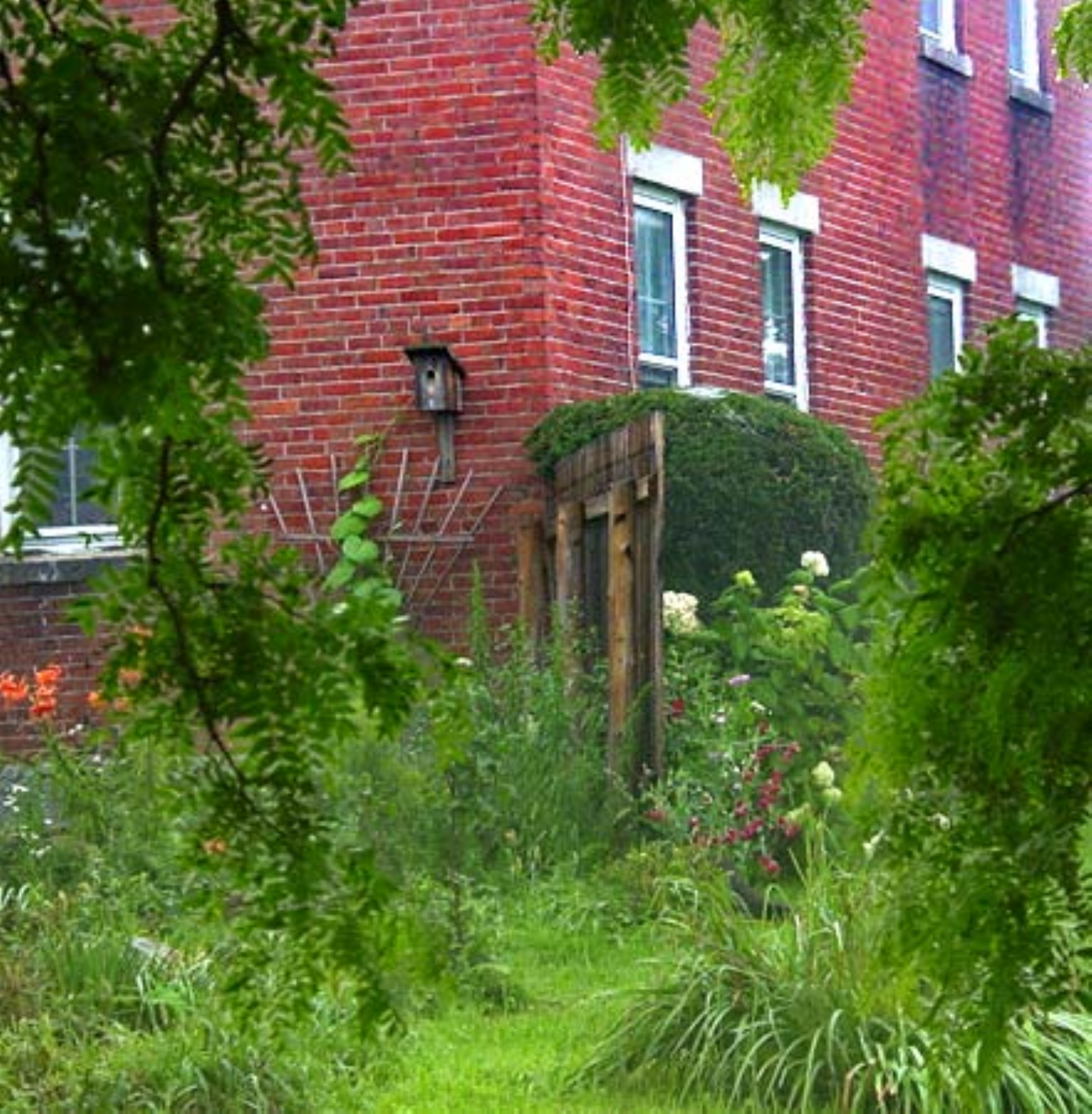} & \hspace{-0.4cm}
			\includegraphics[width = 0.105\textwidth]{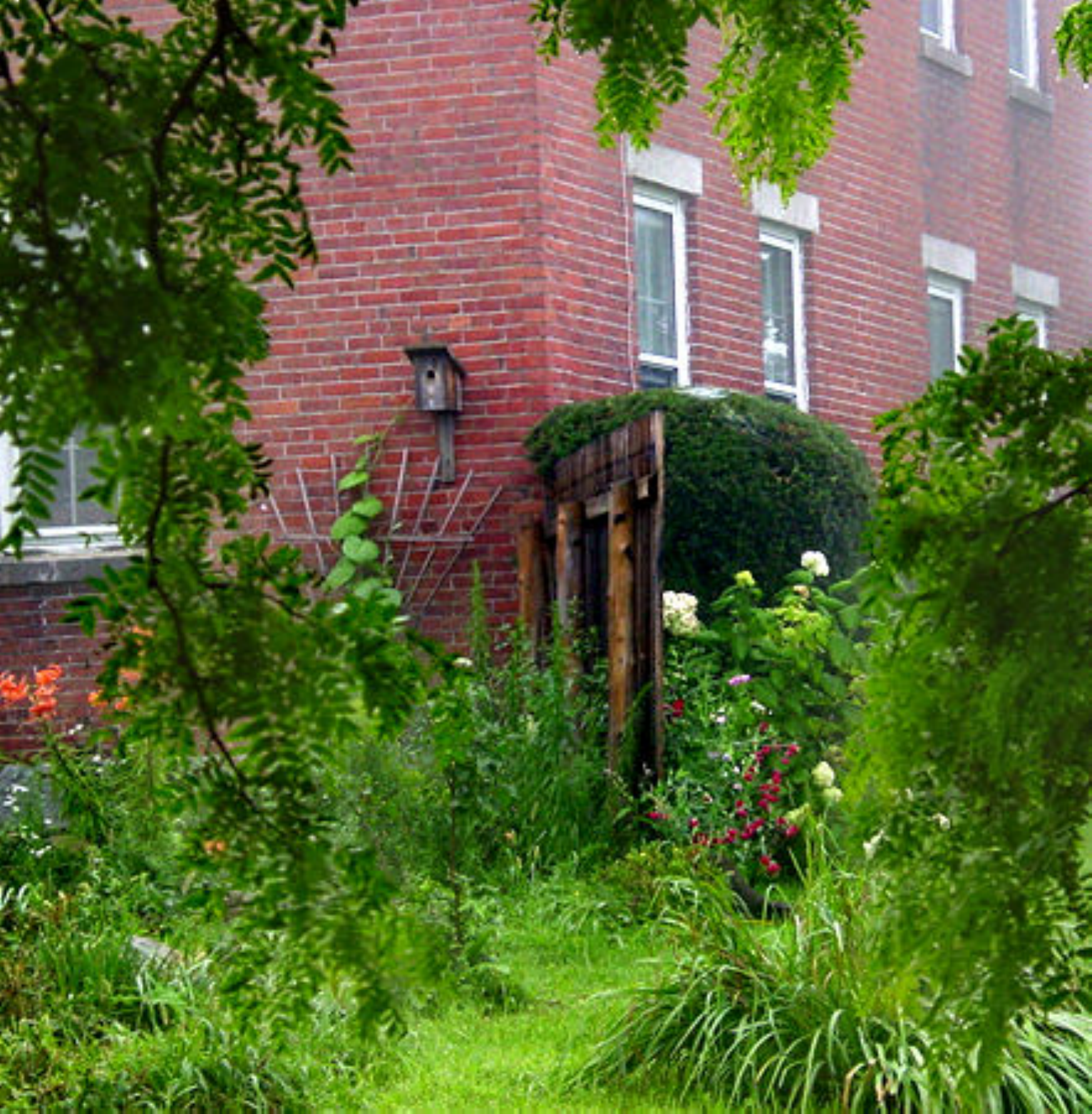} & \hspace{-0.4cm}
			\includegraphics[width = 0.105\textwidth]{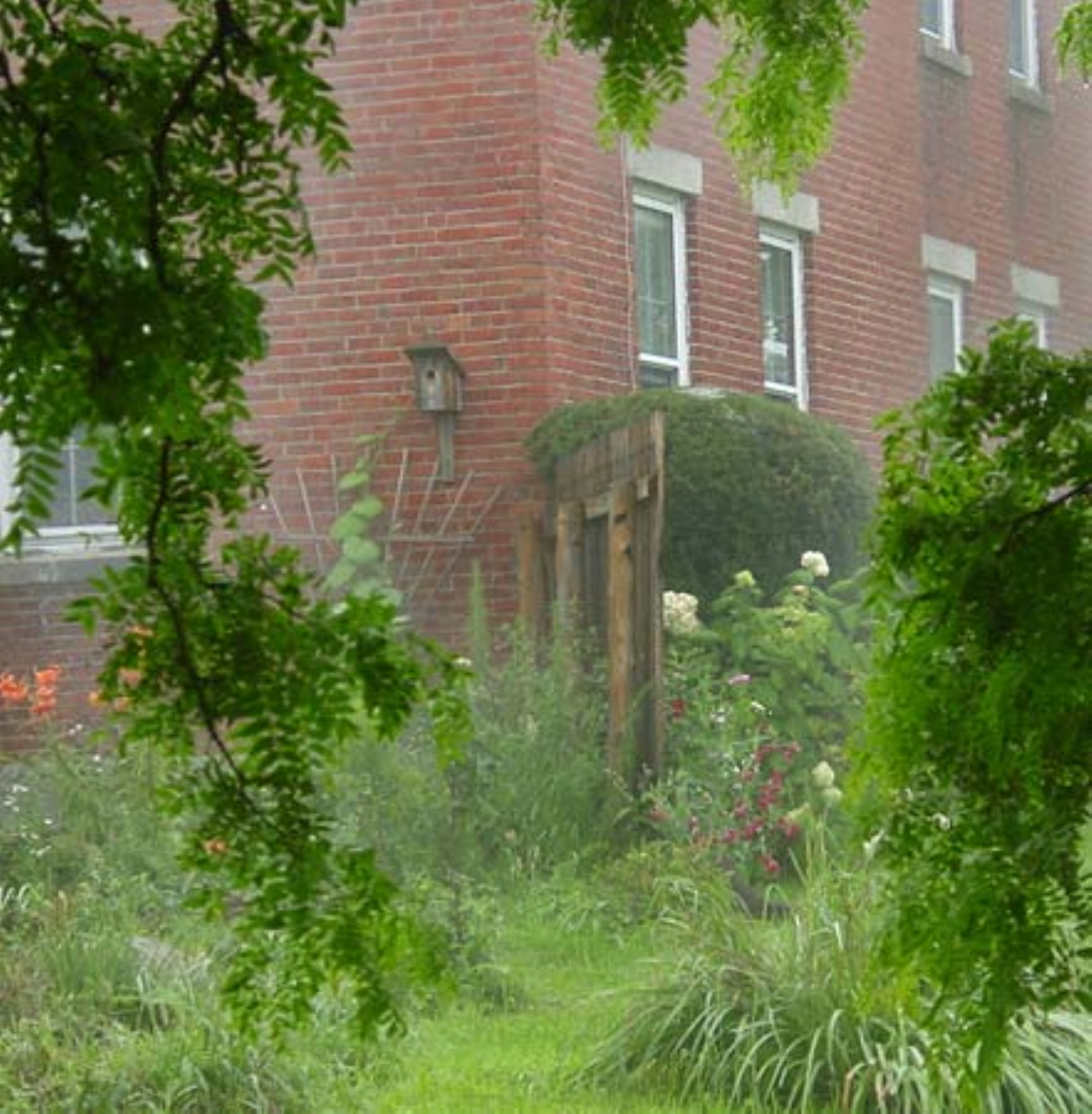} & \hspace{-0.4cm}
			\includegraphics[width = 0.105\textwidth]{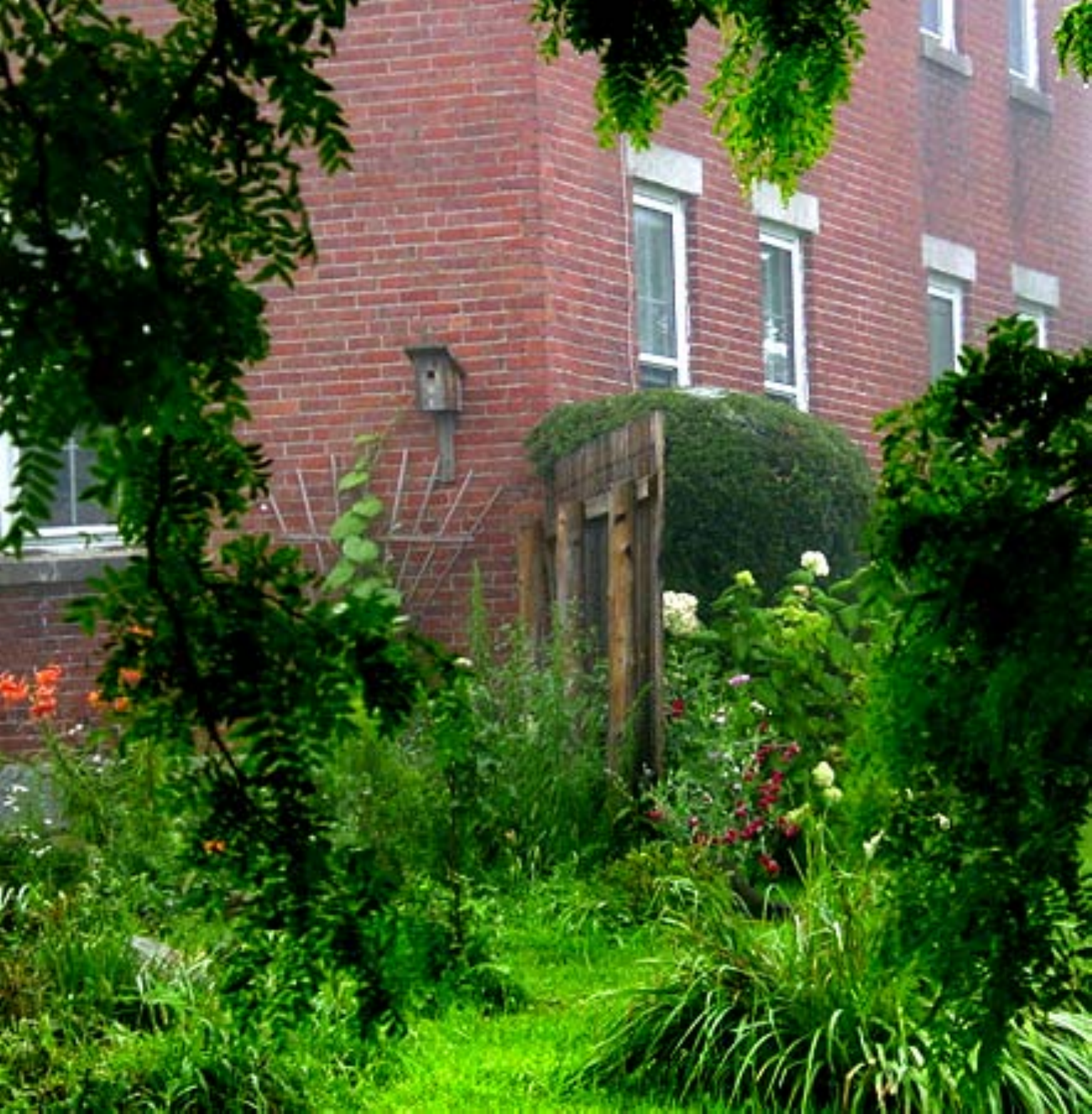} & \hspace{-0.4cm}
			\includegraphics[width = 0.105\textwidth]{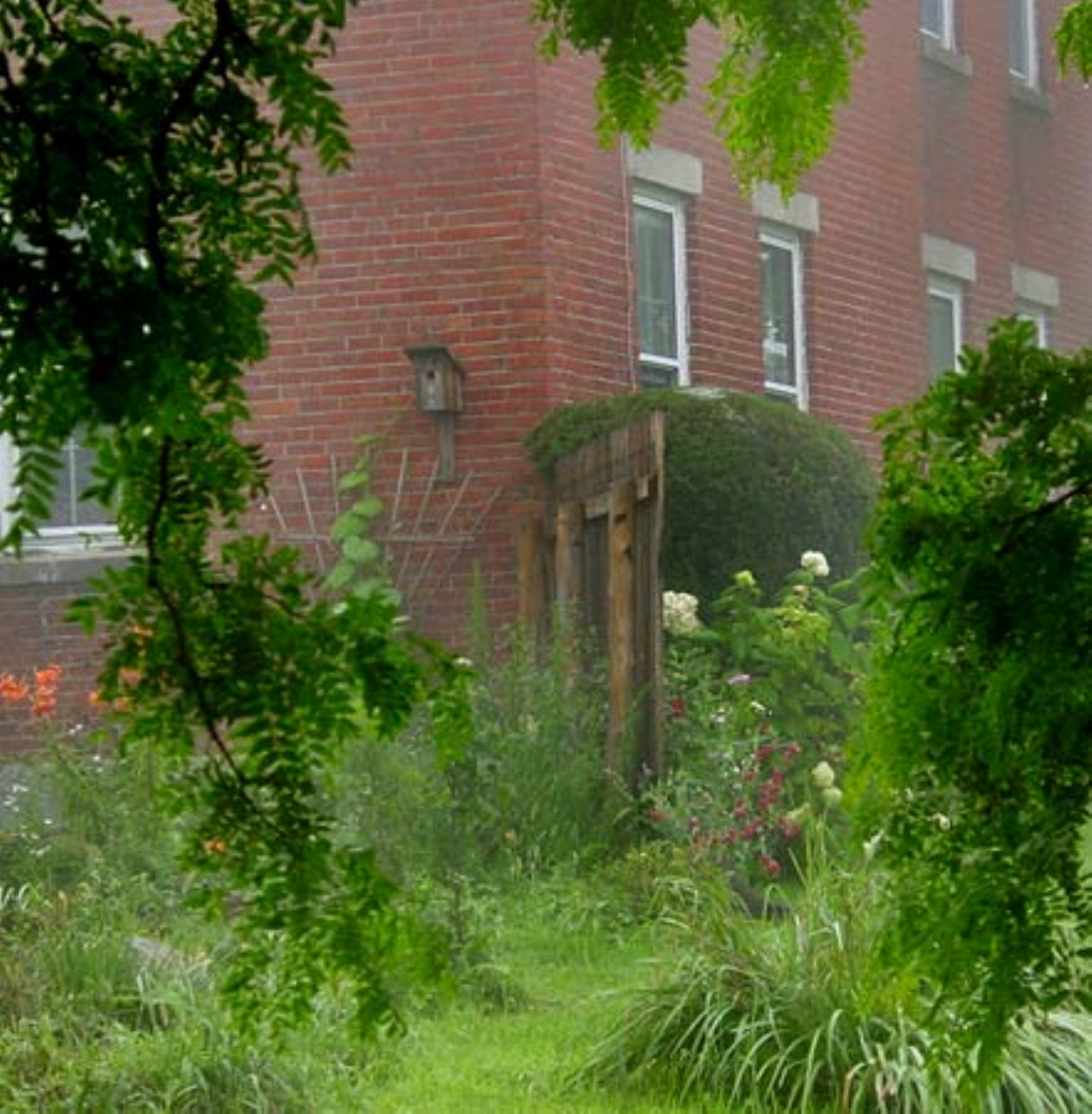} & \hspace{-0.4cm}
			\includegraphics[width = 0.105\textwidth]{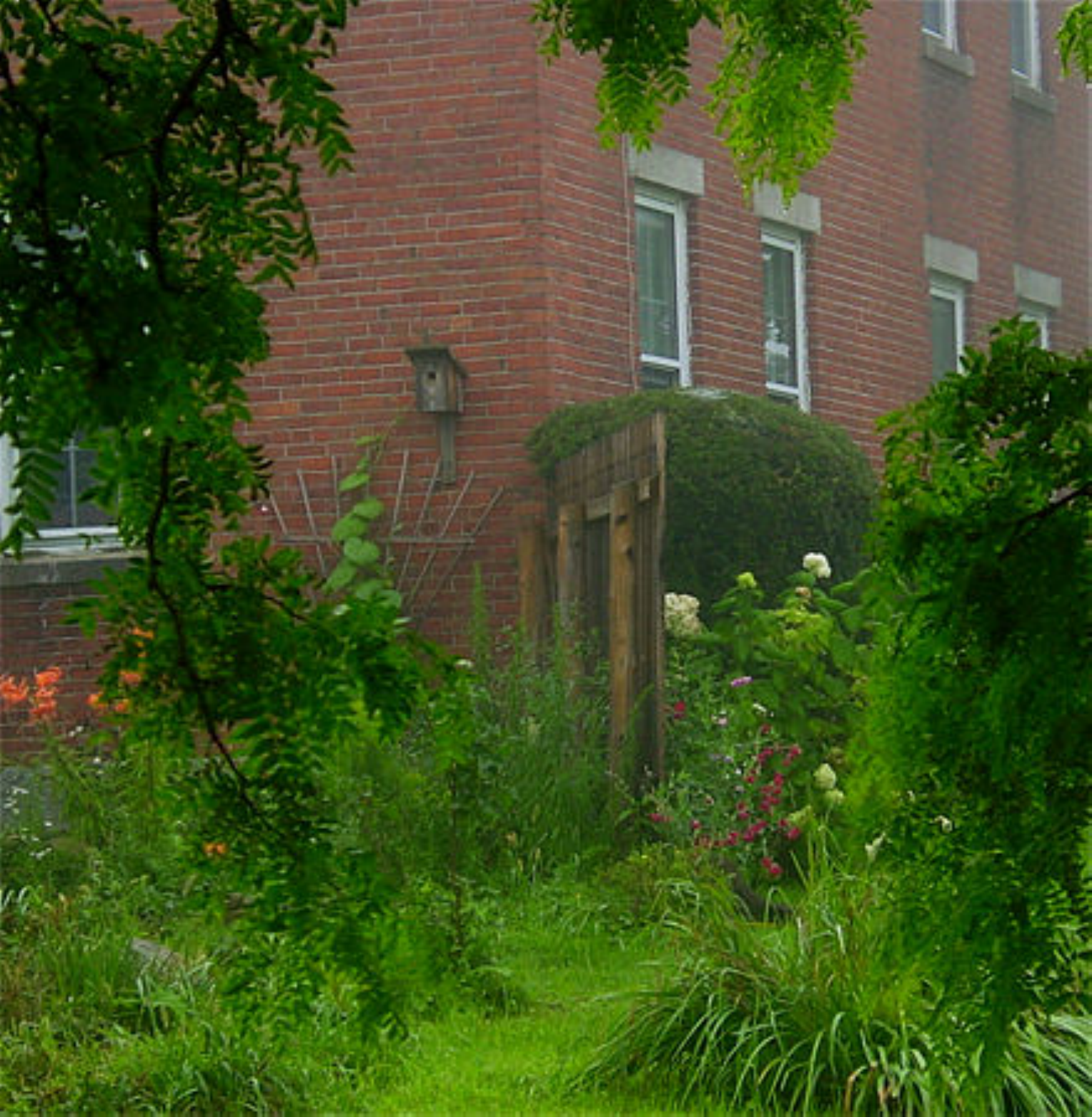} & \hspace{-0.4cm}
			\includegraphics[width = 0.105\textwidth]{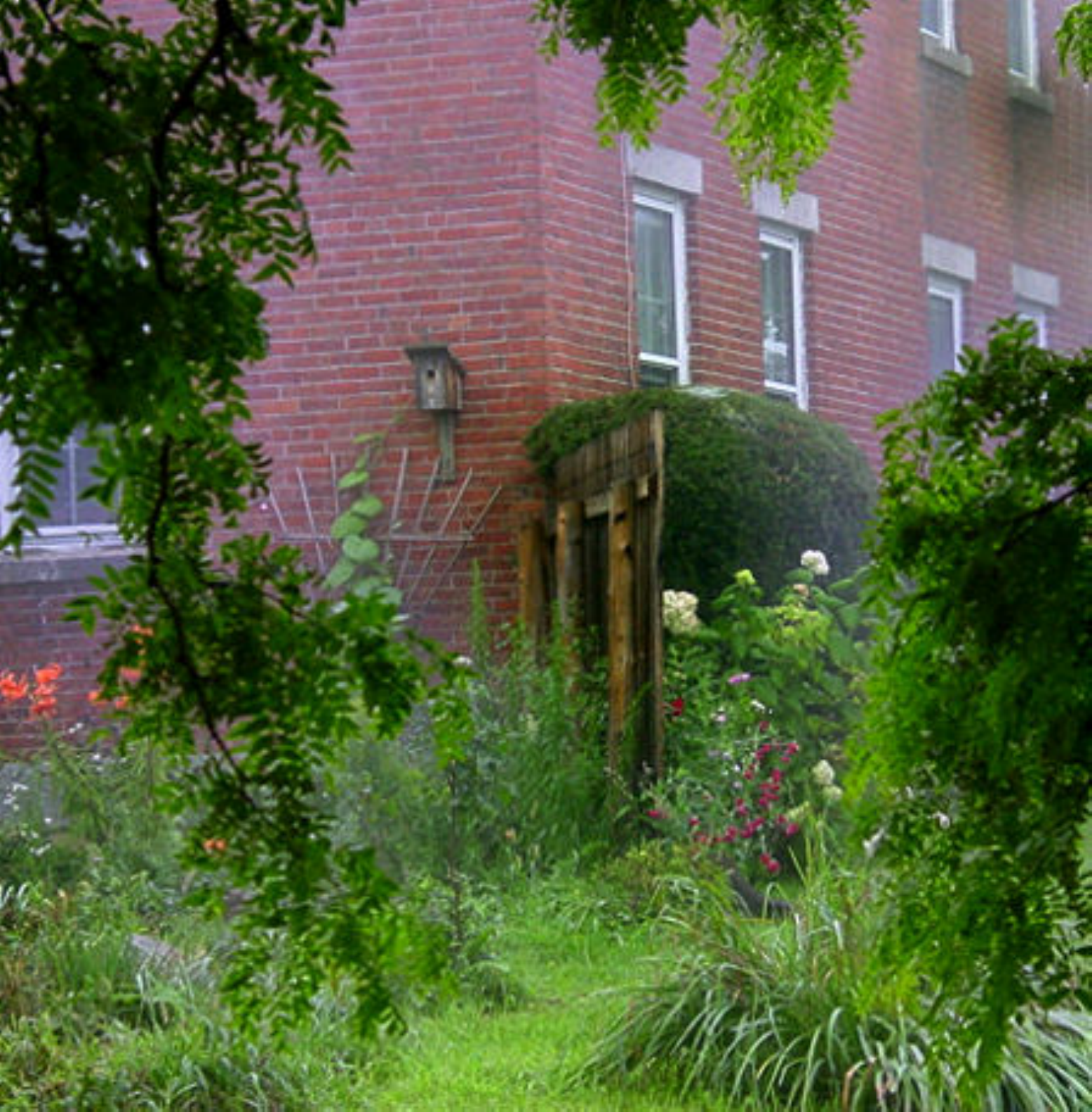}\\
			\includegraphics[width = 0.105\textwidth]{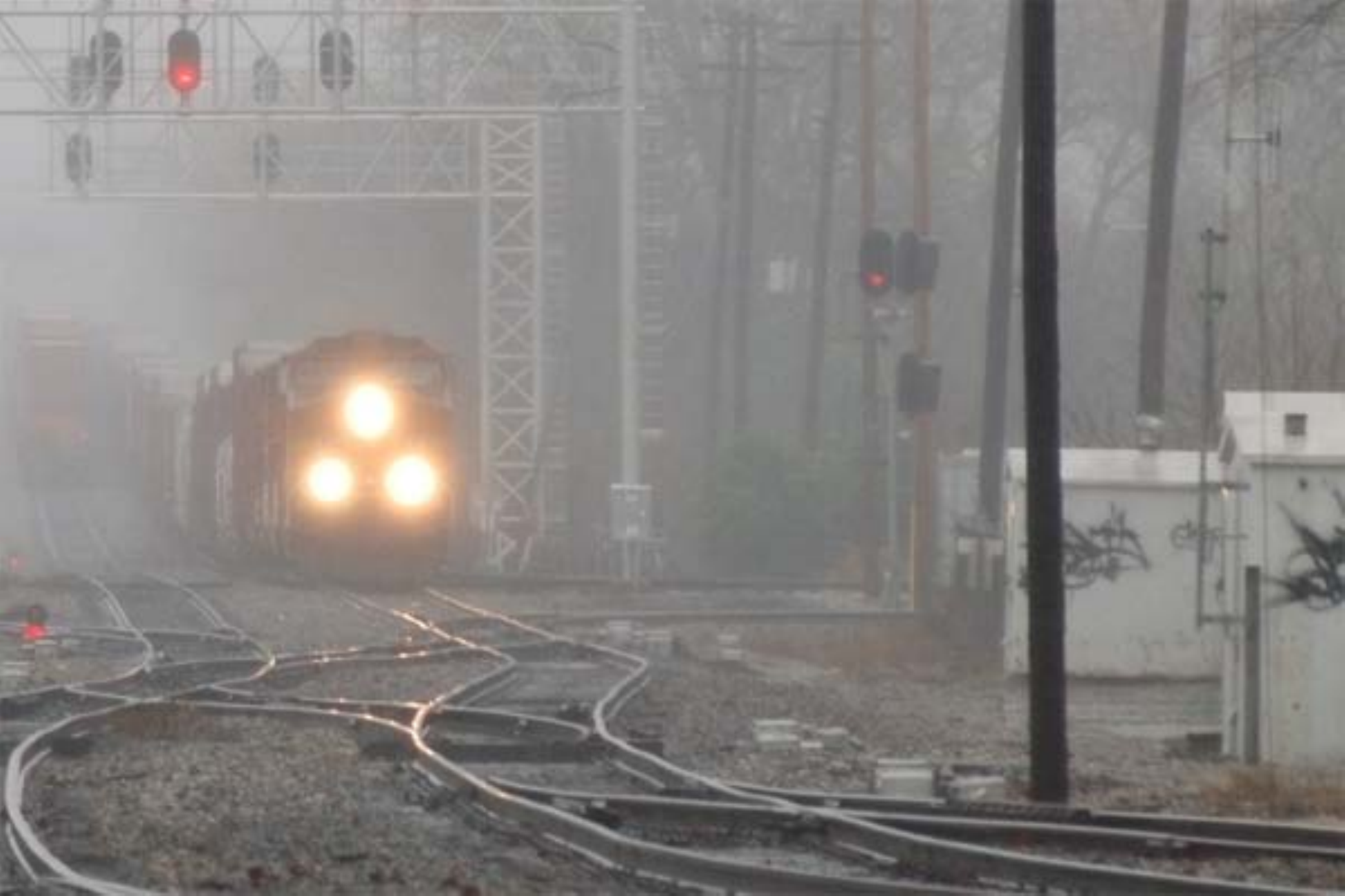} & \hspace{-0.4cm}
			\includegraphics[width = 0.105\textwidth]{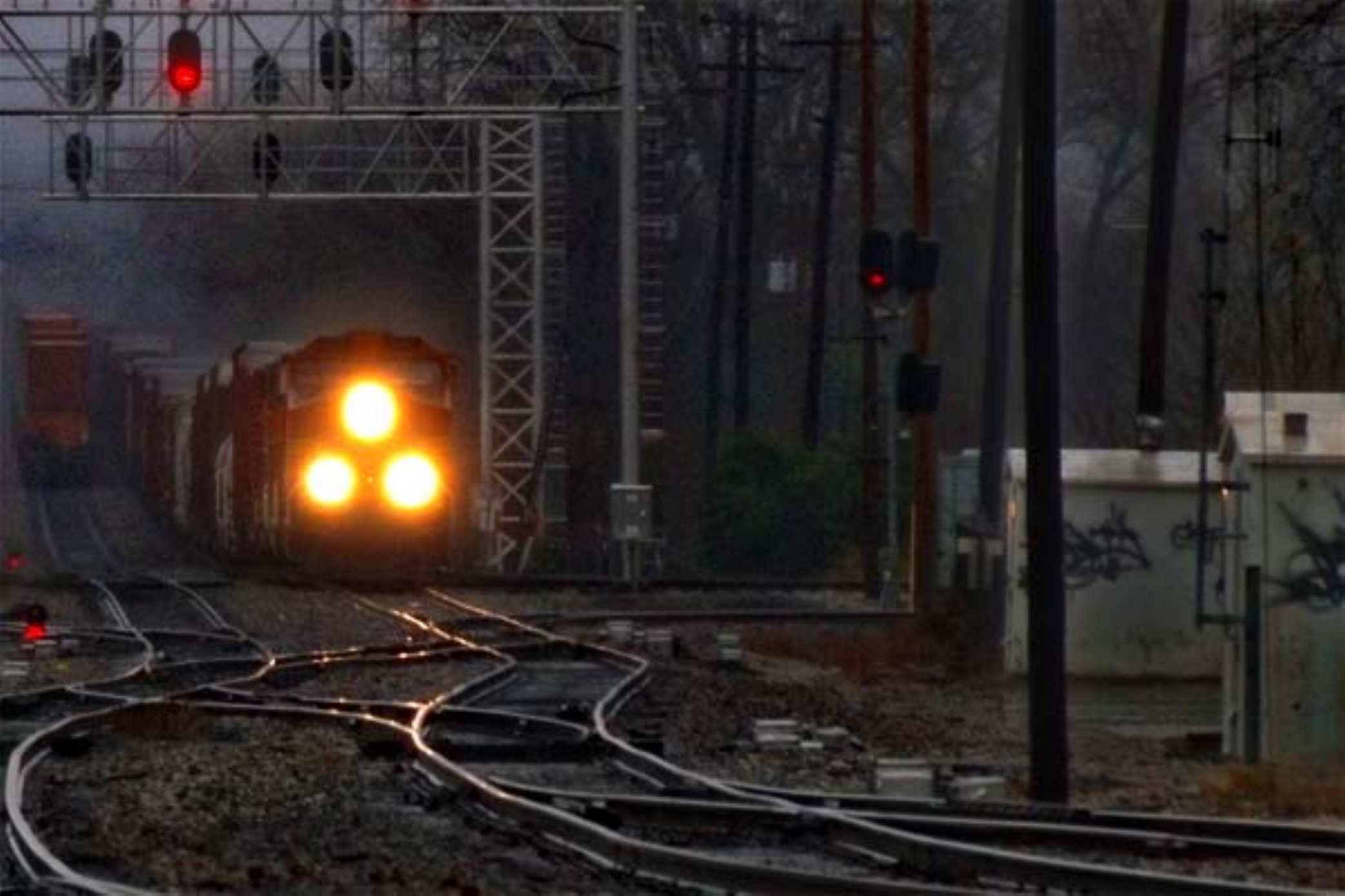} & \hspace{-0.4cm}
			\includegraphics[width = 0.105\textwidth]{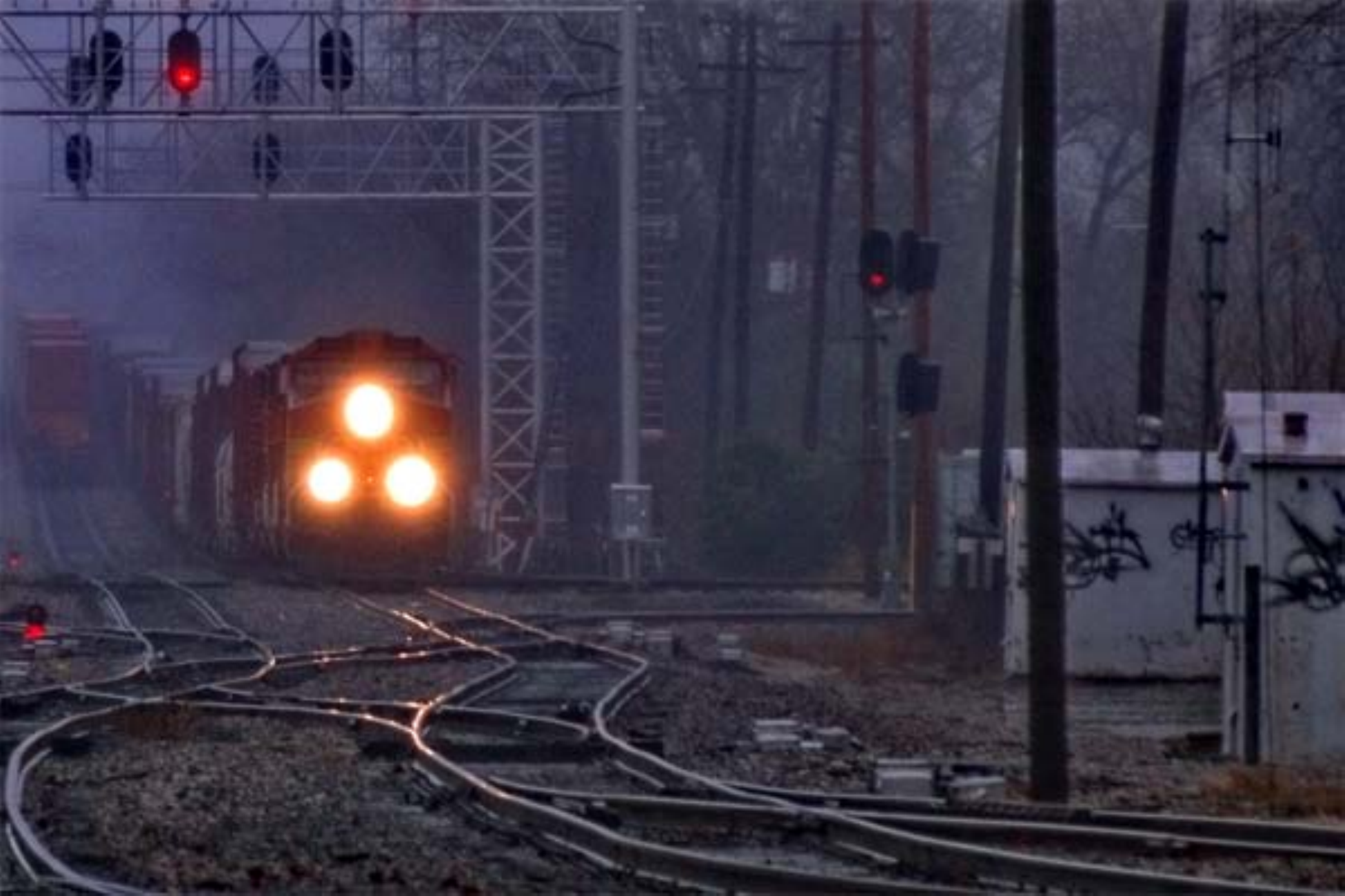} & \hspace{-0.4cm}
			\includegraphics[width = 0.105\textwidth]{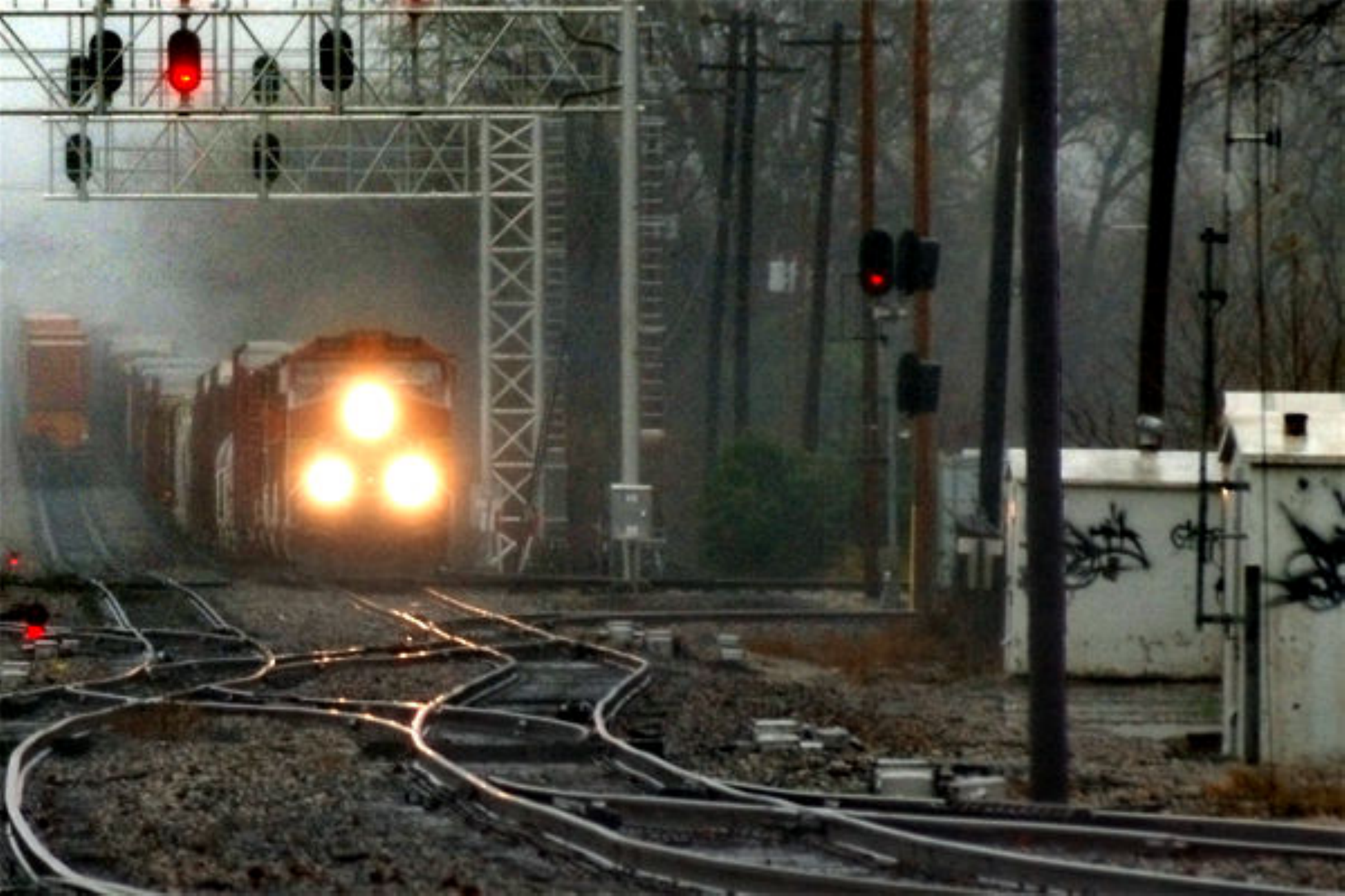} & \hspace{-0.4cm}
			\includegraphics[width = 0.105\textwidth]{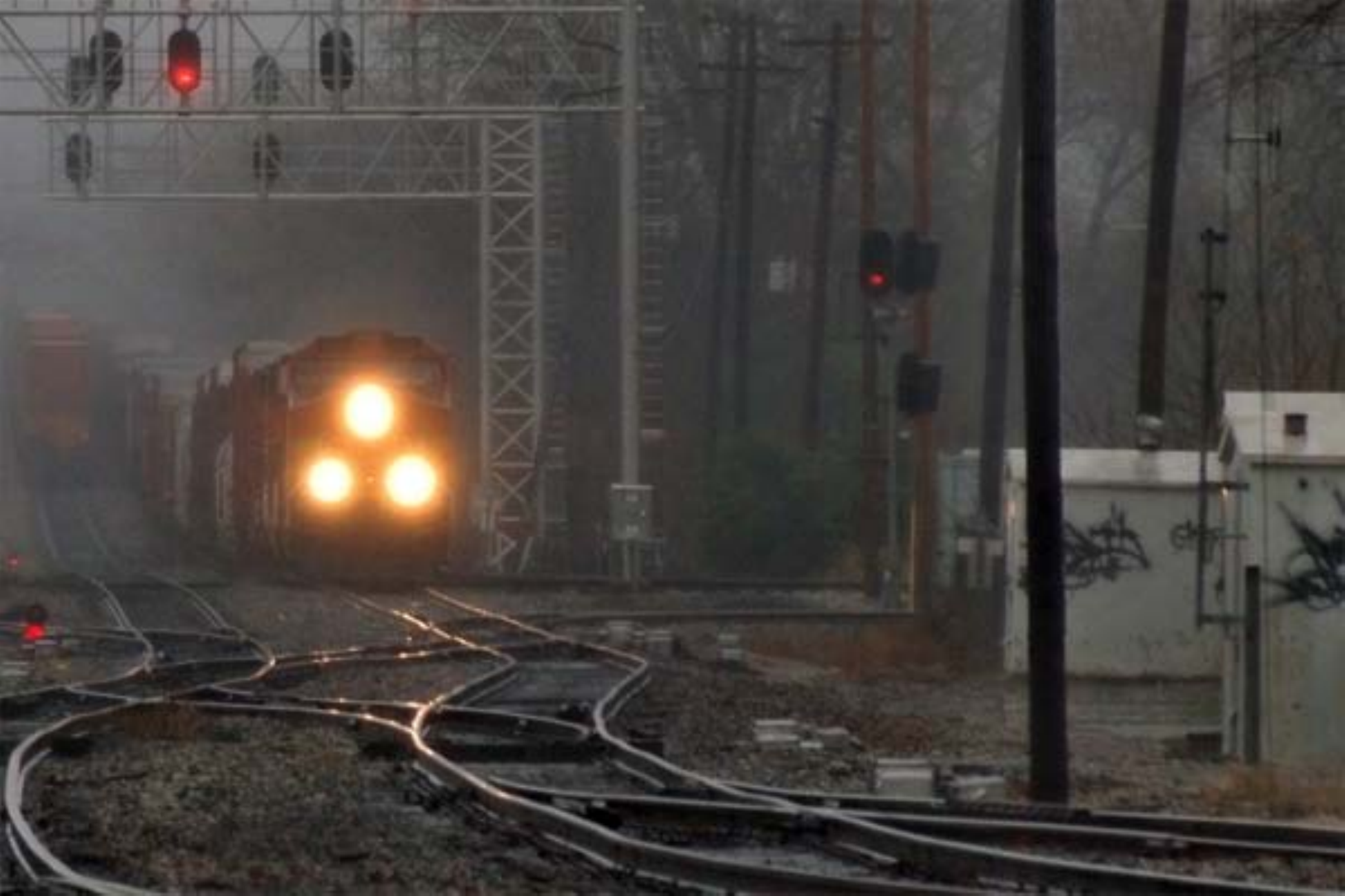} & \hspace{-0.4cm}
			\includegraphics[width = 0.105\textwidth]{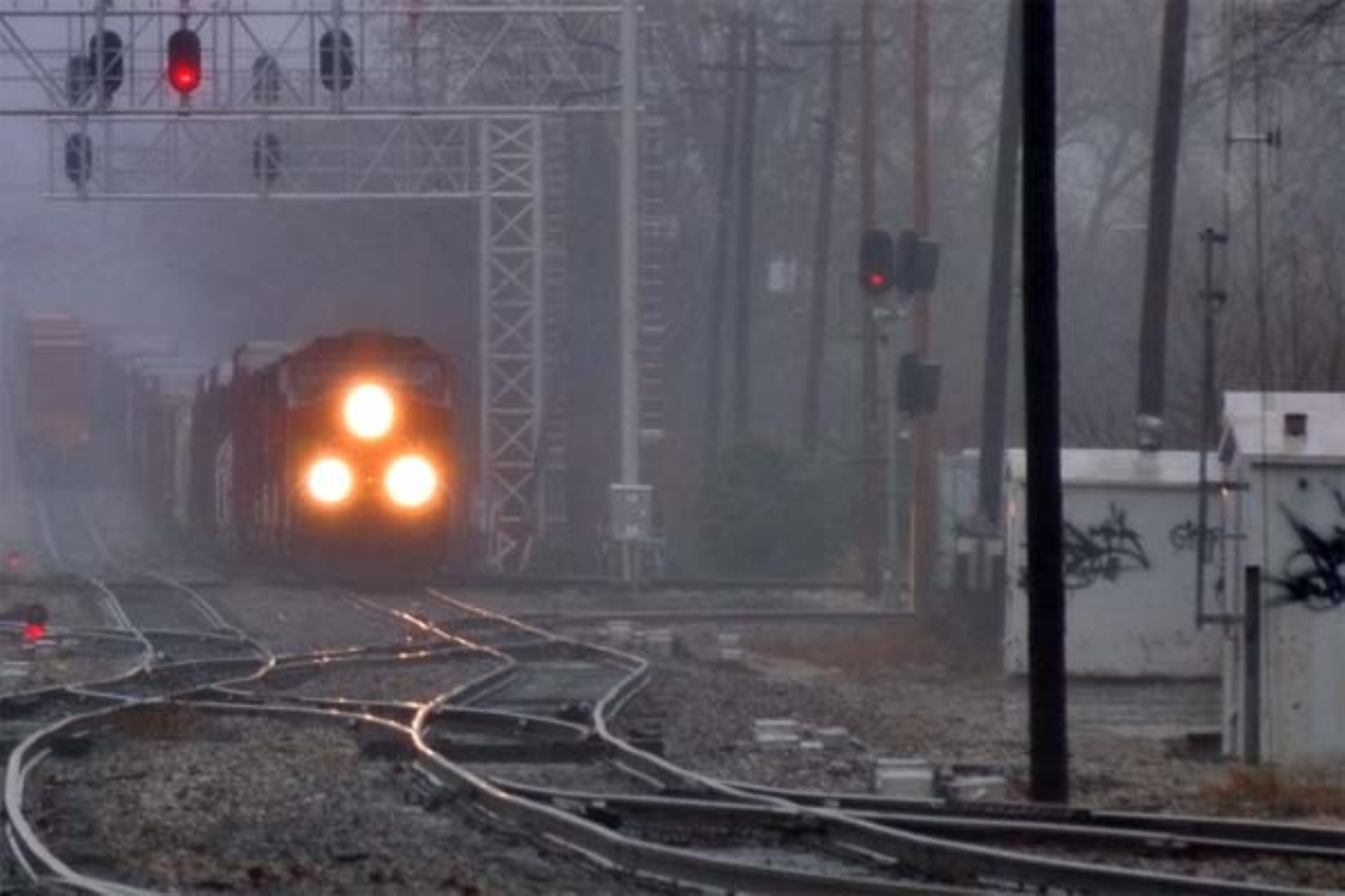} & \hspace{-0.4cm}
			\includegraphics[width = 0.105\textwidth]{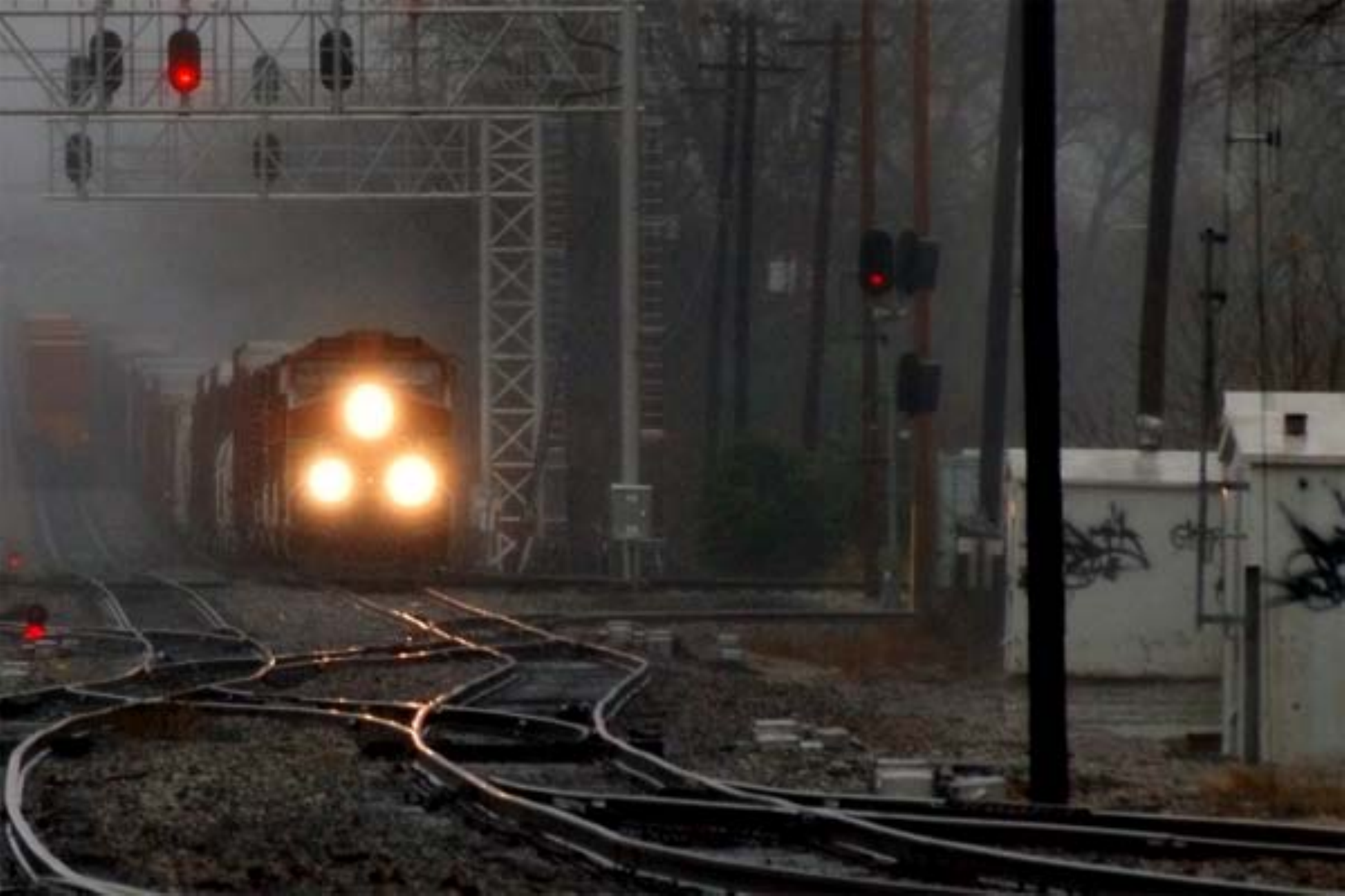} & \hspace{-0.4cm}
			\includegraphics[width = 0.105\textwidth]{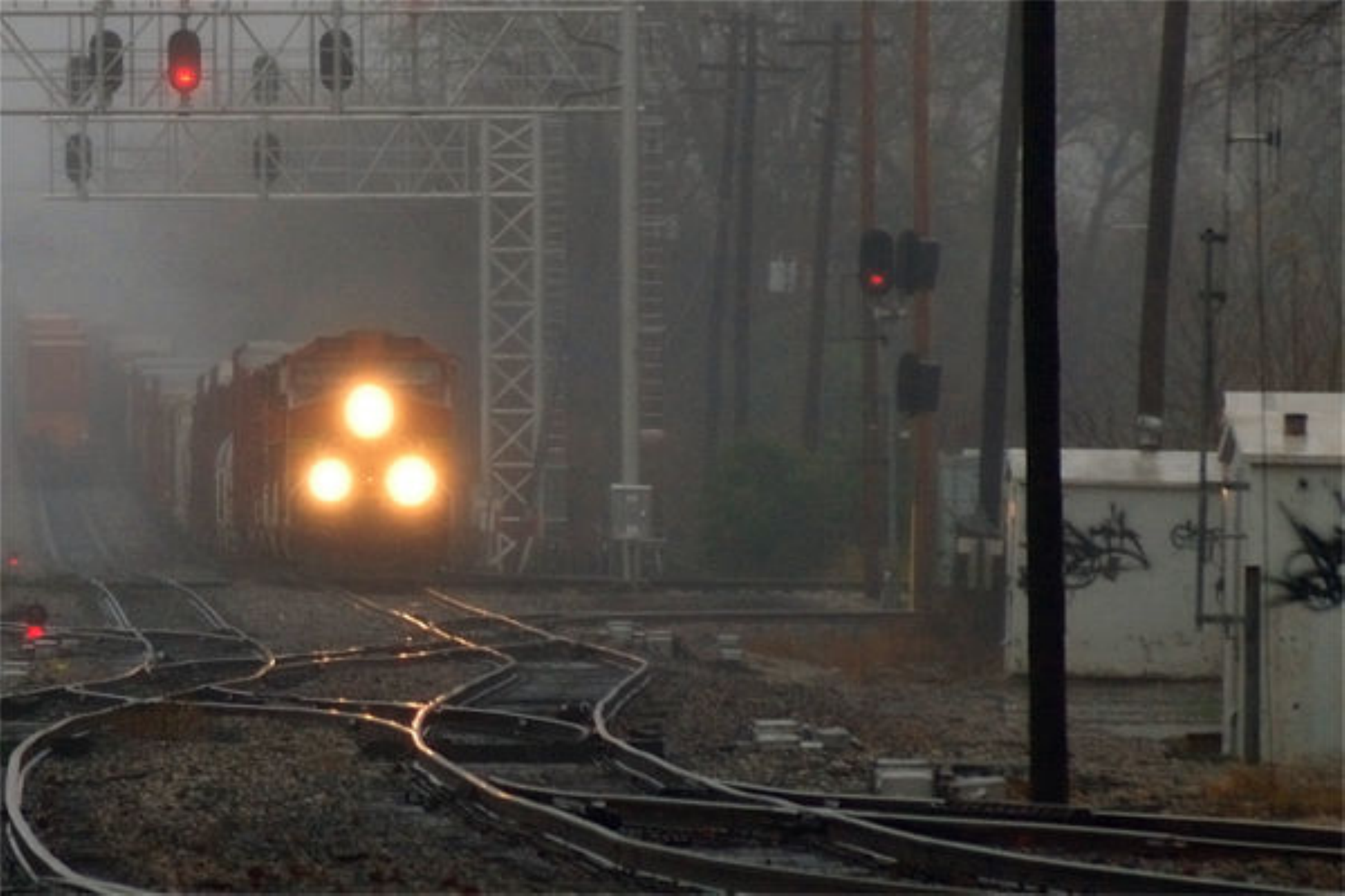} & \hspace{-0.4cm}
			\includegraphics[width = 0.105\textwidth]{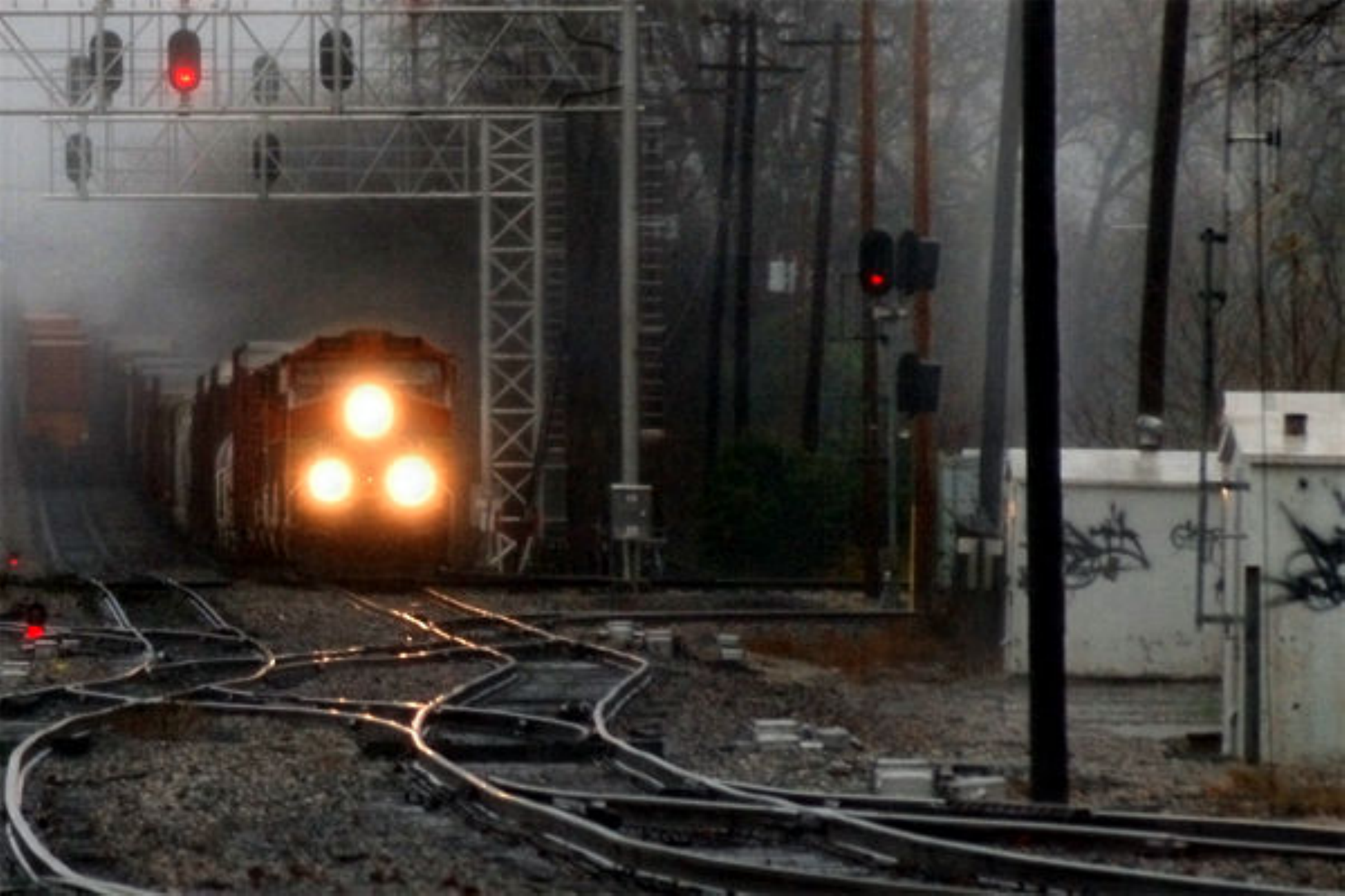}\\
			\includegraphics[width = 0.105\textwidth]{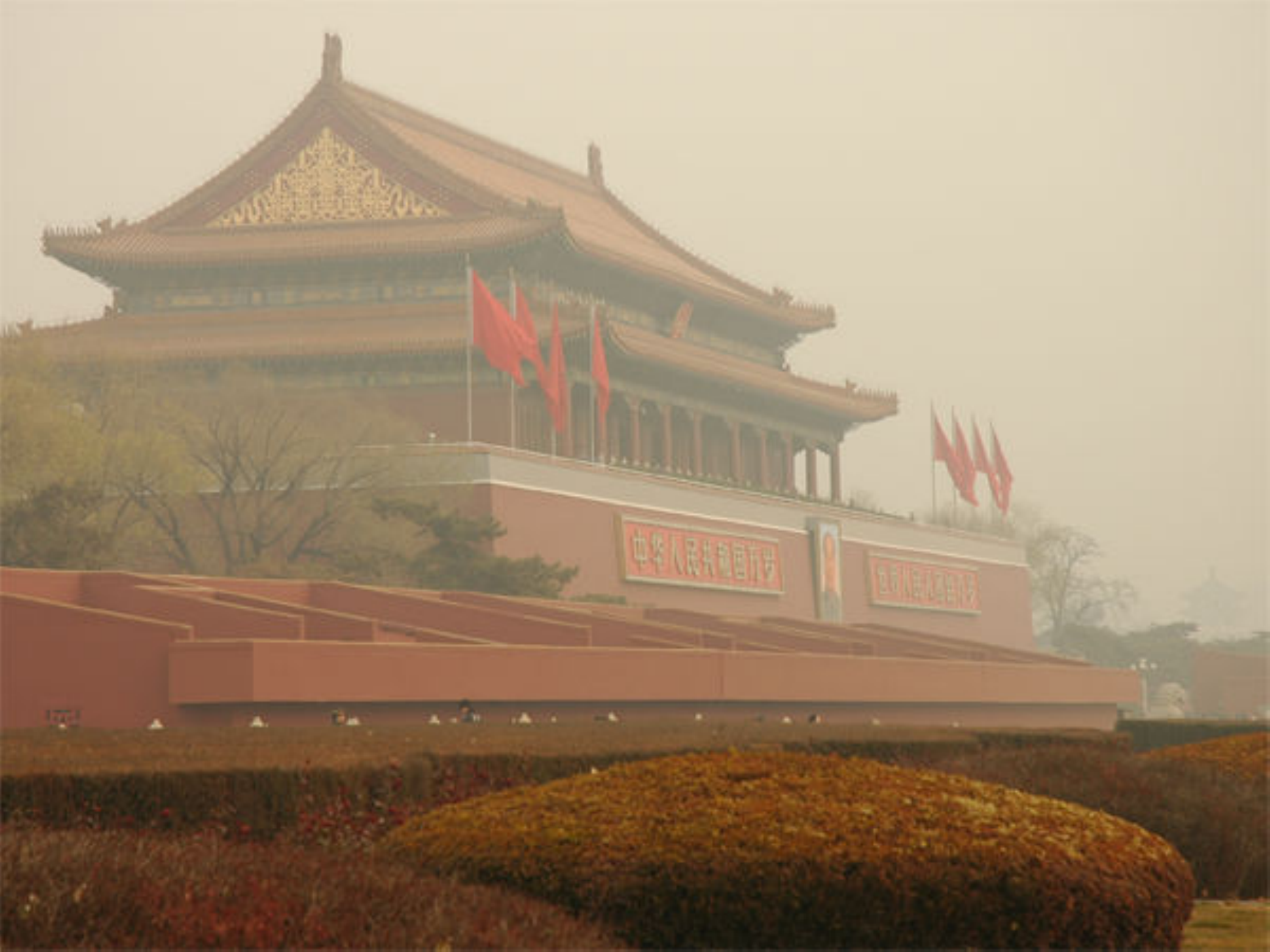} & \hspace{-0.4cm}
			\includegraphics[width = 0.105\textwidth]{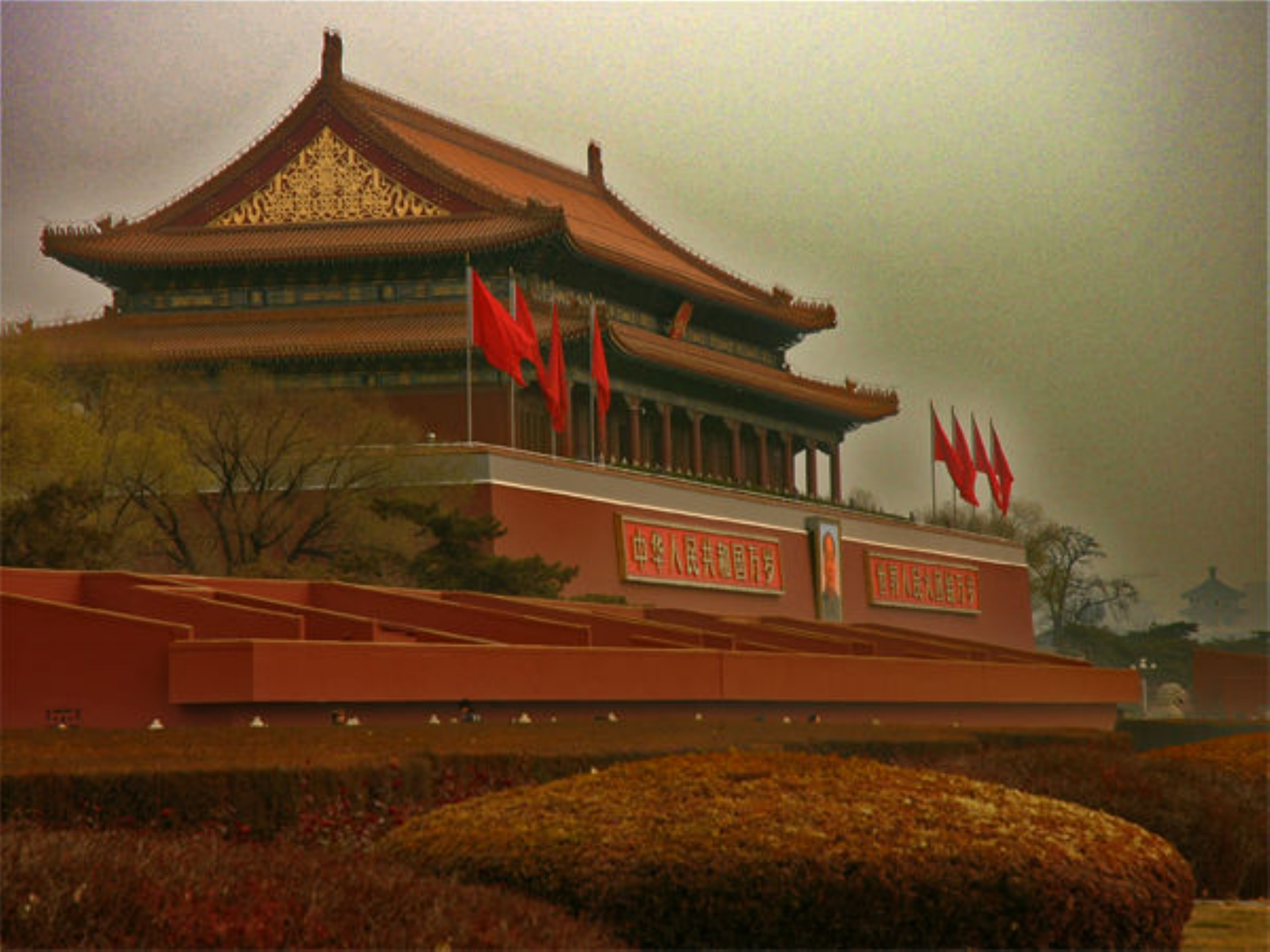} & \hspace{-0.4cm}
			\includegraphics[width = 0.105\textwidth]{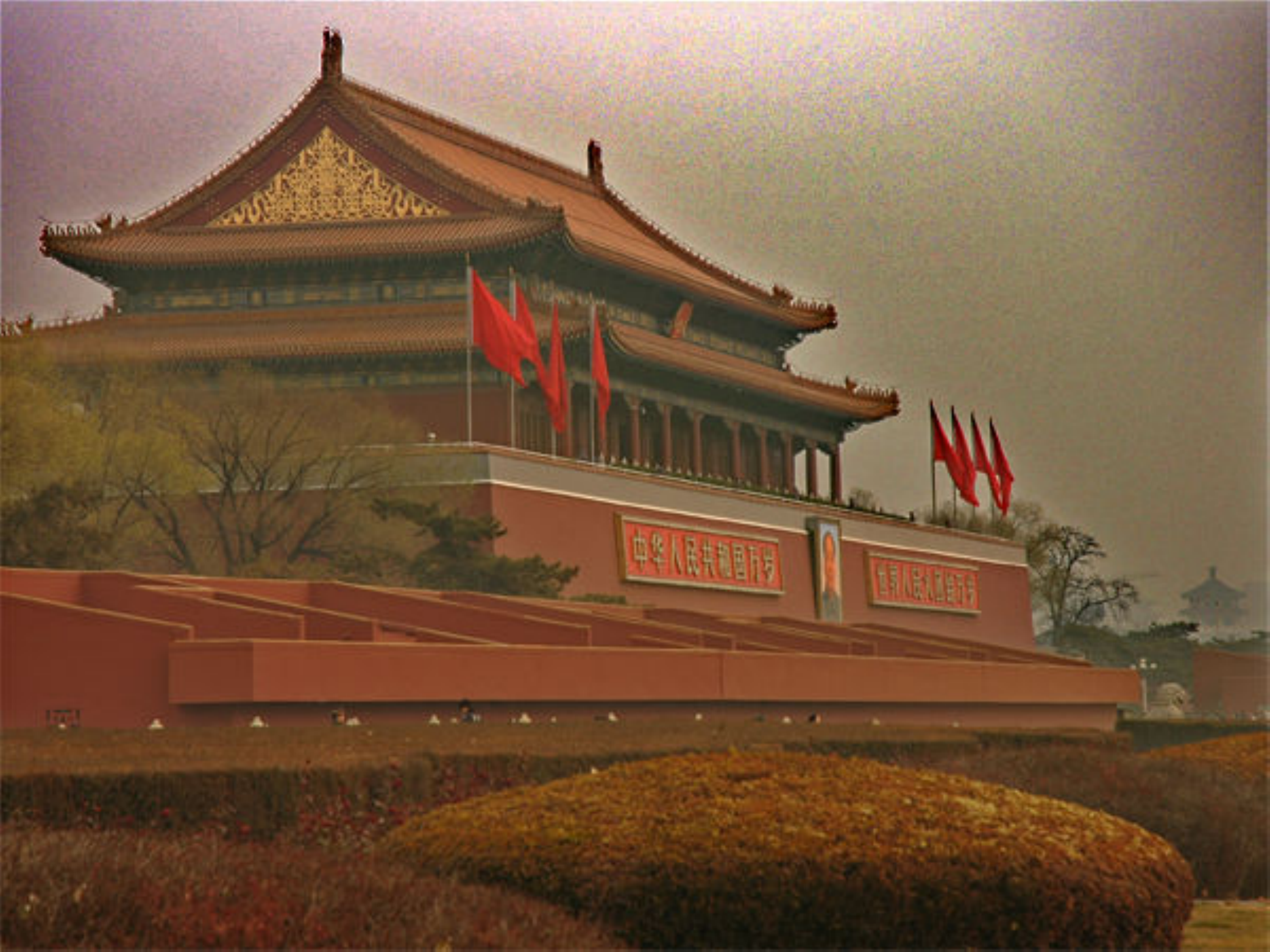} & \hspace{-0.4cm}
			\includegraphics[width = 0.105\textwidth]{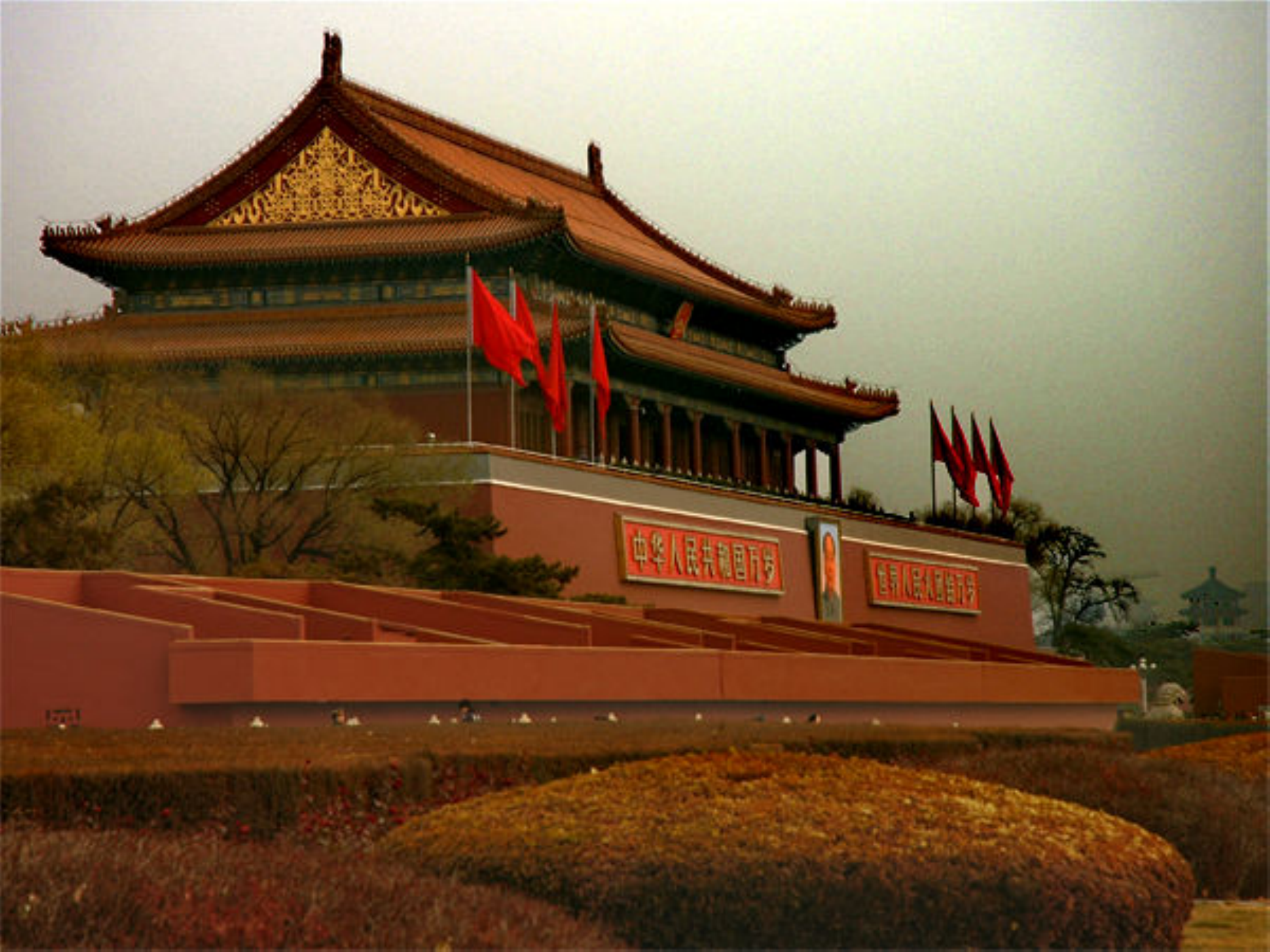} & \hspace{-0.4cm}
			\includegraphics[width = 0.105\textwidth]{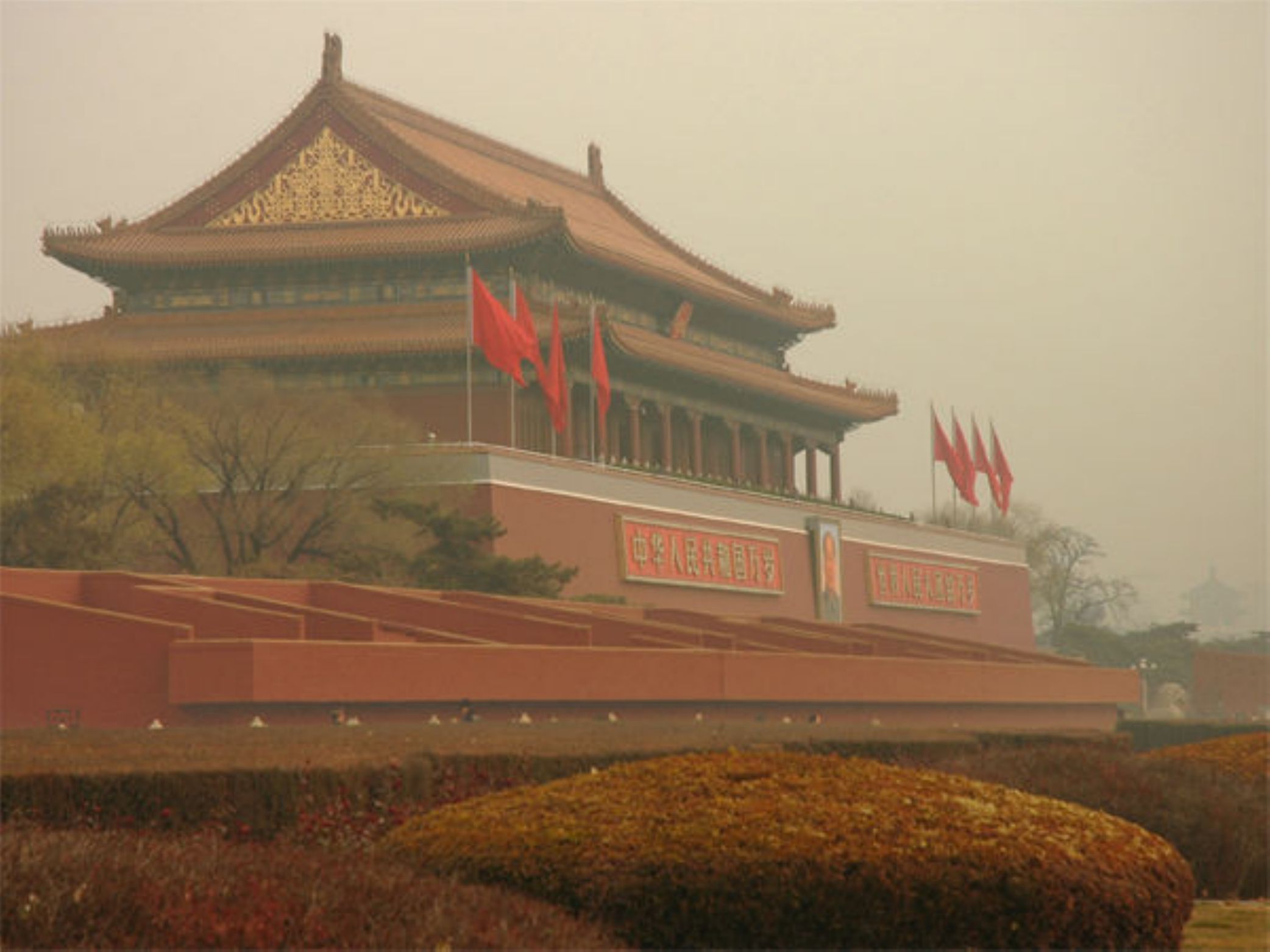} & \hspace{-0.4cm}
			\includegraphics[width = 0.105\textwidth]{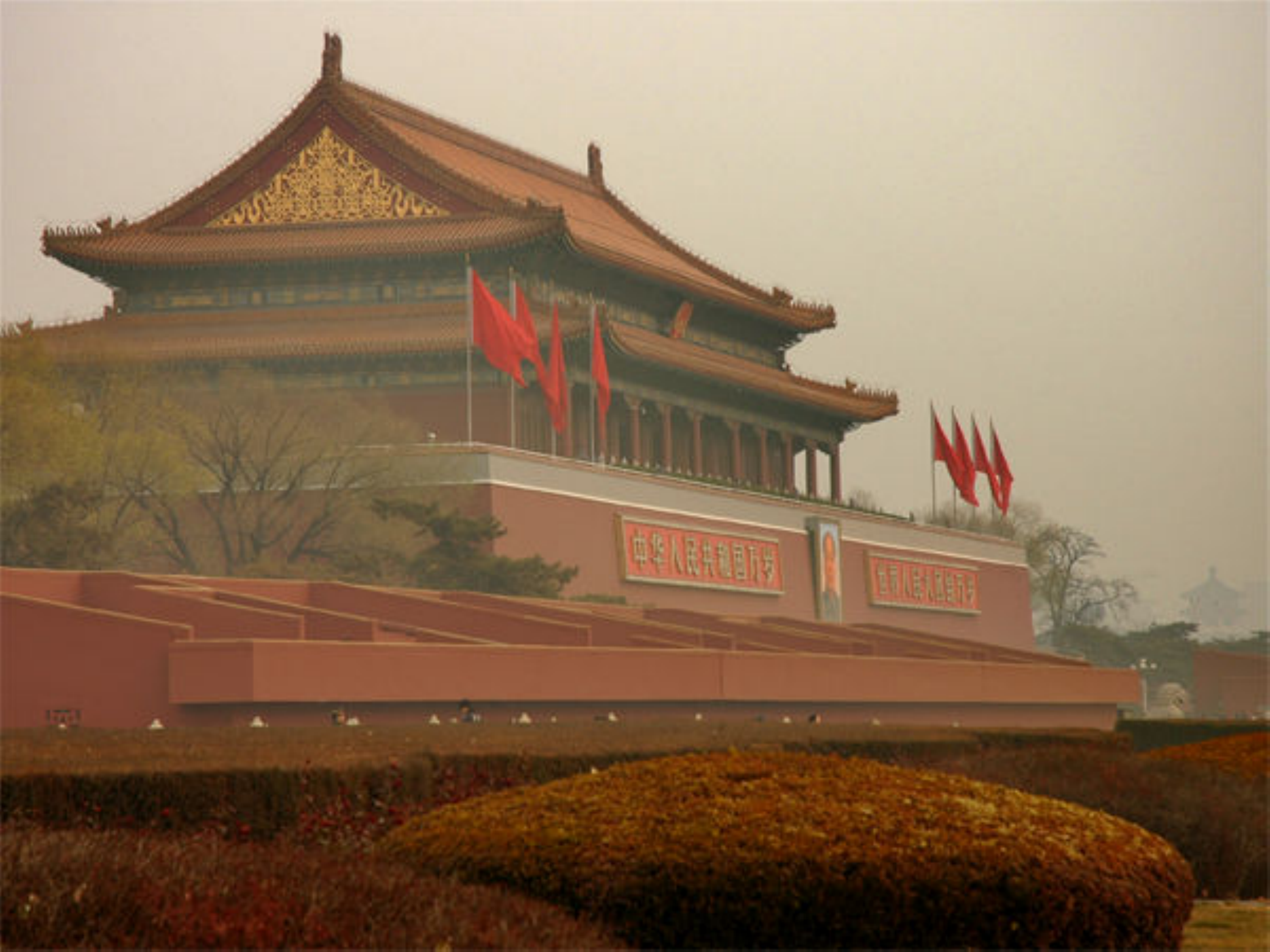} & \hspace{-0.4cm}
			\includegraphics[width = 0.105\textwidth]{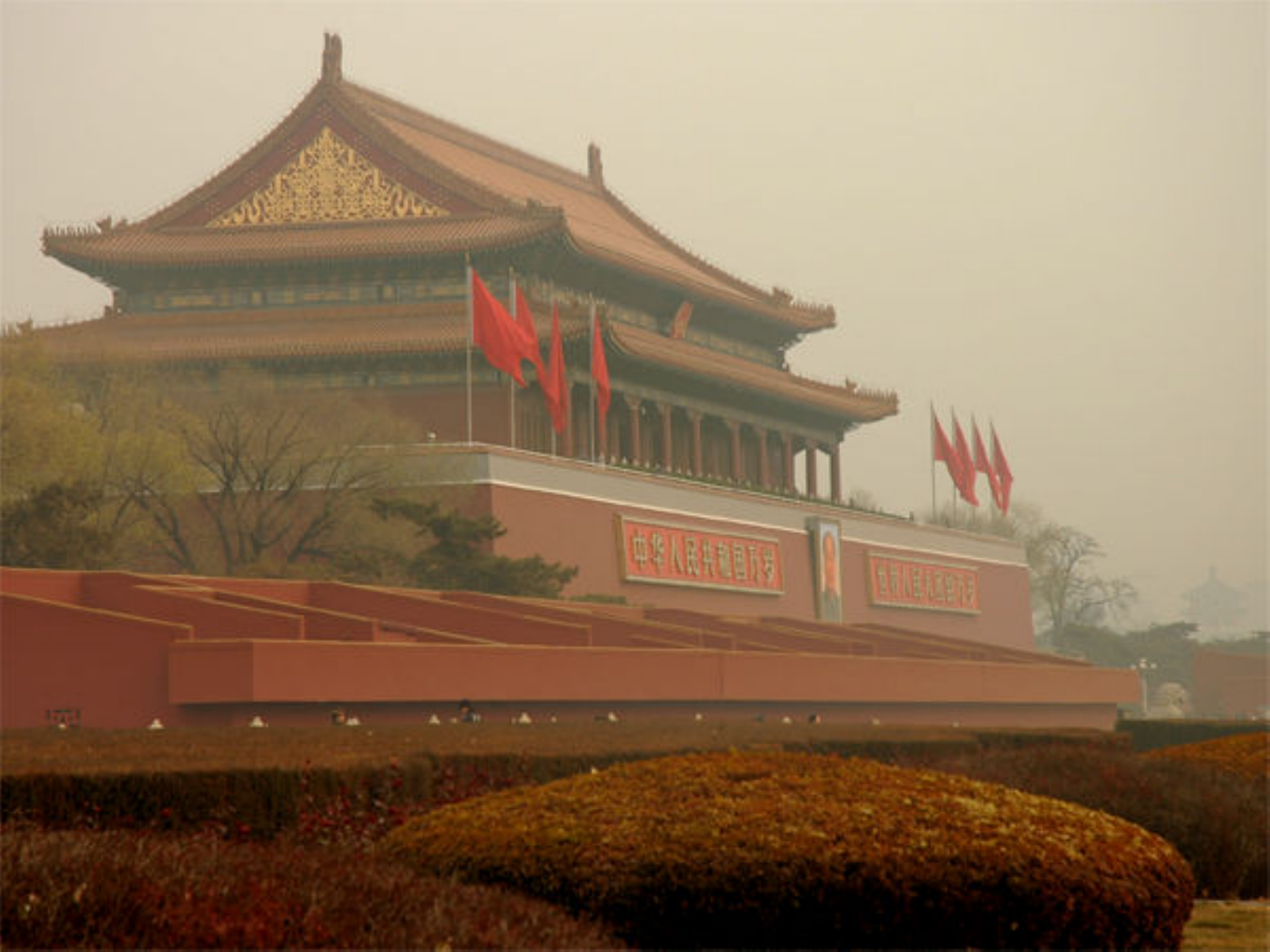} & \hspace{-0.4cm}
			\includegraphics[width = 0.105\textwidth]{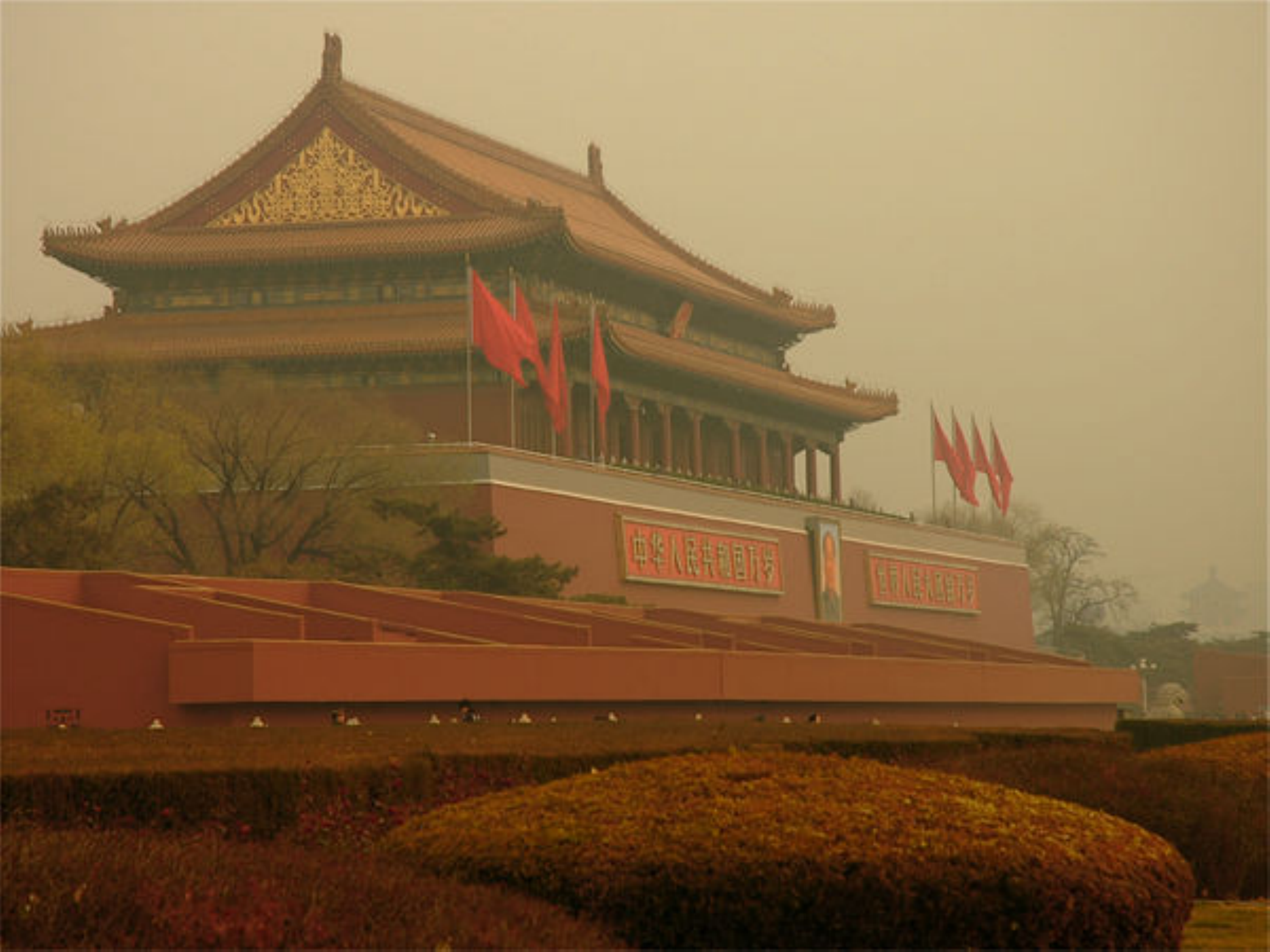} & \hspace{-0.4cm}
			\includegraphics[width = 0.105\textwidth]{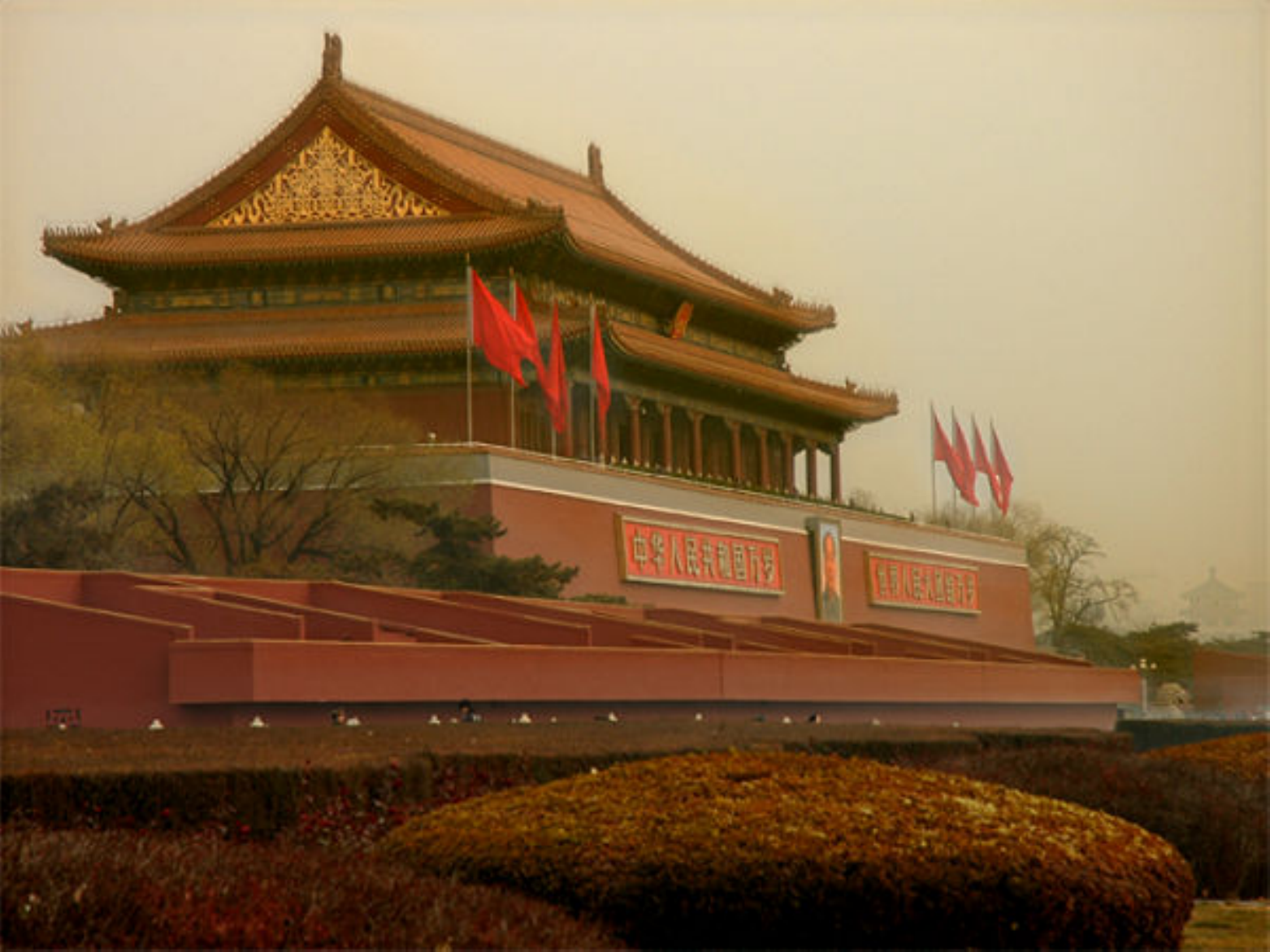}\\
			(a) Hazy inputs & \hspace{-0.4cm}
			(b) DCP~\cite{he2011single} & \hspace{-0.4cm}
			(c) BCCR~\cite{meng2013efficient} & \hspace{-0.4cm}
			(d) NLD~\cite{berman2016non} & \hspace{-0.4cm}
			(e) CAP~\cite{zhu2015fast} & \hspace{-0.4cm}
			(f) MSCNN~\cite{ren2016single} & \hspace{-0.4cm}
			(g) DehazeNet~\cite{cai2016dehazenet} & \hspace{-0.4cm}
			(h) AOD-Net~\cite{li2017aod}  & \hspace{-0.4cm}
			(i) GFN
		\end{tabular}
	\end{center}
	\vspace{-0.2cm}
	\caption{Qualitative comparison of different methods on real-world images. Best viewed on high-resolution
		display.
	}
	\vspace{-0.3cm}
	\label{fig-real-results}
\end{figure*}

Figure~\ref{fig-syn-results} shows some dehazed images by different methods. Since we directly restore the final dehazed image without transmission estimation in our algorithm,
we only compare the final dehazed results with other methods.
The priors based image dehazing methods~\cite{he2011single,meng2013efficient,berman2016non} overestimate
the haze thickness, so the dehazed results tend to be darker than the ground truth
images and contain color distortions in some regions, \eg, the desks in the second row and the wall in the last row in Figure~\ref{fig-syn-results}(b)-(d).
We note that the dehazed results by CAP~\cite{zhu2015fast}, DehazeNet~\cite{cai2016dehazenet}, MSCNN~\cite{ren2016single} and AOD-Net~\cite{li2017aod} methods are similar as shown in
Figure~\ref{fig-syn-results}(e)-(h).
Although the dehazed results by CAP, DehazeNet, MSCNN and AOD-Net
are closer to ground truth than the results by~\cite{he2011single,meng2013efficient,berman2016non},
there are still some remaining haze as shown in Figure~\ref{fig-syn-results}(e)-(h).

In contrast, the dehazed results generated by our approach in Figure~\ref{fig-syn-results}(i) are close to the ground truth
haze-free images in Figure~\ref{fig-syn-results}(j).
Overall, the dehazed results by the proposed algorithm have higher
visual quality and fewer color distortions.
The qualitative results are also reflected
by the quantitative PSNR and SSIM metrics in Table~\ref{tab-psnr-ssim}.
%

In addition, to further test the dehazing effect on different haze concentration,
we use three scattering coefficient $\beta=0.8, 1.0$ and $1.2$ to synthesize three haze concentration on the 49 testing images, respectively.
%
%
As shown in Table~\ref{tab-psnr-ssim}, our method without adversarial loss performs favorably against the state-of-the-art image dehazing methods~\cite{he2011single,meng2013efficient,berman2016non,zhu2015fast,ren2016single,cai2016dehazenet,li2017aod} on all of these haze concentrations.
However, if we use adversarial loss, the network can still recover better dehazed results
than without adding adversarial loss in terms of SSIM in some cases.
Although the SSIM values by~\cite{li2017aod} are close to ours in some cases, the PSNR generated by our method are higher than~\cite{li2017aod} by up to 2dB, especially for heavy haze concentration images.

\begin{table}[htbp]
	\caption{Average PSNR/SSIM of dehazed results on the SOTS dataset from \textbf{RESIDE}.}
	\vspace{-0.5cm}
	\begin{center}\scriptsize{
			\begin{tabular}{ccccc}
				\toprule
				NLD \cite{berman2016non}  & MSCNN \cite{ren2016single}  & DehazeNet \cite{cai2016dehazenet} & AOD-Net \cite{li2017aod} & GFN \\
				\midrule
				17.27/0.75  & 17.57/0.81 &  21.14/0.85  &  19.06/0.85  & \textbf{22.30}/\textbf{0.88}\\
				\bottomrule
		\end{tabular}}
		\label{tab-reside}
	\end{center}
	\vspace{-0.8cm}
\end{table}
\vspace{-0.2cm}
{\flushleft \textbf{RESIDE dataset.}} Recently, a dehazing benchmark is proposed in \cite{li2017reside}, which is an extended version of our data in Table \ref{tab-psnr-ssim}. We further evaluate our method on the RESIDE dataset in Table \ref{tab-reside}. As shown, our method performs favorably against other competitors \cite{berman2016non,cai2016dehazenet,li2017aod,ren2016single} in this dataset.

\subsection{Evaluation on Real Images}
To further evaluate the proposed method, we use the real image dataset in Fattal~\cite{fattal2014dehazing} and compare
with different state-of-the-art methods.
Figure~\ref{fig-real-results} shows the qualitative comparison of results
with the seven state-of-the-art dehazing algorithms~\cite{he2011singlecvpr,meng2013efficient,berman2016non,ren2016single,cai2016dehazenet,li2017aod} on challenging real-world images.
Figure~\ref{fig-real-results}(a) shows the hazy images to be dehazed.
Figure~\ref{fig-real-results}(b)-(h) shows the results of DCP~\cite{he2011singlecvpr}, BCCR~\cite{meng2013efficient}, NLD~\cite{berman2016non}, CAP~\cite{zhu2015fast}, MSCNN~\cite{ren2016single}, DehazeNet~\cite{cai2016dehazenet} and AOD-Net~\cite{li2017aod}, respectively.
The results generated by the proposed algorithm are given in Figure~\ref{fig-real-results}(i).
As shown in Figure~\ref{fig-real-results}(b)-(d), most of the haze is removed by DCP, BCCR and NLD methods, and the details of the scenes and objects are well restored. However, the results significantly suffer from over-enhancement (for instance, the sky region of the first and second images are much darker than it should be as shown in Figure~\ref{fig-real-results}(b)-(d), and there are some color distortions in the second and last images in Figure~\ref{fig-real-results}(c) and (d)). This is because
these algorithms are based on hand-crafted priors which
have an inherent problem of overestimating the transmission
as discussed in~\cite{he2011singlecvpr,zhu2015fast}.
The results of CAP~\cite{zhu2015fast} do not have the
over-estimation problem and maintain the original colors of
the objects as shown in Figure~\ref{fig-real-results}(e). But have
some remaining haze in the dehazed results. For example, the third image.
The dehazed results by MSCNN~\cite{ren2016single} and DehazeNet~\cite{cai2016dehazenet}
have a similar problem as~\cite{zhu2015fast} tends to have some remaining haze.
Especially the last image in Figure~\ref{fig-real-results}(f) and the first image in Figure~\ref{fig-real-results}(g).
The method of AOD-Net~\cite{li2017aod} generates relatively clear results, but the images in first three rows are still dark than ours, while the results in last two rows still have some remaining haze as shown in Figure~\ref{fig-real-results}(h).
In contrast, the dehazed results by our method are clear and the details of
the scenes are enhanced moderately.
%
%
\begin{figure}[t]\scriptsize
	\begin{center}
		\begin{tabular}{@{}cccc@{}}
			\includegraphics[width = 0.11\textwidth]{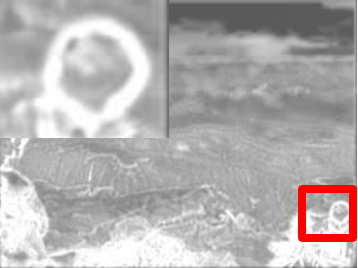} & \hspace{-0.4cm}
			\includegraphics[width = 0.11\textwidth]{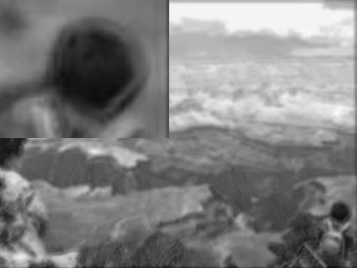} & \hspace{-0.4cm}
			\includegraphics[width = 0.11\textwidth]{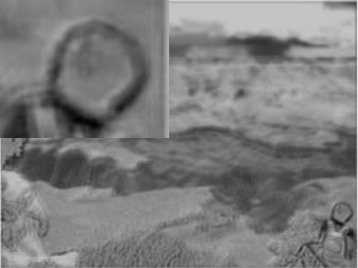} & \hspace{-0.4cm}
			\includegraphics[width = 0.11\textwidth]{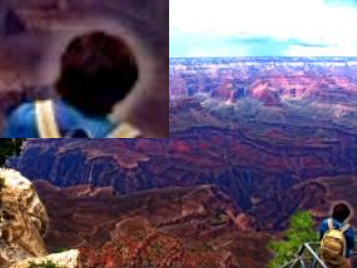} \\
			\includegraphics[width = 0.11\textwidth]{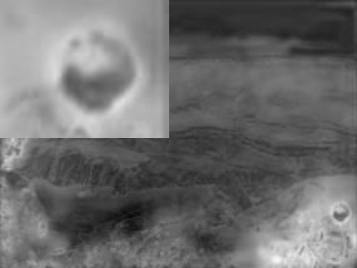} & \hspace{-0.4cm}
			\includegraphics[width = 0.11\textwidth]{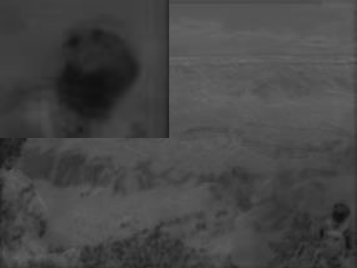} & \hspace{-0.4cm}
			\includegraphics[width = 0.11\textwidth]{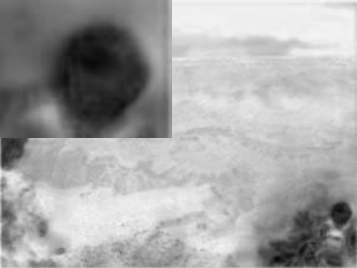} & \hspace{-0.4cm}
			\includegraphics[width = 0.11\textwidth]{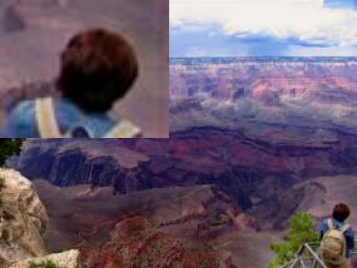} \\
			(a) Maps of $\mathbf{I}_{wb}$ & \hspace{-0.4cm}
			(b) Maps of $\mathbf{I}_{ce}$ & \hspace{-0.4cm}
			(c) Maps of $\mathbf{I}_{gc}$ & \hspace{-0.4cm}
			(d) GFN
		\end{tabular}
	\end{center}
	\vspace{-0.1cm}
	\caption{Effectiveness of the proposed multi-scale approach. The first and second rows are the results by single and multi-scale networks, respectively.
		The zoomed-in regions are shown in the left-top corner in each image.
	}
	\vspace{-0.1cm}
	\label{fig-multiscale}
\end{figure}

\vspace{-3mm}
\section{Analysis and Discussions}
\vspace{-1mm}
\subsection{Effectiveness of Multi-Scale Network}
In this section we analyze how the multi-scale network helps refine dehazed results.
The recovered images from coarser-scale network provide
additional information in the finer-scale net, which can greatly improve the final dehazed results.
We show the dehazed results generated by only using the finest-scale and the proposed multi-scale networks in Figure~\ref{fig-multiscale}.
Figure~\ref{fig-multiscale} shows that dehazed results and corresponding confidence maps. The first row is the dehazed results by only using the finest scale network and the second row is the results by the proposed multi-scale approach.
As shown in the first row in Figure~\ref{fig-multiscale}(a) and (c), there are obvious halo around the \textit{head of the person} in the confidence maps, so the final dehazed result in the first row Figure~\ref{fig-multiscale}(d) has the halo artifacts.
In contrast, the dehazed results generated by the proposed multi-scale approach has a more clean edge as shown in the second row in Figure~\ref{fig-multiscale}(d).

\begin{figure}[t]\scriptsize
	\begin{center}
		\begin{tabular}{@{}cccc@{}}
			\includegraphics[width = 0.115\textwidth]{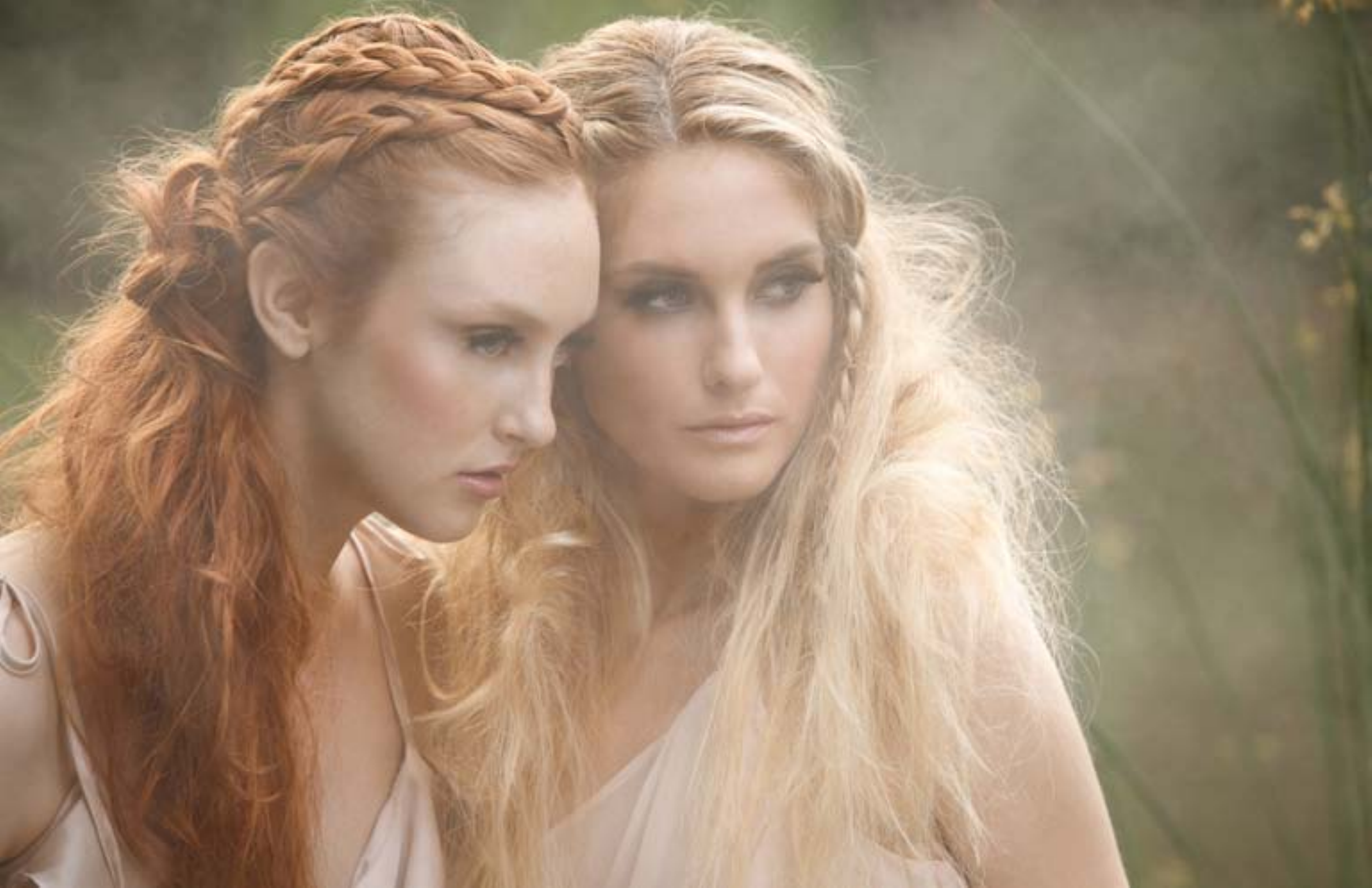} & \hspace{-0.4cm}
			\includegraphics[width = 0.115\textwidth]{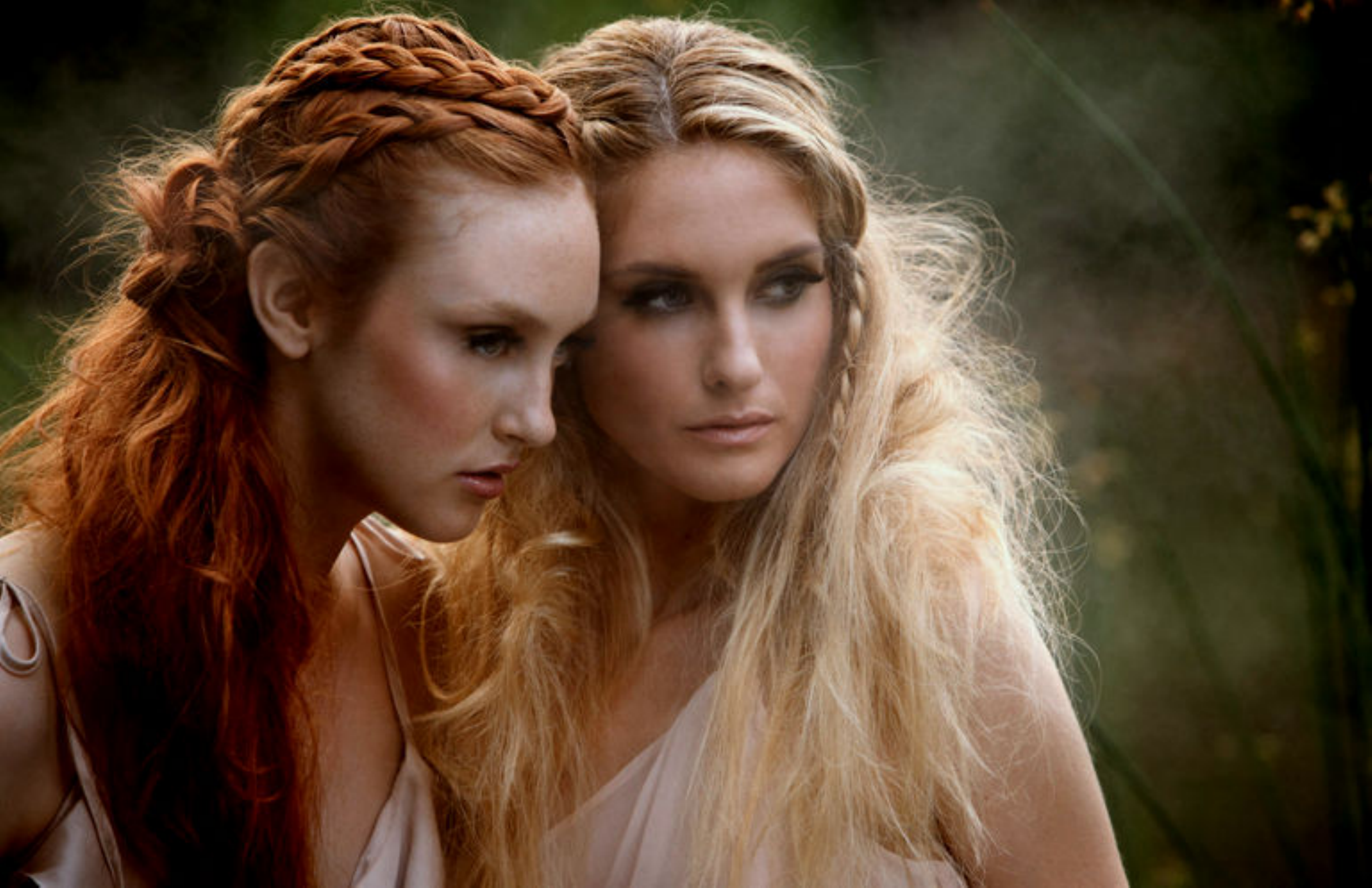} & \hspace{-0.4cm}
			\includegraphics[width = 0.115\textwidth]{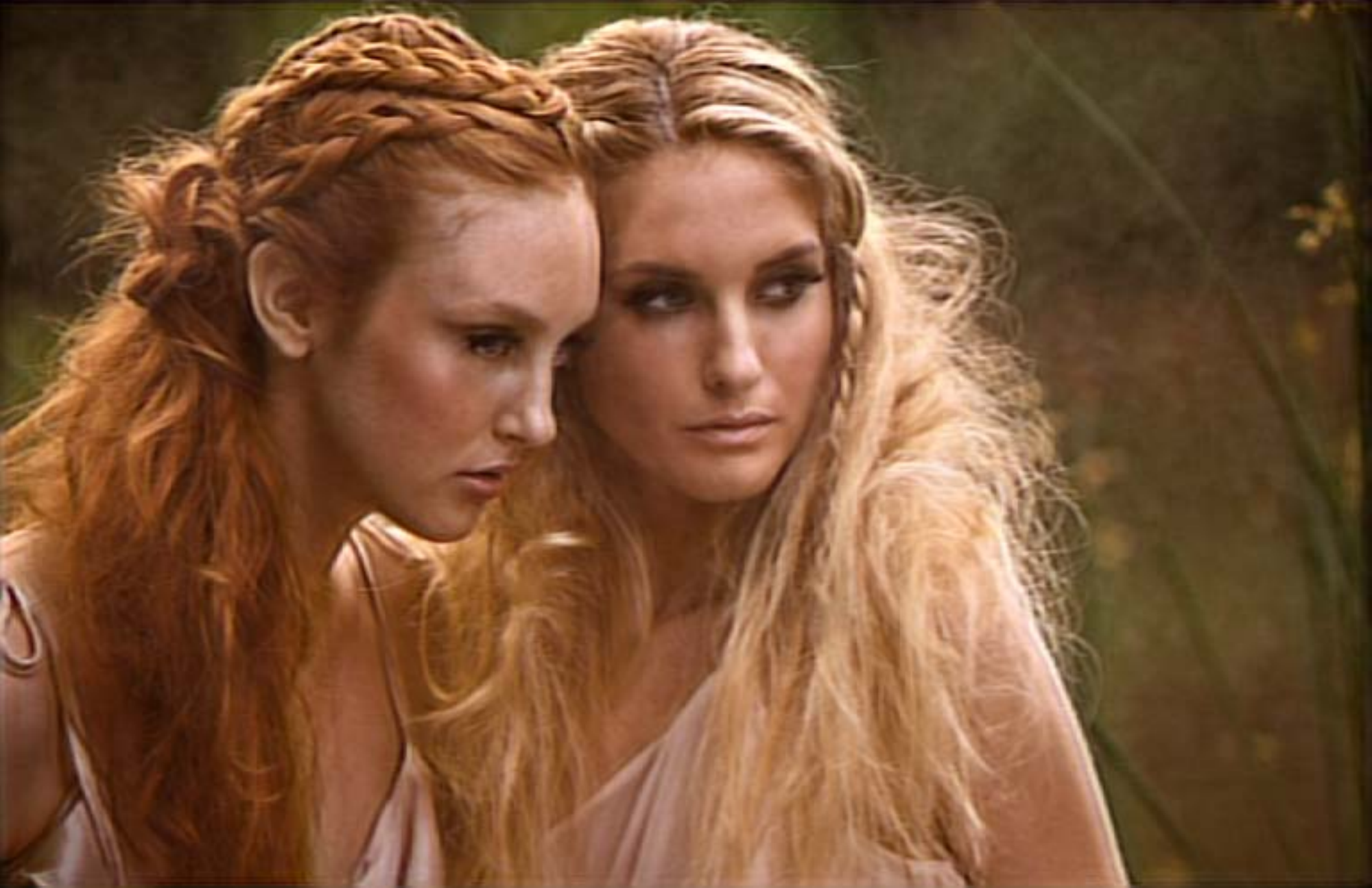} & \hspace{-0.4cm}
			\includegraphics[width = 0.115\textwidth]{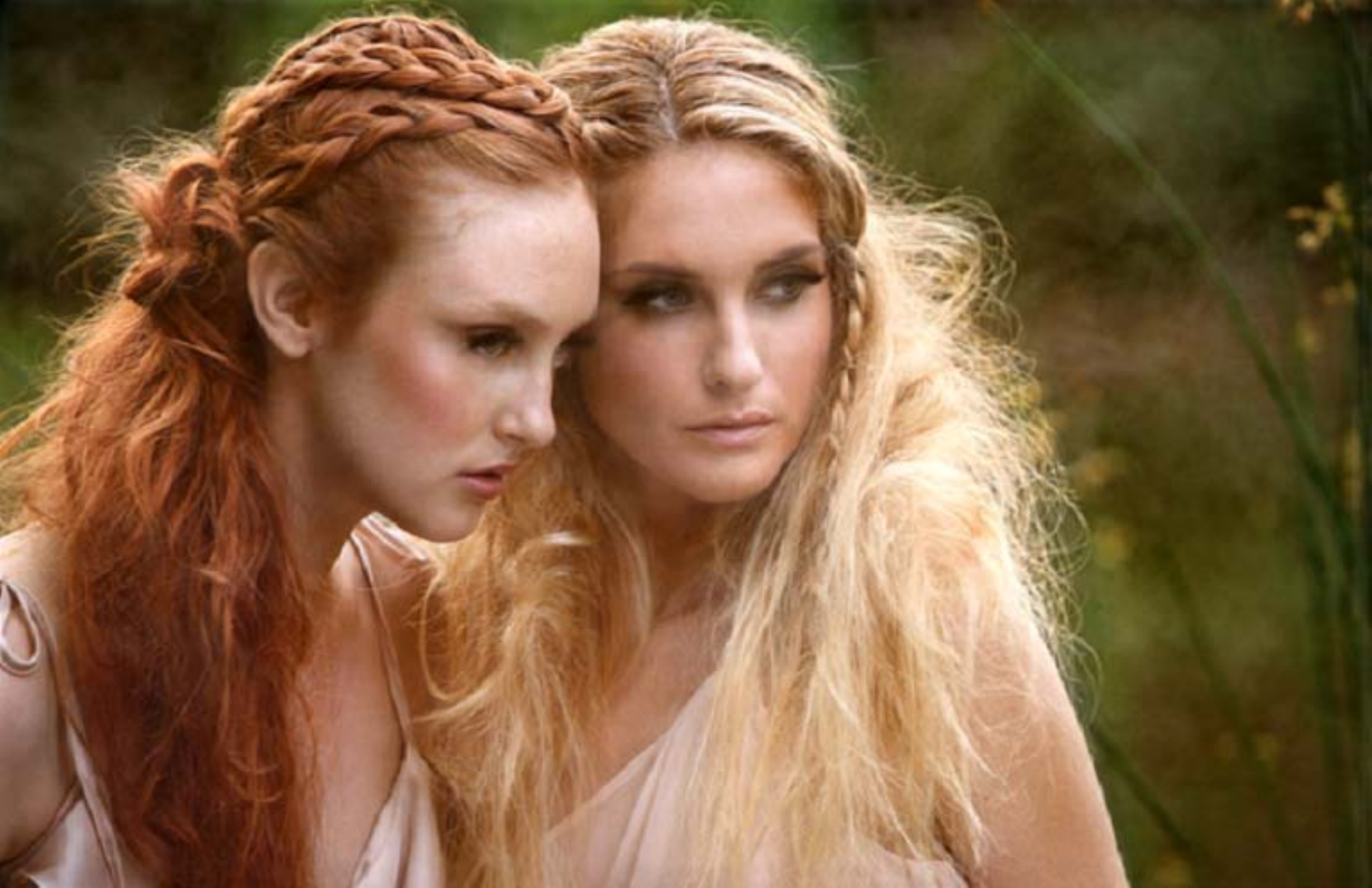} \\
			\includegraphics[width = 0.115\textwidth]{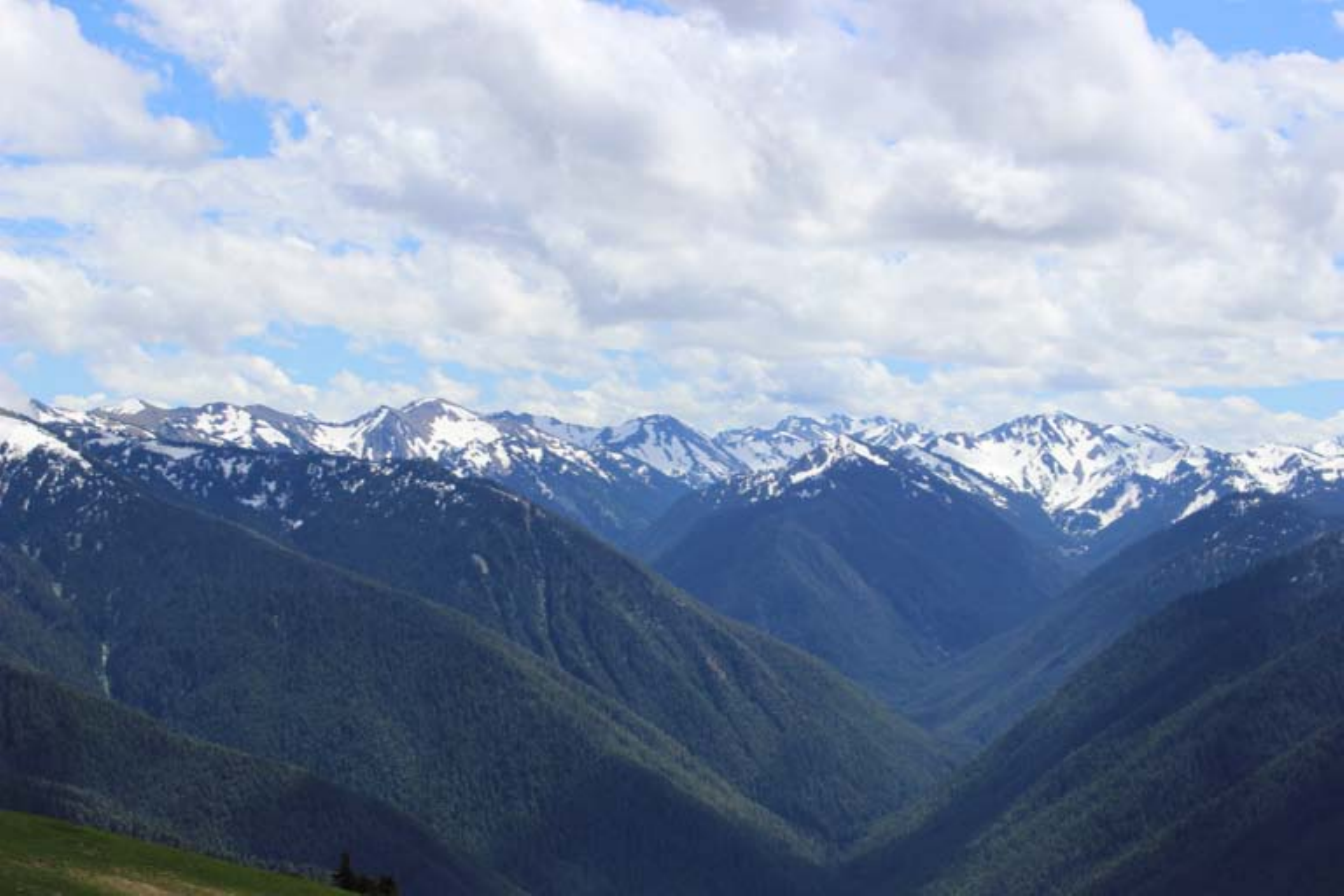} & \hspace{-0.4cm}
			\includegraphics[width = 0.115\textwidth]{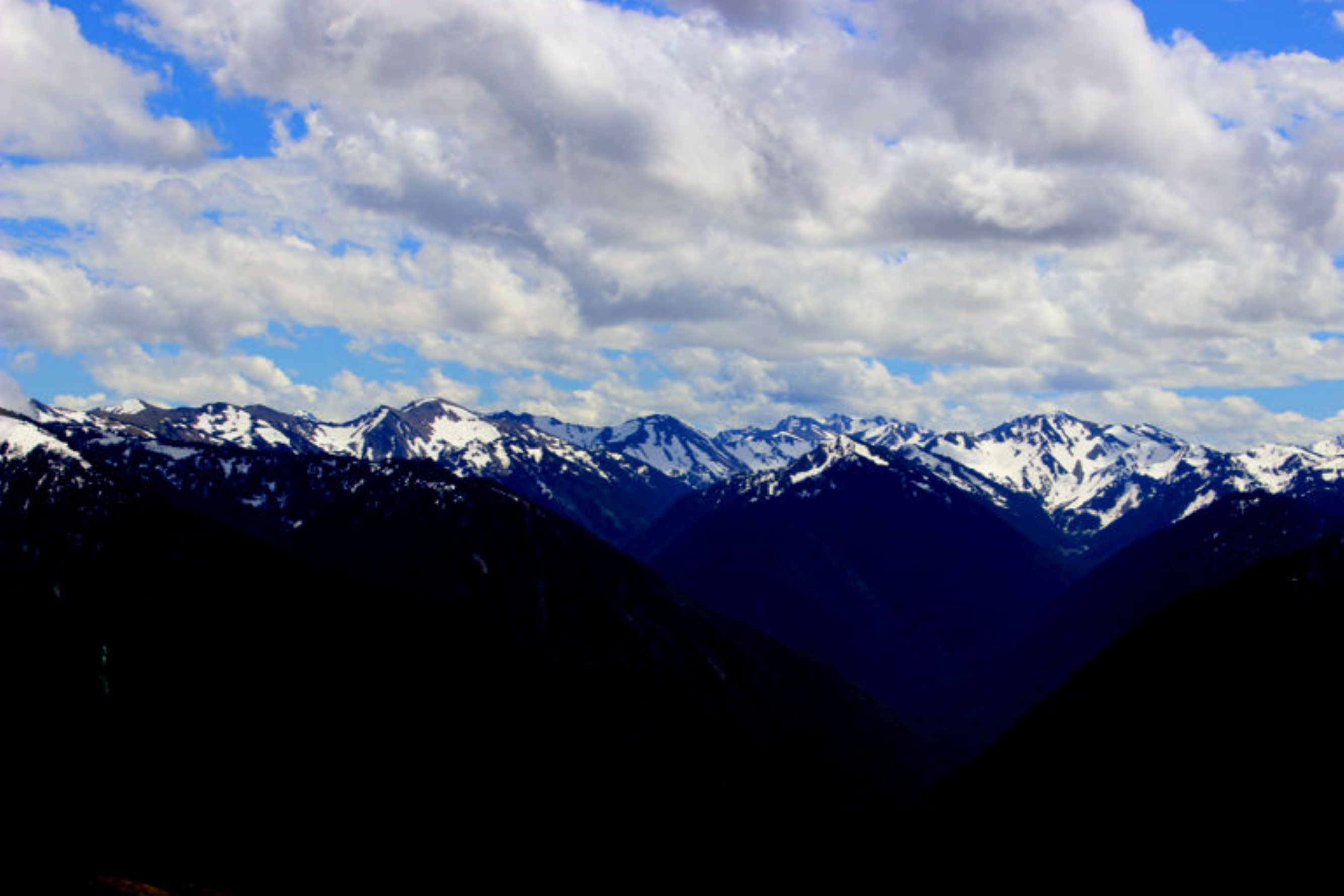} & \hspace{-0.4cm}
			\includegraphics[width = 0.115\textwidth]{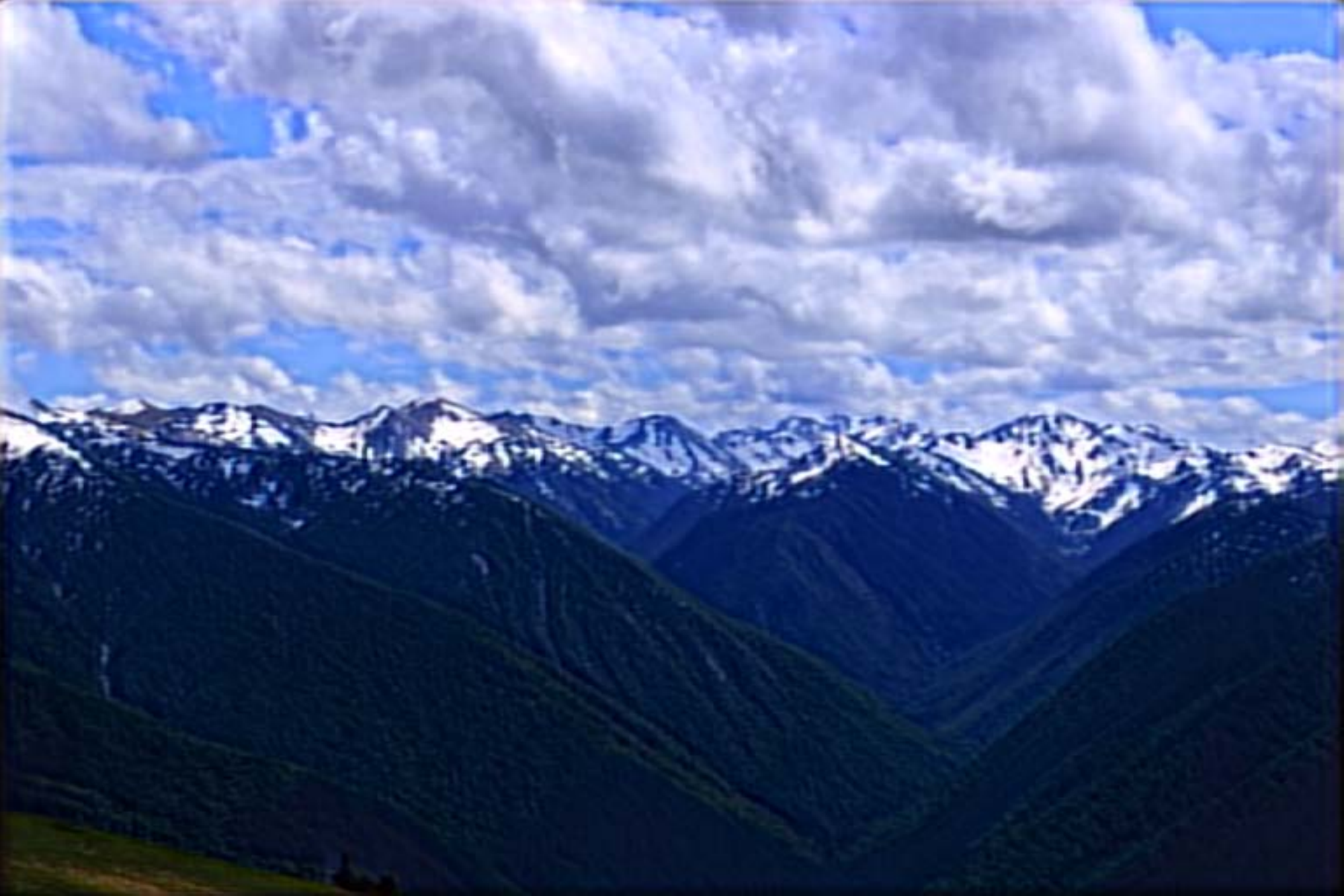} & \hspace{-0.4cm}
			\includegraphics[width = 0.115\textwidth]{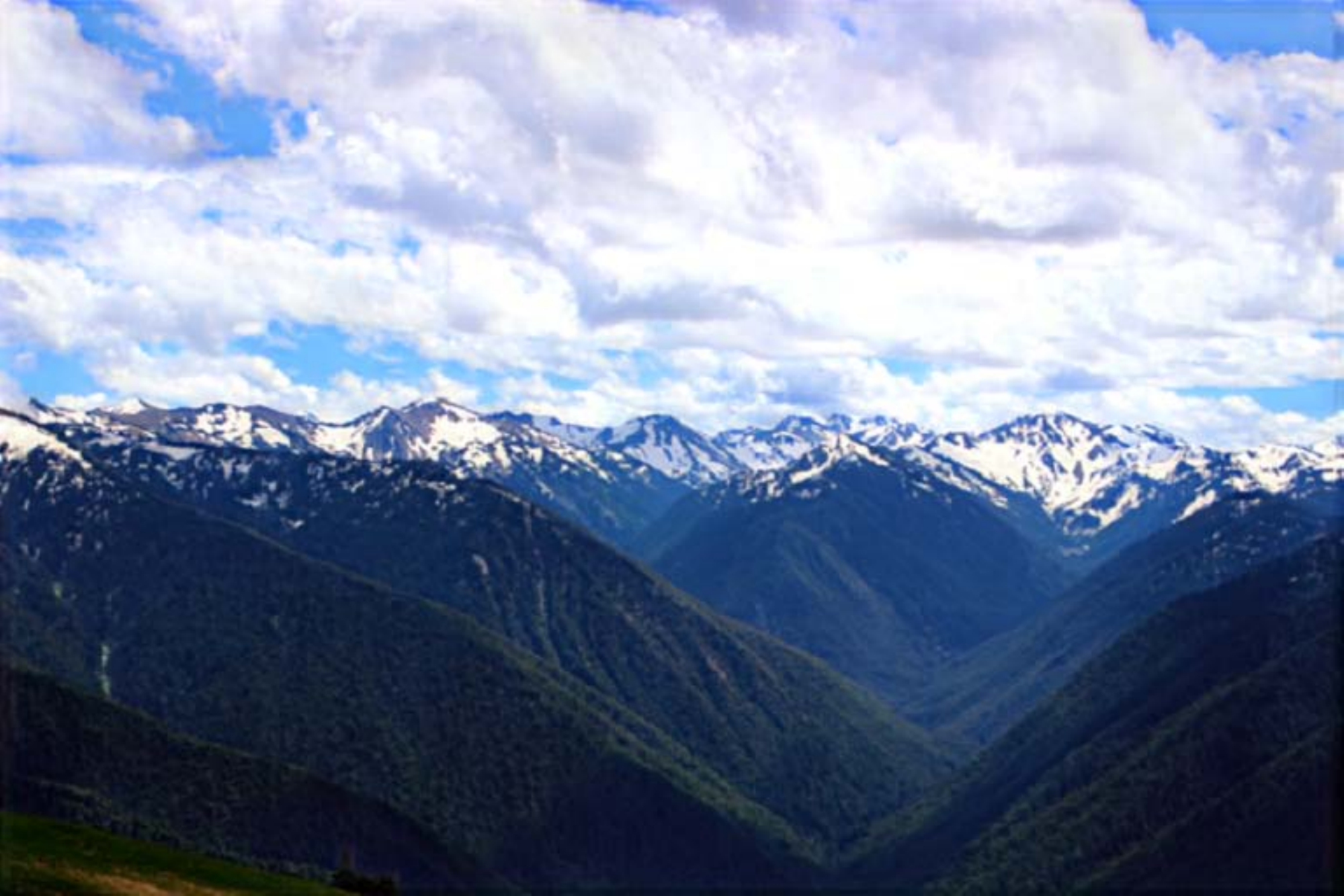} \\
			(a) Hazy inputs & \hspace{-0.4cm}
			(b) Without gating & \hspace{-0.4cm}
			(c) Without fusion & \hspace{-0.4cm}
			(d) GFN
		\end{tabular}
	\end{center}
	\vspace{-0.3cm}
	\caption{Effectiveness of the gated fusion network.
	}
	\vspace{-0.6cm}
	\label{fig-fusion}
\end{figure}

\subsection{Effectiveness of Gating Strategy}
\label{sec-fusion}
Image fusion is a method to blend several images into a single one by
retaining only the most useful features.
To blend effectively the information of the derived inputs, we filter their important information by computing corresponding confidence maps.
Consequently, in our gated fusion network, the derived inputs
are gated by three pixel-wise confidence maps that aim to preserve the regions with
good visibility.
Our fusion network has two advantages:
the first one is that it can reduce patch-based artifacts (\eg dark channel prior~\cite{he2011singlecvpr})
by single pixel operations, and the other one is that it can eliminate the
influence caused by transmission and atmospheric light estimation.

To show the effectiveness of fusion network, we also train an end-to-end network without fusion process. This network has the same architecture as DFN except the input is hazy image and output is dehazed result without confidence maps learning.
In addition, we also conduct a experiment based on equivalent fusion strategy, \ie, all the three derived inputs are weighted equally using $1/3$.
Figure~\ref{fig-fusion} shows visual comparisons of on two real-world examples with different settings. In these examples, the approach without gating generates very dark images in Figure~\ref{fig-fusion}(b), and the method without fusion strategy generates results with color distortion and dark regions as shown in Figure~\ref{fig-fusion}(c).
In contrast, our results recover most scene details and maintain the original
colors.
%

\subsection{Limitations}
\begin{figure}[t]\scriptsize
	\begin{center}
		\begin{tabular}{@{}cccc@{}}
			\includegraphics[width = 0.11\textwidth]{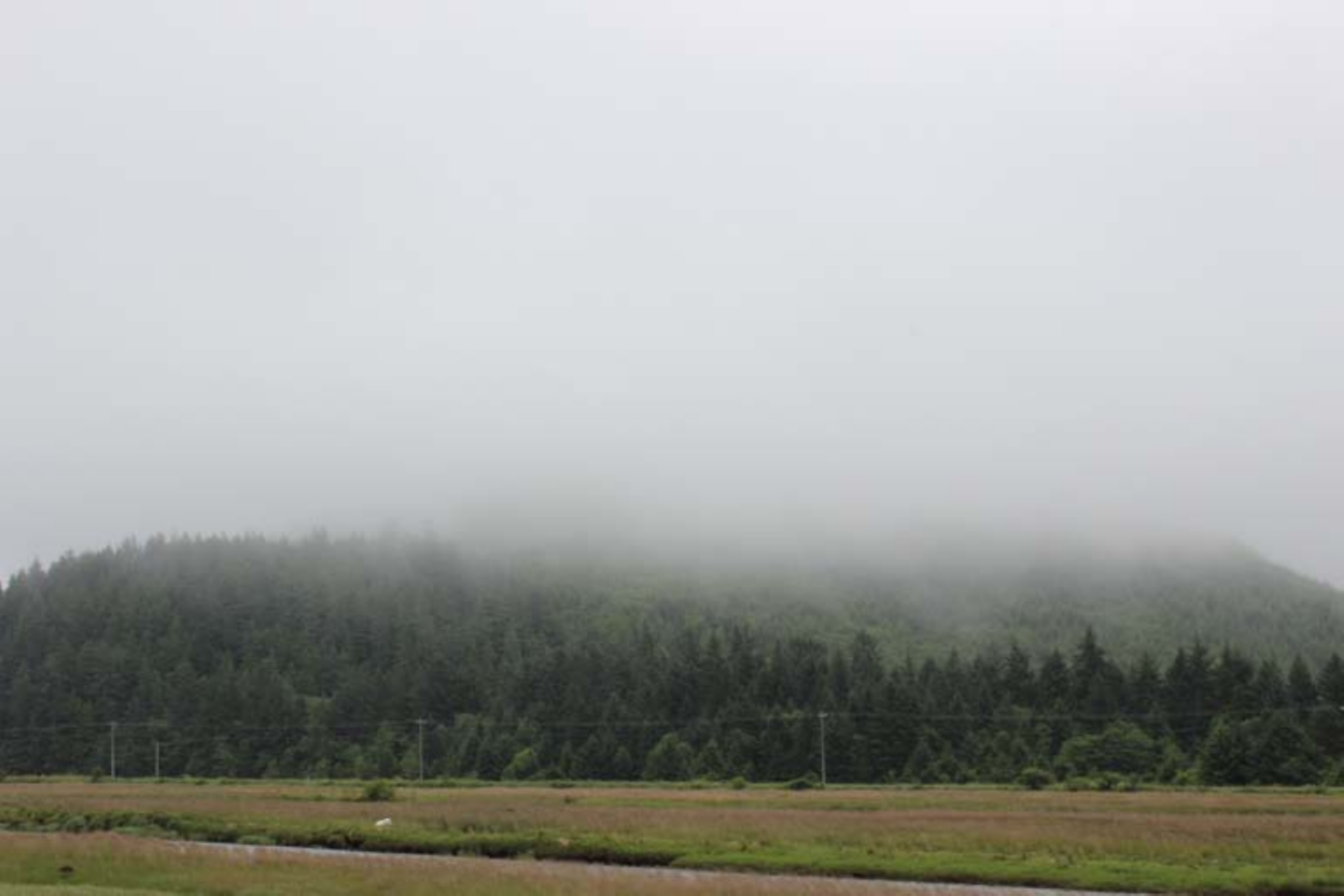} & \hspace{-0.4cm}
			\includegraphics[width = 0.11\textwidth]{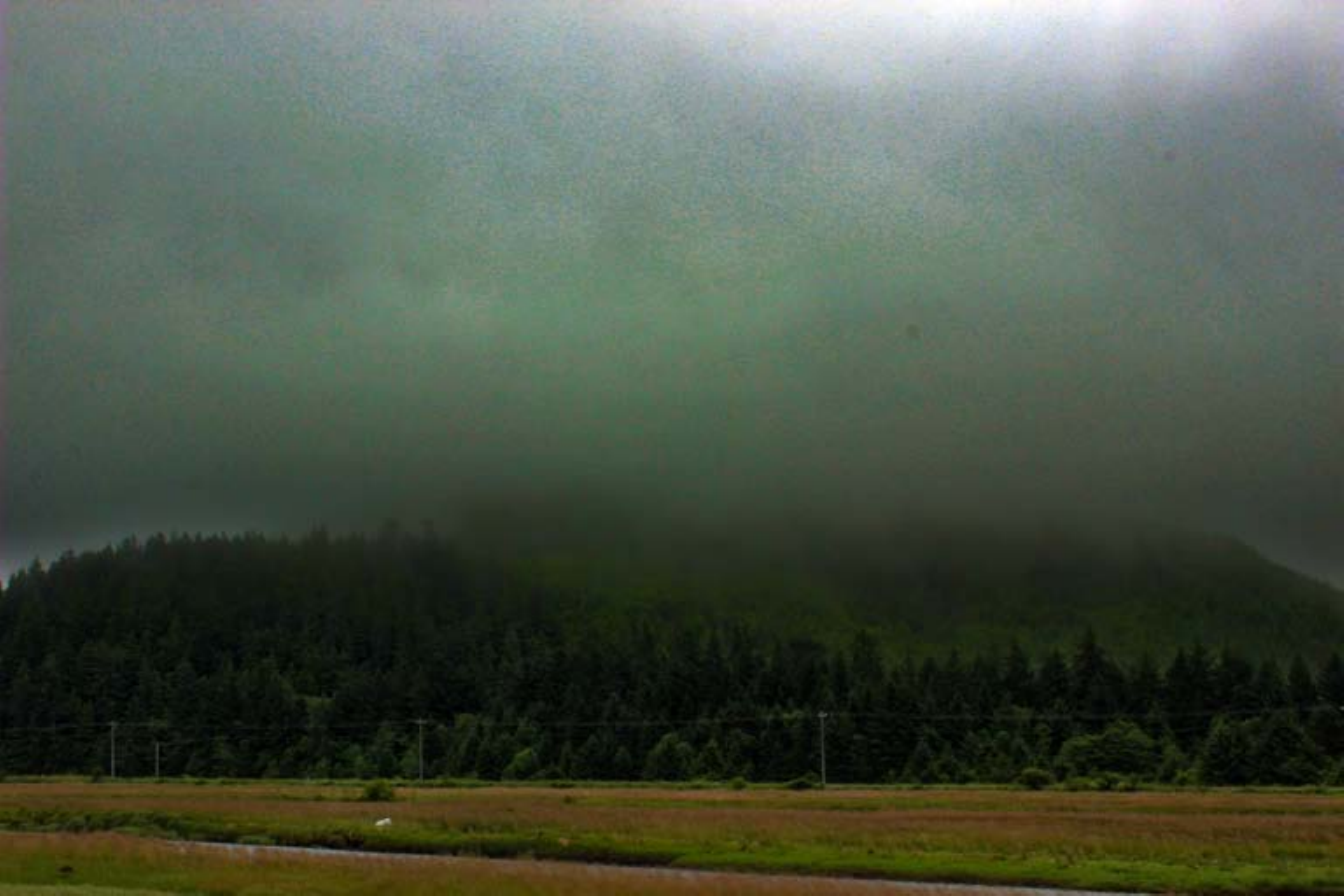} & \hspace{-0.4cm}
			\includegraphics[width = 0.11\textwidth]{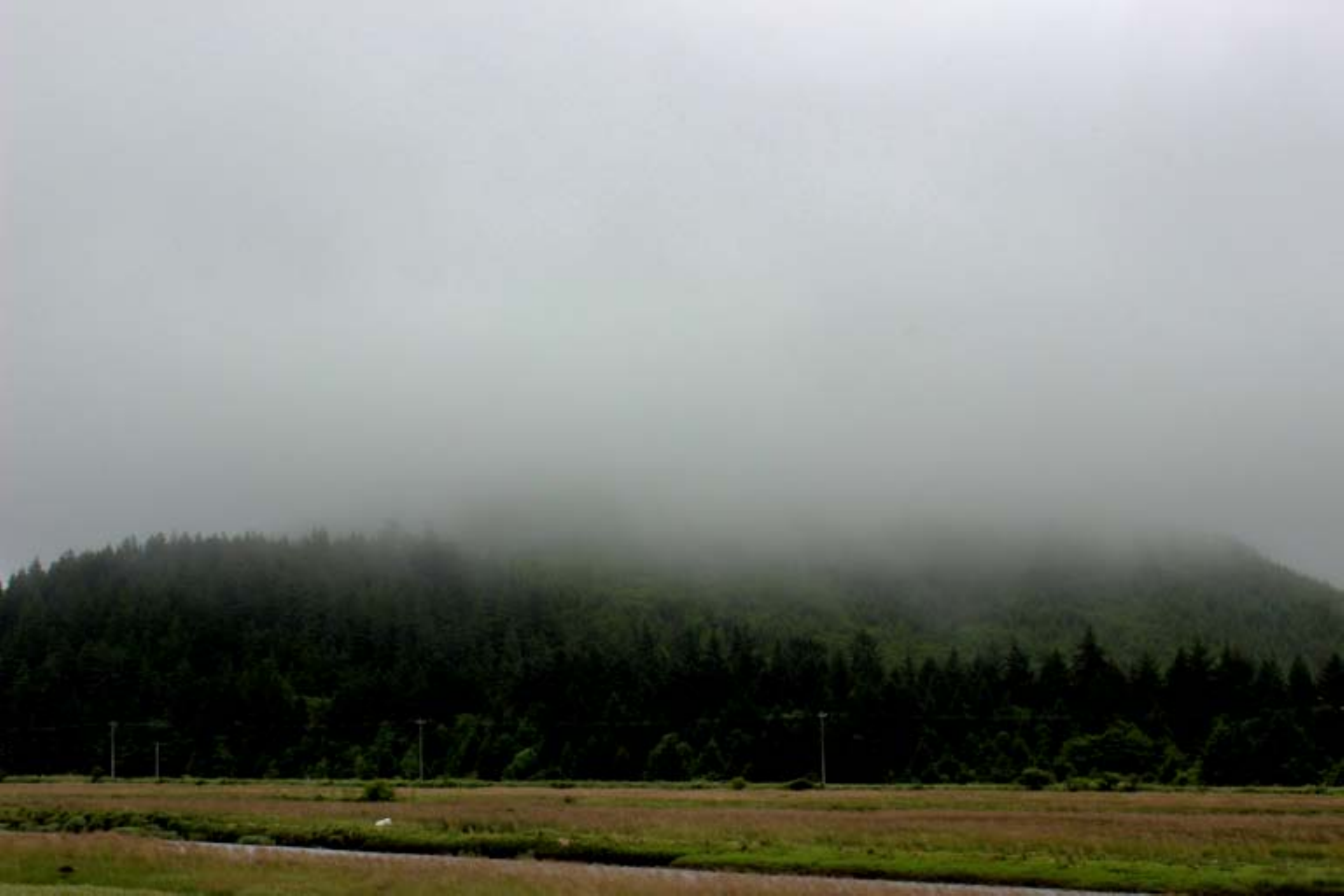} & \hspace{-0.4cm}
			\includegraphics[width = 0.11\textwidth]{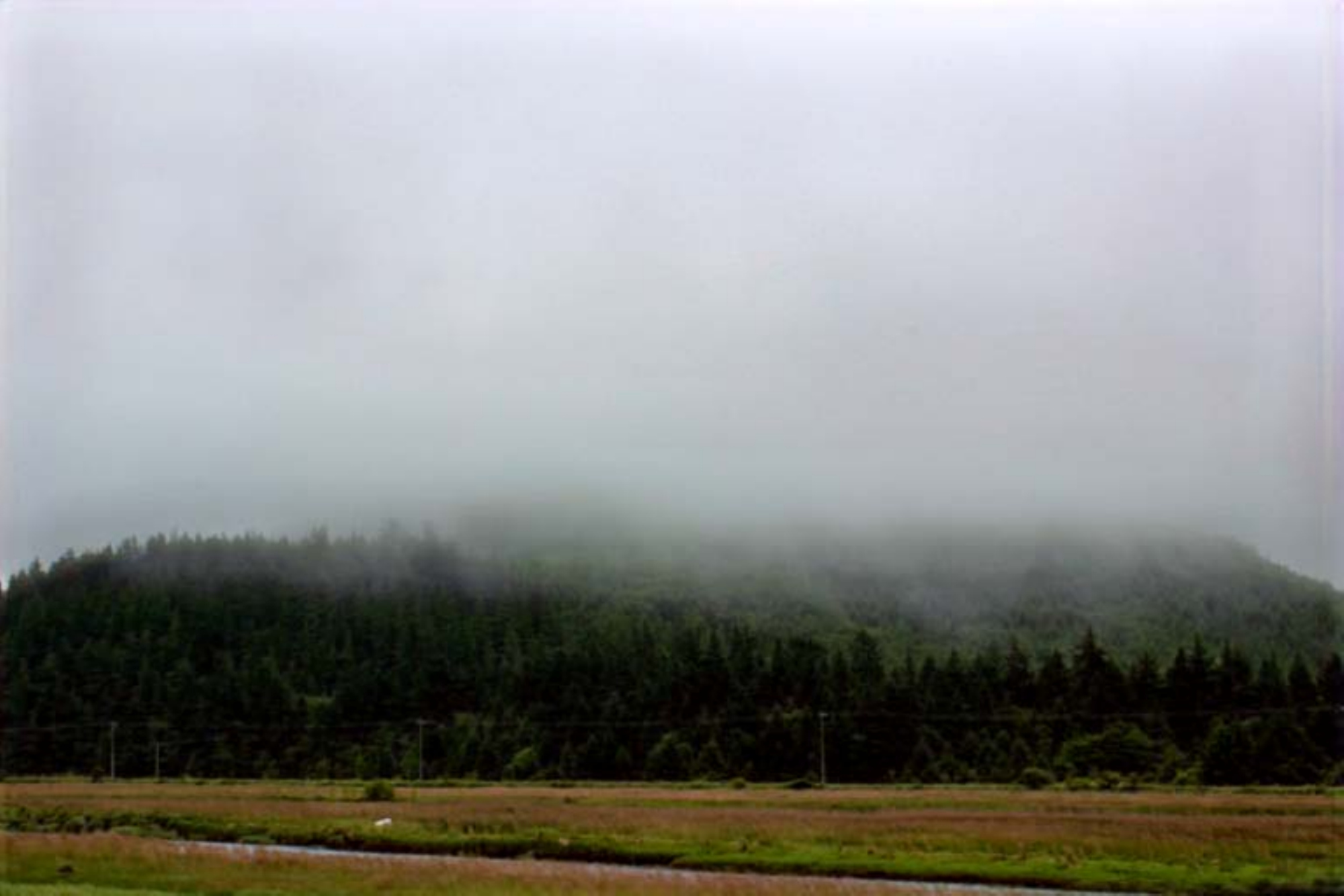} \\
			(a) Hazy input & \hspace{-0.4cm}
			(b) DCP~\cite{he2011singlecvpr} & \hspace{-0.4cm}
			(b) DehazeNet~\cite{cai2016dehazenet} & \hspace{-0.4cm}
			(d) GFN
		\end{tabular}
	\end{center}
	\vspace{-0.1cm}
	\caption{A failure case for a thick foggy image.
	}
	\vspace{-0.5cm}
	\label{fig-Limitations}
\end{figure}

The proposed DFN performs well in general natural images.
However, as the previous methods \cite{ren2016single,cai2016dehazenet}, a limitation of our method
is that the DFN cannot handle corrupted images with very large fog as shown
in Figure~\ref{fig-Limitations}.
As heavy haze seriously interferes the atmospheric light (which is not a constant), the hazy model does not hold for such examples.
Figure~\ref{fig-Limitations}(d) shows an example where the proposed method does not generate a clear image.
Future work will consider this problem with haze-free reference retrieval based on an effective deep neural network model.

%
\vspace{-1mm}
\section{Conclusions}
\vspace{-1mm}
In this paper, we addressed the single image dehazing problem via a multi-scale gated fusion network (GFN), a fusion based encoder-decoder architecture, by learning confidence maps for derived inputs.
Compared with previous methods which impose restrictions on scene transmission and atmospheric light, our proposed GFN is easy to implement and reproduce since the proposed
approach does not rely on the estimations of transmission and atmospheric light.
In the approach, we first applied white balance method to recover the scene color, and then generated two contrast enhanced images for better visibility.
Third, we carried out the GFN to estimate the confidence map for each derived input.
Finally, we used the confidence maps and derived inputs to render the final dehazed result.
The experimental results on synthetic and real-world images demonstrate the effectiveness of the proposed approach.

\vspace{-1mm}
{\flushleft \textbf{Acknowledgments.}} This work is supported in part by National Key Research and Development Plan (No.2016YFB0800603), National Natural Science Foundation of China (No.U1636214, 61733007), Beijing Natural Science Foundation (No.4172068). W. Ren and M.-H. Yang are supported by NSF CAREER (No. 1149783) and the Open Project Program of the National Laboratory of Pattern Recognition (NLPR).


{\small
	\bibliographystyle{ieee}
	\bibliography{dehazing}
}

\end{document}